\documentclass{article}
\usepackage{float}
\usepackage{iclr2027_conference,times}

\usepackage{amsmath,amsfonts,bm}

\def\eqref#1{equation~\ref{#1}}

\def\1{\bm{1}}

\DeclareMathAlphabet{\mathsfit}{\encodingdefault}{\sfdefault}{m}{sl}
\SetMathAlphabet{\mathsfit}{bold}{\encodingdefault}{\sfdefault}{bx}{n}

\newif\ifanon
\anontrue
\newcommand{\method}{MAEM}

\makeatletter
\@ifpackageloaded{xcolor}{}{\@ifpackageloaded{color}{}{\RequirePackage{xcolor}}}
\newif\ifN@draft \N@drafttrue
\newcommand{\Ndraft}{\N@drafttrue}
\newcommand{\Nfinal}{\N@draftfalse}
\newcommand{\N@missing}[1]{\textcolor{red}{??#1??}}
\newcommand{\N@pendingtext}{\textperiodcentered\textperiodcentered\textperiodcentered}
\newcommand{\Npendingtext}[1]{\renewcommand{\N@pendingtext}{#1}}
\newcommand{\N@pending}{\textcolor{gray}{\N@pendingtext}}
\newcommand{\N@ifrow}[2]{%
  \expandafter\ifx\csname Nstatus@#1\endcsname\relax#2\else\N@pending\fi}
\newcommand{\N@sup}[1]{\ifmmode^{\textnormal{\tiny #1}}\else\textsuperscript{\tiny #1}\fi}
\def\N@st@placeholder{placeholder}
\def\N@st@provisional{provisional}
\newcommand{\N@mark}[1]{%
  \ifN@draft
    \expandafter\ifx\csname Nstatus@#1\endcsname\relax\else
      \expandafter\let\expandafter\N@st\csname Nstatus@#1\endcsname
      \ifx\N@st\N@st@placeholder\N@sup{p}\else
        \ifx\N@st\N@st@provisional\N@sup{\textasciitilde}\fi
      \fi
    \fi
  \fi}
\newcommand{\N}[1]{%
  \@ifundefined{N@#1}{\N@ifrow{#1}{\N@missing{#1}}}%
                     {\csname N@#1\endcsname\N@mark{#1}}}
\newcommand{\Nse}[1]{%
  \@ifundefined{Nse@#1}{\N@ifrow{#1}{\N@missing{#1:se}}}{\csname Nse@#1\endcsname}}
\newcommand{\Nn}[1]{%
  \@ifundefined{Nn@#1}{\N@ifrow{#1}{\N@missing{#1:n}}}{\csname Nn@#1\endcsname}}
\newcommand{\Nci}[1]{%
  \@ifundefined{Nlo@#1}{\N@ifrow{#1}{\N@missing{#1:ci}}}{%
    \@ifundefined{Nhi@#1}{\N@ifrow{#1}{\N@missing{#1:ci}}}{%
      [\csname Nlo@#1\endcsname, \csname Nhi@#1\endcsname]}}}
\@ifundefined{fpeval}{\RequirePackage{xfp}}{}
\newcommand{\Npct}[1]{%
  \@ifundefined{N@#1}{\N@ifrow{#1}{\N@missing{#1:pct}}}%
    {\fpeval{round(100*\csname N@#1\endcsname,0)}\%\N@mark{#1}}}
\newcommand{\Nr}[1]{%
  \@ifundefined{N@#1}{\N@ifrow{#1}{\N@missing{#1:r}}}%
    {\fpeval{round(\csname N@#1\endcsname,2)}\N@mark{#1}}}
\makeatother
\expandafter\gdef\csname N@fid.ra.ex.cos.bo1\endcsname{0.7590}
\expandafter\gdef\csname Nse@fid.ra.ex.cos.bo1\endcsname{0.0060}
\expandafter\gdef\csname Nn@fid.ra.ex.cos.bo1\endcsname{486}
\expandafter\gdef\csname Nstatus@fid.ra.ex.cos.bo1\endcsname{final}
\expandafter\gdef\csname N@fid.ra.ex.cos.bo8\endcsname{0.8219}
\expandafter\gdef\csname Nse@fid.ra.ex.cos.bo8\endcsname{0.0051}
\expandafter\gdef\csname Nn@fid.ra.ex.cos.bo8\endcsname{486}
\expandafter\gdef\csname Nstatus@fid.ra.ex.cos.bo8\endcsname{final}
\expandafter\gdef\csname N@fid.ra.corp10.cos\endcsname{0.5720}
\expandafter\gdef\csname Nse@fid.ra.corp10.cos\endcsname{0.0052}
\expandafter\gdef\csname Nn@fid.ra.corp10.cos\endcsname{486}
\expandafter\gdef\csname Nstatus@fid.ra.corp10.cos\endcsname{final}
\expandafter\gdef\csname N@fid.ra.gcg.cos\endcsname{0.3904}
\expandafter\gdef\csname Nse@fid.ra.gcg.cos\endcsname{0.0218}
\expandafter\gdef\csname Nlo@fid.ra.gcg.cos\endcsname{0.3477}
\expandafter\gdef\csname Nhi@fid.ra.gcg.cos\endcsname{0.4332}
\expandafter\gdef\csname Nn@fid.ra.gcg.cos\endcsname{64}
\expandafter\gdef\csname Nstatus@fid.ra.gcg.cos\endcsname{final}
\expandafter\gdef\csname N@fid.ra.epo.cos\endcsname{0.3961}
\expandafter\gdef\csname Nse@fid.ra.epo.cos\endcsname{0.0188}
\expandafter\gdef\csname Nlo@fid.ra.epo.cos\endcsname{0.3592}
\expandafter\gdef\csname Nhi@fid.ra.epo.cos\endcsname{0.4330}
\expandafter\gdef\csname Nn@fid.ra.epo.cos\endcsname{64}
\expandafter\gdef\csname Nstatus@fid.ra.epo.cos\endcsname{final}
\expandafter\gdef\csname N@fid.ra.base.cos.bo8\endcsname{0.1914}
\expandafter\gdef\csname Nse@fid.ra.base.cos.bo8\endcsname{0.0037}
\expandafter\gdef\csname Nn@fid.ra.base.cos.bo8\endcsname{486}
\expandafter\gdef\csname Nstatus@fid.ra.base.cos.bo8\endcsname{final}
\expandafter\gdef\csname N@fid.rnd.ex.cos.bo1\endcsname{0.0485}
\expandafter\gdef\csname Nse@fid.rnd.ex.cos.bo1\endcsname{0.0003}
\expandafter\gdef\csname Nn@fid.rnd.ex.cos.bo1\endcsname{512}
\expandafter\gdef\csname Nstatus@fid.rnd.ex.cos.bo1\endcsname{final}
\expandafter\gdef\csname N@sae.l131k.ex.ratio.bo1.q1\endcsname{0.1211}
\expandafter\gdef\csname Nse@sae.l131k.ex.ratio.bo1.q1\endcsname{0.0437}
\expandafter\gdef\csname Nlo@sae.l131k.ex.ratio.bo1.q1\endcsname{0.0455}
\expandafter\gdef\csname Nhi@sae.l131k.ex.ratio.bo1.q1\endcsname{0.2063}
\expandafter\gdef\csname Nn@sae.l131k.ex.ratio.bo1.q1\endcsname{128}
\expandafter\gdef\csname Nstatus@sae.l131k.ex.ratio.bo1.q1\endcsname{final}
\expandafter\gdef\csname N@sae.l131k.ex.ratio.bo1.q2\endcsname{0.8122}
\expandafter\gdef\csname Nse@sae.l131k.ex.ratio.bo1.q2\endcsname{0.0151}
\expandafter\gdef\csname Nlo@sae.l131k.ex.ratio.bo1.q2\endcsname{0.7864}
\expandafter\gdef\csname Nhi@sae.l131k.ex.ratio.bo1.q2\endcsname{0.8416}
\expandafter\gdef\csname Nn@sae.l131k.ex.ratio.bo1.q2\endcsname{128}
\expandafter\gdef\csname Nstatus@sae.l131k.ex.ratio.bo1.q2\endcsname{final}
\expandafter\gdef\csname N@sae.l131k.ex.ratio.bo1.q3\endcsname{0.7647}
\expandafter\gdef\csname Nse@sae.l131k.ex.ratio.bo1.q3\endcsname{0.0196}
\expandafter\gdef\csname Nlo@sae.l131k.ex.ratio.bo1.q3\endcsname{0.7364}
\expandafter\gdef\csname Nhi@sae.l131k.ex.ratio.bo1.q3\endcsname{0.8034}
\expandafter\gdef\csname Nn@sae.l131k.ex.ratio.bo1.q3\endcsname{128}
\expandafter\gdef\csname Nstatus@sae.l131k.ex.ratio.bo1.q3\endcsname{final}
\expandafter\gdef\csname N@sae.l131k.ex.ratio.bo1.q4\endcsname{0.7209}
\expandafter\gdef\csname Nse@sae.l131k.ex.ratio.bo1.q4\endcsname{0.0300}
\expandafter\gdef\csname Nlo@sae.l131k.ex.ratio.bo1.q4\endcsname{0.6698}
\expandafter\gdef\csname Nhi@sae.l131k.ex.ratio.bo1.q4\endcsname{0.7703}
\expandafter\gdef\csname Nn@sae.l131k.ex.ratio.bo1.q4\endcsname{128}
\expandafter\gdef\csname Nstatus@sae.l131k.ex.ratio.bo1.q4\endcsname{final}
\expandafter\gdef\csname N@sae.l131k.ex.ratio.bo1\endcsname{0.7210}
\expandafter\gdef\csname Nse@sae.l131k.ex.ratio.bo1\endcsname{0.0143}
\expandafter\gdef\csname Nlo@sae.l131k.ex.ratio.bo1\endcsname{0.6900}
\expandafter\gdef\csname Nhi@sae.l131k.ex.ratio.bo1\endcsname{0.7513}
\expandafter\gdef\csname Nn@sae.l131k.ex.ratio.bo1\endcsname{512}
\expandafter\gdef\csname Nstatus@sae.l131k.ex.ratio.bo1\endcsname{final}
\expandafter\gdef\csname N@sae.l131k.ex.ratio.bo64\endcsname{0.9678}
\expandafter\gdef\csname Nse@sae.l131k.ex.ratio.bo64\endcsname{0.0163}
\expandafter\gdef\csname Nlo@sae.l131k.ex.ratio.bo64\endcsname{0.9353}
\expandafter\gdef\csname Nhi@sae.l131k.ex.ratio.bo64\endcsname{1.0046}
\expandafter\gdef\csname Nn@sae.l131k.ex.ratio.bo64\endcsname{512}
\expandafter\gdef\csname Nstatus@sae.l131k.ex.ratio.bo64\endcsname{final}
\expandafter\gdef\csname N@sae.l131k.ex.fired.bo1\endcsname{0.8319}
\expandafter\gdef\csname Nse@sae.l131k.ex.fired.bo1\endcsname{0.0150}
\expandafter\gdef\csname Nlo@sae.l131k.ex.fired.bo1\endcsname{0.8021}
\expandafter\gdef\csname Nhi@sae.l131k.ex.fired.bo1\endcsname{0.8608}
\expandafter\gdef\csname Nn@sae.l131k.ex.fired.bo1\endcsname{512}
\expandafter\gdef\csname Nstatus@sae.l131k.ex.fired.bo1\endcsname{final}
\expandafter\gdef\csname N@ood.en.diff.cos.bo8\endcsname{0.2420}
\expandafter\gdef\csname Nse@ood.en.diff.cos.bo8\endcsname{0.0052}
\expandafter\gdef\csname Nlo@ood.en.diff.cos.bo8\endcsname{0.2318}
\expandafter\gdef\csname Nhi@ood.en.diff.cos.bo8\endcsname{0.2522}
\expandafter\gdef\csname Nn@ood.en.diff.cos.bo8\endcsname{512}
\expandafter\gdef\csname Nstatus@ood.en.diff.cos.bo8\endcsname{final}
\expandafter\gdef\csname N@ood.python.diff.cos.bo8\endcsname{0.0477}
\expandafter\gdef\csname Nse@ood.python.diff.cos.bo8\endcsname{0.0071}
\expandafter\gdef\csname Nlo@ood.python.diff.cos.bo8\endcsname{0.0344}
\expandafter\gdef\csname Nhi@ood.python.diff.cos.bo8\endcsname{0.0608}
\expandafter\gdef\csname Nn@ood.python.diff.cos.bo8\endcsname{512}
\expandafter\gdef\csname Nstatus@ood.python.diff.cos.bo8\endcsname{final}
\expandafter\gdef\csname N@ood.tha.diff.cos.bo8\endcsname{-0.0486}
\expandafter\gdef\csname Nse@ood.tha.diff.cos.bo8\endcsname{0.0055}
\expandafter\gdef\csname Nlo@ood.tha.diff.cos.bo8\endcsname{-0.0592}
\expandafter\gdef\csname Nhi@ood.tha.diff.cos.bo8\endcsname{-0.0380}
\expandafter\gdef\csname Nn@ood.tha.diff.cos.bo8\endcsname{512}
\expandafter\gdef\csname Nstatus@ood.tha.diff.cos.bo8\endcsname{final}
\expandafter\gdef\csname N@ood.isabelle.diff.cos.bo8\endcsname{-0.0889}
\expandafter\gdef\csname Nse@ood.isabelle.diff.cos.bo8\endcsname{0.0051}
\expandafter\gdef\csname Nlo@ood.isabelle.diff.cos.bo8\endcsname{-0.0990}
\expandafter\gdef\csname Nhi@ood.isabelle.diff.cos.bo8\endcsname{-0.0788}
\expandafter\gdef\csname Nn@ood.isabelle.diff.cos.bo8\endcsname{512}
\expandafter\gdef\csname Nstatus@ood.isabelle.diff.cos.bo8\endcsname{final}
\expandafter\gdef\csname N@ood.en.corp10.cos\endcsname{0.5857}
\expandafter\gdef\csname Nn@ood.en.corp10.cos\endcsname{512}
\expandafter\gdef\csname Nstatus@ood.en.corp10.cos\endcsname{final}
\expandafter\gdef\csname N@ood.en.base.cos.bo8\endcsname{0.1882}
\expandafter\gdef\csname Nn@ood.en.base.cos.bo8\endcsname{512}
\expandafter\gdef\csname Nstatus@ood.en.base.cos.bo8\endcsname{final}
\expandafter\gdef\csname N@ood.python.base.cos.bo8\endcsname{0.2216}
\expandafter\gdef\csname Nn@ood.python.base.cos.bo8\endcsname{512}
\expandafter\gdef\csname Nstatus@ood.python.base.cos.bo8\endcsname{final}
\expandafter\gdef\csname N@ai.l131k.c16.det\endcsname{0.810}
\expandafter\gdef\csname Nlo@ai.l131k.c16.det\endcsname{0.801}
\expandafter\gdef\csname Nhi@ai.l131k.c16.det\endcsname{0.819}
\expandafter\gdef\csname Nn@ai.l131k.c16.det\endcsname{506}
\expandafter\gdef\csname Nstatus@ai.l131k.c16.det\endcsname{final}
\expandafter\gdef\csname N@ai.l131k.c16.fuzz\endcsname{0.775}
\expandafter\gdef\csname Nlo@ai.l131k.c16.fuzz\endcsname{0.766}
\expandafter\gdef\csname Nhi@ai.l131k.c16.fuzz\endcsname{0.784}
\expandafter\gdef\csname Nn@ai.l131k.c16.fuzz\endcsname{506}
\expandafter\gdef\csname Nstatus@ai.l131k.c16.fuzz\endcsname{final}
\expandafter\gdef\csname N@ai.l131k.mtop16.det\endcsname{0.680}
\expandafter\gdef\csname Nlo@ai.l131k.mtop16.det\endcsname{0.668}
\expandafter\gdef\csname Nhi@ai.l131k.mtop16.det\endcsname{0.691}
\expandafter\gdef\csname Nn@ai.l131k.mtop16.det\endcsname{456}
\expandafter\gdef\csname Nstatus@ai.l131k.mtop16.det\endcsname{final}
\expandafter\gdef\csname N@ai.l131k.mtop16.fuzz\endcsname{0.644}
\expandafter\gdef\csname Nlo@ai.l131k.mtop16.fuzz\endcsname{0.633}
\expandafter\gdef\csname Nhi@ai.l131k.mtop16.fuzz\endcsname{0.655}
\expandafter\gdef\csname Nn@ai.l131k.mtop16.fuzz\endcsname{456}
\expandafter\gdef\csname Nstatus@ai.l131k.mtop16.fuzz\endcsname{final}
\expandafter\gdef\csname N@ai.l131k.nla4.det\endcsname{0.668}
\expandafter\gdef\csname Nlo@ai.l131k.nla4.det\endcsname{0.655}
\expandafter\gdef\csname Nhi@ai.l131k.nla4.det\endcsname{0.681}
\expandafter\gdef\csname Nn@ai.l131k.nla4.det\endcsname{471}
\expandafter\gdef\csname Nstatus@ai.l131k.nla4.det\endcsname{final}
\expandafter\gdef\csname N@ai.l131k.nla4.fuzz\endcsname{0.639}
\expandafter\gdef\csname Nlo@ai.l131k.nla4.fuzz\endcsname{0.628}
\expandafter\gdef\csname Nhi@ai.l131k.nla4.fuzz\endcsname{0.651}
\expandafter\gdef\csname Nn@ai.l131k.nla4.fuzz\endcsname{470}
\expandafter\gdef\csname Nstatus@ai.l131k.nla4.fuzz\endcsname{final}
\expandafter\gdef\csname N@ai.l131k.floor.det\endcsname{0.504}
\expandafter\gdef\csname Nlo@ai.l131k.floor.det\endcsname{0.500}
\expandafter\gdef\csname Nhi@ai.l131k.floor.det\endcsname{0.507}
\expandafter\gdef\csname Nn@ai.l131k.floor.det\endcsname{506}
\expandafter\gdef\csname Nstatus@ai.l131k.floor.det\endcsname{final}
\expandafter\gdef\csname N@ai.l131k.floor.fuzz\endcsname{0.503}
\expandafter\gdef\csname Nlo@ai.l131k.floor.fuzz\endcsname{0.501}
\expandafter\gdef\csname Nhi@ai.l131k.floor.fuzz\endcsname{0.505}
\expandafter\gdef\csname Nn@ai.l131k.floor.fuzz\endcsname{506}
\expandafter\gdef\csname Nstatus@ai.l131k.floor.fuzz\endcsname{final}
\expandafter\gdef\csname N@ai.l131k.c16.tpr\endcsname{0.708}
\expandafter\gdef\csname Nlo@ai.l131k.c16.tpr\endcsname{0.693}
\expandafter\gdef\csname Nhi@ai.l131k.c16.tpr\endcsname{0.722}
\expandafter\gdef\csname Nn@ai.l131k.c16.tpr\endcsname{506}
\expandafter\gdef\csname Nstatus@ai.l131k.c16.tpr\endcsname{final}
\expandafter\gdef\csname N@ai.l131k.diff.det\endcsname{-0.132}
\expandafter\gdef\csname Nlo@ai.l131k.diff.det\endcsname{-0.143}
\expandafter\gdef\csname Nhi@ai.l131k.diff.det\endcsname{-0.123}
\expandafter\gdef\csname Nn@ai.l131k.diff.det\endcsname{453}
\expandafter\gdef\csname Nstatus@ai.l131k.diff.det\endcsname{final}
\expandafter\gdef\csname N@ai.l131k.diffnla.det\endcsname{-0.003}
\expandafter\gdef\csname Nlo@ai.l131k.diffnla.det\endcsname{-0.015}
\expandafter\gdef\csname Nhi@ai.l131k.diffnla.det\endcsname{0.010}
\expandafter\gdef\csname Nn@ai.l131k.diffnla.det\endcsname{423}
\expandafter\gdef\csname Nstatus@ai.l131k.diffnla.det\endcsname{final}
\expandafter\gdef\csname N@ai.l131k.mcos16.det\endcsname{0.690}
\expandafter\gdef\csname Nlo@ai.l131k.mcos16.det\endcsname{0.678}
\expandafter\gdef\csname Nhi@ai.l131k.mcos16.det\endcsname{0.702}
\expandafter\gdef\csname Nn@ai.l131k.mcos16.det\endcsname{449}
\expandafter\gdef\csname Nstatus@ai.l131k.mcos16.det\endcsname{final}
\expandafter\gdef\csname N@ai.l131k.mjac16.det\endcsname{0.694}
\expandafter\gdef\csname Nlo@ai.l131k.mjac16.det\endcsname{0.681}
\expandafter\gdef\csname Nhi@ai.l131k.mjac16.det\endcsname{0.706}
\expandafter\gdef\csname Nn@ai.l131k.mjac16.det\endcsname{444}
\expandafter\gdef\csname Nstatus@ai.l131k.mjac16.det\endcsname{final}
\expandafter\gdef\csname N@ai.l131k.mtop16.tpr\endcsname{0.375}
\expandafter\gdef\csname Nlo@ai.l131k.mtop16.tpr\endcsname{0.351}
\expandafter\gdef\csname Nhi@ai.l131k.mtop16.tpr\endcsname{0.398}
\expandafter\gdef\csname Nn@ai.l131k.mtop16.tpr\endcsname{456}
\expandafter\gdef\csname Nstatus@ai.l131k.mtop16.tpr\endcsname{final}
\expandafter\gdef\csname N@ai.l131k.null.det\endcsname{0.797}
\expandafter\gdef\csname Nlo@ai.l131k.null.det\endcsname{0.786}
\expandafter\gdef\csname Nhi@ai.l131k.null.det\endcsname{0.806}
\expandafter\gdef\csname Nn@ai.l131k.null.det\endcsname{477}
\expandafter\gdef\csname Nstatus@ai.l131k.null.det\endcsname{final}
\expandafter\gdef\csname N@fid.ra.corp10.slope\endcsname{0.0251}
\expandafter\gdef\csname Nse@fid.ra.corp10.slope\endcsname{0.0010}
\expandafter\gdef\csname Nn@fid.ra.corp10.slope\endcsname{486}
\expandafter\gdef\csname Nstatus@fid.ra.corp10.slope\endcsname{final}
\expandafter\gdef\csname N@fid.ra.corp1p25.cos\endcsname{0.4965}
\expandafter\gdef\csname Nse@fid.ra.corp1p25.cos\endcsname{0.0048}
\expandafter\gdef\csname Nn@fid.ra.corp1p25.cos\endcsname{486}
\expandafter\gdef\csname Nstatus@fid.ra.corp1p25.cos\endcsname{final}
\expandafter\gdef\csname N@fid.ra.diff.cos.bo8\endcsname{0.2499}
\expandafter\gdef\csname Nse@fid.ra.diff.cos.bo8\endcsname{0.0052}
\expandafter\gdef\csname Nlo@fid.ra.diff.cos.bo8\endcsname{0.2397}
\expandafter\gdef\csname Nhi@fid.ra.diff.cos.bo8\endcsname{0.2602}
\expandafter\gdef\csname Nn@fid.ra.diff.cos.bo8\endcsname{486}
\expandafter\gdef\csname Nstatus@fid.ra.diff.cos.bo8\endcsname{final}
\expandafter\gdef\csname N@fid.ra.ex.win.bo8\endcsname{0.9712}
\expandafter\gdef\csname Nse@fid.ra.ex.win.bo8\endcsname{0.0077}
\expandafter\gdef\csname Nn@fid.ra.ex.win.bo8\endcsname{486}
\expandafter\gdef\csname Nstatus@fid.ra.ex.win.bo8\endcsname{final}
\expandafter\gdef\csname N@ood.cmn.diff.cos.bo8\endcsname{0.1476}
\expandafter\gdef\csname Nse@ood.cmn.diff.cos.bo8\endcsname{0.0057}
\expandafter\gdef\csname Nlo@ood.cmn.diff.cos.bo8\endcsname{0.1364}
\expandafter\gdef\csname Nhi@ood.cmn.diff.cos.bo8\endcsname{0.1590}
\expandafter\gdef\csname Nn@ood.cmn.diff.cos.bo8\endcsname{512}
\expandafter\gdef\csname Nstatus@ood.cmn.diff.cos.bo8\endcsname{final}
\expandafter\gdef\csname N@ood.conj.diff.nexceed\endcsname{12}
\expandafter\gdef\csname Nn@ood.conj.diff.nexceed\endcsname{21}
\expandafter\gdef\csname Nstatus@ood.conj.diff.nexceed\endcsname{final}
\expandafter\gdef\csname N@ood.conj.diff.ninconcl\endcsname{2}
\expandafter\gdef\csname Nn@ood.conj.diff.ninconcl\endcsname{21}
\expandafter\gdef\csname Nstatus@ood.conj.diff.ninconcl\endcsname{final}
\expandafter\gdef\csname N@ood.conj.diff.nreversed\endcsname{7}
\expandafter\gdef\csname Nn@ood.conj.diff.nreversed\endcsname{21}
\expandafter\gdef\csname Nstatus@ood.conj.diff.nreversed\endcsname{final}
\expandafter\gdef\csname N@ood.zh.diff.cos.bo8\endcsname{0.1831}
\expandafter\gdef\csname Nse@ood.zh.diff.cos.bo8\endcsname{0.0059}
\expandafter\gdef\csname Nlo@ood.zh.diff.cos.bo8\endcsname{0.1715}
\expandafter\gdef\csname Nhi@ood.zh.diff.cos.bo8\endcsname{0.1945}
\expandafter\gdef\csname Nn@ood.zh.diff.cos.bo8\endcsname{512}
\expandafter\gdef\csname Nstatus@ood.zh.diff.cos.bo8\endcsname{final}
\expandafter\gdef\csname N@sae.l131k.ex.fired.bo8.q1\endcsname{0.6249}
\expandafter\gdef\csname Nse@sae.l131k.ex.fired.bo8.q1\endcsname{0.0394}
\expandafter\gdef\csname Nlo@sae.l131k.ex.fired.bo8.q1\endcsname{0.5471}
\expandafter\gdef\csname Nhi@sae.l131k.ex.fired.bo8.q1\endcsname{0.7019}
\expandafter\gdef\csname Nn@sae.l131k.ex.fired.bo8.q1\endcsname{128}
\expandafter\gdef\csname Nstatus@sae.l131k.ex.fired.bo8.q1\endcsname{final}
\expandafter\gdef\csname N@sae.l131k.ex.fired.bo8.q4\endcsname{0.9780}
\expandafter\gdef\csname Nse@sae.l131k.ex.fired.bo8.q4\endcsname{0.0111}
\expandafter\gdef\csname Nlo@sae.l131k.ex.fired.bo8.q4\endcsname{0.9534}
\expandafter\gdef\csname Nhi@sae.l131k.ex.fired.bo8.q4\endcsname{0.9968}
\expandafter\gdef\csname Nn@sae.l131k.ex.fired.bo8.q4\endcsname{128}
\expandafter\gdef\csname Nstatus@sae.l131k.ex.fired.bo8.q4\endcsname{final}
\expandafter\gdef\csname N@sae.l131k.ex.ratio.bo8.q1\endcsname{0.2349}
\expandafter\gdef\csname Nse@sae.l131k.ex.ratio.bo8.q1\endcsname{0.0759}
\expandafter\gdef\csname Nlo@sae.l131k.ex.ratio.bo8.q1\endcsname{0.1139}
\expandafter\gdef\csname Nhi@sae.l131k.ex.ratio.bo8.q1\endcsname{0.4301}
\expandafter\gdef\csname Nn@sae.l131k.ex.ratio.bo8.q1\endcsname{128}
\expandafter\gdef\csname Nstatus@sae.l131k.ex.ratio.bo8.q1\endcsname{final}
\expandafter\gdef\csname Nstatus@ai.l131k.all.usd\endcsname{pending}
\expandafter\gdef\csname Nstatus@ai.l131k.all.usdcell\endcsname{pending}
\expandafter\gdef\csname N@ai.l131k.c16.det.q1\endcsname{0.774}
\expandafter\gdef\csname Nlo@ai.l131k.c16.det.q1\endcsname{0.755}
\expandafter\gdef\csname Nhi@ai.l131k.c16.det.q1\endcsname{0.792}
\expandafter\gdef\csname Nn@ai.l131k.c16.det.q1\endcsname{128}
\expandafter\gdef\csname Nstatus@ai.l131k.c16.det.q1\endcsname{final}
\expandafter\gdef\csname N@ai.l131k.c16.det.q2\endcsname{0.831}
\expandafter\gdef\csname Nlo@ai.l131k.c16.det.q2\endcsname{0.819}
\expandafter\gdef\csname Nhi@ai.l131k.c16.det.q2\endcsname{0.843}
\expandafter\gdef\csname Nn@ai.l131k.c16.det.q2\endcsname{126}
\expandafter\gdef\csname Nstatus@ai.l131k.c16.det.q2\endcsname{final}
\expandafter\gdef\csname N@ai.l131k.c16.det.q3\endcsname{0.841}
\expandafter\gdef\csname Nlo@ai.l131k.c16.det.q3\endcsname{0.827}
\expandafter\gdef\csname Nhi@ai.l131k.c16.det.q3\endcsname{0.854}
\expandafter\gdef\csname Nn@ai.l131k.c16.det.q3\endcsname{127}
\expandafter\gdef\csname Nstatus@ai.l131k.c16.det.q3\endcsname{final}
\expandafter\gdef\csname N@ai.l131k.c16.det.q4\endcsname{0.795}
\expandafter\gdef\csname Nlo@ai.l131k.c16.det.q4\endcsname{0.774}
\expandafter\gdef\csname Nhi@ai.l131k.c16.det.q4\endcsname{0.815}
\expandafter\gdef\csname Nn@ai.l131k.c16.det.q4\endcsname{125}
\expandafter\gdef\csname Nstatus@ai.l131k.c16.det.q4\endcsname{final}
\expandafter\gdef\csname N@ai.l131k.c16.tnr\endcsname{0.912}
\expandafter\gdef\csname Nlo@ai.l131k.c16.tnr\endcsname{0.900}
\expandafter\gdef\csname Nhi@ai.l131k.c16.tnr\endcsname{0.924}
\expandafter\gdef\csname Nn@ai.l131k.c16.tnr\endcsname{506}
\expandafter\gdef\csname Nstatus@ai.l131k.c16.tnr\endcsname{final}
\expandafter\gdef\csname N@ai.l131k.c16.tpr.b1\endcsname{0.372}
\expandafter\gdef\csname Nlo@ai.l131k.c16.tpr.b1\endcsname{0.345}
\expandafter\gdef\csname Nhi@ai.l131k.c16.tpr.b1\endcsname{0.401}
\expandafter\gdef\csname Nn@ai.l131k.c16.tpr.b1\endcsname{487}
\expandafter\gdef\csname Nstatus@ai.l131k.c16.tpr.b1\endcsname{final}
\expandafter\gdef\csname N@ai.l131k.c16.tpr.b2\endcsname{0.664}
\expandafter\gdef\csname Nlo@ai.l131k.c16.tpr.b2\endcsname{0.638}
\expandafter\gdef\csname Nhi@ai.l131k.c16.tpr.b2\endcsname{0.689}
\expandafter\gdef\csname Nn@ai.l131k.c16.tpr.b2\endcsname{500}
\expandafter\gdef\csname Nstatus@ai.l131k.c16.tpr.b2\endcsname{final}
\expandafter\gdef\csname N@ai.l131k.c16.tpr.b3\endcsname{0.842}
\expandafter\gdef\csname Nlo@ai.l131k.c16.tpr.b3\endcsname{0.819}
\expandafter\gdef\csname Nhi@ai.l131k.c16.tpr.b3\endcsname{0.862}
\expandafter\gdef\csname Nn@ai.l131k.c16.tpr.b3\endcsname{476}
\expandafter\gdef\csname Nstatus@ai.l131k.c16.tpr.b3\endcsname{final}
\expandafter\gdef\csname N@ai.l131k.c16.tpr.b4\endcsname{0.925}
\expandafter\gdef\csname Nlo@ai.l131k.c16.tpr.b4\endcsname{0.909}
\expandafter\gdef\csname Nhi@ai.l131k.c16.tpr.b4\endcsname{0.941}
\expandafter\gdef\csname Nn@ai.l131k.c16.tpr.b4\endcsname{506}
\expandafter\gdef\csname Nstatus@ai.l131k.c16.tpr.b4\endcsname{final}
\expandafter\gdef\csname Nstatus@ai.l131k.diff.det.q1\endcsname{pending}
\expandafter\gdef\csname Nstatus@ai.l131k.diff.det.q2\endcsname{pending}
\expandafter\gdef\csname Nstatus@ai.l131k.diff.det.q3\endcsname{pending}
\expandafter\gdef\csname Nstatus@ai.l131k.diff.det.q4\endcsname{pending}
\expandafter\gdef\csname Nstatus@ai.l131k.diffnla.det.q1\endcsname{pending}
\expandafter\gdef\csname Nstatus@ai.l131k.diffnla.det.q2\endcsname{pending}
\expandafter\gdef\csname Nstatus@ai.l131k.diffnla.det.q3\endcsname{pending}
\expandafter\gdef\csname Nstatus@ai.l131k.diffnla.det.q4\endcsname{pending}
\expandafter\gdef\csname N@ai.l131k.floor.tnr\endcsname{0.941}
\expandafter\gdef\csname Nlo@ai.l131k.floor.tnr\endcsname{0.929}
\expandafter\gdef\csname Nhi@ai.l131k.floor.tnr\endcsname{0.952}
\expandafter\gdef\csname Nn@ai.l131k.floor.tnr\endcsname{506}
\expandafter\gdef\csname Nstatus@ai.l131k.floor.tnr\endcsname{final}
\expandafter\gdef\csname N@ai.l131k.floor.tpr.b1\endcsname{0.066}
\expandafter\gdef\csname Nlo@ai.l131k.floor.tpr.b1\endcsname{0.051}
\expandafter\gdef\csname Nhi@ai.l131k.floor.tpr.b1\endcsname{0.083}
\expandafter\gdef\csname Nn@ai.l131k.floor.tpr.b1\endcsname{487}
\expandafter\gdef\csname Nstatus@ai.l131k.floor.tpr.b1\endcsname{final}
\expandafter\gdef\csname N@ai.l131k.floor.tpr.b2\endcsname{0.070}
\expandafter\gdef\csname Nlo@ai.l131k.floor.tpr.b2\endcsname{0.056}
\expandafter\gdef\csname Nhi@ai.l131k.floor.tpr.b2\endcsname{0.087}
\expandafter\gdef\csname Nn@ai.l131k.floor.tpr.b2\endcsname{500}
\expandafter\gdef\csname Nstatus@ai.l131k.floor.tpr.b2\endcsname{final}
\expandafter\gdef\csname N@ai.l131k.floor.tpr.b3\endcsname{0.070}
\expandafter\gdef\csname Nlo@ai.l131k.floor.tpr.b3\endcsname{0.053}
\expandafter\gdef\csname Nhi@ai.l131k.floor.tpr.b3\endcsname{0.088}
\expandafter\gdef\csname Nn@ai.l131k.floor.tpr.b3\endcsname{478}
\expandafter\gdef\csname Nstatus@ai.l131k.floor.tpr.b3\endcsname{final}
\expandafter\gdef\csname N@ai.l131k.floor.tpr.b4\endcsname{0.064}
\expandafter\gdef\csname Nlo@ai.l131k.floor.tpr.b4\endcsname{0.049}
\expandafter\gdef\csname Nhi@ai.l131k.floor.tpr.b4\endcsname{0.080}
\expandafter\gdef\csname Nn@ai.l131k.floor.tpr.b4\endcsname{506}
\expandafter\gdef\csname Nstatus@ai.l131k.floor.tpr.b4\endcsname{final}
\expandafter\gdef\csname N@ai.l131k.jnull.det\endcsname{0.811}
\expandafter\gdef\csname Nlo@ai.l131k.jnull.det\endcsname{0.802}
\expandafter\gdef\csname Nhi@ai.l131k.jnull.det\endcsname{0.819}
\expandafter\gdef\csname Nn@ai.l131k.jnull.det\endcsname{506}
\expandafter\gdef\csname Nstatus@ai.l131k.jnull.det\endcsname{final}
\expandafter\gdef\csname N@ai.l131k.jnull.fuzz\endcsname{0.774}
\expandafter\gdef\csname Nlo@ai.l131k.jnull.fuzz\endcsname{0.765}
\expandafter\gdef\csname Nhi@ai.l131k.jnull.fuzz\endcsname{0.783}
\expandafter\gdef\csname Nn@ai.l131k.jnull.fuzz\endcsname{506}
\expandafter\gdef\csname Nstatus@ai.l131k.jnull.fuzz\endcsname{final}
\expandafter\gdef\csname N@ai.l131k.mcos16.fuzz\endcsname{0.654}
\expandafter\gdef\csname Nlo@ai.l131k.mcos16.fuzz\endcsname{0.643}
\expandafter\gdef\csname Nhi@ai.l131k.mcos16.fuzz\endcsname{0.665}
\expandafter\gdef\csname Nn@ai.l131k.mcos16.fuzz\endcsname{449}
\expandafter\gdef\csname Nstatus@ai.l131k.mcos16.fuzz\endcsname{final}
\expandafter\gdef\csname N@ai.l131k.mjac16.fuzz\endcsname{0.655}
\expandafter\gdef\csname Nlo@ai.l131k.mjac16.fuzz\endcsname{0.644}
\expandafter\gdef\csname Nhi@ai.l131k.mjac16.fuzz\endcsname{0.665}
\expandafter\gdef\csname Nn@ai.l131k.mjac16.fuzz\endcsname{444}
\expandafter\gdef\csname Nstatus@ai.l131k.mjac16.fuzz\endcsname{final}
\expandafter\gdef\csname N@ai.l131k.mtop16.det.q1\endcsname{0.612}
\expandafter\gdef\csname Nlo@ai.l131k.mtop16.det.q1\endcsname{0.590}
\expandafter\gdef\csname Nhi@ai.l131k.mtop16.det.q1\endcsname{0.637}
\expandafter\gdef\csname Nn@ai.l131k.mtop16.det.q1\endcsname{97}
\expandafter\gdef\csname Nstatus@ai.l131k.mtop16.det.q1\endcsname{final}
\expandafter\gdef\csname N@ai.l131k.mtop16.det.q2\endcsname{0.718}
\expandafter\gdef\csname Nlo@ai.l131k.mtop16.det.q2\endcsname{0.698}
\expandafter\gdef\csname Nhi@ai.l131k.mtop16.det.q2\endcsname{0.738}
\expandafter\gdef\csname Nn@ai.l131k.mtop16.det.q2\endcsname{122}
\expandafter\gdef\csname Nstatus@ai.l131k.mtop16.det.q2\endcsname{final}
\expandafter\gdef\csname N@ai.l131k.mtop16.det.q3\endcsname{0.718}
\expandafter\gdef\csname Nlo@ai.l131k.mtop16.det.q3\endcsname{0.696}
\expandafter\gdef\csname Nhi@ai.l131k.mtop16.det.q3\endcsname{0.740}
\expandafter\gdef\csname Nn@ai.l131k.mtop16.det.q3\endcsname{120}
\expandafter\gdef\csname Nstatus@ai.l131k.mtop16.det.q3\endcsname{final}
\expandafter\gdef\csname N@ai.l131k.mtop16.det.q4\endcsname{0.659}
\expandafter\gdef\csname Nlo@ai.l131k.mtop16.det.q4\endcsname{0.636}
\expandafter\gdef\csname Nhi@ai.l131k.mtop16.det.q4\endcsname{0.680}
\expandafter\gdef\csname Nn@ai.l131k.mtop16.det.q4\endcsname{117}
\expandafter\gdef\csname Nstatus@ai.l131k.mtop16.det.q4\endcsname{final}
\expandafter\gdef\csname N@ai.l131k.mtop16.detrc\endcsname{0.660}
\expandafter\gdef\csname Nlo@ai.l131k.mtop16.detrc\endcsname{0.649}
\expandafter\gdef\csname Nhi@ai.l131k.mtop16.detrc\endcsname{0.673}
\expandafter\gdef\csname Nn@ai.l131k.mtop16.detrc\endcsname{512}
\expandafter\gdef\csname Nstatus@ai.l131k.mtop16.detrc\endcsname{final}
\expandafter\gdef\csname N@ai.l131k.mtop16.fuzzrc\endcsname{0.628}
\expandafter\gdef\csname Nlo@ai.l131k.mtop16.fuzzrc\endcsname{0.618}
\expandafter\gdef\csname Nhi@ai.l131k.mtop16.fuzzrc\endcsname{0.640}
\expandafter\gdef\csname Nn@ai.l131k.mtop16.fuzzrc\endcsname{512}
\expandafter\gdef\csname Nstatus@ai.l131k.mtop16.fuzzrc\endcsname{final}
\expandafter\gdef\csname N@ai.l131k.mtop16.tnr\endcsname{0.986}
\expandafter\gdef\csname Nlo@ai.l131k.mtop16.tnr\endcsname{0.982}
\expandafter\gdef\csname Nhi@ai.l131k.mtop16.tnr\endcsname{0.989}
\expandafter\gdef\csname Nn@ai.l131k.mtop16.tnr\endcsname{456}
\expandafter\gdef\csname Nstatus@ai.l131k.mtop16.tnr\endcsname{final}
\expandafter\gdef\csname N@ai.l131k.mtop16.tpr.b1\endcsname{0.093}
\expandafter\gdef\csname Nlo@ai.l131k.mtop16.tpr.b1\endcsname{0.077}
\expandafter\gdef\csname Nhi@ai.l131k.mtop16.tpr.b1\endcsname{0.110}
\expandafter\gdef\csname Nn@ai.l131k.mtop16.tpr.b1\endcsname{438}
\expandafter\gdef\csname Nstatus@ai.l131k.mtop16.tpr.b1\endcsname{final}
\expandafter\gdef\csname N@ai.l131k.mtop16.tpr.b2\endcsname{0.291}
\expandafter\gdef\csname Nlo@ai.l131k.mtop16.tpr.b2\endcsname{0.263}
\expandafter\gdef\csname Nhi@ai.l131k.mtop16.tpr.b2\endcsname{0.319}
\expandafter\gdef\csname Nn@ai.l131k.mtop16.tpr.b2\endcsname{452}
\expandafter\gdef\csname Nstatus@ai.l131k.mtop16.tpr.b2\endcsname{final}
\expandafter\gdef\csname N@ai.l131k.mtop16.tpr.b3\endcsname{0.501}
\expandafter\gdef\csname Nlo@ai.l131k.mtop16.tpr.b3\endcsname{0.467}
\expandafter\gdef\csname Nhi@ai.l131k.mtop16.tpr.b3\endcsname{0.535}
\expandafter\gdef\csname Nn@ai.l131k.mtop16.tpr.b3\endcsname{436}
\expandafter\gdef\csname Nstatus@ai.l131k.mtop16.tpr.b3\endcsname{final}
\expandafter\gdef\csname N@ai.l131k.mtop16.tpr.b4\endcsname{0.635}
\expandafter\gdef\csname Nlo@ai.l131k.mtop16.tpr.b4\endcsname{0.600}
\expandafter\gdef\csname Nhi@ai.l131k.mtop16.tpr.b4\endcsname{0.672}
\expandafter\gdef\csname Nn@ai.l131k.mtop16.tpr.b4\endcsname{456}
\expandafter\gdef\csname Nstatus@ai.l131k.mtop16.tpr.b4\endcsname{final}
\expandafter\gdef\csname N@ai.l131k.nla1.det\endcsname{0.675}
\expandafter\gdef\csname Nlo@ai.l131k.nla1.det\endcsname{0.661}
\expandafter\gdef\csname Nhi@ai.l131k.nla1.det\endcsname{0.688}
\expandafter\gdef\csname Nn@ai.l131k.nla1.det\endcsname{469}
\expandafter\gdef\csname Nstatus@ai.l131k.nla1.det\endcsname{final}
\expandafter\gdef\csname N@ai.l131k.nla1.fuzz\endcsname{0.648}
\expandafter\gdef\csname Nlo@ai.l131k.nla1.fuzz\endcsname{0.636}
\expandafter\gdef\csname Nhi@ai.l131k.nla1.fuzz\endcsname{0.660}
\expandafter\gdef\csname Nn@ai.l131k.nla1.fuzz\endcsname{469}
\expandafter\gdef\csname Nstatus@ai.l131k.nla1.fuzz\endcsname{final}
\expandafter\gdef\csname N@ai.l131k.nla4.detrc\endcsname{0.655}
\expandafter\gdef\csname Nlo@ai.l131k.nla4.detrc\endcsname{0.642}
\expandafter\gdef\csname Nhi@ai.l131k.nla4.detrc\endcsname{0.667}
\expandafter\gdef\csname Nn@ai.l131k.nla4.detrc\endcsname{512}
\expandafter\gdef\csname Nstatus@ai.l131k.nla4.detrc\endcsname{final}
\expandafter\gdef\csname N@ai.l131k.nla4.fuzzrc\endcsname{0.628}
\expandafter\gdef\csname Nlo@ai.l131k.nla4.fuzzrc\endcsname{0.617}
\expandafter\gdef\csname Nhi@ai.l131k.nla4.fuzzrc\endcsname{0.640}
\expandafter\gdef\csname Nn@ai.l131k.nla4.fuzzrc\endcsname{511}
\expandafter\gdef\csname Nstatus@ai.l131k.nla4.fuzzrc\endcsname{final}
\expandafter\gdef\csname N@ai.l131k.nlab.det\endcsname{0.533}
\expandafter\gdef\csname Nlo@ai.l131k.nlab.det\endcsname{0.528}
\expandafter\gdef\csname Nhi@ai.l131k.nlab.det\endcsname{0.538}
\expandafter\gdef\csname Nn@ai.l131k.nlab.det\endcsname{503}
\expandafter\gdef\csname Nstatus@ai.l131k.nlab.det\endcsname{final}
\expandafter\gdef\csname N@ai.l131k.nlab.fuzz\endcsname{0.534}
\expandafter\gdef\csname Nlo@ai.l131k.nlab.fuzz\endcsname{0.529}
\expandafter\gdef\csname Nhi@ai.l131k.nlab.fuzz\endcsname{0.540}
\expandafter\gdef\csname Nn@ai.l131k.nlab.fuzz\endcsname{483}
\expandafter\gdef\csname Nstatus@ai.l131k.nlab.fuzz\endcsname{final}
\expandafter\gdef\csname N@ai.l131k.null.det.q1\endcsname{0.710}
\expandafter\gdef\csname Nlo@ai.l131k.null.det.q1\endcsname{0.689}
\expandafter\gdef\csname Nhi@ai.l131k.null.det.q1\endcsname{0.732}
\expandafter\gdef\csname Nn@ai.l131k.null.det.q1\endcsname{99}
\expandafter\gdef\csname Nstatus@ai.l131k.null.det.q1\endcsname{final}
\expandafter\gdef\csname N@ai.l131k.null.det.q2\endcsname{0.795}
\expandafter\gdef\csname Nlo@ai.l131k.null.det.q2\endcsname{0.778}
\expandafter\gdef\csname Nhi@ai.l131k.null.det.q2\endcsname{0.811}
\expandafter\gdef\csname Nn@ai.l131k.null.det.q2\endcsname{126}
\expandafter\gdef\csname Nstatus@ai.l131k.null.det.q2\endcsname{final}
\expandafter\gdef\csname N@ai.l131k.null.det.q3\endcsname{0.847}
\expandafter\gdef\csname Nlo@ai.l131k.null.det.q3\endcsname{0.831}
\expandafter\gdef\csname Nhi@ai.l131k.null.det.q3\endcsname{0.862}
\expandafter\gdef\csname Nn@ai.l131k.null.det.q3\endcsname{127}
\expandafter\gdef\csname Nstatus@ai.l131k.null.det.q3\endcsname{final}
\expandafter\gdef\csname N@ai.l131k.null.det.q4\endcsname{0.816}
\expandafter\gdef\csname Nlo@ai.l131k.null.det.q4\endcsname{0.793}
\expandafter\gdef\csname Nhi@ai.l131k.null.det.q4\endcsname{0.837}
\expandafter\gdef\csname Nn@ai.l131k.null.det.q4\endcsname{125}
\expandafter\gdef\csname Nstatus@ai.l131k.null.det.q4\endcsname{final}
\expandafter\gdef\csname N@ai.l131k.null.fuzz\endcsname{0.762}
\expandafter\gdef\csname Nlo@ai.l131k.null.fuzz\endcsname{0.751}
\expandafter\gdef\csname Nhi@ai.l131k.null.fuzz\endcsname{0.772}
\expandafter\gdef\csname Nn@ai.l131k.null.fuzz\endcsname{477}
\expandafter\gdef\csname Nstatus@ai.l131k.null.fuzz\endcsname{final}
\expandafter\gdef\csname Nstatus@fid.ra.all.usd\endcsname{pending}
\expandafter\gdef\csname Nstatus@fid.ra.base.dcorp.bo8\endcsname{pending}
\expandafter\gdef\csname Nstatus@fid.ra.base.injcos\endcsname{pending}
\expandafter\gdef\csname Nstatus@fid.ra.base.win.bo8\endcsname{pending}
\expandafter\gdef\csname N@fid.ra.corp10.gpus\endcsname{6.3}
\expandafter\gdef\csname Nn@fid.ra.corp10.gpus\endcsname{512}
\expandafter\gdef\csname Nstatus@fid.ra.corp10.gpus\endcsname{final}
\expandafter\gdef\csname N@fid.ra.corp2p5.cos\endcsname{0.5215}
\expandafter\gdef\csname Nse@fid.ra.corp2p5.cos\endcsname{0.0050}
\expandafter\gdef\csname Nn@fid.ra.corp2p5.cos\endcsname{486}
\expandafter\gdef\csname Nstatus@fid.ra.corp2p5.cos\endcsname{final}
\expandafter\gdef\csname N@fid.ra.corp5.cos\endcsname{0.5458}
\expandafter\gdef\csname Nse@fid.ra.corp5.cos\endcsname{0.0049}
\expandafter\gdef\csname Nn@fid.ra.corp5.cos\endcsname{486}
\expandafter\gdef\csname Nstatus@fid.ra.corp5.cos\endcsname{final}
\expandafter\gdef\csname N@fid.ra.diff.cos.bo1\endcsname{0.1870}
\expandafter\gdef\csname Nse@fid.ra.diff.cos.bo1\endcsname{0.0057}
\expandafter\gdef\csname Nlo@fid.ra.diff.cos.bo1\endcsname{0.1757}
\expandafter\gdef\csname Nhi@fid.ra.diff.cos.bo1\endcsname{0.1981}
\expandafter\gdef\csname Nn@fid.ra.diff.cos.bo1\endcsname{486}
\expandafter\gdef\csname Nstatus@fid.ra.diff.cos.bo1\endcsname{final}
\expandafter\gdef\csname N@fid.ra.epo.dcorp\endcsname{-0.1663}
\expandafter\gdef\csname Nse@fid.ra.epo.dcorp\endcsname{0.0180}
\expandafter\gdef\csname Nlo@fid.ra.epo.dcorp\endcsname{-0.2017}
\expandafter\gdef\csname Nhi@fid.ra.epo.dcorp\endcsname{-0.1312}
\expandafter\gdef\csname Nn@fid.ra.epo.dcorp\endcsname{64}
\expandafter\gdef\csname Nstatus@fid.ra.epo.dcorp\endcsname{final}
\expandafter\gdef\csname N@fid.ra.epo.dex\endcsname{0.4158}
\expandafter\gdef\csname Nse@fid.ra.epo.dex\endcsname{0.0194}
\expandafter\gdef\csname Nlo@fid.ra.epo.dex\endcsname{0.3777}
\expandafter\gdef\csname Nhi@fid.ra.epo.dex\endcsname{0.4538}
\expandafter\gdef\csname Nn@fid.ra.epo.dex\endcsname{64}
\expandafter\gdef\csname Nstatus@fid.ra.epo.dex\endcsname{final}
\expandafter\gdef\csname N@fid.ra.epo.gpus\endcsname{932.0}
\expandafter\gdef\csname Nn@fid.ra.epo.gpus\endcsname{64}
\expandafter\gdef\csname Nstatus@fid.ra.epo.gpus\endcsname{final}
\expandafter\gdef\csname N@fid.ra.epo.win\endcsname{0.1250}
\expandafter\gdef\csname Nse@fid.ra.epo.win\endcsname{0.0425}
\expandafter\gdef\csname Nn@fid.ra.epo.win\endcsname{64}
\expandafter\gdef\csname Nstatus@fid.ra.epo.win\endcsname{final}
\expandafter\gdef\csname N@fid.ra.ex.cos.bo64\endcsname{0.8498}
\expandafter\gdef\csname Nse@fid.ra.ex.cos.bo64\endcsname{0.0043}
\expandafter\gdef\csname Nlo@fid.ra.ex.cos.bo64\endcsname{0.8411}
\expandafter\gdef\csname Nhi@fid.ra.ex.cos.bo64\endcsname{0.8582}
\expandafter\gdef\csname Nn@fid.ra.ex.cos.bo64\endcsname{486}
\expandafter\gdef\csname Nstatus@fid.ra.ex.cos.bo64\endcsname{final}
\expandafter\gdef\csname Nstatus@fid.ra.ex.gpus\endcsname{pending}
\expandafter\gdef\csname Nstatus@fid.ra.ex.injcos\endcsname{pending}
\expandafter\gdef\csname Nstatus@fid.ra.ex.lastfrac\endcsname{pending}
\expandafter\gdef\csname N@fid.ra.ex.ndocs\endcsname{425}
\expandafter\gdef\csname Nn@fid.ra.ex.ndocs\endcsname{486}
\expandafter\gdef\csname Nstatus@fid.ra.ex.ndocs\endcsname{final}
\expandafter\gdef\csname N@fid.ra.ex.win.bo1\endcsname{0.9177}
\expandafter\gdef\csname Nse@fid.ra.ex.win.bo1\endcsname{0.0127}
\expandafter\gdef\csname Nlo@fid.ra.ex.win.bo1\endcsname{0.8916}
\expandafter\gdef\csname Nhi@fid.ra.ex.win.bo1\endcsname{0.9413}
\expandafter\gdef\csname Nn@fid.ra.ex.win.bo1\endcsname{486}
\expandafter\gdef\csname Nstatus@fid.ra.ex.win.bo1\endcsname{final}
\expandafter\gdef\csname N@fid.ra.gcg.dcorp\endcsname{-0.1720}
\expandafter\gdef\csname Nse@fid.ra.gcg.dcorp\endcsname{0.0226}
\expandafter\gdef\csname Nlo@fid.ra.gcg.dcorp\endcsname{-0.2167}
\expandafter\gdef\csname Nhi@fid.ra.gcg.dcorp\endcsname{-0.1279}
\expandafter\gdef\csname Nn@fid.ra.gcg.dcorp\endcsname{64}
\expandafter\gdef\csname Nstatus@fid.ra.gcg.dcorp\endcsname{final}
\expandafter\gdef\csname N@fid.ra.gcg.dex\endcsname{0.4214}
\expandafter\gdef\csname Nse@fid.ra.gcg.dex\endcsname{0.0201}
\expandafter\gdef\csname Nlo@fid.ra.gcg.dex\endcsname{0.3820}
\expandafter\gdef\csname Nhi@fid.ra.gcg.dex\endcsname{0.4609}
\expandafter\gdef\csname Nn@fid.ra.gcg.dex\endcsname{64}
\expandafter\gdef\csname Nstatus@fid.ra.gcg.dex\endcsname{final}
\expandafter\gdef\csname N@fid.ra.gcg.gpus\endcsname{274.5}
\expandafter\gdef\csname Nn@fid.ra.gcg.gpus\endcsname{64}
\expandafter\gdef\csname Nstatus@fid.ra.gcg.gpus\endcsname{final}
\expandafter\gdef\csname N@fid.ra.gcg.win\endcsname{0.2031}
\expandafter\gdef\csname Nse@fid.ra.gcg.win\endcsname{0.0510}
\expandafter\gdef\csname Nn@fid.ra.gcg.win\endcsname{64}
\expandafter\gdef\csname Nstatus@fid.ra.gcg.win\endcsname{final}
\expandafter\gdef\csname Nstatus@fid.ra.mu.cos\endcsname{pending}
\expandafter\gdef\csname N@fid.ra.nla.cos.bo1\endcsname{0.5672}
\expandafter\gdef\csname Nse@fid.ra.nla.cos.bo1\endcsname{0.0059}
\expandafter\gdef\csname Nn@fid.ra.nla.cos.bo1\endcsname{486}
\expandafter\gdef\csname Nstatus@fid.ra.nla.cos.bo1\endcsname{final}
\expandafter\gdef\csname Nstatus@fid.ra.nla.dcorp.bo1\endcsname{pending}
\expandafter\gdef\csname N@fid.ra.nla.dex\endcsname{0.2547}
\expandafter\gdef\csname Nse@fid.ra.nla.dex\endcsname{0.0051}
\expandafter\gdef\csname Nlo@fid.ra.nla.dex\endcsname{0.2447}
\expandafter\gdef\csname Nhi@fid.ra.nla.dex\endcsname{0.2647}
\expandafter\gdef\csname Nn@fid.ra.nla.dex\endcsname{486}
\expandafter\gdef\csname Nstatus@fid.ra.nla.dex\endcsname{final}
\expandafter\gdef\csname Nstatus@fid.ra.nla.win.bo1\endcsname{pending}
\expandafter\gdef\csname N@fid.ra.ps.cos.l14\endcsname{0.2351}
\expandafter\gdef\csname Nse@fid.ra.ps.cos.l14\endcsname{0.0041}
\expandafter\gdef\csname Nn@fid.ra.ps.cos.l14\endcsname{486}
\expandafter\gdef\csname Nstatus@fid.ra.ps.cos.l14\endcsname{final}
\expandafter\gdef\csname N@fid.ra.ps.cos.l21\endcsname{0.2274}
\expandafter\gdef\csname Nse@fid.ra.ps.cos.l21\endcsname{0.0039}
\expandafter\gdef\csname Nn@fid.ra.ps.cos.l21\endcsname{486}
\expandafter\gdef\csname Nstatus@fid.ra.ps.cos.l21\endcsname{final}
\expandafter\gdef\csname N@fid.ra.ps.cos.l42\endcsname{0.2173}
\expandafter\gdef\csname Nse@fid.ra.ps.cos.l42\endcsname{0.0036}
\expandafter\gdef\csname Nn@fid.ra.ps.cos.l42\endcsname{486}
\expandafter\gdef\csname Nstatus@fid.ra.ps.cos.l42\endcsname{final}
\expandafter\gdef\csname N@fid.ra.ps.cos.l8\endcsname{0.2289}
\expandafter\gdef\csname Nse@fid.ra.ps.cos.l8\endcsname{0.0041}
\expandafter\gdef\csname Nn@fid.ra.ps.cos.l8\endcsname{486}
\expandafter\gdef\csname Nstatus@fid.ra.ps.cos.l8\endcsname{final}
\expandafter\gdef\csname N@fid.ra.ps.dfloor.l14\endcsname{+0.0600}
\expandafter\gdef\csname Nse@fid.ra.ps.dfloor.l14\endcsname{0.0031}
\expandafter\gdef\csname Nlo@fid.ra.ps.dfloor.l14\endcsname{0.0538}
\expandafter\gdef\csname Nhi@fid.ra.ps.dfloor.l14\endcsname{0.0663}
\expandafter\gdef\csname Nn@fid.ra.ps.dfloor.l14\endcsname{486}
\expandafter\gdef\csname Nstatus@fid.ra.ps.dfloor.l14\endcsname{final}
\expandafter\gdef\csname N@fid.ra.ps.dfloor.l21\endcsname{+0.0522}
\expandafter\gdef\csname Nse@fid.ra.ps.dfloor.l21\endcsname{0.0028}
\expandafter\gdef\csname Nlo@fid.ra.ps.dfloor.l21\endcsname{0.0467}
\expandafter\gdef\csname Nhi@fid.ra.ps.dfloor.l21\endcsname{0.0579}
\expandafter\gdef\csname Nn@fid.ra.ps.dfloor.l21\endcsname{486}
\expandafter\gdef\csname Nstatus@fid.ra.ps.dfloor.l21\endcsname{final}
\expandafter\gdef\csname N@fid.ra.ps.dfloor.l42\endcsname{+0.0421}
\expandafter\gdef\csname Nse@fid.ra.ps.dfloor.l42\endcsname{0.0025}
\expandafter\gdef\csname Nlo@fid.ra.ps.dfloor.l42\endcsname{0.0374}
\expandafter\gdef\csname Nhi@fid.ra.ps.dfloor.l42\endcsname{0.0471}
\expandafter\gdef\csname Nn@fid.ra.ps.dfloor.l42\endcsname{486}
\expandafter\gdef\csname Nstatus@fid.ra.ps.dfloor.l42\endcsname{final}
\expandafter\gdef\csname N@fid.ra.ps.dfloor.l8\endcsname{+0.0537}
\expandafter\gdef\csname Nse@fid.ra.ps.dfloor.l8\endcsname{0.0031}
\expandafter\gdef\csname Nlo@fid.ra.ps.dfloor.l8\endcsname{0.0476}
\expandafter\gdef\csname Nhi@fid.ra.ps.dfloor.l8\endcsname{0.0600}
\expandafter\gdef\csname Nn@fid.ra.ps.dfloor.l8\endcsname{486}
\expandafter\gdef\csname Nstatus@fid.ra.ps.dfloor.l8\endcsname{final}
\expandafter\gdef\csname N@fid.ra.psfloor.cos.l14\endcsname{0.1752}
\expandafter\gdef\csname Nse@fid.ra.psfloor.cos.l14\endcsname{0.0029}
\expandafter\gdef\csname Nn@fid.ra.psfloor.cos.l14\endcsname{486}
\expandafter\gdef\csname Nstatus@fid.ra.psfloor.cos.l14\endcsname{final}
\expandafter\gdef\csname N@fid.ra.psfloor.cos.l21\endcsname{0.1752}
\expandafter\gdef\csname Nse@fid.ra.psfloor.cos.l21\endcsname{0.0029}
\expandafter\gdef\csname Nn@fid.ra.psfloor.cos.l21\endcsname{486}
\expandafter\gdef\csname Nstatus@fid.ra.psfloor.cos.l21\endcsname{final}
\expandafter\gdef\csname N@fid.ra.psfloor.cos.l42\endcsname{0.1752}
\expandafter\gdef\csname Nse@fid.ra.psfloor.cos.l42\endcsname{0.0029}
\expandafter\gdef\csname Nn@fid.ra.psfloor.cos.l42\endcsname{486}
\expandafter\gdef\csname Nstatus@fid.ra.psfloor.cos.l42\endcsname{final}
\expandafter\gdef\csname N@fid.ra.psfloor.cos.l8\endcsname{0.1752}
\expandafter\gdef\csname Nse@fid.ra.psfloor.cos.l8\endcsname{0.0029}
\expandafter\gdef\csname Nn@fid.ra.psfloor.cos.l8\endcsname{486}
\expandafter\gdef\csname Nstatus@fid.ra.psfloor.cos.l8\endcsname{final}
\expandafter\gdef\csname N@fid.ra64.ex.cos.bo1\endcsname{0.7459}
\expandafter\gdef\csname Nse@fid.ra64.ex.cos.bo1\endcsname{0.0191}
\expandafter\gdef\csname Nlo@fid.ra64.ex.cos.bo1\endcsname{0.7084}
\expandafter\gdef\csname Nhi@fid.ra64.ex.cos.bo1\endcsname{0.7834}
\expandafter\gdef\csname Nn@fid.ra64.ex.cos.bo1\endcsname{64}
\expandafter\gdef\csname Nstatus@fid.ra64.ex.cos.bo1\endcsname{final}
\expandafter\gdef\csname N@fid.ra64.ex.cos.bo8\endcsname{0.8119}
\expandafter\gdef\csname Nse@fid.ra64.ex.cos.bo8\endcsname{0.0165}
\expandafter\gdef\csname Nlo@fid.ra64.ex.cos.bo8\endcsname{0.7795}
\expandafter\gdef\csname Nhi@fid.ra64.ex.cos.bo8\endcsname{0.8443}
\expandafter\gdef\csname Nn@fid.ra64.ex.cos.bo8\endcsname{64}
\expandafter\gdef\csname Nstatus@fid.ra64.ex.cos.bo8\endcsname{final}
\expandafter\gdef\csname N@fid.ra64.ex.dcorp.bo8\endcsname{0.2494}
\expandafter\gdef\csname Nse@fid.ra64.ex.dcorp.bo8\endcsname{0.0150}
\expandafter\gdef\csname Nlo@fid.ra64.ex.dcorp.bo8\endcsname{0.2204}
\expandafter\gdef\csname Nhi@fid.ra64.ex.dcorp.bo8\endcsname{0.2771}
\expandafter\gdef\csname Nn@fid.ra64.ex.dcorp.bo8\endcsname{64}
\expandafter\gdef\csname Nstatus@fid.ra64.ex.dcorp.bo8\endcsname{final}
\expandafter\gdef\csname N@fid.ra64.ex.win.bo8\endcsname{0.9688}
\expandafter\gdef\csname Nse@fid.ra64.ex.win.bo8\endcsname{0.0218}
\expandafter\gdef\csname Nn@fid.ra64.ex.win.bo8\endcsname{64}
\expandafter\gdef\csname Nstatus@fid.ra64.ex.win.bo8\endcsname{final}
\expandafter\gdef\csname Nstatus@fid.rnd.corp10.cos\endcsname{pending}
\expandafter\gdef\csname Nstatus@ood.all.ex.usd\endcsname{pending}
\expandafter\gdef\csname N@ood.arb.base.bpb\endcsname{0.6400}
\expandafter\gdef\csname Nn@ood.arb.base.bpb\endcsname{512}
\expandafter\gdef\csname Nstatus@ood.arb.base.bpb\endcsname{final}
\expandafter\gdef\csname N@ood.arb.base.cos.bo8\endcsname{0.1927}
\expandafter\gdef\csname Nn@ood.arb.base.cos.bo8\endcsname{512}
\expandafter\gdef\csname Nstatus@ood.arb.base.cos.bo8\endcsname{final}
\expandafter\gdef\csname N@ood.arb.corp10.cos\endcsname{0.7096}
\expandafter\gdef\csname Nn@ood.arb.corp10.cos\endcsname{512}
\expandafter\gdef\csname Nstatus@ood.arb.corp10.cos\endcsname{final}
\expandafter\gdef\csname N@ood.arb.corp10.lid\endcsname{0.837}
\expandafter\gdef\csname Nn@ood.arb.corp10.lid\endcsname{512}
\expandafter\gdef\csname Nstatus@ood.arb.corp10.lid\endcsname{final}
\expandafter\gdef\csname N@ood.arb.corp10.mtok\endcsname{10}
\expandafter\gdef\csname Nstatus@ood.arb.corp10.mtok\endcsname{final}
\expandafter\gdef\csname N@ood.arb.diff.cos.bo8\endcsname{0.0720}
\expandafter\gdef\csname Nse@ood.arb.diff.cos.bo8\endcsname{0.0049}
\expandafter\gdef\csname Nlo@ood.arb.diff.cos.bo8\endcsname{0.0623}
\expandafter\gdef\csname Nhi@ood.arb.diff.cos.bo8\endcsname{0.0816}
\expandafter\gdef\csname Nn@ood.arb.diff.cos.bo8\endcsname{512}
\expandafter\gdef\csname Nstatus@ood.arb.diff.cos.bo8\endcsname{final}
\expandafter\gdef\csname N@ood.arb.diff.verdict\endcsname{exceeds}
\expandafter\gdef\csname Nn@ood.arb.diff.verdict\endcsname{512}
\expandafter\gdef\csname Nstatus@ood.arb.diff.verdict\endcsname{final}
\expandafter\gdef\csname Nstatus@ood.arb.ex.cos.bo1\endcsname{pending}
\expandafter\gdef\csname N@ood.arb.ex.cos.bo8\endcsname{0.7816}
\expandafter\gdef\csname Nn@ood.arb.ex.cos.bo8\endcsname{512}
\expandafter\gdef\csname Nstatus@ood.arb.ex.cos.bo8\endcsname{final}
\expandafter\gdef\csname N@ood.arb.ex.lid\endcsname{0.707}
\expandafter\gdef\csname Nn@ood.arb.ex.lid\endcsname{512}
\expandafter\gdef\csname Nstatus@ood.arb.ex.lid\endcsname{final}
\expandafter\gdef\csname N@ood.arxiv.base.bpb\endcsname{0.6590}
\expandafter\gdef\csname Nn@ood.arxiv.base.bpb\endcsname{512}
\expandafter\gdef\csname Nstatus@ood.arxiv.base.bpb\endcsname{final}
\expandafter\gdef\csname N@ood.arxiv.base.cos.bo8\endcsname{0.2038}
\expandafter\gdef\csname Nn@ood.arxiv.base.cos.bo8\endcsname{512}
\expandafter\gdef\csname Nstatus@ood.arxiv.base.cos.bo8\endcsname{final}
\expandafter\gdef\csname N@ood.arxiv.corp10.cos\endcsname{0.6533}
\expandafter\gdef\csname Nn@ood.arxiv.corp10.cos\endcsname{512}
\expandafter\gdef\csname Nstatus@ood.arxiv.corp10.cos\endcsname{final}
\expandafter\gdef\csname N@ood.arxiv.corp10.lid\endcsname{0.013}
\expandafter\gdef\csname Nn@ood.arxiv.corp10.lid\endcsname{512}
\expandafter\gdef\csname Nstatus@ood.arxiv.corp10.lid\endcsname{final}
\expandafter\gdef\csname N@ood.arxiv.corp10.mtok\endcsname{10}
\expandafter\gdef\csname Nstatus@ood.arxiv.corp10.mtok\endcsname{final}
\expandafter\gdef\csname N@ood.arxiv.diff.cos.bo8\endcsname{0.1358}
\expandafter\gdef\csname Nse@ood.arxiv.diff.cos.bo8\endcsname{0.0062}
\expandafter\gdef\csname Nlo@ood.arxiv.diff.cos.bo8\endcsname{0.1239}
\expandafter\gdef\csname Nhi@ood.arxiv.diff.cos.bo8\endcsname{0.1476}
\expandafter\gdef\csname Nn@ood.arxiv.diff.cos.bo8\endcsname{512}
\expandafter\gdef\csname Nstatus@ood.arxiv.diff.cos.bo8\endcsname{final}
\expandafter\gdef\csname N@ood.arxiv.diff.verdict\endcsname{exceeds}
\expandafter\gdef\csname Nn@ood.arxiv.diff.verdict\endcsname{512}
\expandafter\gdef\csname Nstatus@ood.arxiv.diff.verdict\endcsname{final}
\expandafter\gdef\csname Nstatus@ood.arxiv.ex.cos.bo1\endcsname{pending}
\expandafter\gdef\csname N@ood.arxiv.ex.cos.bo8\endcsname{0.7891}
\expandafter\gdef\csname Nn@ood.arxiv.ex.cos.bo8\endcsname{512}
\expandafter\gdef\csname Nstatus@ood.arxiv.ex.cos.bo8\endcsname{final}
\expandafter\gdef\csname N@ood.arxiv.ex.lid\endcsname{n/m}
\expandafter\gdef\csname Nn@ood.arxiv.ex.lid\endcsname{512}
\expandafter\gdef\csname Nstatus@ood.arxiv.ex.lid\endcsname{final}
\expandafter\gdef\csname N@ood.c.base.bpb\endcsname{0.3287}
\expandafter\gdef\csname Nn@ood.c.base.bpb\endcsname{512}
\expandafter\gdef\csname Nstatus@ood.c.base.bpb\endcsname{final}
\expandafter\gdef\csname N@ood.c.base.cos.bo8\endcsname{0.2178}
\expandafter\gdef\csname Nn@ood.c.base.cos.bo8\endcsname{512}
\expandafter\gdef\csname Nstatus@ood.c.base.cos.bo8\endcsname{final}
\expandafter\gdef\csname N@ood.c.corp10.cos\endcsname{0.7109}
\expandafter\gdef\csname Nn@ood.c.corp10.cos\endcsname{512}
\expandafter\gdef\csname Nstatus@ood.c.corp10.cos\endcsname{final}
\expandafter\gdef\csname N@ood.c.corp10.lid\endcsname{0.132}
\expandafter\gdef\csname Nn@ood.c.corp10.lid\endcsname{512}
\expandafter\gdef\csname Nstatus@ood.c.corp10.lid\endcsname{final}
\expandafter\gdef\csname N@ood.c.corp10.mtok\endcsname{10}
\expandafter\gdef\csname Nstatus@ood.c.corp10.mtok\endcsname{final}
\expandafter\gdef\csname N@ood.c.diff.cos.bo8\endcsname{0.0262}
\expandafter\gdef\csname Nse@ood.c.diff.cos.bo8\endcsname{0.0062}
\expandafter\gdef\csname Nlo@ood.c.diff.cos.bo8\endcsname{0.0134}
\expandafter\gdef\csname Nhi@ood.c.diff.cos.bo8\endcsname{0.0388}
\expandafter\gdef\csname Nn@ood.c.diff.cos.bo8\endcsname{512}
\expandafter\gdef\csname Nstatus@ood.c.diff.cos.bo8\endcsname{final}
\expandafter\gdef\csname N@ood.c.diff.verdict\endcsname{exceeds}
\expandafter\gdef\csname Nn@ood.c.diff.verdict\endcsname{512}
\expandafter\gdef\csname Nstatus@ood.c.diff.verdict\endcsname{final}
\expandafter\gdef\csname Nstatus@ood.c.ex.cos.bo1\endcsname{pending}
\expandafter\gdef\csname N@ood.c.ex.cos.bo8\endcsname{0.7371}
\expandafter\gdef\csname Nn@ood.c.ex.cos.bo8\endcsname{512}
\expandafter\gdef\csname Nstatus@ood.c.ex.cos.bo8\endcsname{final}
\expandafter\gdef\csname N@ood.c.ex.lid\endcsname{n/m}
\expandafter\gdef\csname Nn@ood.c.ex.lid\endcsname{512}
\expandafter\gdef\csname Nstatus@ood.c.ex.lid\endcsname{final}
\expandafter\gdef\csname N@ood.ces.base.bpb\endcsname{0.9541}
\expandafter\gdef\csname Nn@ood.ces.base.bpb\endcsname{512}
\expandafter\gdef\csname Nstatus@ood.ces.base.bpb\endcsname{final}
\expandafter\gdef\csname N@ood.ces.base.cos.bo8\endcsname{0.1788}
\expandafter\gdef\csname Nn@ood.ces.base.cos.bo8\endcsname{512}
\expandafter\gdef\csname Nstatus@ood.ces.base.cos.bo8\endcsname{final}
\expandafter\gdef\csname N@ood.ces.corp10.cos\endcsname{0.6884}
\expandafter\gdef\csname Nn@ood.ces.corp10.cos\endcsname{512}
\expandafter\gdef\csname Nstatus@ood.ces.corp10.cos\endcsname{final}
\expandafter\gdef\csname N@ood.ces.corp10.lid\endcsname{0.847}
\expandafter\gdef\csname Nn@ood.ces.corp10.lid\endcsname{512}
\expandafter\gdef\csname Nstatus@ood.ces.corp10.lid\endcsname{final}
\expandafter\gdef\csname N@ood.ces.corp10.mtok\endcsname{10}
\expandafter\gdef\csname Nstatus@ood.ces.corp10.mtok\endcsname{final}
\expandafter\gdef\csname N@ood.ces.diff.cos.bo8\endcsname{0.0847}
\expandafter\gdef\csname Nse@ood.ces.diff.cos.bo8\endcsname{0.0053}
\expandafter\gdef\csname Nlo@ood.ces.diff.cos.bo8\endcsname{0.0740}
\expandafter\gdef\csname Nhi@ood.ces.diff.cos.bo8\endcsname{0.0950}
\expandafter\gdef\csname Nn@ood.ces.diff.cos.bo8\endcsname{512}
\expandafter\gdef\csname Nstatus@ood.ces.diff.cos.bo8\endcsname{final}
\expandafter\gdef\csname N@ood.ces.diff.verdict\endcsname{exceeds}
\expandafter\gdef\csname Nn@ood.ces.diff.verdict\endcsname{512}
\expandafter\gdef\csname Nstatus@ood.ces.diff.verdict\endcsname{final}
\expandafter\gdef\csname Nstatus@ood.ces.ex.cos.bo1\endcsname{pending}
\expandafter\gdef\csname N@ood.ces.ex.cos.bo8\endcsname{0.7731}
\expandafter\gdef\csname Nn@ood.ces.ex.cos.bo8\endcsname{512}
\expandafter\gdef\csname Nstatus@ood.ces.ex.cos.bo8\endcsname{final}
\expandafter\gdef\csname N@ood.ces.ex.lid\endcsname{0.600}
\expandafter\gdef\csname Nn@ood.ces.ex.lid\endcsname{512}
\expandafter\gdef\csname Nstatus@ood.ces.ex.lid\endcsname{final}
\expandafter\gdef\csname N@ood.cmn.base.bpb\endcsname{1.1189}
\expandafter\gdef\csname Nn@ood.cmn.base.bpb\endcsname{512}
\expandafter\gdef\csname Nstatus@ood.cmn.base.bpb\endcsname{final}
\expandafter\gdef\csname N@ood.cmn.base.cos.bo8\endcsname{0.1970}
\expandafter\gdef\csname Nn@ood.cmn.base.cos.bo8\endcsname{512}
\expandafter\gdef\csname Nstatus@ood.cmn.base.cos.bo8\endcsname{final}
\expandafter\gdef\csname N@ood.cmn.corp10.cos\endcsname{0.6595}
\expandafter\gdef\csname Nn@ood.cmn.corp10.cos\endcsname{512}
\expandafter\gdef\csname Nstatus@ood.cmn.corp10.cos\endcsname{final}
\expandafter\gdef\csname N@ood.cmn.corp10.lid\endcsname{0.881}
\expandafter\gdef\csname Nn@ood.cmn.corp10.lid\endcsname{512}
\expandafter\gdef\csname Nstatus@ood.cmn.corp10.lid\endcsname{final}
\expandafter\gdef\csname N@ood.cmn.corp10.mtok\endcsname{10}
\expandafter\gdef\csname Nstatus@ood.cmn.corp10.mtok\endcsname{final}
\expandafter\gdef\csname N@ood.cmn.diff.verdict\endcsname{exceeds}
\expandafter\gdef\csname Nn@ood.cmn.diff.verdict\endcsname{512}
\expandafter\gdef\csname Nstatus@ood.cmn.diff.verdict\endcsname{final}
\expandafter\gdef\csname Nstatus@ood.cmn.ex.cos.bo1\endcsname{pending}
\expandafter\gdef\csname N@ood.cmn.ex.cos.bo8\endcsname{0.8071}
\expandafter\gdef\csname Nn@ood.cmn.ex.cos.bo8\endcsname{512}
\expandafter\gdef\csname Nstatus@ood.cmn.ex.cos.bo8\endcsname{final}
\expandafter\gdef\csname N@ood.cmn.ex.lid\endcsname{0.754}
\expandafter\gdef\csname Nn@ood.cmn.ex.lid\endcsname{512}
\expandafter\gdef\csname Nstatus@ood.cmn.ex.lid\endcsname{final}
\expandafter\gdef\csname N@ood.ell.base.bpb\endcsname{0.5237}
\expandafter\gdef\csname Nn@ood.ell.base.bpb\endcsname{512}
\expandafter\gdef\csname Nstatus@ood.ell.base.bpb\endcsname{final}
\expandafter\gdef\csname N@ood.ell.base.cos.bo8\endcsname{0.1854}
\expandafter\gdef\csname Nn@ood.ell.base.cos.bo8\endcsname{512}
\expandafter\gdef\csname Nstatus@ood.ell.base.cos.bo8\endcsname{final}
\expandafter\gdef\csname N@ood.ell.corp10.cos\endcsname{0.7036}
\expandafter\gdef\csname Nn@ood.ell.corp10.cos\endcsname{512}
\expandafter\gdef\csname Nstatus@ood.ell.corp10.cos\endcsname{final}
\expandafter\gdef\csname N@ood.ell.corp10.lid\endcsname{0.843}
\expandafter\gdef\csname Nn@ood.ell.corp10.lid\endcsname{512}
\expandafter\gdef\csname Nstatus@ood.ell.corp10.lid\endcsname{final}
\expandafter\gdef\csname N@ood.ell.corp10.mtok\endcsname{10}
\expandafter\gdef\csname Nstatus@ood.ell.corp10.mtok\endcsname{final}
\expandafter\gdef\csname N@ood.ell.diff.cos.bo8\endcsname{0.0112}
\expandafter\gdef\csname Nse@ood.ell.diff.cos.bo8\endcsname{0.0060}
\expandafter\gdef\csname Nlo@ood.ell.diff.cos.bo8\endcsname{-0.0002}
\expandafter\gdef\csname Nhi@ood.ell.diff.cos.bo8\endcsname{0.0230}
\expandafter\gdef\csname Nn@ood.ell.diff.cos.bo8\endcsname{512}
\expandafter\gdef\csname Nstatus@ood.ell.diff.cos.bo8\endcsname{final}
\expandafter\gdef\csname N@ood.ell.diff.verdict\endcsname{inconclusive}
\expandafter\gdef\csname Nn@ood.ell.diff.verdict\endcsname{512}
\expandafter\gdef\csname Nstatus@ood.ell.diff.verdict\endcsname{final}
\expandafter\gdef\csname Nstatus@ood.ell.ex.cos.bo1\endcsname{pending}
\expandafter\gdef\csname N@ood.ell.ex.cos.bo8\endcsname{0.7148}
\expandafter\gdef\csname Nn@ood.ell.ex.cos.bo8\endcsname{512}
\expandafter\gdef\csname Nstatus@ood.ell.ex.cos.bo8\endcsname{final}
\expandafter\gdef\csname N@ood.ell.ex.lid\endcsname{0.355}
\expandafter\gdef\csname Nn@ood.ell.ex.lid\endcsname{512}
\expandafter\gdef\csname Nstatus@ood.ell.ex.lid\endcsname{final}
\expandafter\gdef\csname N@ood.en.base.bpb\endcsname{0.6887}
\expandafter\gdef\csname Nn@ood.en.base.bpb\endcsname{512}
\expandafter\gdef\csname Nstatus@ood.en.base.bpb\endcsname{final}
\expandafter\gdef\csname N@ood.en.corp10.lid\endcsname{0.859}
\expandafter\gdef\csname Nn@ood.en.corp10.lid\endcsname{512}
\expandafter\gdef\csname Nstatus@ood.en.corp10.lid\endcsname{final}
\expandafter\gdef\csname N@ood.en.corp10.mtok\endcsname{10}
\expandafter\gdef\csname Nstatus@ood.en.corp10.mtok\endcsname{final}
\expandafter\gdef\csname N@ood.en.diff.verdict\endcsname{exceeds}
\expandafter\gdef\csname Nn@ood.en.diff.verdict\endcsname{512}
\expandafter\gdef\csname Nstatus@ood.en.diff.verdict\endcsname{final}
\expandafter\gdef\csname Nstatus@ood.en.ex.cos.bo1\endcsname{pending}
\expandafter\gdef\csname N@ood.en.ex.cos.bo8\endcsname{0.8277}
\expandafter\gdef\csname Nn@ood.en.ex.cos.bo8\endcsname{512}
\expandafter\gdef\csname Nstatus@ood.en.ex.cos.bo8\endcsname{final}
\expandafter\gdef\csname N@ood.en.ex.lid\endcsname{0.990}
\expandafter\gdef\csname Nn@ood.en.ex.lid\endcsname{512}
\expandafter\gdef\csname Nstatus@ood.en.ex.lid\endcsname{final}
\expandafter\gdef\csname Nstatus@ood.formulas.base.bpb\endcsname{pending}
\expandafter\gdef\csname Nstatus@ood.formulas.base.cos.bo8\endcsname{pending}
\expandafter\gdef\csname Nstatus@ood.formulas.corp10.cos\endcsname{pending}
\expandafter\gdef\csname Nstatus@ood.formulas.corp10.lid\endcsname{pending}
\expandafter\gdef\csname Nstatus@ood.formulas.corp10.mtok\endcsname{pending}
\expandafter\gdef\csname Nstatus@ood.formulas.diff.cos.bo8\endcsname{pending}
\expandafter\gdef\csname Nstatus@ood.formulas.diff.verdict\endcsname{pending}
\expandafter\gdef\csname Nstatus@ood.formulas.ex.cos.bo1\endcsname{pending}
\expandafter\gdef\csname Nstatus@ood.formulas.ex.cos.bo8\endcsname{pending}
\expandafter\gdef\csname Nstatus@ood.formulas.ex.lid\endcsname{pending}
\expandafter\gdef\csname N@ood.go.base.bpb\endcsname{0.3033}
\expandafter\gdef\csname Nn@ood.go.base.bpb\endcsname{512}
\expandafter\gdef\csname Nstatus@ood.go.base.bpb\endcsname{final}
\expandafter\gdef\csname N@ood.go.base.cos.bo8\endcsname{0.2164}
\expandafter\gdef\csname Nn@ood.go.base.cos.bo8\endcsname{512}
\expandafter\gdef\csname Nstatus@ood.go.base.cos.bo8\endcsname{final}
\expandafter\gdef\csname N@ood.go.corp10.cos\endcsname{0.7137}
\expandafter\gdef\csname Nn@ood.go.corp10.cos\endcsname{512}
\expandafter\gdef\csname Nstatus@ood.go.corp10.cos\endcsname{final}
\expandafter\gdef\csname N@ood.go.corp10.lid\endcsname{0.009}
\expandafter\gdef\csname Nn@ood.go.corp10.lid\endcsname{512}
\expandafter\gdef\csname Nstatus@ood.go.corp10.lid\endcsname{final}
\expandafter\gdef\csname N@ood.go.corp10.mtok\endcsname{10}
\expandafter\gdef\csname Nstatus@ood.go.corp10.mtok\endcsname{final}
\expandafter\gdef\csname N@ood.go.diff.cos.bo8\endcsname{-0.0172}
\expandafter\gdef\csname Nse@ood.go.diff.cos.bo8\endcsname{0.0066}
\expandafter\gdef\csname Nlo@ood.go.diff.cos.bo8\endcsname{-0.0299}
\expandafter\gdef\csname Nhi@ood.go.diff.cos.bo8\endcsname{-0.0046}
\expandafter\gdef\csname Nn@ood.go.diff.cos.bo8\endcsname{512}
\expandafter\gdef\csname Nstatus@ood.go.diff.cos.bo8\endcsname{final}
\expandafter\gdef\csname N@ood.go.diff.verdict\endcsname{reversed}
\expandafter\gdef\csname Nn@ood.go.diff.verdict\endcsname{512}
\expandafter\gdef\csname Nstatus@ood.go.diff.verdict\endcsname{final}
\expandafter\gdef\csname Nstatus@ood.go.ex.cos.bo1\endcsname{pending}
\expandafter\gdef\csname N@ood.go.ex.cos.bo8\endcsname{0.6965}
\expandafter\gdef\csname Nn@ood.go.ex.cos.bo8\endcsname{512}
\expandafter\gdef\csname Nstatus@ood.go.ex.cos.bo8\endcsname{final}
\expandafter\gdef\csname N@ood.go.ex.lid\endcsname{n/m}
\expandafter\gdef\csname Nn@ood.go.ex.lid\endcsname{512}
\expandafter\gdef\csname Nstatus@ood.go.ex.lid\endcsname{final}
\expandafter\gdef\csname N@ood.haskell.base.bpb\endcsname{0.4769}
\expandafter\gdef\csname Nn@ood.haskell.base.bpb\endcsname{512}
\expandafter\gdef\csname Nstatus@ood.haskell.base.bpb\endcsname{final}
\expandafter\gdef\csname N@ood.haskell.base.cos.bo8\endcsname{0.2018}
\expandafter\gdef\csname Nn@ood.haskell.base.cos.bo8\endcsname{512}
\expandafter\gdef\csname Nstatus@ood.haskell.base.cos.bo8\endcsname{final}
\expandafter\gdef\csname N@ood.haskell.corp10.cos\endcsname{0.7209}
\expandafter\gdef\csname Nn@ood.haskell.corp10.cos\endcsname{512}
\expandafter\gdef\csname Nstatus@ood.haskell.corp10.cos\endcsname{final}
\expandafter\gdef\csname N@ood.haskell.corp10.lid\endcsname{0.079}
\expandafter\gdef\csname Nn@ood.haskell.corp10.lid\endcsname{512}
\expandafter\gdef\csname Nstatus@ood.haskell.corp10.lid\endcsname{final}
\expandafter\gdef\csname N@ood.haskell.corp10.mtok\endcsname{10}
\expandafter\gdef\csname Nstatus@ood.haskell.corp10.mtok\endcsname{final}
\expandafter\gdef\csname N@ood.haskell.diff.cos.bo8\endcsname{-0.0132}
\expandafter\gdef\csname Nse@ood.haskell.diff.cos.bo8\endcsname{0.0053}
\expandafter\gdef\csname Nlo@ood.haskell.diff.cos.bo8\endcsname{-0.0237}
\expandafter\gdef\csname Nhi@ood.haskell.diff.cos.bo8\endcsname{-0.0023}
\expandafter\gdef\csname Nn@ood.haskell.diff.cos.bo8\endcsname{512}
\expandafter\gdef\csname Nstatus@ood.haskell.diff.cos.bo8\endcsname{final}
\expandafter\gdef\csname N@ood.haskell.diff.verdict\endcsname{reversed}
\expandafter\gdef\csname Nn@ood.haskell.diff.verdict\endcsname{512}
\expandafter\gdef\csname Nstatus@ood.haskell.diff.verdict\endcsname{final}
\expandafter\gdef\csname Nstatus@ood.haskell.ex.cos.bo1\endcsname{pending}
\expandafter\gdef\csname N@ood.haskell.ex.cos.bo8\endcsname{0.7076}
\expandafter\gdef\csname Nn@ood.haskell.ex.cos.bo8\endcsname{512}
\expandafter\gdef\csname Nstatus@ood.haskell.ex.cos.bo8\endcsname{final}
\expandafter\gdef\csname N@ood.haskell.ex.lid\endcsname{n/m}
\expandafter\gdef\csname Nn@ood.haskell.ex.lid\endcsname{512}
\expandafter\gdef\csname Nstatus@ood.haskell.ex.lid\endcsname{final}
\expandafter\gdef\csname N@ood.hin.base.bpb\endcsname{1.2128}
\expandafter\gdef\csname Nn@ood.hin.base.bpb\endcsname{512}
\expandafter\gdef\csname Nstatus@ood.hin.base.bpb\endcsname{final}
\expandafter\gdef\csname N@ood.hin.base.cos.bo8\endcsname{0.2097}
\expandafter\gdef\csname Nn@ood.hin.base.cos.bo8\endcsname{512}
\expandafter\gdef\csname Nstatus@ood.hin.base.cos.bo8\endcsname{final}
\expandafter\gdef\csname N@ood.hin.corp10.cos\endcsname{0.7850}
\expandafter\gdef\csname Nn@ood.hin.corp10.cos\endcsname{512}
\expandafter\gdef\csname Nstatus@ood.hin.corp10.cos\endcsname{final}
\expandafter\gdef\csname N@ood.hin.corp10.lid\endcsname{0.629}
\expandafter\gdef\csname Nn@ood.hin.corp10.lid\endcsname{512}
\expandafter\gdef\csname Nstatus@ood.hin.corp10.lid\endcsname{final}
\expandafter\gdef\csname N@ood.hin.corp10.mtok\endcsname{10}
\expandafter\gdef\csname Nstatus@ood.hin.corp10.mtok\endcsname{final}
\expandafter\gdef\csname N@ood.hin.diff.cos.bo8\endcsname{-0.0327}
\expandafter\gdef\csname Nse@ood.hin.diff.cos.bo8\endcsname{0.0042}
\expandafter\gdef\csname Nlo@ood.hin.diff.cos.bo8\endcsname{-0.0410}
\expandafter\gdef\csname Nhi@ood.hin.diff.cos.bo8\endcsname{-0.0246}
\expandafter\gdef\csname Nn@ood.hin.diff.cos.bo8\endcsname{512}
\expandafter\gdef\csname Nstatus@ood.hin.diff.cos.bo8\endcsname{final}
\expandafter\gdef\csname N@ood.hin.diff.verdict\endcsname{reversed}
\expandafter\gdef\csname Nn@ood.hin.diff.verdict\endcsname{512}
\expandafter\gdef\csname Nstatus@ood.hin.diff.verdict\endcsname{final}
\expandafter\gdef\csname Nstatus@ood.hin.ex.cos.bo1\endcsname{pending}
\expandafter\gdef\csname N@ood.hin.ex.cos.bo8\endcsname{0.7523}
\expandafter\gdef\csname Nn@ood.hin.ex.cos.bo8\endcsname{512}
\expandafter\gdef\csname Nstatus@ood.hin.ex.cos.bo8\endcsname{final}
\expandafter\gdef\csname N@ood.hin.ex.lid\endcsname{0.703}
\expandafter\gdef\csname Nn@ood.hin.ex.lid\endcsname{512}
\expandafter\gdef\csname Nstatus@ood.hin.ex.lid\endcsname{final}
\expandafter\gdef\csname N@ood.isabelle.base.bpb\endcsname{0.7448}
\expandafter\gdef\csname Nn@ood.isabelle.base.bpb\endcsname{512}
\expandafter\gdef\csname Nstatus@ood.isabelle.base.bpb\endcsname{final}
\expandafter\gdef\csname N@ood.isabelle.base.cos.bo8\endcsname{0.2218}
\expandafter\gdef\csname Nn@ood.isabelle.base.cos.bo8\endcsname{512}
\expandafter\gdef\csname Nstatus@ood.isabelle.base.cos.bo8\endcsname{final}
\expandafter\gdef\csname N@ood.isabelle.corp10.cos\endcsname{0.7710}
\expandafter\gdef\csname Nn@ood.isabelle.corp10.cos\endcsname{512}
\expandafter\gdef\csname Nstatus@ood.isabelle.corp10.cos\endcsname{final}
\expandafter\gdef\csname N@ood.isabelle.corp10.lid\endcsname{0.025}
\expandafter\gdef\csname Nn@ood.isabelle.corp10.lid\endcsname{512}
\expandafter\gdef\csname Nstatus@ood.isabelle.corp10.lid\endcsname{final}
\expandafter\gdef\csname N@ood.isabelle.corp10.mtok\endcsname{10}
\expandafter\gdef\csname Nstatus@ood.isabelle.corp10.mtok\endcsname{final}
\expandafter\gdef\csname N@ood.isabelle.diff.verdict\endcsname{reversed}
\expandafter\gdef\csname Nn@ood.isabelle.diff.verdict\endcsname{512}
\expandafter\gdef\csname Nstatus@ood.isabelle.diff.verdict\endcsname{final}
\expandafter\gdef\csname Nstatus@ood.isabelle.ex.cos.bo1\endcsname{pending}
\expandafter\gdef\csname N@ood.isabelle.ex.cos.bo8\endcsname{0.6821}
\expandafter\gdef\csname Nn@ood.isabelle.ex.cos.bo8\endcsname{512}
\expandafter\gdef\csname Nstatus@ood.isabelle.ex.cos.bo8\endcsname{final}
\expandafter\gdef\csname N@ood.isabelle.ex.lid\endcsname{n/m}
\expandafter\gdef\csname Nn@ood.isabelle.ex.lid\endcsname{512}
\expandafter\gdef\csname Nstatus@ood.isabelle.ex.lid\endcsname{final}
\expandafter\gdef\csname N@ood.javascript.base.bpb\endcsname{0.4221}
\expandafter\gdef\csname Nn@ood.javascript.base.bpb\endcsname{512}
\expandafter\gdef\csname Nstatus@ood.javascript.base.bpb\endcsname{final}
\expandafter\gdef\csname N@ood.javascript.base.cos.bo8\endcsname{0.2233}
\expandafter\gdef\csname Nn@ood.javascript.base.cos.bo8\endcsname{512}
\expandafter\gdef\csname Nstatus@ood.javascript.base.cos.bo8\endcsname{final}
\expandafter\gdef\csname N@ood.javascript.corp10.cos\endcsname{0.6694}
\expandafter\gdef\csname Nn@ood.javascript.corp10.cos\endcsname{512}
\expandafter\gdef\csname Nstatus@ood.javascript.corp10.cos\endcsname{final}
\expandafter\gdef\csname N@ood.javascript.corp10.lid\endcsname{0.111}
\expandafter\gdef\csname Nn@ood.javascript.corp10.lid\endcsname{512}
\expandafter\gdef\csname Nstatus@ood.javascript.corp10.lid\endcsname{final}
\expandafter\gdef\csname N@ood.javascript.corp10.mtok\endcsname{10}
\expandafter\gdef\csname Nstatus@ood.javascript.corp10.mtok\endcsname{final}
\expandafter\gdef\csname N@ood.javascript.diff.cos.bo8\endcsname{0.0299}
\expandafter\gdef\csname Nse@ood.javascript.diff.cos.bo8\endcsname{0.0074}
\expandafter\gdef\csname Nlo@ood.javascript.diff.cos.bo8\endcsname{0.0157}
\expandafter\gdef\csname Nhi@ood.javascript.diff.cos.bo8\endcsname{0.0440}
\expandafter\gdef\csname Nn@ood.javascript.diff.cos.bo8\endcsname{512}
\expandafter\gdef\csname Nstatus@ood.javascript.diff.cos.bo8\endcsname{final}
\expandafter\gdef\csname N@ood.javascript.diff.verdict\endcsname{exceeds}
\expandafter\gdef\csname Nn@ood.javascript.diff.verdict\endcsname{512}
\expandafter\gdef\csname Nstatus@ood.javascript.diff.verdict\endcsname{final}
\expandafter\gdef\csname Nstatus@ood.javascript.ex.cos.bo1\endcsname{pending}
\expandafter\gdef\csname N@ood.javascript.ex.cos.bo8\endcsname{0.6993}
\expandafter\gdef\csname Nn@ood.javascript.ex.cos.bo8\endcsname{512}
\expandafter\gdef\csname Nstatus@ood.javascript.ex.cos.bo8\endcsname{final}
\expandafter\gdef\csname N@ood.javascript.ex.lid\endcsname{n/m}
\expandafter\gdef\csname Nn@ood.javascript.ex.lid\endcsname{512}
\expandafter\gdef\csname Nstatus@ood.javascript.ex.lid\endcsname{final}
\expandafter\gdef\csname N@ood.jpn.base.bpb\endcsname{0.8309}
\expandafter\gdef\csname Nn@ood.jpn.base.bpb\endcsname{512}
\expandafter\gdef\csname Nstatus@ood.jpn.base.bpb\endcsname{final}
\expandafter\gdef\csname N@ood.jpn.base.cos.bo8\endcsname{0.2038}
\expandafter\gdef\csname Nn@ood.jpn.base.cos.bo8\endcsname{512}
\expandafter\gdef\csname Nstatus@ood.jpn.base.cos.bo8\endcsname{final}
\expandafter\gdef\csname N@ood.jpn.corp10.cos\endcsname{0.6914}
\expandafter\gdef\csname Nn@ood.jpn.corp10.cos\endcsname{512}
\expandafter\gdef\csname Nstatus@ood.jpn.corp10.cos\endcsname{final}
\expandafter\gdef\csname N@ood.jpn.corp10.lid\endcsname{0.551}
\expandafter\gdef\csname Nn@ood.jpn.corp10.lid\endcsname{512}
\expandafter\gdef\csname Nstatus@ood.jpn.corp10.lid\endcsname{final}
\expandafter\gdef\csname N@ood.jpn.corp10.mtok\endcsname{10}
\expandafter\gdef\csname Nstatus@ood.jpn.corp10.mtok\endcsname{final}
\expandafter\gdef\csname N@ood.jpn.diff.cos.bo8\endcsname{0.0931}
\expandafter\gdef\csname Nse@ood.jpn.diff.cos.bo8\endcsname{0.0046}
\expandafter\gdef\csname Nlo@ood.jpn.diff.cos.bo8\endcsname{0.0838}
\expandafter\gdef\csname Nhi@ood.jpn.diff.cos.bo8\endcsname{0.1022}
\expandafter\gdef\csname Nn@ood.jpn.diff.cos.bo8\endcsname{512}
\expandafter\gdef\csname Nstatus@ood.jpn.diff.cos.bo8\endcsname{final}
\expandafter\gdef\csname N@ood.jpn.diff.verdict\endcsname{exceeds}
\expandafter\gdef\csname Nn@ood.jpn.diff.verdict\endcsname{512}
\expandafter\gdef\csname Nstatus@ood.jpn.diff.verdict\endcsname{final}
\expandafter\gdef\csname Nstatus@ood.jpn.ex.cos.bo1\endcsname{pending}
\expandafter\gdef\csname N@ood.jpn.ex.cos.bo8\endcsname{0.7845}
\expandafter\gdef\csname Nn@ood.jpn.ex.cos.bo8\endcsname{512}
\expandafter\gdef\csname Nstatus@ood.jpn.ex.cos.bo8\endcsname{final}
\expandafter\gdef\csname N@ood.jpn.ex.lid\endcsname{0.590}
\expandafter\gdef\csname Nn@ood.jpn.ex.lid\endcsname{512}
\expandafter\gdef\csname Nstatus@ood.jpn.ex.lid\endcsname{final}
\expandafter\gdef\csname N@ood.lean.base.bpb\endcsname{0.4810}
\expandafter\gdef\csname Nn@ood.lean.base.bpb\endcsname{512}
\expandafter\gdef\csname Nstatus@ood.lean.base.bpb\endcsname{final}
\expandafter\gdef\csname N@ood.lean.base.cos.bo8\endcsname{0.1998}
\expandafter\gdef\csname Nn@ood.lean.base.cos.bo8\endcsname{512}
\expandafter\gdef\csname Nstatus@ood.lean.base.cos.bo8\endcsname{final}
\expandafter\gdef\csname N@ood.lean.corp10.cos\endcsname{0.7923}
\expandafter\gdef\csname Nn@ood.lean.corp10.cos\endcsname{512}
\expandafter\gdef\csname Nstatus@ood.lean.corp10.cos\endcsname{final}
\expandafter\gdef\csname N@ood.lean.corp10.lid\endcsname{0.070}
\expandafter\gdef\csname Nn@ood.lean.corp10.lid\endcsname{512}
\expandafter\gdef\csname Nstatus@ood.lean.corp10.lid\endcsname{final}
\expandafter\gdef\csname N@ood.lean.corp10.mtok\endcsname{10}
\expandafter\gdef\csname Nstatus@ood.lean.corp10.mtok\endcsname{final}
\expandafter\gdef\csname N@ood.lean.diff.cos.bo8\endcsname{-0.0606}
\expandafter\gdef\csname Nse@ood.lean.diff.cos.bo8\endcsname{0.0054}
\expandafter\gdef\csname Nlo@ood.lean.diff.cos.bo8\endcsname{-0.0709}
\expandafter\gdef\csname Nhi@ood.lean.diff.cos.bo8\endcsname{-0.0502}
\expandafter\gdef\csname Nn@ood.lean.diff.cos.bo8\endcsname{512}
\expandafter\gdef\csname Nstatus@ood.lean.diff.cos.bo8\endcsname{final}
\expandafter\gdef\csname N@ood.lean.diff.verdict\endcsname{reversed}
\expandafter\gdef\csname Nn@ood.lean.diff.verdict\endcsname{512}
\expandafter\gdef\csname Nstatus@ood.lean.diff.verdict\endcsname{final}
\expandafter\gdef\csname Nstatus@ood.lean.ex.cos.bo1\endcsname{pending}
\expandafter\gdef\csname N@ood.lean.ex.cos.bo8\endcsname{0.7317}
\expandafter\gdef\csname Nn@ood.lean.ex.cos.bo8\endcsname{512}
\expandafter\gdef\csname Nstatus@ood.lean.ex.cos.bo8\endcsname{final}
\expandafter\gdef\csname N@ood.lean.ex.lid\endcsname{n/m}
\expandafter\gdef\csname Nn@ood.lean.ex.lid\endcsname{512}
\expandafter\gdef\csname Nstatus@ood.lean.ex.lid\endcsname{final}
\expandafter\gdef\csname N@ood.owm.base.bpb\endcsname{0.7097}
\expandafter\gdef\csname Nn@ood.owm.base.bpb\endcsname{512}
\expandafter\gdef\csname Nstatus@ood.owm.base.bpb\endcsname{final}
\expandafter\gdef\csname N@ood.owm.base.cos.bo8\endcsname{0.2053}
\expandafter\gdef\csname Nn@ood.owm.base.cos.bo8\endcsname{512}
\expandafter\gdef\csname Nstatus@ood.owm.base.cos.bo8\endcsname{final}
\expandafter\gdef\csname N@ood.owm.corp10.cos\endcsname{0.6009}
\expandafter\gdef\csname Nn@ood.owm.corp10.cos\endcsname{512}
\expandafter\gdef\csname Nstatus@ood.owm.corp10.cos\endcsname{final}
\expandafter\gdef\csname N@ood.owm.corp10.lid\endcsname{0.754}
\expandafter\gdef\csname Nn@ood.owm.corp10.lid\endcsname{512}
\expandafter\gdef\csname Nstatus@ood.owm.corp10.lid\endcsname{final}
\expandafter\gdef\csname N@ood.owm.corp10.mtok\endcsname{10}
\expandafter\gdef\csname Nstatus@ood.owm.corp10.mtok\endcsname{final}
\expandafter\gdef\csname N@ood.owm.diff.cos.bo8\endcsname{0.1722}
\expandafter\gdef\csname Nse@ood.owm.diff.cos.bo8\endcsname{0.0067}
\expandafter\gdef\csname Nlo@ood.owm.diff.cos.bo8\endcsname{0.1593}
\expandafter\gdef\csname Nhi@ood.owm.diff.cos.bo8\endcsname{0.1853}
\expandafter\gdef\csname Nn@ood.owm.diff.cos.bo8\endcsname{512}
\expandafter\gdef\csname Nstatus@ood.owm.diff.cos.bo8\endcsname{final}
\expandafter\gdef\csname N@ood.owm.diff.verdict\endcsname{exceeds}
\expandafter\gdef\csname Nn@ood.owm.diff.verdict\endcsname{512}
\expandafter\gdef\csname Nstatus@ood.owm.diff.verdict\endcsname{final}
\expandafter\gdef\csname Nstatus@ood.owm.ex.cos.bo1\endcsname{pending}
\expandafter\gdef\csname N@ood.owm.ex.cos.bo8\endcsname{0.7731}
\expandafter\gdef\csname Nn@ood.owm.ex.cos.bo8\endcsname{512}
\expandafter\gdef\csname Nstatus@ood.owm.ex.cos.bo8\endcsname{final}
\expandafter\gdef\csname N@ood.owm.ex.lid\endcsname{0.953}
\expandafter\gdef\csname Nn@ood.owm.ex.lid\endcsname{512}
\expandafter\gdef\csname Nstatus@ood.owm.ex.lid\endcsname{final}
\expandafter\gdef\csname N@ood.python.base.bpb\endcsname{0.3946}
\expandafter\gdef\csname Nn@ood.python.base.bpb\endcsname{512}
\expandafter\gdef\csname Nstatus@ood.python.base.bpb\endcsname{final}
\expandafter\gdef\csname N@ood.python.corp10.cos\endcsname{0.6713}
\expandafter\gdef\csname Nn@ood.python.corp10.cos\endcsname{512}
\expandafter\gdef\csname Nstatus@ood.python.corp10.cos\endcsname{final}
\expandafter\gdef\csname N@ood.python.corp10.lid\endcsname{0.053}
\expandafter\gdef\csname Nn@ood.python.corp10.lid\endcsname{512}
\expandafter\gdef\csname Nstatus@ood.python.corp10.lid\endcsname{final}
\expandafter\gdef\csname N@ood.python.corp10.mtok\endcsname{10}
\expandafter\gdef\csname Nstatus@ood.python.corp10.mtok\endcsname{final}
\expandafter\gdef\csname N@ood.python.diff.verdict\endcsname{exceeds}
\expandafter\gdef\csname Nn@ood.python.diff.verdict\endcsname{512}
\expandafter\gdef\csname Nstatus@ood.python.diff.verdict\endcsname{final}
\expandafter\gdef\csname Nstatus@ood.python.ex.cos.bo1\endcsname{pending}
\expandafter\gdef\csname N@ood.python.ex.cos.bo8\endcsname{0.7190}
\expandafter\gdef\csname Nn@ood.python.ex.cos.bo8\endcsname{512}
\expandafter\gdef\csname Nstatus@ood.python.ex.cos.bo8\endcsname{final}
\expandafter\gdef\csname N@ood.python.ex.lid\endcsname{n/m}
\expandafter\gdef\csname Nn@ood.python.ex.lid\endcsname{512}
\expandafter\gdef\csname Nstatus@ood.python.ex.lid\endcsname{final}
\expandafter\gdef\csname N@ood.rus.base.bpb\endcsname{0.5232}
\expandafter\gdef\csname Nn@ood.rus.base.bpb\endcsname{512}
\expandafter\gdef\csname Nstatus@ood.rus.base.bpb\endcsname{final}
\expandafter\gdef\csname N@ood.rus.base.cos.bo8\endcsname{0.1953}
\expandafter\gdef\csname Nn@ood.rus.base.cos.bo8\endcsname{512}
\expandafter\gdef\csname Nstatus@ood.rus.base.cos.bo8\endcsname{final}
\expandafter\gdef\csname N@ood.rus.corp10.cos\endcsname{0.6712}
\expandafter\gdef\csname Nn@ood.rus.corp10.cos\endcsname{512}
\expandafter\gdef\csname Nstatus@ood.rus.corp10.cos\endcsname{final}
\expandafter\gdef\csname N@ood.rus.corp10.lid\endcsname{0.787}
\expandafter\gdef\csname Nn@ood.rus.corp10.lid\endcsname{512}
\expandafter\gdef\csname Nstatus@ood.rus.corp10.lid\endcsname{final}
\expandafter\gdef\csname N@ood.rus.corp10.mtok\endcsname{10}
\expandafter\gdef\csname Nstatus@ood.rus.corp10.mtok\endcsname{final}
\expandafter\gdef\csname N@ood.rus.diff.cos.bo8\endcsname{0.1314}
\expandafter\gdef\csname Nse@ood.rus.diff.cos.bo8\endcsname{0.0049}
\expandafter\gdef\csname Nlo@ood.rus.diff.cos.bo8\endcsname{0.1217}
\expandafter\gdef\csname Nhi@ood.rus.diff.cos.bo8\endcsname{0.1413}
\expandafter\gdef\csname Nn@ood.rus.diff.cos.bo8\endcsname{512}
\expandafter\gdef\csname Nstatus@ood.rus.diff.cos.bo8\endcsname{final}
\expandafter\gdef\csname N@ood.rus.diff.verdict\endcsname{exceeds}
\expandafter\gdef\csname Nn@ood.rus.diff.verdict\endcsname{512}
\expandafter\gdef\csname Nstatus@ood.rus.diff.verdict\endcsname{final}
\expandafter\gdef\csname Nstatus@ood.rus.ex.cos.bo1\endcsname{pending}
\expandafter\gdef\csname N@ood.rus.ex.cos.bo8\endcsname{0.8026}
\expandafter\gdef\csname Nn@ood.rus.ex.cos.bo8\endcsname{512}
\expandafter\gdef\csname Nstatus@ood.rus.ex.cos.bo8\endcsname{final}
\expandafter\gdef\csname N@ood.rus.ex.lid\endcsname{0.773}
\expandafter\gdef\csname Nn@ood.rus.ex.lid\endcsname{512}
\expandafter\gdef\csname Nstatus@ood.rus.ex.lid\endcsname{final}
\expandafter\gdef\csname N@ood.rust.base.bpb\endcsname{0.4129}
\expandafter\gdef\csname Nn@ood.rust.base.bpb\endcsname{512}
\expandafter\gdef\csname Nstatus@ood.rust.base.bpb\endcsname{final}
\expandafter\gdef\csname N@ood.rust.base.cos.bo8\endcsname{0.2077}
\expandafter\gdef\csname Nn@ood.rust.base.cos.bo8\endcsname{512}
\expandafter\gdef\csname Nstatus@ood.rust.base.cos.bo8\endcsname{final}
\expandafter\gdef\csname N@ood.rust.corp10.cos\endcsname{0.6889}
\expandafter\gdef\csname Nn@ood.rust.corp10.cos\endcsname{512}
\expandafter\gdef\csname Nstatus@ood.rust.corp10.cos\endcsname{final}
\expandafter\gdef\csname N@ood.rust.corp10.lid\endcsname{0.224}
\expandafter\gdef\csname Nn@ood.rust.corp10.lid\endcsname{512}
\expandafter\gdef\csname Nstatus@ood.rust.corp10.lid\endcsname{final}
\expandafter\gdef\csname N@ood.rust.corp10.mtok\endcsname{10}
\expandafter\gdef\csname Nstatus@ood.rust.corp10.mtok\endcsname{final}
\expandafter\gdef\csname N@ood.rust.diff.cos.bo8\endcsname{0.0134}
\expandafter\gdef\csname Nse@ood.rust.diff.cos.bo8\endcsname{0.0062}
\expandafter\gdef\csname Nlo@ood.rust.diff.cos.bo8\endcsname{0.0014}
\expandafter\gdef\csname Nhi@ood.rust.diff.cos.bo8\endcsname{0.0256}
\expandafter\gdef\csname Nn@ood.rust.diff.cos.bo8\endcsname{512}
\expandafter\gdef\csname Nstatus@ood.rust.diff.cos.bo8\endcsname{final}
\expandafter\gdef\csname N@ood.rust.diff.verdict\endcsname{exceeds}
\expandafter\gdef\csname Nn@ood.rust.diff.verdict\endcsname{512}
\expandafter\gdef\csname Nstatus@ood.rust.diff.verdict\endcsname{final}
\expandafter\gdef\csname Nstatus@ood.rust.ex.cos.bo1\endcsname{pending}
\expandafter\gdef\csname N@ood.rust.ex.cos.bo8\endcsname{0.7023}
\expandafter\gdef\csname Nn@ood.rust.ex.cos.bo8\endcsname{512}
\expandafter\gdef\csname Nstatus@ood.rust.ex.cos.bo8\endcsname{final}
\expandafter\gdef\csname N@ood.rust.ex.lid\endcsname{n/m}
\expandafter\gdef\csname Nn@ood.rust.ex.lid\endcsname{512}
\expandafter\gdef\csname Nstatus@ood.rust.ex.lid\endcsname{final}
\expandafter\gdef\csname N@ood.shell.base.bpb\endcsname{0.5522}
\expandafter\gdef\csname Nn@ood.shell.base.bpb\endcsname{512}
\expandafter\gdef\csname Nstatus@ood.shell.base.bpb\endcsname{final}
\expandafter\gdef\csname N@ood.shell.base.cos.bo8\endcsname{0.2393}
\expandafter\gdef\csname Nn@ood.shell.base.cos.bo8\endcsname{512}
\expandafter\gdef\csname Nstatus@ood.shell.base.cos.bo8\endcsname{final}
\expandafter\gdef\csname N@ood.shell.corp10.cos\endcsname{0.7119}
\expandafter\gdef\csname Nn@ood.shell.corp10.cos\endcsname{512}
\expandafter\gdef\csname Nstatus@ood.shell.corp10.cos\endcsname{final}
\expandafter\gdef\csname N@ood.shell.corp10.lid\endcsname{0.026}
\expandafter\gdef\csname Nn@ood.shell.corp10.lid\endcsname{512}
\expandafter\gdef\csname Nstatus@ood.shell.corp10.lid\endcsname{final}
\expandafter\gdef\csname N@ood.shell.corp10.mtok\endcsname{4}
\expandafter\gdef\csname Nstatus@ood.shell.corp10.mtok\endcsname{final}
\expandafter\gdef\csname N@ood.shell.diff.cos.bo8\endcsname{0.0110}
\expandafter\gdef\csname Nse@ood.shell.diff.cos.bo8\endcsname{0.0063}
\expandafter\gdef\csname Nlo@ood.shell.diff.cos.bo8\endcsname{-0.0013}
\expandafter\gdef\csname Nhi@ood.shell.diff.cos.bo8\endcsname{0.0231}
\expandafter\gdef\csname Nn@ood.shell.diff.cos.bo8\endcsname{512}
\expandafter\gdef\csname Nstatus@ood.shell.diff.cos.bo8\endcsname{final}
\expandafter\gdef\csname N@ood.shell.diff.verdict\endcsname{inconclusive}
\expandafter\gdef\csname Nn@ood.shell.diff.verdict\endcsname{512}
\expandafter\gdef\csname Nstatus@ood.shell.diff.verdict\endcsname{final}
\expandafter\gdef\csname Nstatus@ood.shell.ex.cos.bo1\endcsname{pending}
\expandafter\gdef\csname N@ood.shell.ex.cos.bo8\endcsname{0.7229}
\expandafter\gdef\csname Nn@ood.shell.ex.cos.bo8\endcsname{512}
\expandafter\gdef\csname Nstatus@ood.shell.ex.cos.bo8\endcsname{final}
\expandafter\gdef\csname N@ood.shell.ex.lid\endcsname{n/m}
\expandafter\gdef\csname Nn@ood.shell.ex.lid\endcsname{512}
\expandafter\gdef\csname Nstatus@ood.shell.ex.lid\endcsname{final}
\expandafter\gdef\csname N@ood.sql.base.bpb\endcsname{0.4527}
\expandafter\gdef\csname Nn@ood.sql.base.bpb\endcsname{512}
\expandafter\gdef\csname Nstatus@ood.sql.base.bpb\endcsname{final}
\expandafter\gdef\csname N@ood.sql.base.cos.bo8\endcsname{0.2397}
\expandafter\gdef\csname Nn@ood.sql.base.cos.bo8\endcsname{512}
\expandafter\gdef\csname Nstatus@ood.sql.base.cos.bo8\endcsname{final}
\expandafter\gdef\csname N@ood.sql.corp10.cos\endcsname{0.7161}
\expandafter\gdef\csname Nn@ood.sql.corp10.cos\endcsname{512}
\expandafter\gdef\csname Nstatus@ood.sql.corp10.cos\endcsname{final}
\expandafter\gdef\csname N@ood.sql.corp10.lid\endcsname{0.010}
\expandafter\gdef\csname Nn@ood.sql.corp10.lid\endcsname{512}
\expandafter\gdef\csname Nstatus@ood.sql.corp10.lid\endcsname{final}
\expandafter\gdef\csname N@ood.sql.corp10.mtok\endcsname{10}
\expandafter\gdef\csname Nstatus@ood.sql.corp10.mtok\endcsname{final}
\expandafter\gdef\csname N@ood.sql.diff.cos.bo8\endcsname{-0.0153}
\expandafter\gdef\csname Nse@ood.sql.diff.cos.bo8\endcsname{0.0068}
\expandafter\gdef\csname Nlo@ood.sql.diff.cos.bo8\endcsname{-0.0285}
\expandafter\gdef\csname Nhi@ood.sql.diff.cos.bo8\endcsname{-0.0021}
\expandafter\gdef\csname Nn@ood.sql.diff.cos.bo8\endcsname{512}
\expandafter\gdef\csname Nstatus@ood.sql.diff.cos.bo8\endcsname{final}
\expandafter\gdef\csname N@ood.sql.diff.verdict\endcsname{reversed}
\expandafter\gdef\csname Nn@ood.sql.diff.verdict\endcsname{512}
\expandafter\gdef\csname Nstatus@ood.sql.diff.verdict\endcsname{final}
\expandafter\gdef\csname Nstatus@ood.sql.ex.cos.bo1\endcsname{pending}
\expandafter\gdef\csname N@ood.sql.ex.cos.bo8\endcsname{0.7008}
\expandafter\gdef\csname Nn@ood.sql.ex.cos.bo8\endcsname{512}
\expandafter\gdef\csname Nstatus@ood.sql.ex.cos.bo8\endcsname{final}
\expandafter\gdef\csname N@ood.sql.ex.lid\endcsname{n/m}
\expandafter\gdef\csname Nn@ood.sql.ex.lid\endcsname{512}
\expandafter\gdef\csname Nstatus@ood.sql.ex.lid\endcsname{final}
\expandafter\gdef\csname N@ood.tha.base.bpb\endcsname{1.0167}
\expandafter\gdef\csname Nn@ood.tha.base.bpb\endcsname{512}
\expandafter\gdef\csname Nstatus@ood.tha.base.bpb\endcsname{final}
\expandafter\gdef\csname N@ood.tha.base.cos.bo8\endcsname{0.2167}
\expandafter\gdef\csname Nn@ood.tha.base.cos.bo8\endcsname{512}
\expandafter\gdef\csname Nstatus@ood.tha.base.cos.bo8\endcsname{final}
\expandafter\gdef\csname N@ood.tha.corp10.cos\endcsname{0.7906}
\expandafter\gdef\csname Nn@ood.tha.corp10.cos\endcsname{512}
\expandafter\gdef\csname Nstatus@ood.tha.corp10.cos\endcsname{final}
\expandafter\gdef\csname N@ood.tha.corp10.lid\endcsname{0.722}
\expandafter\gdef\csname Nn@ood.tha.corp10.lid\endcsname{512}
\expandafter\gdef\csname Nstatus@ood.tha.corp10.lid\endcsname{final}
\expandafter\gdef\csname N@ood.tha.corp10.mtok\endcsname{10}
\expandafter\gdef\csname Nstatus@ood.tha.corp10.mtok\endcsname{final}
\expandafter\gdef\csname N@ood.tha.diff.verdict\endcsname{reversed}
\expandafter\gdef\csname Nn@ood.tha.diff.verdict\endcsname{512}
\expandafter\gdef\csname Nstatus@ood.tha.diff.verdict\endcsname{final}
\expandafter\gdef\csname Nstatus@ood.tha.ex.cos.bo1\endcsname{pending}
\expandafter\gdef\csname N@ood.tha.ex.cos.bo8\endcsname{0.7420}
\expandafter\gdef\csname Nn@ood.tha.ex.cos.bo8\endcsname{512}
\expandafter\gdef\csname Nstatus@ood.tha.ex.cos.bo8\endcsname{final}
\expandafter\gdef\csname N@ood.tha.ex.lid\endcsname{0.779}
\expandafter\gdef\csname Nn@ood.tha.ex.lid\endcsname{512}
\expandafter\gdef\csname Nstatus@ood.tha.ex.lid\endcsname{final}
\expandafter\gdef\csname N@ood.zh.base.bpb\endcsname{0.9030}
\expandafter\gdef\csname Nn@ood.zh.base.bpb\endcsname{512}
\expandafter\gdef\csname Nstatus@ood.zh.base.bpb\endcsname{final}
\expandafter\gdef\csname N@ood.zh.base.cos.bo8\endcsname{0.1835}
\expandafter\gdef\csname Nn@ood.zh.base.cos.bo8\endcsname{512}
\expandafter\gdef\csname Nstatus@ood.zh.base.cos.bo8\endcsname{final}
\expandafter\gdef\csname N@ood.zh.corp10.cos\endcsname{0.6260}
\expandafter\gdef\csname Nn@ood.zh.corp10.cos\endcsname{512}
\expandafter\gdef\csname Nstatus@ood.zh.corp10.cos\endcsname{final}
\expandafter\gdef\csname N@ood.zh.corp10.lid\endcsname{0.714}
\expandafter\gdef\csname Nn@ood.zh.corp10.lid\endcsname{512}
\expandafter\gdef\csname Nstatus@ood.zh.corp10.lid\endcsname{final}
\expandafter\gdef\csname N@ood.zh.corp10.mtok\endcsname{10}
\expandafter\gdef\csname Nstatus@ood.zh.corp10.mtok\endcsname{final}
\expandafter\gdef\csname N@ood.zh.diff.verdict\endcsname{exceeds}
\expandafter\gdef\csname Nn@ood.zh.diff.verdict\endcsname{512}
\expandafter\gdef\csname Nstatus@ood.zh.diff.verdict\endcsname{final}
\expandafter\gdef\csname Nstatus@ood.zh.ex.cos.bo1\endcsname{pending}
\expandafter\gdef\csname N@ood.zh.ex.cos.bo8\endcsname{0.8091}
\expandafter\gdef\csname Nn@ood.zh.ex.cos.bo8\endcsname{512}
\expandafter\gdef\csname Nstatus@ood.zh.ex.cos.bo8\endcsname{final}
\expandafter\gdef\csname N@ood.zh.ex.lid\endcsname{0.725}
\expandafter\gdef\csname Nn@ood.zh.ex.lid\endcsname{512}
\expandafter\gdef\csname Nstatus@ood.zh.ex.lid\endcsname{final}
\expandafter\gdef\csname N@sae.l131k.ex.fired.bo1.q1\endcsname{0.4965}
\expandafter\gdef\csname Nse@sae.l131k.ex.fired.bo1.q1\endcsname{0.0398}
\expandafter\gdef\csname Nlo@sae.l131k.ex.fired.bo1.q1\endcsname{0.4183}
\expandafter\gdef\csname Nhi@sae.l131k.ex.fired.bo1.q1\endcsname{0.5743}
\expandafter\gdef\csname Nn@sae.l131k.ex.fired.bo1.q1\endcsname{128}
\expandafter\gdef\csname Nstatus@sae.l131k.ex.fired.bo1.q1\endcsname{final}
\expandafter\gdef\csname N@sae.l131k.ex.fired.bo1.q2\endcsname{0.9297}
\expandafter\gdef\csname Nse@sae.l131k.ex.fired.bo1.q2\endcsname{0.0200}
\expandafter\gdef\csname Nlo@sae.l131k.ex.fired.bo1.q2\endcsname{0.8881}
\expandafter\gdef\csname Nhi@sae.l131k.ex.fired.bo1.q2\endcsname{0.9657}
\expandafter\gdef\csname Nn@sae.l131k.ex.fired.bo1.q2\endcsname{128}
\expandafter\gdef\csname Nstatus@sae.l131k.ex.fired.bo1.q2\endcsname{final}
\expandafter\gdef\csname N@sae.l131k.ex.fired.bo1.q3\endcsname{0.9716}
\expandafter\gdef\csname Nse@sae.l131k.ex.fired.bo1.q3\endcsname{0.0130}
\expandafter\gdef\csname Nlo@sae.l131k.ex.fired.bo1.q3\endcsname{0.9429}
\expandafter\gdef\csname Nhi@sae.l131k.ex.fired.bo1.q3\endcsname{0.9938}
\expandafter\gdef\csname Nn@sae.l131k.ex.fired.bo1.q3\endcsname{128}
\expandafter\gdef\csname Nstatus@sae.l131k.ex.fired.bo1.q3\endcsname{final}
\expandafter\gdef\csname N@sae.l131k.ex.fired.bo1.q4\endcsname{0.9301}
\expandafter\gdef\csname Nse@sae.l131k.ex.fired.bo1.q4\endcsname{0.0181}
\expandafter\gdef\csname Nlo@sae.l131k.ex.fired.bo1.q4\endcsname{0.8917}
\expandafter\gdef\csname Nhi@sae.l131k.ex.fired.bo1.q4\endcsname{0.9626}
\expandafter\gdef\csname Nn@sae.l131k.ex.fired.bo1.q4\endcsname{128}
\expandafter\gdef\csname Nstatus@sae.l131k.ex.fired.bo1.q4\endcsname{final}
\expandafter\gdef\csname N@sae.l131k.ex.fired.bo64.q1\endcsname{0.7578}
\expandafter\gdef\csname Nse@sae.l131k.ex.fired.bo64.q1\endcsname{0.0382}
\expandafter\gdef\csname Nlo@sae.l131k.ex.fired.bo64.q1\endcsname{0.6742}
\expandafter\gdef\csname Nhi@sae.l131k.ex.fired.bo64.q1\endcsname{0.8291}
\expandafter\gdef\csname Nn@sae.l131k.ex.fired.bo64.q1\endcsname{128}
\expandafter\gdef\csname Nstatus@sae.l131k.ex.fired.bo64.q1\endcsname{final}
\expandafter\gdef\csname N@sae.l131k.ex.fired.bo64.q2\endcsname{0.9922}
\expandafter\gdef\csname Nse@sae.l131k.ex.fired.bo64.q2\endcsname{0.0078}
\expandafter\gdef\csname Nlo@sae.l131k.ex.fired.bo64.q2\endcsname{0.9572}
\expandafter\gdef\csname Nhi@sae.l131k.ex.fired.bo64.q2\endcsname{0.9998}
\expandafter\gdef\csname Nn@sae.l131k.ex.fired.bo64.q2\endcsname{128}
\expandafter\gdef\csname Nstatus@sae.l131k.ex.fired.bo64.q2\endcsname{final}
\expandafter\gdef\csname N@sae.l131k.ex.fired.bo64.q3\endcsname{0.9922}
\expandafter\gdef\csname Nse@sae.l131k.ex.fired.bo64.q3\endcsname{0.0078}
\expandafter\gdef\csname Nlo@sae.l131k.ex.fired.bo64.q3\endcsname{0.9572}
\expandafter\gdef\csname Nhi@sae.l131k.ex.fired.bo64.q3\endcsname{0.9998}
\expandafter\gdef\csname Nn@sae.l131k.ex.fired.bo64.q3\endcsname{128}
\expandafter\gdef\csname Nstatus@sae.l131k.ex.fired.bo64.q3\endcsname{final}
\expandafter\gdef\csname N@sae.l131k.ex.fired.bo64.q4\endcsname{1.0000}
\expandafter\gdef\csname Nlo@sae.l131k.ex.fired.bo64.q4\endcsname{0.9716}
\expandafter\gdef\csname Nhi@sae.l131k.ex.fired.bo64.q4\endcsname{1.0000}
\expandafter\gdef\csname Nn@sae.l131k.ex.fired.bo64.q4\endcsname{128}
\expandafter\gdef\csname Nstatus@sae.l131k.ex.fired.bo64.q4\endcsname{final}
\expandafter\gdef\csname N@sae.l131k.ex.fired.bo8.q2\endcsname{0.9683}
\expandafter\gdef\csname Nse@sae.l131k.ex.fired.bo8.q2\endcsname{0.0124}
\expandafter\gdef\csname Nlo@sae.l131k.ex.fired.bo8.q2\endcsname{0.9414}
\expandafter\gdef\csname Nhi@sae.l131k.ex.fired.bo8.q2\endcsname{0.9901}
\expandafter\gdef\csname Nn@sae.l131k.ex.fired.bo8.q2\endcsname{128}
\expandafter\gdef\csname Nstatus@sae.l131k.ex.fired.bo8.q2\endcsname{final}
\expandafter\gdef\csname N@sae.l131k.ex.fired.bo8.q3\endcsname{0.9838}
\expandafter\gdef\csname Nse@sae.l131k.ex.fired.bo8.q3\endcsname{0.0098}
\expandafter\gdef\csname Nlo@sae.l131k.ex.fired.bo8.q3\endcsname{0.9624}
\expandafter\gdef\csname Nhi@sae.l131k.ex.fired.bo8.q3\endcsname{1.0000}
\expandafter\gdef\csname Nn@sae.l131k.ex.fired.bo8.q3\endcsname{128}
\expandafter\gdef\csname Nstatus@sae.l131k.ex.fired.bo8.q3\endcsname{final}
\expandafter\gdef\csname N@sae.l131k.ex.ratio.bo64.q1\endcsname{0.3684}
\expandafter\gdef\csname Nse@sae.l131k.ex.ratio.bo64.q1\endcsname{0.0888}
\expandafter\gdef\csname Nlo@sae.l131k.ex.ratio.bo64.q1\endcsname{0.2336}
\expandafter\gdef\csname Nhi@sae.l131k.ex.ratio.bo64.q1\endcsname{0.6239}
\expandafter\gdef\csname Nn@sae.l131k.ex.ratio.bo64.q1\endcsname{128}
\expandafter\gdef\csname Nstatus@sae.l131k.ex.ratio.bo64.q1\endcsname{final}
\expandafter\gdef\csname N@sae.l131k.ex.ratio.bo64.q2\endcsname{1.0393}
\expandafter\gdef\csname Nse@sae.l131k.ex.ratio.bo64.q2\endcsname{0.0209}
\expandafter\gdef\csname Nlo@sae.l131k.ex.ratio.bo64.q2\endcsname{0.9900}
\expandafter\gdef\csname Nhi@sae.l131k.ex.ratio.bo64.q2\endcsname{1.0789}
\expandafter\gdef\csname Nn@sae.l131k.ex.ratio.bo64.q2\endcsname{128}
\expandafter\gdef\csname Nstatus@sae.l131k.ex.ratio.bo64.q2\endcsname{final}
\expandafter\gdef\csname N@sae.l131k.ex.ratio.bo64.q3\endcsname{1.0009}
\expandafter\gdef\csname Nse@sae.l131k.ex.ratio.bo64.q3\endcsname{0.0254}
\expandafter\gdef\csname Nlo@sae.l131k.ex.ratio.bo64.q3\endcsname{0.9586}
\expandafter\gdef\csname Nhi@sae.l131k.ex.ratio.bo64.q3\endcsname{1.0634}
\expandafter\gdef\csname Nn@sae.l131k.ex.ratio.bo64.q3\endcsname{128}
\expandafter\gdef\csname Nstatus@sae.l131k.ex.ratio.bo64.q3\endcsname{final}
\expandafter\gdef\csname N@sae.l131k.ex.ratio.bo64.q4\endcsname{1.0100}
\expandafter\gdef\csname Nse@sae.l131k.ex.ratio.bo64.q4\endcsname{0.0236}
\expandafter\gdef\csname Nlo@sae.l131k.ex.ratio.bo64.q4\endcsname{0.9529}
\expandafter\gdef\csname Nhi@sae.l131k.ex.ratio.bo64.q4\endcsname{1.0495}
\expandafter\gdef\csname Nn@sae.l131k.ex.ratio.bo64.q4\endcsname{128}
\expandafter\gdef\csname Nstatus@sae.l131k.ex.ratio.bo64.q4\endcsname{final}
\expandafter\gdef\csname N@sae.l131k.ex.ratio.bo8.q2\endcsname{0.9403}
\expandafter\gdef\csname Nse@sae.l131k.ex.ratio.bo8.q2\endcsname{0.0174}
\expandafter\gdef\csname Nlo@sae.l131k.ex.ratio.bo8.q2\endcsname{0.9076}
\expandafter\gdef\csname Nhi@sae.l131k.ex.ratio.bo8.q2\endcsname{0.9775}
\expandafter\gdef\csname Nn@sae.l131k.ex.ratio.bo8.q2\endcsname{128}
\expandafter\gdef\csname Nstatus@sae.l131k.ex.ratio.bo8.q2\endcsname{final}
\expandafter\gdef\csname N@sae.l131k.ex.ratio.bo8.q3\endcsname{0.9167}
\expandafter\gdef\csname Nse@sae.l131k.ex.ratio.bo8.q3\endcsname{0.0189}
\expandafter\gdef\csname Nlo@sae.l131k.ex.ratio.bo8.q3\endcsname{0.8785}
\expandafter\gdef\csname Nhi@sae.l131k.ex.ratio.bo8.q3\endcsname{0.9622}
\expandafter\gdef\csname Nn@sae.l131k.ex.ratio.bo8.q3\endcsname{128}
\expandafter\gdef\csname Nstatus@sae.l131k.ex.ratio.bo8.q3\endcsname{final}
\expandafter\gdef\csname N@sae.l131k.ex.ratio.bo8.q4\endcsname{0.8859}
\expandafter\gdef\csname Nse@sae.l131k.ex.ratio.bo8.q4\endcsname{0.0202}
\expandafter\gdef\csname Nlo@sae.l131k.ex.ratio.bo8.q4\endcsname{0.8485}
\expandafter\gdef\csname Nhi@sae.l131k.ex.ratio.bo8.q4\endcsname{0.9246}
\expandafter\gdef\csname Nn@sae.l131k.ex.ratio.bo8.q4\endcsname{128}
\expandafter\gdef\csname Nstatus@sae.l131k.ex.ratio.bo8.q4\endcsname{final}
\expandafter\gdef\csname N@sae.s2mdec.ex.fired.bo1.q1\endcsname{0.0571}
\expandafter\gdef\csname Nn@sae.s2mdec.ex.fired.bo1.q1\endcsname{128}
\expandafter\gdef\csname Nstatus@sae.s2mdec.ex.fired.bo1.q1\endcsname{final}
\expandafter\gdef\csname N@sae.s2mdec.ex.fired.bo1.q2\endcsname{0.0980}
\expandafter\gdef\csname Nn@sae.s2mdec.ex.fired.bo1.q2\endcsname{128}
\expandafter\gdef\csname Nstatus@sae.s2mdec.ex.fired.bo1.q2\endcsname{final}
\expandafter\gdef\csname N@sae.s2mdec.ex.fired.bo1.q3\endcsname{0.2063}
\expandafter\gdef\csname Nn@sae.s2mdec.ex.fired.bo1.q3\endcsname{128}
\expandafter\gdef\csname Nstatus@sae.s2mdec.ex.fired.bo1.q3\endcsname{final}
\expandafter\gdef\csname N@sae.s2mdec.ex.fired.bo1.q4\endcsname{0.5493}
\expandafter\gdef\csname Nn@sae.s2mdec.ex.fired.bo1.q4\endcsname{128}
\expandafter\gdef\csname Nstatus@sae.s2mdec.ex.fired.bo1.q4\endcsname{final}
\expandafter\gdef\csname N@sae.s2mdec.ex.fired.bo8.q1\endcsname{0.2385}
\expandafter\gdef\csname Nn@sae.s2mdec.ex.fired.bo8.q1\endcsname{128}
\expandafter\gdef\csname Nstatus@sae.s2mdec.ex.fired.bo8.q1\endcsname{final}
\expandafter\gdef\csname N@sae.s2mdec.ex.fired.bo8.q2\endcsname{0.3331}
\expandafter\gdef\csname Nn@sae.s2mdec.ex.fired.bo8.q2\endcsname{128}
\expandafter\gdef\csname Nstatus@sae.s2mdec.ex.fired.bo8.q2\endcsname{final}
\expandafter\gdef\csname N@sae.s2mdec.ex.fired.bo8.q3\endcsname{0.5799}
\expandafter\gdef\csname Nn@sae.s2mdec.ex.fired.bo8.q3\endcsname{128}
\expandafter\gdef\csname Nstatus@sae.s2mdec.ex.fired.bo8.q3\endcsname{final}
\expandafter\gdef\csname N@sae.s2mdec.ex.fired.bo8.q4\endcsname{0.8447}
\expandafter\gdef\csname Nn@sae.s2mdec.ex.fired.bo8.q4\endcsname{128}
\expandafter\gdef\csname Nstatus@sae.s2mdec.ex.fired.bo8.q4\endcsname{final}
\expandafter\gdef\csname Nstatus@sae.s2mdec.ex.ratio.bo1.q1\endcsname{pending}
\expandafter\gdef\csname Nstatus@sae.s2mdec.ex.ratio.bo1.q2\endcsname{pending}
\expandafter\gdef\csname Nstatus@sae.s2mdec.ex.ratio.bo1.q3\endcsname{pending}
\expandafter\gdef\csname Nstatus@sae.s2mdec.ex.ratio.bo1.q4\endcsname{pending}
\expandafter\gdef\csname Nstatus@sae.s2mdec.ex.ratio.bo8.q1\endcsname{pending}
\expandafter\gdef\csname Nstatus@sae.s2mdec.ex.ratio.bo8.q2\endcsname{pending}
\expandafter\gdef\csname Nstatus@sae.s2mdec.ex.ratio.bo8.q3\endcsname{pending}
\expandafter\gdef\csname Nstatus@sae.s2mdec.ex.ratio.bo8.q4\endcsname{pending}
\expandafter\gdef\csname N@sae.s2menc.ex.fired.bo1.q1\endcsname{0.0336}
\expandafter\gdef\csname Nn@sae.s2menc.ex.fired.bo1.q1\endcsname{128}
\expandafter\gdef\csname Nstatus@sae.s2menc.ex.fired.bo1.q1\endcsname{final}
\expandafter\gdef\csname N@sae.s2menc.ex.fired.bo1.q2\endcsname{0.0822}
\expandafter\gdef\csname Nn@sae.s2menc.ex.fired.bo1.q2\endcsname{128}
\expandafter\gdef\csname Nstatus@sae.s2menc.ex.fired.bo1.q2\endcsname{final}
\expandafter\gdef\csname N@sae.s2menc.ex.fired.bo1.q3\endcsname{0.1663}
\expandafter\gdef\csname Nn@sae.s2menc.ex.fired.bo1.q3\endcsname{128}
\expandafter\gdef\csname Nstatus@sae.s2menc.ex.fired.bo1.q3\endcsname{final}
\expandafter\gdef\csname N@sae.s2menc.ex.fired.bo1.q4\endcsname{0.4995}
\expandafter\gdef\csname Nn@sae.s2menc.ex.fired.bo1.q4\endcsname{128}
\expandafter\gdef\csname Nstatus@sae.s2menc.ex.fired.bo1.q4\endcsname{final}
\expandafter\gdef\csname N@sae.s2menc.ex.fired.bo8.q1\endcsname{0.1776}
\expandafter\gdef\csname Nn@sae.s2menc.ex.fired.bo8.q1\endcsname{128}
\expandafter\gdef\csname Nstatus@sae.s2menc.ex.fired.bo8.q1\endcsname{final}
\expandafter\gdef\csname N@sae.s2menc.ex.fired.bo8.q2\endcsname{0.3335}
\expandafter\gdef\csname Nn@sae.s2menc.ex.fired.bo8.q2\endcsname{128}
\expandafter\gdef\csname Nstatus@sae.s2menc.ex.fired.bo8.q2\endcsname{final}
\expandafter\gdef\csname N@sae.s2menc.ex.fired.bo8.q3\endcsname{0.5205}
\expandafter\gdef\csname Nn@sae.s2menc.ex.fired.bo8.q3\endcsname{128}
\expandafter\gdef\csname Nstatus@sae.s2menc.ex.fired.bo8.q3\endcsname{final}
\expandafter\gdef\csname N@sae.s2menc.ex.fired.bo8.q4\endcsname{0.7898}
\expandafter\gdef\csname Nn@sae.s2menc.ex.fired.bo8.q4\endcsname{128}
\expandafter\gdef\csname Nstatus@sae.s2menc.ex.fired.bo8.q4\endcsname{final}
\expandafter\gdef\csname Nstatus@sae.s2menc.ex.ratio.bo1.q1\endcsname{pending}
\expandafter\gdef\csname Nstatus@sae.s2menc.ex.ratio.bo1.q2\endcsname{pending}
\expandafter\gdef\csname Nstatus@sae.s2menc.ex.ratio.bo1.q3\endcsname{pending}
\expandafter\gdef\csname Nstatus@sae.s2menc.ex.ratio.bo1.q4\endcsname{pending}
\expandafter\gdef\csname Nstatus@sae.s2menc.ex.ratio.bo8.q1\endcsname{pending}
\expandafter\gdef\csname Nstatus@sae.s2menc.ex.ratio.bo8.q2\endcsname{pending}
\expandafter\gdef\csname Nstatus@sae.s2menc.ex.ratio.bo8.q3\endcsname{pending}
\expandafter\gdef\csname Nstatus@sae.s2menc.ex.ratio.bo8.q4\endcsname{pending}
\expandafter\gdef\csname N@sae.l131k.corp10.fired.q1\endcsname{0.9922}
\expandafter\gdef\csname Nn@sae.l131k.corp10.fired.q1\endcsname{128}
\expandafter\gdef\csname Nstatus@sae.l131k.corp10.fired.q1\endcsname{final}
\expandafter\gdef\csname N@sae.l131k.corp10.fired.q2\endcsname{1.0000}
\expandafter\gdef\csname Nn@sae.l131k.corp10.fired.q2\endcsname{128}
\expandafter\gdef\csname Nstatus@sae.l131k.corp10.fired.q2\endcsname{final}
\expandafter\gdef\csname N@sae.l131k.corp10.fired.q3\endcsname{1.0000}
\expandafter\gdef\csname Nn@sae.l131k.corp10.fired.q3\endcsname{128}
\expandafter\gdef\csname Nstatus@sae.l131k.corp10.fired.q3\endcsname{final}
\expandafter\gdef\csname N@sae.l131k.corp10.fired.q4\endcsname{1.0000}
\expandafter\gdef\csname Nn@sae.l131k.corp10.fired.q4\endcsname{128}
\expandafter\gdef\csname Nstatus@sae.l131k.corp10.fired.q4\endcsname{final}
\expandafter\gdef\csname N@ai.l131k.c16.refused\endcsname{6}
\expandafter\gdef\csname Nn@ai.l131k.c16.refused\endcsname{512}
\expandafter\gdef\csname Nstatus@ai.l131k.c16.refused\endcsname{final}
\expandafter\gdef\csname N@ai.l131k.mtop16.refused\endcsname{56}
\expandafter\gdef\csname Nn@ai.l131k.mtop16.refused\endcsname{512}
\expandafter\gdef\csname Nstatus@ai.l131k.mtop16.refused\endcsname{final}
\expandafter\gdef\csname N@ai.l131k.mjac16.refused\endcsname{68}
\expandafter\gdef\csname Nn@ai.l131k.mjac16.refused\endcsname{512}
\expandafter\gdef\csname Nstatus@ai.l131k.mjac16.refused\endcsname{final}
\expandafter\gdef\csname N@ai.l131k.mcos16.refused\endcsname{63}
\expandafter\gdef\csname Nn@ai.l131k.mcos16.refused\endcsname{512}
\expandafter\gdef\csname Nstatus@ai.l131k.mcos16.refused\endcsname{final}
\expandafter\gdef\csname N@ai.l131k.nla4.refused\endcsname{41}
\expandafter\gdef\csname Nn@ai.l131k.nla4.refused\endcsname{512}
\expandafter\gdef\csname Nstatus@ai.l131k.nla4.refused\endcsname{final}
\expandafter\gdef\csname N@sae.l131k.ex.fired.bo8\endcsname{0.8887}
\expandafter\gdef\csname Nse@sae.l131k.ex.fired.bo8\endcsname{0.0128}
\expandafter\gdef\csname Nlo@sae.l131k.ex.fired.bo8\endcsname{0.8630}
\expandafter\gdef\csname Nhi@sae.l131k.ex.fired.bo8\endcsname{0.9130}
\expandafter\gdef\csname Nn@sae.l131k.ex.fired.bo8\endcsname{512}
\expandafter\gdef\csname Nstatus@sae.l131k.ex.fired.bo8\endcsname{final}
\expandafter\gdef\csname N@sae.l131k.ex.nofire.bo8\endcsname{0.1133}
\expandafter\gdef\csname Nn@sae.l131k.ex.nofire.bo8\endcsname{512}
\expandafter\gdef\csname Nstatus@sae.l131k.ex.nofire.bo8\endcsname{final}
\expandafter\gdef\csname N@ai.l131k.c16.detrc\endcsname{0.807}
\expandafter\gdef\csname Nlo@ai.l131k.c16.detrc\endcsname{0.798}
\expandafter\gdef\csname Nhi@ai.l131k.c16.detrc\endcsname{0.815}
\expandafter\gdef\csname Nn@ai.l131k.c16.detrc\endcsname{512}
\expandafter\gdef\csname Nstatus@ai.l131k.c16.detrc\endcsname{final}
\expandafter\gdef\csname N@ai.l131k.c16.fuzzrc\endcsname{0.772}
\expandafter\gdef\csname Nlo@ai.l131k.c16.fuzzrc\endcsname{0.763}
\expandafter\gdef\csname Nhi@ai.l131k.c16.fuzzrc\endcsname{0.781}
\expandafter\gdef\csname Nn@ai.l131k.c16.fuzzrc\endcsname{512}
\expandafter\gdef\csname Nstatus@ai.l131k.c16.fuzzrc\endcsname{final}
\expandafter\gdef\csname N@ai.l131k.null.tpr\endcsname{0.675}
\expandafter\gdef\csname Nlo@ai.l131k.null.tpr\endcsname{0.656}
\expandafter\gdef\csname Nhi@ai.l131k.null.tpr\endcsname{0.693}
\expandafter\gdef\csname Nn@ai.l131k.null.tpr\endcsname{477}
\expandafter\gdef\csname Nstatus@ai.l131k.null.tpr\endcsname{final}
\expandafter\gdef\csname N@ai.l131k.null.tnr\endcsname{0.907}
\expandafter\gdef\csname Nlo@ai.l131k.null.tnr\endcsname{0.894}
\expandafter\gdef\csname Nhi@ai.l131k.null.tnr\endcsname{0.919}
\expandafter\gdef\csname Nn@ai.l131k.null.tnr\endcsname{506}
\expandafter\gdef\csname Nstatus@ai.l131k.null.tnr\endcsname{final}
\expandafter\gdef\csname N@ai.l131k.jnull.tpr\endcsname{0.706}
\expandafter\gdef\csname Nlo@ai.l131k.jnull.tpr\endcsname{0.691}
\expandafter\gdef\csname Nhi@ai.l131k.jnull.tpr\endcsname{0.721}
\expandafter\gdef\csname Nn@ai.l131k.jnull.tpr\endcsname{506}
\expandafter\gdef\csname Nstatus@ai.l131k.jnull.tpr\endcsname{final}
\expandafter\gdef\csname N@ai.l131k.jnull.tnr\endcsname{0.916}
\expandafter\gdef\csname Nlo@ai.l131k.jnull.tnr\endcsname{0.903}
\expandafter\gdef\csname Nhi@ai.l131k.jnull.tnr\endcsname{0.926}
\expandafter\gdef\csname Nn@ai.l131k.jnull.tnr\endcsname{506}
\expandafter\gdef\csname Nstatus@ai.l131k.jnull.tnr\endcsname{final}
\expandafter\gdef\csname N@ai.l131k.mcos16.tpr\endcsname{0.398}
\expandafter\gdef\csname Nlo@ai.l131k.mcos16.tpr\endcsname{0.373}
\expandafter\gdef\csname Nhi@ai.l131k.mcos16.tpr\endcsname{0.422}
\expandafter\gdef\csname Nn@ai.l131k.mcos16.tpr\endcsname{449}
\expandafter\gdef\csname Nstatus@ai.l131k.mcos16.tpr\endcsname{final}
\expandafter\gdef\csname N@ai.l131k.mcos16.tnr\endcsname{0.982}
\expandafter\gdef\csname Nlo@ai.l131k.mcos16.tnr\endcsname{0.978}
\expandafter\gdef\csname Nhi@ai.l131k.mcos16.tnr\endcsname{0.986}
\expandafter\gdef\csname Nn@ai.l131k.mcos16.tnr\endcsname{449}
\expandafter\gdef\csname Nstatus@ai.l131k.mcos16.tnr\endcsname{final}
\expandafter\gdef\csname N@ai.l131k.mcos16.detrc\endcsname{0.667}
\expandafter\gdef\csname Nlo@ai.l131k.mcos16.detrc\endcsname{0.655}
\expandafter\gdef\csname Nhi@ai.l131k.mcos16.detrc\endcsname{0.679}
\expandafter\gdef\csname Nn@ai.l131k.mcos16.detrc\endcsname{512}
\expandafter\gdef\csname Nstatus@ai.l131k.mcos16.detrc\endcsname{final}
\expandafter\gdef\csname N@ai.l131k.mcos16.fuzzrc\endcsname{0.635}
\expandafter\gdef\csname Nlo@ai.l131k.mcos16.fuzzrc\endcsname{0.624}
\expandafter\gdef\csname Nhi@ai.l131k.mcos16.fuzzrc\endcsname{0.646}
\expandafter\gdef\csname Nn@ai.l131k.mcos16.fuzzrc\endcsname{512}
\expandafter\gdef\csname Nstatus@ai.l131k.mcos16.fuzzrc\endcsname{final}
\expandafter\gdef\csname N@ai.l131k.mjac16.tpr\endcsname{0.403}
\expandafter\gdef\csname Nlo@ai.l131k.mjac16.tpr\endcsname{0.378}
\expandafter\gdef\csname Nhi@ai.l131k.mjac16.tpr\endcsname{0.428}
\expandafter\gdef\csname Nn@ai.l131k.mjac16.tpr\endcsname{444}
\expandafter\gdef\csname Nstatus@ai.l131k.mjac16.tpr\endcsname{final}
\expandafter\gdef\csname N@ai.l131k.mjac16.tnr\endcsname{0.984}
\expandafter\gdef\csname Nlo@ai.l131k.mjac16.tnr\endcsname{0.980}
\expandafter\gdef\csname Nhi@ai.l131k.mjac16.tnr\endcsname{0.988}
\expandafter\gdef\csname Nn@ai.l131k.mjac16.tnr\endcsname{444}
\expandafter\gdef\csname Nstatus@ai.l131k.mjac16.tnr\endcsname{final}
\expandafter\gdef\csname N@ai.l131k.mjac16.detrc\endcsname{0.668}
\expandafter\gdef\csname Nlo@ai.l131k.mjac16.detrc\endcsname{0.657}
\expandafter\gdef\csname Nhi@ai.l131k.mjac16.detrc\endcsname{0.680}
\expandafter\gdef\csname Nn@ai.l131k.mjac16.detrc\endcsname{512}
\expandafter\gdef\csname Nstatus@ai.l131k.mjac16.detrc\endcsname{final}
\expandafter\gdef\csname N@ai.l131k.mjac16.fuzzrc\endcsname{0.634}
\expandafter\gdef\csname Nlo@ai.l131k.mjac16.fuzzrc\endcsname{0.624}
\expandafter\gdef\csname Nhi@ai.l131k.mjac16.fuzzrc\endcsname{0.645}
\expandafter\gdef\csname Nn@ai.l131k.mjac16.fuzzrc\endcsname{512}
\expandafter\gdef\csname Nstatus@ai.l131k.mjac16.fuzzrc\endcsname{final}
\expandafter\gdef\csname N@ai.l131k.nla4.tpr\endcsname{0.373}
\expandafter\gdef\csname Nlo@ai.l131k.nla4.tpr\endcsname{0.348}
\expandafter\gdef\csname Nhi@ai.l131k.nla4.tpr\endcsname{0.398}
\expandafter\gdef\csname Nn@ai.l131k.nla4.tpr\endcsname{471}
\expandafter\gdef\csname Nstatus@ai.l131k.nla4.tpr\endcsname{final}
\expandafter\gdef\csname N@ai.l131k.nla4.tnr\endcsname{0.963}
\expandafter\gdef\csname Nlo@ai.l131k.nla4.tnr\endcsname{0.955}
\expandafter\gdef\csname Nhi@ai.l131k.nla4.tnr\endcsname{0.970}
\expandafter\gdef\csname Nn@ai.l131k.nla4.tnr\endcsname{471}
\expandafter\gdef\csname Nstatus@ai.l131k.nla4.tnr\endcsname{final}
\expandafter\gdef\csname N@ai.l131k.nla1.tpr\endcsname{0.426}
\expandafter\gdef\csname Nlo@ai.l131k.nla1.tpr\endcsname{0.400}
\expandafter\gdef\csname Nhi@ai.l131k.nla1.tpr\endcsname{0.452}
\expandafter\gdef\csname Nn@ai.l131k.nla1.tpr\endcsname{469}
\expandafter\gdef\csname Nstatus@ai.l131k.nla1.tpr\endcsname{final}
\expandafter\gdef\csname N@ai.l131k.nla1.tnr\endcsname{0.923}
\expandafter\gdef\csname Nlo@ai.l131k.nla1.tnr\endcsname{0.912}
\expandafter\gdef\csname Nhi@ai.l131k.nla1.tnr\endcsname{0.934}
\expandafter\gdef\csname Nn@ai.l131k.nla1.tnr\endcsname{469}
\expandafter\gdef\csname Nstatus@ai.l131k.nla1.tnr\endcsname{final}
\expandafter\gdef\csname N@ai.l131k.nla1.detrc\endcsname{0.660}
\expandafter\gdef\csname Nlo@ai.l131k.nla1.detrc\endcsname{0.647}
\expandafter\gdef\csname Nhi@ai.l131k.nla1.detrc\endcsname{0.673}
\expandafter\gdef\csname Nn@ai.l131k.nla1.detrc\endcsname{512}
\expandafter\gdef\csname Nstatus@ai.l131k.nla1.detrc\endcsname{final}
\expandafter\gdef\csname N@ai.l131k.nla1.fuzzrc\endcsname{0.635}
\expandafter\gdef\csname Nlo@ai.l131k.nla1.fuzzrc\endcsname{0.623}
\expandafter\gdef\csname Nhi@ai.l131k.nla1.fuzzrc\endcsname{0.647}
\expandafter\gdef\csname Nn@ai.l131k.nla1.fuzzrc\endcsname{512}
\expandafter\gdef\csname Nstatus@ai.l131k.nla1.fuzzrc\endcsname{final}
\expandafter\gdef\csname N@ai.l131k.nlab.tpr\endcsname{0.068}
\expandafter\gdef\csname Nlo@ai.l131k.nlab.tpr\endcsname{0.058}
\expandafter\gdef\csname Nhi@ai.l131k.nlab.tpr\endcsname{0.080}
\expandafter\gdef\csname Nn@ai.l131k.nlab.tpr\endcsname{503}
\expandafter\gdef\csname Nstatus@ai.l131k.nlab.tpr\endcsname{final}
\expandafter\gdef\csname N@ai.l131k.nlab.tnr\endcsname{0.997}
\expandafter\gdef\csname Nlo@ai.l131k.nlab.tnr\endcsname{0.996}
\expandafter\gdef\csname Nhi@ai.l131k.nlab.tnr\endcsname{0.998}
\expandafter\gdef\csname Nn@ai.l131k.nlab.tnr\endcsname{503}
\expandafter\gdef\csname Nstatus@ai.l131k.nlab.tnr\endcsname{final}
\expandafter\gdef\csname N@ai.l131k.floor.tpr\endcsname{0.066}
\expandafter\gdef\csname Nlo@ai.l131k.floor.tpr\endcsname{0.054}
\expandafter\gdef\csname Nhi@ai.l131k.floor.tpr\endcsname{0.081}
\expandafter\gdef\csname Nn@ai.l131k.floor.tpr\endcsname{506}
\expandafter\gdef\csname Nstatus@ai.l131k.floor.tpr\endcsname{final}
\expandafter\gdef\csname N@ai.l131k.diff.fuzz\endcsname{-0.130}
\expandafter\gdef\csname Nlo@ai.l131k.diff.fuzz\endcsname{-0.141}
\expandafter\gdef\csname Nhi@ai.l131k.diff.fuzz\endcsname{-0.119}
\expandafter\gdef\csname Nn@ai.l131k.diff.fuzz\endcsname{453}
\expandafter\gdef\csname Nstatus@ai.l131k.diff.fuzz\endcsname{final}
\expandafter\gdef\csname N@ai.l131k.diffnla.fuzz\endcsname{-0.008}
\expandafter\gdef\csname Nlo@ai.l131k.diffnla.fuzz\endcsname{-0.019}
\expandafter\gdef\csname Nhi@ai.l131k.diffnla.fuzz\endcsname{0.004}
\expandafter\gdef\csname Nn@ai.l131k.diffnla.fuzz\endcsname{422}
\expandafter\gdef\csname Nstatus@ai.l131k.diffnla.fuzz\endcsname{final}
\expandafter\gdef\csname N@ai.l131k.jnull.det.q1\endcsname{0.779}
\expandafter\gdef\csname Nlo@ai.l131k.jnull.det.q1\endcsname{0.760}
\expandafter\gdef\csname Nhi@ai.l131k.jnull.det.q1\endcsname{0.796}
\expandafter\gdef\csname Nn@ai.l131k.jnull.det.q1\endcsname{128}
\expandafter\gdef\csname Nstatus@ai.l131k.jnull.det.q1\endcsname{final}
\expandafter\gdef\csname N@ai.l131k.jnull.det.q2\endcsname{0.835}
\expandafter\gdef\csname Nlo@ai.l131k.jnull.det.q2\endcsname{0.824}
\expandafter\gdef\csname Nhi@ai.l131k.jnull.det.q2\endcsname{0.846}
\expandafter\gdef\csname Nn@ai.l131k.jnull.det.q2\endcsname{126}
\expandafter\gdef\csname Nstatus@ai.l131k.jnull.det.q2\endcsname{final}
\expandafter\gdef\csname N@ai.l131k.jnull.det.q3\endcsname{0.838}
\expandafter\gdef\csname Nlo@ai.l131k.jnull.det.q3\endcsname{0.823}
\expandafter\gdef\csname Nhi@ai.l131k.jnull.det.q3\endcsname{0.853}
\expandafter\gdef\csname Nn@ai.l131k.jnull.det.q3\endcsname{127}
\expandafter\gdef\csname Nstatus@ai.l131k.jnull.det.q3\endcsname{final}
\expandafter\gdef\csname N@ai.l131k.jnull.det.q4\endcsname{0.793}
\expandafter\gdef\csname Nlo@ai.l131k.jnull.det.q4\endcsname{0.772}
\expandafter\gdef\csname Nhi@ai.l131k.jnull.det.q4\endcsname{0.812}
\expandafter\gdef\csname Nn@ai.l131k.jnull.det.q4\endcsname{125}
\expandafter\gdef\csname Nstatus@ai.l131k.jnull.det.q4\endcsname{final}
\expandafter\gdef\csname N@ai.l131k.mcos16.det.q1\endcsname{0.610}
\expandafter\gdef\csname Nlo@ai.l131k.mcos16.det.q1\endcsname{0.587}
\expandafter\gdef\csname Nhi@ai.l131k.mcos16.det.q1\endcsname{0.633}
\expandafter\gdef\csname Nn@ai.l131k.mcos16.det.q1\endcsname{98}
\expandafter\gdef\csname Nstatus@ai.l131k.mcos16.det.q1\endcsname{final}
\expandafter\gdef\csname N@ai.l131k.mcos16.det.q2\endcsname{0.718}
\expandafter\gdef\csname Nlo@ai.l131k.mcos16.det.q2\endcsname{0.697}
\expandafter\gdef\csname Nhi@ai.l131k.mcos16.det.q2\endcsname{0.739}
\expandafter\gdef\csname Nn@ai.l131k.mcos16.det.q2\endcsname{118}
\expandafter\gdef\csname Nstatus@ai.l131k.mcos16.det.q2\endcsname{final}
\expandafter\gdef\csname N@ai.l131k.mcos16.det.q3\endcsname{0.739}
\expandafter\gdef\csname Nlo@ai.l131k.mcos16.det.q3\endcsname{0.717}
\expandafter\gdef\csname Nhi@ai.l131k.mcos16.det.q3\endcsname{0.761}
\expandafter\gdef\csname Nn@ai.l131k.mcos16.det.q3\endcsname{117}
\expandafter\gdef\csname Nstatus@ai.l131k.mcos16.det.q3\endcsname{final}
\expandafter\gdef\csname N@ai.l131k.mcos16.det.q4\endcsname{0.681}
\expandafter\gdef\csname Nlo@ai.l131k.mcos16.det.q4\endcsname{0.656}
\expandafter\gdef\csname Nhi@ai.l131k.mcos16.det.q4\endcsname{0.706}
\expandafter\gdef\csname Nn@ai.l131k.mcos16.det.q4\endcsname{116}
\expandafter\gdef\csname Nstatus@ai.l131k.mcos16.det.q4\endcsname{final}
\expandafter\gdef\csname N@ai.l131k.mjac16.det.q1\endcsname{0.625}
\expandafter\gdef\csname Nlo@ai.l131k.mjac16.det.q1\endcsname{0.600}
\expandafter\gdef\csname Nhi@ai.l131k.mjac16.det.q1\endcsname{0.648}
\expandafter\gdef\csname Nn@ai.l131k.mjac16.det.q1\endcsname{93}
\expandafter\gdef\csname Nstatus@ai.l131k.mjac16.det.q1\endcsname{final}
\expandafter\gdef\csname N@ai.l131k.mjac16.det.q2\endcsname{0.708}
\expandafter\gdef\csname Nlo@ai.l131k.mjac16.det.q2\endcsname{0.688}
\expandafter\gdef\csname Nhi@ai.l131k.mjac16.det.q2\endcsname{0.729}
\expandafter\gdef\csname Nn@ai.l131k.mjac16.det.q2\endcsname{118}
\expandafter\gdef\csname Nstatus@ai.l131k.mjac16.det.q2\endcsname{final}
\expandafter\gdef\csname N@ai.l131k.mjac16.det.q3\endcsname{0.744}
\expandafter\gdef\csname Nlo@ai.l131k.mjac16.det.q3\endcsname{0.721}
\expandafter\gdef\csname Nhi@ai.l131k.mjac16.det.q3\endcsname{0.765}
\expandafter\gdef\csname Nn@ai.l131k.mjac16.det.q3\endcsname{118}
\expandafter\gdef\csname Nstatus@ai.l131k.mjac16.det.q3\endcsname{final}
\expandafter\gdef\csname N@ai.l131k.mjac16.det.q4\endcsname{0.683}
\expandafter\gdef\csname Nlo@ai.l131k.mjac16.det.q4\endcsname{0.657}
\expandafter\gdef\csname Nhi@ai.l131k.mjac16.det.q4\endcsname{0.710}
\expandafter\gdef\csname Nn@ai.l131k.mjac16.det.q4\endcsname{115}
\expandafter\gdef\csname Nstatus@ai.l131k.mjac16.det.q4\endcsname{final}
\expandafter\gdef\csname N@ai.l131k.nla4.det.q1\endcsname{0.560}
\expandafter\gdef\csname Nlo@ai.l131k.nla4.det.q1\endcsname{0.542}
\expandafter\gdef\csname Nhi@ai.l131k.nla4.det.q1\endcsname{0.579}
\expandafter\gdef\csname Nn@ai.l131k.nla4.det.q1\endcsname{122}
\expandafter\gdef\csname Nstatus@ai.l131k.nla4.det.q1\endcsname{final}
\expandafter\gdef\csname N@ai.l131k.nla4.det.q2\endcsname{0.700}
\expandafter\gdef\csname Nlo@ai.l131k.nla4.det.q2\endcsname{0.675}
\expandafter\gdef\csname Nhi@ai.l131k.nla4.det.q2\endcsname{0.724}
\expandafter\gdef\csname Nn@ai.l131k.nla4.det.q2\endcsname{118}
\expandafter\gdef\csname Nstatus@ai.l131k.nla4.det.q2\endcsname{final}
\expandafter\gdef\csname N@ai.l131k.nla4.det.q3\endcsname{0.725}
\expandafter\gdef\csname Nlo@ai.l131k.nla4.det.q3\endcsname{0.702}
\expandafter\gdef\csname Nhi@ai.l131k.nla4.det.q3\endcsname{0.749}
\expandafter\gdef\csname Nn@ai.l131k.nla4.det.q3\endcsname{112}
\expandafter\gdef\csname Nstatus@ai.l131k.nla4.det.q3\endcsname{final}
\expandafter\gdef\csname N@ai.l131k.nla4.det.q4\endcsname{0.694}
\expandafter\gdef\csname Nlo@ai.l131k.nla4.det.q4\endcsname{0.669}
\expandafter\gdef\csname Nhi@ai.l131k.nla4.det.q4\endcsname{0.717}
\expandafter\gdef\csname Nn@ai.l131k.nla4.det.q4\endcsname{119}
\expandafter\gdef\csname Nstatus@ai.l131k.nla4.det.q4\endcsname{final}
\expandafter\gdef\csname N@ai.l131k.nla1.det.q1\endcsname{0.574}
\expandafter\gdef\csname Nlo@ai.l131k.nla1.det.q1\endcsname{0.554}
\expandafter\gdef\csname Nhi@ai.l131k.nla1.det.q1\endcsname{0.596}
\expandafter\gdef\csname Nn@ai.l131k.nla1.det.q1\endcsname{123}
\expandafter\gdef\csname Nstatus@ai.l131k.nla1.det.q1\endcsname{final}
\expandafter\gdef\csname N@ai.l131k.nla1.det.q2\endcsname{0.704}
\expandafter\gdef\csname Nlo@ai.l131k.nla1.det.q2\endcsname{0.679}
\expandafter\gdef\csname Nhi@ai.l131k.nla1.det.q2\endcsname{0.730}
\expandafter\gdef\csname Nn@ai.l131k.nla1.det.q2\endcsname{118}
\expandafter\gdef\csname Nstatus@ai.l131k.nla1.det.q2\endcsname{final}
\expandafter\gdef\csname N@ai.l131k.nla1.det.q3\endcsname{0.733}
\expandafter\gdef\csname Nlo@ai.l131k.nla1.det.q3\endcsname{0.706}
\expandafter\gdef\csname Nhi@ai.l131k.nla1.det.q3\endcsname{0.761}
\expandafter\gdef\csname Nn@ai.l131k.nla1.det.q3\endcsname{113}
\expandafter\gdef\csname Nstatus@ai.l131k.nla1.det.q3\endcsname{final}
\expandafter\gdef\csname N@ai.l131k.nla1.det.q4\endcsname{0.695}
\expandafter\gdef\csname Nlo@ai.l131k.nla1.det.q4\endcsname{0.669}
\expandafter\gdef\csname Nhi@ai.l131k.nla1.det.q4\endcsname{0.723}
\expandafter\gdef\csname Nn@ai.l131k.nla1.det.q4\endcsname{115}
\expandafter\gdef\csname Nstatus@ai.l131k.nla1.det.q4\endcsname{final}
\expandafter\gdef\csname N@ai.l131k.nlab.det.q1\endcsname{0.516}
\expandafter\gdef\csname Nlo@ai.l131k.nlab.det.q1\endcsname{0.510}
\expandafter\gdef\csname Nhi@ai.l131k.nlab.det.q1\endcsname{0.523}
\expandafter\gdef\csname Nn@ai.l131k.nlab.det.q1\endcsname{128}
\expandafter\gdef\csname Nstatus@ai.l131k.nlab.det.q1\endcsname{final}
\expandafter\gdef\csname N@ai.l131k.nlab.det.q2\endcsname{0.546}
\expandafter\gdef\csname Nlo@ai.l131k.nlab.det.q2\endcsname{0.535}
\expandafter\gdef\csname Nhi@ai.l131k.nlab.det.q2\endcsname{0.558}
\expandafter\gdef\csname Nn@ai.l131k.nlab.det.q2\endcsname{125}
\expandafter\gdef\csname Nstatus@ai.l131k.nlab.det.q2\endcsname{final}
\expandafter\gdef\csname N@ai.l131k.nlab.det.q3\endcsname{0.538}
\expandafter\gdef\csname Nlo@ai.l131k.nlab.det.q3\endcsname{0.527}
\expandafter\gdef\csname Nhi@ai.l131k.nlab.det.q3\endcsname{0.550}
\expandafter\gdef\csname Nn@ai.l131k.nlab.det.q3\endcsname{125}
\expandafter\gdef\csname Nstatus@ai.l131k.nlab.det.q3\endcsname{final}
\expandafter\gdef\csname N@ai.l131k.nlab.det.q4\endcsname{0.531}
\expandafter\gdef\csname Nlo@ai.l131k.nlab.det.q4\endcsname{0.521}
\expandafter\gdef\csname Nhi@ai.l131k.nlab.det.q4\endcsname{0.542}
\expandafter\gdef\csname Nn@ai.l131k.nlab.det.q4\endcsname{125}
\expandafter\gdef\csname Nstatus@ai.l131k.nlab.det.q4\endcsname{final}
\expandafter\gdef\csname N@ai.l131k.floor.det.q1\endcsname{0.509}
\expandafter\gdef\csname Nlo@ai.l131k.floor.det.q1\endcsname{0.502}
\expandafter\gdef\csname Nhi@ai.l131k.floor.det.q1\endcsname{0.518}
\expandafter\gdef\csname Nn@ai.l131k.floor.det.q1\endcsname{125}
\expandafter\gdef\csname Nstatus@ai.l131k.floor.det.q1\endcsname{final}
\expandafter\gdef\csname N@ai.l131k.floor.det.q2\endcsname{0.500}
\expandafter\gdef\csname Nlo@ai.l131k.floor.det.q2\endcsname{0.495}
\expandafter\gdef\csname Nhi@ai.l131k.floor.det.q2\endcsname{0.505}
\expandafter\gdef\csname Nn@ai.l131k.floor.det.q2\endcsname{126}
\expandafter\gdef\csname Nstatus@ai.l131k.floor.det.q2\endcsname{final}
\expandafter\gdef\csname N@ai.l131k.floor.det.q3\endcsname{0.502}
\expandafter\gdef\csname Nlo@ai.l131k.floor.det.q3\endcsname{0.494}
\expandafter\gdef\csname Nhi@ai.l131k.floor.det.q3\endcsname{0.510}
\expandafter\gdef\csname Nn@ai.l131k.floor.det.q3\endcsname{127}
\expandafter\gdef\csname Nstatus@ai.l131k.floor.det.q3\endcsname{final}
\expandafter\gdef\csname N@ai.l131k.floor.det.q4\endcsname{0.504}
\expandafter\gdef\csname Nlo@ai.l131k.floor.det.q4\endcsname{0.497}
\expandafter\gdef\csname Nhi@ai.l131k.floor.det.q4\endcsname{0.510}
\expandafter\gdef\csname Nn@ai.l131k.floor.det.q4\endcsname{128}
\expandafter\gdef\csname Nstatus@ai.l131k.floor.det.q4\endcsname{final}
\expandafter\gdef\csname N@sae.l131k.ex.win.bo1\endcsname{0.1191}
\expandafter\gdef\csname Nn@sae.l131k.ex.win.bo1\endcsname{512}
\expandafter\gdef\csname Nstatus@sae.l131k.ex.win.bo1\endcsname{final}
\expandafter\gdef\csname N@sae.l131k.ex.win.bo8\endcsname{0.2988}
\expandafter\gdef\csname Nn@sae.l131k.ex.win.bo8\endcsname{512}
\expandafter\gdef\csname Nstatus@sae.l131k.ex.win.bo8\endcsname{final}
\expandafter\gdef\csname N@sae.l131k.ex.win.bo64\endcsname{0.4609}
\expandafter\gdef\csname Nn@sae.l131k.ex.win.bo64\endcsname{512}
\expandafter\gdef\csname Nstatus@sae.l131k.ex.win.bo64\endcsname{final}
\expandafter\gdef\csname N@sae.l131k.ex.neverfire\endcsname{0.0645}
\expandafter\gdef\csname Nn@sae.l131k.ex.neverfire\endcsname{512}
\expandafter\gdef\csname Nstatus@sae.l131k.ex.neverfire\endcsname{final}
\expandafter\gdef\csname N@ood.ces.ex.vspan\endcsname{0.0020}
\expandafter\gdef\csname Nn@ood.ces.ex.vspan\endcsname{512}
\expandafter\gdef\csname Nstatus@ood.ces.ex.vspan\endcsname{final}
\expandafter\gdef\csname N@ood.rus.ex.vspan\endcsname{0.0098}
\expandafter\gdef\csname Nn@ood.rus.ex.vspan\endcsname{512}
\expandafter\gdef\csname Nstatus@ood.rus.ex.vspan\endcsname{final}
\expandafter\gdef\csname N@ood.ell.ex.vspan\endcsname{0.0039}
\expandafter\gdef\csname Nn@ood.ell.ex.vspan\endcsname{512}
\expandafter\gdef\csname Nstatus@ood.ell.ex.vspan\endcsname{final}
\expandafter\gdef\csname N@ood.arb.ex.vspan\endcsname{0.0098}
\expandafter\gdef\csname Nn@ood.arb.ex.vspan\endcsname{512}
\expandafter\gdef\csname Nstatus@ood.arb.ex.vspan\endcsname{final}
\expandafter\gdef\csname N@ood.hin.ex.vspan\endcsname{0.0039}
\expandafter\gdef\csname Nn@ood.hin.ex.vspan\endcsname{512}
\expandafter\gdef\csname Nstatus@ood.hin.ex.vspan\endcsname{final}
\expandafter\gdef\csname N@ood.tha.ex.vspan\endcsname{0.0195}
\expandafter\gdef\csname Nn@ood.tha.ex.vspan\endcsname{512}
\expandafter\gdef\csname Nstatus@ood.tha.ex.vspan\endcsname{final}
\expandafter\gdef\csname N@ood.cmn.ex.vspan\endcsname{0.0137}
\expandafter\gdef\csname Nn@ood.cmn.ex.vspan\endcsname{512}
\expandafter\gdef\csname Nstatus@ood.cmn.ex.vspan\endcsname{final}
\expandafter\gdef\csname N@ood.jpn.ex.vspan\endcsname{0.0059}
\expandafter\gdef\csname Nn@ood.jpn.ex.vspan\endcsname{512}
\expandafter\gdef\csname Nstatus@ood.jpn.ex.vspan\endcsname{final}
\expandafter\gdef\csname N@ood.zh.ex.vspan\endcsname{0.0059}
\expandafter\gdef\csname Nn@ood.zh.ex.vspan\endcsname{512}
\expandafter\gdef\csname Nstatus@ood.zh.ex.vspan\endcsname{final}
\expandafter\gdef\csname N@ood.en.ex.vspan\endcsname{0.0039}
\expandafter\gdef\csname Nn@ood.en.ex.vspan\endcsname{512}
\expandafter\gdef\csname Nstatus@ood.en.ex.vspan\endcsname{final}
\expandafter\gdef\csname N@ood.python.ex.vspan\endcsname{0.0625}
\expandafter\gdef\csname Nn@ood.python.ex.vspan\endcsname{512}
\expandafter\gdef\csname Nstatus@ood.python.ex.vspan\endcsname{final}
\expandafter\gdef\csname N@ood.javascript.ex.vspan\endcsname{0.0410}
\expandafter\gdef\csname Nn@ood.javascript.ex.vspan\endcsname{512}
\expandafter\gdef\csname Nstatus@ood.javascript.ex.vspan\endcsname{final}
\expandafter\gdef\csname N@ood.c.ex.vspan\endcsname{0.1152}
\expandafter\gdef\csname Nn@ood.c.ex.vspan\endcsname{512}
\expandafter\gdef\csname Nstatus@ood.c.ex.vspan\endcsname{final}
\expandafter\gdef\csname N@ood.rust.ex.vspan\endcsname{0.0684}
\expandafter\gdef\csname Nn@ood.rust.ex.vspan\endcsname{512}
\expandafter\gdef\csname Nstatus@ood.rust.ex.vspan\endcsname{final}
\expandafter\gdef\csname N@ood.go.ex.vspan\endcsname{0.0977}
\expandafter\gdef\csname Nn@ood.go.ex.vspan\endcsname{512}
\expandafter\gdef\csname Nstatus@ood.go.ex.vspan\endcsname{final}
\expandafter\gdef\csname N@ood.haskell.ex.vspan\endcsname{0.0410}
\expandafter\gdef\csname Nn@ood.haskell.ex.vspan\endcsname{512}
\expandafter\gdef\csname Nstatus@ood.haskell.ex.vspan\endcsname{final}
\expandafter\gdef\csname N@ood.sql.ex.vspan\endcsname{0.0957}
\expandafter\gdef\csname Nn@ood.sql.ex.vspan\endcsname{512}
\expandafter\gdef\csname Nstatus@ood.sql.ex.vspan\endcsname{final}
\expandafter\gdef\csname N@ood.shell.ex.vspan\endcsname{0.1016}
\expandafter\gdef\csname Nn@ood.shell.ex.vspan\endcsname{512}
\expandafter\gdef\csname Nstatus@ood.shell.ex.vspan\endcsname{final}
\expandafter\gdef\csname N@ood.owm.ex.vspan\endcsname{0.0059}
\expandafter\gdef\csname Nn@ood.owm.ex.vspan\endcsname{512}
\expandafter\gdef\csname Nstatus@ood.owm.ex.vspan\endcsname{final}
\expandafter\gdef\csname N@ood.arxiv.ex.vspan\endcsname{0.0020}
\expandafter\gdef\csname Nn@ood.arxiv.ex.vspan\endcsname{512}
\expandafter\gdef\csname Nstatus@ood.arxiv.ex.vspan\endcsname{final}
\expandafter\gdef\csname N@ood.lean.ex.vspan\endcsname{0.3066}
\expandafter\gdef\csname Nn@ood.lean.ex.vspan\endcsname{512}
\expandafter\gdef\csname Nstatus@ood.lean.ex.vspan\endcsname{final}
\expandafter\gdef\csname N@ood.isabelle.ex.vspan\endcsname{0.3770}
\expandafter\gdef\csname Nn@ood.isabelle.ex.vspan\endcsname{512}
\expandafter\gdef\csname Nstatus@ood.isabelle.ex.vspan\endcsname{final}
\expandafter\gdef\csname N@ai.l131k.c16.tnr.b0\endcsname{0.912}
\expandafter\gdef\csname Nlo@ai.l131k.c16.tnr.b0\endcsname{0.900}
\expandafter\gdef\csname Nhi@ai.l131k.c16.tnr.b0\endcsname{0.923}
\expandafter\gdef\csname Nn@ai.l131k.c16.tnr.b0\endcsname{506}
\expandafter\gdef\csname Nstatus@ai.l131k.c16.tnr.b0\endcsname{final}
\expandafter\gdef\csname N@ai.l131k.c16.tpr.btop\endcsname{0.877}
\expandafter\gdef\csname Nlo@ai.l131k.c16.tpr.btop\endcsname{0.832}
\expandafter\gdef\csname Nhi@ai.l131k.c16.tpr.btop\endcsname{0.918}
\expandafter\gdef\csname Nn@ai.l131k.c16.tpr.btop\endcsname{125}
\expandafter\gdef\csname Nstatus@ai.l131k.c16.tpr.btop\endcsname{final}
\expandafter\gdef\csname N@ai.l131k.c16.fuzztnr.b0\endcsname{0.980}
\expandafter\gdef\csname Nlo@ai.l131k.c16.fuzztnr.b0\endcsname{0.974}
\expandafter\gdef\csname Nhi@ai.l131k.c16.fuzztnr.b0\endcsname{0.985}
\expandafter\gdef\csname Nn@ai.l131k.c16.fuzztnr.b0\endcsname{506}
\expandafter\gdef\csname Nstatus@ai.l131k.c16.fuzztnr.b0\endcsname{final}
\expandafter\gdef\csname N@ai.l131k.c16.fuzztpr.b1\endcsname{0.286}
\expandafter\gdef\csname Nlo@ai.l131k.c16.fuzztpr.b1\endcsname{0.262}
\expandafter\gdef\csname Nhi@ai.l131k.c16.fuzztpr.b1\endcsname{0.312}
\expandafter\gdef\csname Nn@ai.l131k.c16.fuzztpr.b1\endcsname{487}
\expandafter\gdef\csname Nstatus@ai.l131k.c16.fuzztpr.b1\endcsname{final}
\expandafter\gdef\csname N@ai.l131k.c16.fuzztpr.b2\endcsname{0.548}
\expandafter\gdef\csname Nlo@ai.l131k.c16.fuzztpr.b2\endcsname{0.522}
\expandafter\gdef\csname Nhi@ai.l131k.c16.fuzztpr.b2\endcsname{0.573}
\expandafter\gdef\csname Nn@ai.l131k.c16.fuzztpr.b2\endcsname{500}
\expandafter\gdef\csname Nstatus@ai.l131k.c16.fuzztpr.b2\endcsname{final}
\expandafter\gdef\csname N@ai.l131k.c16.fuzztpr.b3\endcsname{0.686}
\expandafter\gdef\csname Nlo@ai.l131k.c16.fuzztpr.b3\endcsname{0.660}
\expandafter\gdef\csname Nhi@ai.l131k.c16.fuzztpr.b3\endcsname{0.713}
\expandafter\gdef\csname Nn@ai.l131k.c16.fuzztpr.b3\endcsname{476}
\expandafter\gdef\csname Nstatus@ai.l131k.c16.fuzztpr.b3\endcsname{final}
\expandafter\gdef\csname N@ai.l131k.c16.fuzztpr.b4\endcsname{0.751}
\expandafter\gdef\csname Nlo@ai.l131k.c16.fuzztpr.b4\endcsname{0.726}
\expandafter\gdef\csname Nhi@ai.l131k.c16.fuzztpr.b4\endcsname{0.776}
\expandafter\gdef\csname Nn@ai.l131k.c16.fuzztpr.b4\endcsname{505}
\expandafter\gdef\csname Nstatus@ai.l131k.c16.fuzztpr.b4\endcsname{final}
\expandafter\gdef\csname N@ai.l131k.c16.fuzztpr.btop\endcsname{0.741}
\expandafter\gdef\csname Nlo@ai.l131k.c16.fuzztpr.btop\endcsname{0.685}
\expandafter\gdef\csname Nhi@ai.l131k.c16.fuzztpr.btop\endcsname{0.791}
\expandafter\gdef\csname Nn@ai.l131k.c16.fuzztpr.btop\endcsname{125}
\expandafter\gdef\csname Nstatus@ai.l131k.c16.fuzztpr.btop\endcsname{final}
\expandafter\gdef\csname N@ai.l131k.null.tnr.b0\endcsname{0.907}
\expandafter\gdef\csname Nlo@ai.l131k.null.tnr.b0\endcsname{0.894}
\expandafter\gdef\csname Nhi@ai.l131k.null.tnr.b0\endcsname{0.919}
\expandafter\gdef\csname Nn@ai.l131k.null.tnr.b0\endcsname{506}
\expandafter\gdef\csname Nstatus@ai.l131k.null.tnr.b0\endcsname{final}
\expandafter\gdef\csname N@ai.l131k.null.tpr.b1\endcsname{0.418}
\expandafter\gdef\csname Nlo@ai.l131k.null.tpr.b1\endcsname{0.388}
\expandafter\gdef\csname Nhi@ai.l131k.null.tpr.b1\endcsname{0.449}
\expandafter\gdef\csname Nn@ai.l131k.null.tpr.b1\endcsname{338}
\expandafter\gdef\csname Nstatus@ai.l131k.null.tpr.b1\endcsname{final}
\expandafter\gdef\csname N@ai.l131k.null.tpr.b2\endcsname{0.629}
\expandafter\gdef\csname Nlo@ai.l131k.null.tpr.b2\endcsname{0.601}
\expandafter\gdef\csname Nhi@ai.l131k.null.tpr.b2\endcsname{0.657}
\expandafter\gdef\csname Nn@ai.l131k.null.tpr.b2\endcsname{415}
\expandafter\gdef\csname Nstatus@ai.l131k.null.tpr.b2\endcsname{final}
\expandafter\gdef\csname N@ai.l131k.null.tpr.b3\endcsname{0.820}
\expandafter\gdef\csname Nlo@ai.l131k.null.tpr.b3\endcsname{0.795}
\expandafter\gdef\csname Nhi@ai.l131k.null.tpr.b3\endcsname{0.844}
\expandafter\gdef\csname Nn@ai.l131k.null.tpr.b3\endcsname{346}
\expandafter\gdef\csname Nstatus@ai.l131k.null.tpr.b3\endcsname{final}
\expandafter\gdef\csname N@ai.l131k.null.tpr.b4\endcsname{0.909}
\expandafter\gdef\csname Nlo@ai.l131k.null.tpr.b4\endcsname{0.886}
\expandafter\gdef\csname Nhi@ai.l131k.null.tpr.b4\endcsname{0.931}
\expandafter\gdef\csname Nn@ai.l131k.null.tpr.b4\endcsname{269}
\expandafter\gdef\csname Nstatus@ai.l131k.null.tpr.b4\endcsname{final}
\expandafter\gdef\csname N@ai.l131k.null.tpr.btop\endcsname{0.856}
\expandafter\gdef\csname Nlo@ai.l131k.null.tpr.btop\endcsname{0.821}
\expandafter\gdef\csname Nhi@ai.l131k.null.tpr.btop\endcsname{0.890}
\expandafter\gdef\csname Nn@ai.l131k.null.tpr.btop\endcsname{176}
\expandafter\gdef\csname Nstatus@ai.l131k.null.tpr.btop\endcsname{final}
\expandafter\gdef\csname N@ai.l131k.null.fuzztnr.b0\endcsname{0.978}
\expandafter\gdef\csname Nlo@ai.l131k.null.fuzztnr.b0\endcsname{0.972}
\expandafter\gdef\csname Nhi@ai.l131k.null.fuzztnr.b0\endcsname{0.984}
\expandafter\gdef\csname Nn@ai.l131k.null.fuzztnr.b0\endcsname{506}
\expandafter\gdef\csname Nstatus@ai.l131k.null.fuzztnr.b0\endcsname{final}
\expandafter\gdef\csname N@ai.l131k.null.fuzztpr.b1\endcsname{0.309}
\expandafter\gdef\csname Nlo@ai.l131k.null.fuzztpr.b1\endcsname{0.281}
\expandafter\gdef\csname Nhi@ai.l131k.null.fuzztpr.b1\endcsname{0.338}
\expandafter\gdef\csname Nn@ai.l131k.null.fuzztpr.b1\endcsname{338}
\expandafter\gdef\csname Nstatus@ai.l131k.null.fuzztpr.b1\endcsname{final}
\expandafter\gdef\csname N@ai.l131k.null.fuzztpr.b2\endcsname{0.520}
\expandafter\gdef\csname Nlo@ai.l131k.null.fuzztpr.b2\endcsname{0.491}
\expandafter\gdef\csname Nhi@ai.l131k.null.fuzztpr.b2\endcsname{0.546}
\expandafter\gdef\csname Nn@ai.l131k.null.fuzztpr.b2\endcsname{415}
\expandafter\gdef\csname Nstatus@ai.l131k.null.fuzztpr.b2\endcsname{final}
\expandafter\gdef\csname N@ai.l131k.null.fuzztpr.b3\endcsname{0.665}
\expandafter\gdef\csname Nlo@ai.l131k.null.fuzztpr.b3\endcsname{0.636}
\expandafter\gdef\csname Nhi@ai.l131k.null.fuzztpr.b3\endcsname{0.693}
\expandafter\gdef\csname Nn@ai.l131k.null.fuzztpr.b3\endcsname{346}
\expandafter\gdef\csname Nstatus@ai.l131k.null.fuzztpr.b3\endcsname{final}
\expandafter\gdef\csname N@ai.l131k.null.fuzztpr.b4\endcsname{0.752}
\expandafter\gdef\csname Nlo@ai.l131k.null.fuzztpr.b4\endcsname{0.718}
\expandafter\gdef\csname Nhi@ai.l131k.null.fuzztpr.b4\endcsname{0.784}
\expandafter\gdef\csname Nn@ai.l131k.null.fuzztpr.b4\endcsname{269}
\expandafter\gdef\csname Nstatus@ai.l131k.null.fuzztpr.b4\endcsname{final}
\expandafter\gdef\csname N@ai.l131k.null.fuzztpr.btop\endcsname{0.737}
\expandafter\gdef\csname Nlo@ai.l131k.null.fuzztpr.btop\endcsname{0.691}
\expandafter\gdef\csname Nhi@ai.l131k.null.fuzztpr.btop\endcsname{0.781}
\expandafter\gdef\csname Nn@ai.l131k.null.fuzztpr.btop\endcsname{176}
\expandafter\gdef\csname Nstatus@ai.l131k.null.fuzztpr.btop\endcsname{final}
\expandafter\gdef\csname N@ai.l131k.jnull.tnr.b0\endcsname{0.916}
\expandafter\gdef\csname Nlo@ai.l131k.jnull.tnr.b0\endcsname{0.904}
\expandafter\gdef\csname Nhi@ai.l131k.jnull.tnr.b0\endcsname{0.926}
\expandafter\gdef\csname Nn@ai.l131k.jnull.tnr.b0\endcsname{506}
\expandafter\gdef\csname Nstatus@ai.l131k.jnull.tnr.b0\endcsname{final}
\expandafter\gdef\csname N@ai.l131k.jnull.tpr.b1\endcsname{0.371}
\expandafter\gdef\csname Nlo@ai.l131k.jnull.tpr.b1\endcsname{0.343}
\expandafter\gdef\csname Nhi@ai.l131k.jnull.tpr.b1\endcsname{0.399}
\expandafter\gdef\csname Nn@ai.l131k.jnull.tpr.b1\endcsname{487}
\expandafter\gdef\csname Nstatus@ai.l131k.jnull.tpr.b1\endcsname{final}
\expandafter\gdef\csname N@ai.l131k.jnull.tpr.b2\endcsname{0.660}
\expandafter\gdef\csname Nlo@ai.l131k.jnull.tpr.b2\endcsname{0.635}
\expandafter\gdef\csname Nhi@ai.l131k.jnull.tpr.b2\endcsname{0.686}
\expandafter\gdef\csname Nn@ai.l131k.jnull.tpr.b2\endcsname{500}
\expandafter\gdef\csname Nstatus@ai.l131k.jnull.tpr.b2\endcsname{final}
\expandafter\gdef\csname N@ai.l131k.jnull.tpr.b3\endcsname{0.840}
\expandafter\gdef\csname Nlo@ai.l131k.jnull.tpr.b3\endcsname{0.818}
\expandafter\gdef\csname Nhi@ai.l131k.jnull.tpr.b3\endcsname{0.860}
\expandafter\gdef\csname Nn@ai.l131k.jnull.tpr.b3\endcsname{476}
\expandafter\gdef\csname Nstatus@ai.l131k.jnull.tpr.b3\endcsname{final}
\expandafter\gdef\csname N@ai.l131k.jnull.tpr.b4\endcsname{0.922}
\expandafter\gdef\csname Nlo@ai.l131k.jnull.tpr.b4\endcsname{0.906}
\expandafter\gdef\csname Nhi@ai.l131k.jnull.tpr.b4\endcsname{0.936}
\expandafter\gdef\csname Nn@ai.l131k.jnull.tpr.b4\endcsname{506}
\expandafter\gdef\csname Nstatus@ai.l131k.jnull.tpr.b4\endcsname{final}
\expandafter\gdef\csname N@ai.l131k.jnull.tpr.btop\endcsname{0.862}
\expandafter\gdef\csname Nlo@ai.l131k.jnull.tpr.btop\endcsname{0.815}
\expandafter\gdef\csname Nhi@ai.l131k.jnull.tpr.btop\endcsname{0.905}
\expandafter\gdef\csname Nn@ai.l131k.jnull.tpr.btop\endcsname{125}
\expandafter\gdef\csname Nstatus@ai.l131k.jnull.tpr.btop\endcsname{final}
\expandafter\gdef\csname N@ai.l131k.jnull.fuzztnr.b0\endcsname{0.979}
\expandafter\gdef\csname Nlo@ai.l131k.jnull.fuzztnr.b0\endcsname{0.973}
\expandafter\gdef\csname Nhi@ai.l131k.jnull.fuzztnr.b0\endcsname{0.985}
\expandafter\gdef\csname Nn@ai.l131k.jnull.fuzztnr.b0\endcsname{506}
\expandafter\gdef\csname Nstatus@ai.l131k.jnull.fuzztnr.b0\endcsname{final}
\expandafter\gdef\csname N@ai.l131k.jnull.fuzztpr.b1\endcsname{0.291}
\expandafter\gdef\csname Nlo@ai.l131k.jnull.fuzztpr.b1\endcsname{0.265}
\expandafter\gdef\csname Nhi@ai.l131k.jnull.fuzztpr.b1\endcsname{0.318}
\expandafter\gdef\csname Nn@ai.l131k.jnull.fuzztpr.b1\endcsname{487}
\expandafter\gdef\csname Nstatus@ai.l131k.jnull.fuzztpr.b1\endcsname{final}
\expandafter\gdef\csname N@ai.l131k.jnull.fuzztpr.b2\endcsname{0.537}
\expandafter\gdef\csname Nlo@ai.l131k.jnull.fuzztpr.b2\endcsname{0.510}
\expandafter\gdef\csname Nhi@ai.l131k.jnull.fuzztpr.b2\endcsname{0.563}
\expandafter\gdef\csname Nn@ai.l131k.jnull.fuzztpr.b2\endcsname{500}
\expandafter\gdef\csname Nstatus@ai.l131k.jnull.fuzztpr.b2\endcsname{final}
\expandafter\gdef\csname N@ai.l131k.jnull.fuzztpr.b3\endcsname{0.695}
\expandafter\gdef\csname Nlo@ai.l131k.jnull.fuzztpr.b3\endcsname{0.669}
\expandafter\gdef\csname Nhi@ai.l131k.jnull.fuzztpr.b3\endcsname{0.721}
\expandafter\gdef\csname Nn@ai.l131k.jnull.fuzztpr.b3\endcsname{476}
\expandafter\gdef\csname Nstatus@ai.l131k.jnull.fuzztpr.b3\endcsname{final}
\expandafter\gdef\csname N@ai.l131k.jnull.fuzztpr.b4\endcsname{0.746}
\expandafter\gdef\csname Nlo@ai.l131k.jnull.fuzztpr.b4\endcsname{0.721}
\expandafter\gdef\csname Nhi@ai.l131k.jnull.fuzztpr.b4\endcsname{0.771}
\expandafter\gdef\csname Nn@ai.l131k.jnull.fuzztpr.b4\endcsname{506}
\expandafter\gdef\csname Nstatus@ai.l131k.jnull.fuzztpr.b4\endcsname{final}
\expandafter\gdef\csname N@ai.l131k.jnull.fuzztpr.btop\endcsname{0.726}
\expandafter\gdef\csname Nlo@ai.l131k.jnull.fuzztpr.btop\endcsname{0.669}
\expandafter\gdef\csname Nhi@ai.l131k.jnull.fuzztpr.btop\endcsname{0.783}
\expandafter\gdef\csname Nn@ai.l131k.jnull.fuzztpr.btop\endcsname{125}
\expandafter\gdef\csname Nstatus@ai.l131k.jnull.fuzztpr.btop\endcsname{final}
\expandafter\gdef\csname N@ai.l131k.mtop16.tnr.b0\endcsname{0.986}
\expandafter\gdef\csname Nlo@ai.l131k.mtop16.tnr.b0\endcsname{0.982}
\expandafter\gdef\csname Nhi@ai.l131k.mtop16.tnr.b0\endcsname{0.989}
\expandafter\gdef\csname Nn@ai.l131k.mtop16.tnr.b0\endcsname{456}
\expandafter\gdef\csname Nstatus@ai.l131k.mtop16.tnr.b0\endcsname{final}
\expandafter\gdef\csname N@ai.l131k.mtop16.tpr.btop\endcsname{0.552}
\expandafter\gdef\csname Nlo@ai.l131k.mtop16.tpr.btop\endcsname{0.473}
\expandafter\gdef\csname Nhi@ai.l131k.mtop16.tpr.btop\endcsname{0.623}
\expandafter\gdef\csname Nn@ai.l131k.mtop16.tpr.btop\endcsname{114}
\expandafter\gdef\csname Nstatus@ai.l131k.mtop16.tpr.btop\endcsname{final}
\expandafter\gdef\csname N@ai.l131k.mtop16.fuzztnr.b0\endcsname{0.998}
\expandafter\gdef\csname Nlo@ai.l131k.mtop16.fuzztnr.b0\endcsname{0.997}
\expandafter\gdef\csname Nhi@ai.l131k.mtop16.fuzztnr.b0\endcsname{0.999}
\expandafter\gdef\csname Nn@ai.l131k.mtop16.fuzztnr.b0\endcsname{456}
\expandafter\gdef\csname Nstatus@ai.l131k.mtop16.fuzztnr.b0\endcsname{final}
\expandafter\gdef\csname N@ai.l131k.mtop16.fuzztpr.b1\endcsname{0.073}
\expandafter\gdef\csname Nlo@ai.l131k.mtop16.fuzztpr.b1\endcsname{0.059}
\expandafter\gdef\csname Nhi@ai.l131k.mtop16.fuzztpr.b1\endcsname{0.089}
\expandafter\gdef\csname Nn@ai.l131k.mtop16.fuzztpr.b1\endcsname{438}
\expandafter\gdef\csname Nstatus@ai.l131k.mtop16.fuzztpr.b1\endcsname{final}
\expandafter\gdef\csname N@ai.l131k.mtop16.fuzztpr.b2\endcsname{0.231}
\expandafter\gdef\csname Nlo@ai.l131k.mtop16.fuzztpr.b2\endcsname{0.204}
\expandafter\gdef\csname Nhi@ai.l131k.mtop16.fuzztpr.b2\endcsname{0.257}
\expandafter\gdef\csname Nn@ai.l131k.mtop16.fuzztpr.b2\endcsname{452}
\expandafter\gdef\csname Nstatus@ai.l131k.mtop16.fuzztpr.b2\endcsname{final}
\expandafter\gdef\csname N@ai.l131k.mtop16.fuzztpr.b3\endcsname{0.389}
\expandafter\gdef\csname Nlo@ai.l131k.mtop16.fuzztpr.b3\endcsname{0.358}
\expandafter\gdef\csname Nhi@ai.l131k.mtop16.fuzztpr.b3\endcsname{0.420}
\expandafter\gdef\csname Nn@ai.l131k.mtop16.fuzztpr.b3\endcsname{435}
\expandafter\gdef\csname Nstatus@ai.l131k.mtop16.fuzztpr.b3\endcsname{final}
\expandafter\gdef\csname N@ai.l131k.mtop16.fuzztpr.b4\endcsname{0.487}
\expandafter\gdef\csname Nlo@ai.l131k.mtop16.fuzztpr.b4\endcsname{0.451}
\expandafter\gdef\csname Nhi@ai.l131k.mtop16.fuzztpr.b4\endcsname{0.521}
\expandafter\gdef\csname Nn@ai.l131k.mtop16.fuzztpr.b4\endcsname{454}
\expandafter\gdef\csname Nstatus@ai.l131k.mtop16.fuzztpr.b4\endcsname{final}
\expandafter\gdef\csname N@ai.l131k.mtop16.fuzztpr.btop\endcsname{0.414}
\expandafter\gdef\csname Nlo@ai.l131k.mtop16.fuzztpr.btop\endcsname{0.340}
\expandafter\gdef\csname Nhi@ai.l131k.mtop16.fuzztpr.btop\endcsname{0.486}
\expandafter\gdef\csname Nn@ai.l131k.mtop16.fuzztpr.btop\endcsname{114}
\expandafter\gdef\csname Nstatus@ai.l131k.mtop16.fuzztpr.btop\endcsname{final}
\expandafter\gdef\csname N@ai.l131k.mcos16.tnr.b0\endcsname{0.982}
\expandafter\gdef\csname Nlo@ai.l131k.mcos16.tnr.b0\endcsname{0.978}
\expandafter\gdef\csname Nhi@ai.l131k.mcos16.tnr.b0\endcsname{0.986}
\expandafter\gdef\csname Nn@ai.l131k.mcos16.tnr.b0\endcsname{449}
\expandafter\gdef\csname Nstatus@ai.l131k.mcos16.tnr.b0\endcsname{final}
\expandafter\gdef\csname N@ai.l131k.mcos16.tpr.b1\endcsname{0.114}
\expandafter\gdef\csname Nlo@ai.l131k.mcos16.tpr.b1\endcsname{0.095}
\expandafter\gdef\csname Nhi@ai.l131k.mcos16.tpr.b1\endcsname{0.133}
\expandafter\gdef\csname Nn@ai.l131k.mcos16.tpr.b1\endcsname{432}
\expandafter\gdef\csname Nstatus@ai.l131k.mcos16.tpr.b1\endcsname{final}
\expandafter\gdef\csname N@ai.l131k.mcos16.tpr.b2\endcsname{0.327}
\expandafter\gdef\csname Nlo@ai.l131k.mcos16.tpr.b2\endcsname{0.297}
\expandafter\gdef\csname Nhi@ai.l131k.mcos16.tpr.b2\endcsname{0.356}
\expandafter\gdef\csname Nn@ai.l131k.mcos16.tpr.b2\endcsname{443}
\expandafter\gdef\csname Nstatus@ai.l131k.mcos16.tpr.b2\endcsname{final}
\expandafter\gdef\csname N@ai.l131k.mcos16.tpr.b3\endcsname{0.526}
\expandafter\gdef\csname Nlo@ai.l131k.mcos16.tpr.b3\endcsname{0.491}
\expandafter\gdef\csname Nhi@ai.l131k.mcos16.tpr.b3\endcsname{0.558}
\expandafter\gdef\csname Nn@ai.l131k.mcos16.tpr.b3\endcsname{429}
\expandafter\gdef\csname Nstatus@ai.l131k.mcos16.tpr.b3\endcsname{final}
\expandafter\gdef\csname N@ai.l131k.mcos16.tpr.b4\endcsname{0.643}
\expandafter\gdef\csname Nlo@ai.l131k.mcos16.tpr.b4\endcsname{0.605}
\expandafter\gdef\csname Nhi@ai.l131k.mcos16.tpr.b4\endcsname{0.681}
\expandafter\gdef\csname Nn@ai.l131k.mcos16.tpr.b4\endcsname{449}
\expandafter\gdef\csname Nstatus@ai.l131k.mcos16.tpr.b4\endcsname{final}
\expandafter\gdef\csname N@ai.l131k.mcos16.tpr.btop\endcsname{0.608}
\expandafter\gdef\csname Nlo@ai.l131k.mcos16.tpr.btop\endcsname{0.537}
\expandafter\gdef\csname Nhi@ai.l131k.mcos16.tpr.btop\endcsname{0.677}
\expandafter\gdef\csname Nn@ai.l131k.mcos16.tpr.btop\endcsname{110}
\expandafter\gdef\csname Nstatus@ai.l131k.mcos16.tpr.btop\endcsname{final}
\expandafter\gdef\csname N@ai.l131k.mcos16.fuzztnr.b0\endcsname{0.997}
\expandafter\gdef\csname Nlo@ai.l131k.mcos16.fuzztnr.b0\endcsname{0.995}
\expandafter\gdef\csname Nhi@ai.l131k.mcos16.fuzztnr.b0\endcsname{0.998}
\expandafter\gdef\csname Nn@ai.l131k.mcos16.fuzztnr.b0\endcsname{449}
\expandafter\gdef\csname Nstatus@ai.l131k.mcos16.fuzztnr.b0\endcsname{final}
\expandafter\gdef\csname N@ai.l131k.mcos16.fuzztpr.b1\endcsname{0.087}
\expandafter\gdef\csname Nlo@ai.l131k.mcos16.fuzztpr.b1\endcsname{0.072}
\expandafter\gdef\csname Nhi@ai.l131k.mcos16.fuzztpr.b1\endcsname{0.104}
\expandafter\gdef\csname Nn@ai.l131k.mcos16.fuzztpr.b1\endcsname{432}
\expandafter\gdef\csname Nstatus@ai.l131k.mcos16.fuzztpr.b1\endcsname{final}
\expandafter\gdef\csname N@ai.l131k.mcos16.fuzztpr.b2\endcsname{0.265}
\expandafter\gdef\csname Nlo@ai.l131k.mcos16.fuzztpr.b2\endcsname{0.236}
\expandafter\gdef\csname Nhi@ai.l131k.mcos16.fuzztpr.b2\endcsname{0.292}
\expandafter\gdef\csname Nn@ai.l131k.mcos16.fuzztpr.b2\endcsname{443}
\expandafter\gdef\csname Nstatus@ai.l131k.mcos16.fuzztpr.b2\endcsname{final}
\expandafter\gdef\csname N@ai.l131k.mcos16.fuzztpr.b3\endcsname{0.401}
\expandafter\gdef\csname Nlo@ai.l131k.mcos16.fuzztpr.b3\endcsname{0.367}
\expandafter\gdef\csname Nhi@ai.l131k.mcos16.fuzztpr.b3\endcsname{0.431}
\expandafter\gdef\csname Nn@ai.l131k.mcos16.fuzztpr.b3\endcsname{429}
\expandafter\gdef\csname Nstatus@ai.l131k.mcos16.fuzztpr.b3\endcsname{final}
\expandafter\gdef\csname N@ai.l131k.mcos16.fuzztpr.b4\endcsname{0.497}
\expandafter\gdef\csname Nlo@ai.l131k.mcos16.fuzztpr.b4\endcsname{0.463}
\expandafter\gdef\csname Nhi@ai.l131k.mcos16.fuzztpr.b4\endcsname{0.531}
\expandafter\gdef\csname Nn@ai.l131k.mcos16.fuzztpr.b4\endcsname{449}
\expandafter\gdef\csname Nstatus@ai.l131k.mcos16.fuzztpr.b4\endcsname{final}
\expandafter\gdef\csname N@ai.l131k.mcos16.fuzztpr.btop\endcsname{0.509}
\expandafter\gdef\csname Nlo@ai.l131k.mcos16.fuzztpr.btop\endcsname{0.434}
\expandafter\gdef\csname Nhi@ai.l131k.mcos16.fuzztpr.btop\endcsname{0.580}
\expandafter\gdef\csname Nn@ai.l131k.mcos16.fuzztpr.btop\endcsname{110}
\expandafter\gdef\csname Nstatus@ai.l131k.mcos16.fuzztpr.btop\endcsname{final}
\expandafter\gdef\csname N@ai.l131k.mjac16.tnr.b0\endcsname{0.984}
\expandafter\gdef\csname Nlo@ai.l131k.mjac16.tnr.b0\endcsname{0.981}
\expandafter\gdef\csname Nhi@ai.l131k.mjac16.tnr.b0\endcsname{0.988}
\expandafter\gdef\csname Nn@ai.l131k.mjac16.tnr.b0\endcsname{444}
\expandafter\gdef\csname Nstatus@ai.l131k.mjac16.tnr.b0\endcsname{final}
\expandafter\gdef\csname N@ai.l131k.mjac16.tpr.b1\endcsname{0.105}
\expandafter\gdef\csname Nlo@ai.l131k.mjac16.tpr.b1\endcsname{0.089}
\expandafter\gdef\csname Nhi@ai.l131k.mjac16.tpr.b1\endcsname{0.123}
\expandafter\gdef\csname Nn@ai.l131k.mjac16.tpr.b1\endcsname{427}
\expandafter\gdef\csname Nstatus@ai.l131k.mjac16.tpr.b1\endcsname{final}
\expandafter\gdef\csname N@ai.l131k.mjac16.tpr.b2\endcsname{0.321}
\expandafter\gdef\csname Nlo@ai.l131k.mjac16.tpr.b2\endcsname{0.293}
\expandafter\gdef\csname Nhi@ai.l131k.mjac16.tpr.b2\endcsname{0.352}
\expandafter\gdef\csname Nn@ai.l131k.mjac16.tpr.b2\endcsname{440}
\expandafter\gdef\csname Nstatus@ai.l131k.mjac16.tpr.b2\endcsname{final}
\expandafter\gdef\csname N@ai.l131k.mjac16.tpr.b3\endcsname{0.526}
\expandafter\gdef\csname Nlo@ai.l131k.mjac16.tpr.b3\endcsname{0.489}
\expandafter\gdef\csname Nhi@ai.l131k.mjac16.tpr.b3\endcsname{0.561}
\expandafter\gdef\csname Nn@ai.l131k.mjac16.tpr.b3\endcsname{429}
\expandafter\gdef\csname Nstatus@ai.l131k.mjac16.tpr.b3\endcsname{final}
\expandafter\gdef\csname N@ai.l131k.mjac16.tpr.b4\endcsname{0.657}
\expandafter\gdef\csname Nlo@ai.l131k.mjac16.tpr.b4\endcsname{0.619}
\expandafter\gdef\csname Nhi@ai.l131k.mjac16.tpr.b4\endcsname{0.694}
\expandafter\gdef\csname Nn@ai.l131k.mjac16.tpr.b4\endcsname{444}
\expandafter\gdef\csname Nstatus@ai.l131k.mjac16.tpr.b4\endcsname{final}
\expandafter\gdef\csname N@ai.l131k.mjac16.tpr.btop\endcsname{0.629}
\expandafter\gdef\csname Nlo@ai.l131k.mjac16.tpr.btop\endcsname{0.558}
\expandafter\gdef\csname Nhi@ai.l131k.mjac16.tpr.btop\endcsname{0.698}
\expandafter\gdef\csname Nn@ai.l131k.mjac16.tpr.btop\endcsname{111}
\expandafter\gdef\csname Nstatus@ai.l131k.mjac16.tpr.btop\endcsname{final}
\expandafter\gdef\csname N@ai.l131k.mjac16.fuzztnr.b0\endcsname{0.996}
\expandafter\gdef\csname Nlo@ai.l131k.mjac16.fuzztnr.b0\endcsname{0.994}
\expandafter\gdef\csname Nhi@ai.l131k.mjac16.fuzztnr.b0\endcsname{0.997}
\expandafter\gdef\csname Nn@ai.l131k.mjac16.fuzztnr.b0\endcsname{444}
\expandafter\gdef\csname Nstatus@ai.l131k.mjac16.fuzztnr.b0\endcsname{final}
\expandafter\gdef\csname N@ai.l131k.mjac16.fuzztpr.b1\endcsname{0.088}
\expandafter\gdef\csname Nlo@ai.l131k.mjac16.fuzztpr.b1\endcsname{0.072}
\expandafter\gdef\csname Nhi@ai.l131k.mjac16.fuzztpr.b1\endcsname{0.104}
\expandafter\gdef\csname Nn@ai.l131k.mjac16.fuzztpr.b1\endcsname{427}
\expandafter\gdef\csname Nstatus@ai.l131k.mjac16.fuzztpr.b1\endcsname{final}
\expandafter\gdef\csname N@ai.l131k.mjac16.fuzztpr.b2\endcsname{0.254}
\expandafter\gdef\csname Nlo@ai.l131k.mjac16.fuzztpr.b2\endcsname{0.229}
\expandafter\gdef\csname Nhi@ai.l131k.mjac16.fuzztpr.b2\endcsname{0.280}
\expandafter\gdef\csname Nn@ai.l131k.mjac16.fuzztpr.b2\endcsname{440}
\expandafter\gdef\csname Nstatus@ai.l131k.mjac16.fuzztpr.b2\endcsname{final}
\expandafter\gdef\csname N@ai.l131k.mjac16.fuzztpr.b3\endcsname{0.402}
\expandafter\gdef\csname Nlo@ai.l131k.mjac16.fuzztpr.b3\endcsname{0.368}
\expandafter\gdef\csname Nhi@ai.l131k.mjac16.fuzztpr.b3\endcsname{0.434}
\expandafter\gdef\csname Nn@ai.l131k.mjac16.fuzztpr.b3\endcsname{429}
\expandafter\gdef\csname Nstatus@ai.l131k.mjac16.fuzztpr.b3\endcsname{final}
\expandafter\gdef\csname N@ai.l131k.mjac16.fuzztpr.b4\endcsname{0.517}
\expandafter\gdef\csname Nlo@ai.l131k.mjac16.fuzztpr.b4\endcsname{0.482}
\expandafter\gdef\csname Nhi@ai.l131k.mjac16.fuzztpr.b4\endcsname{0.551}
\expandafter\gdef\csname Nn@ai.l131k.mjac16.fuzztpr.b4\endcsname{443}
\expandafter\gdef\csname Nstatus@ai.l131k.mjac16.fuzztpr.b4\endcsname{final}
\expandafter\gdef\csname N@ai.l131k.mjac16.fuzztpr.btop\endcsname{0.485}
\expandafter\gdef\csname Nlo@ai.l131k.mjac16.fuzztpr.btop\endcsname{0.412}
\expandafter\gdef\csname Nhi@ai.l131k.mjac16.fuzztpr.btop\endcsname{0.556}
\expandafter\gdef\csname Nn@ai.l131k.mjac16.fuzztpr.btop\endcsname{111}
\expandafter\gdef\csname Nstatus@ai.l131k.mjac16.fuzztpr.btop\endcsname{final}
\expandafter\gdef\csname N@ai.l131k.nla4.tnr.b0\endcsname{0.963}
\expandafter\gdef\csname Nlo@ai.l131k.nla4.tnr.b0\endcsname{0.954}
\expandafter\gdef\csname Nhi@ai.l131k.nla4.tnr.b0\endcsname{0.970}
\expandafter\gdef\csname Nn@ai.l131k.nla4.tnr.b0\endcsname{471}
\expandafter\gdef\csname Nstatus@ai.l131k.nla4.tnr.b0\endcsname{final}
\expandafter\gdef\csname N@ai.l131k.nla4.tpr.b1\endcsname{0.137}
\expandafter\gdef\csname Nlo@ai.l131k.nla4.tpr.b1\endcsname{0.116}
\expandafter\gdef\csname Nhi@ai.l131k.nla4.tpr.b1\endcsname{0.158}
\expandafter\gdef\csname Nn@ai.l131k.nla4.tpr.b1\endcsname{453}
\expandafter\gdef\csname Nstatus@ai.l131k.nla4.tpr.b1\endcsname{final}
\expandafter\gdef\csname N@ai.l131k.nla4.tpr.b2\endcsname{0.315}
\expandafter\gdef\csname Nlo@ai.l131k.nla4.tpr.b2\endcsname{0.286}
\expandafter\gdef\csname Nhi@ai.l131k.nla4.tpr.b2\endcsname{0.346}
\expandafter\gdef\csname Nn@ai.l131k.nla4.tpr.b2\endcsname{465}
\expandafter\gdef\csname Nstatus@ai.l131k.nla4.tpr.b2\endcsname{final}
\expandafter\gdef\csname N@ai.l131k.nla4.tpr.b3\endcsname{0.481}
\expandafter\gdef\csname Nlo@ai.l131k.nla4.tpr.b3\endcsname{0.443}
\expandafter\gdef\csname Nhi@ai.l131k.nla4.tpr.b3\endcsname{0.516}
\expandafter\gdef\csname Nn@ai.l131k.nla4.tpr.b3\endcsname{443}
\expandafter\gdef\csname Nstatus@ai.l131k.nla4.tpr.b3\endcsname{final}
\expandafter\gdef\csname N@ai.l131k.nla4.tpr.b4\endcsname{0.566}
\expandafter\gdef\csname Nlo@ai.l131k.nla4.tpr.b4\endcsname{0.529}
\expandafter\gdef\csname Nhi@ai.l131k.nla4.tpr.b4\endcsname{0.604}
\expandafter\gdef\csname Nn@ai.l131k.nla4.tpr.b4\endcsname{471}
\expandafter\gdef\csname Nstatus@ai.l131k.nla4.tpr.b4\endcsname{final}
\expandafter\gdef\csname N@ai.l131k.nla4.tpr.btop\endcsname{0.590}
\expandafter\gdef\csname Nlo@ai.l131k.nla4.tpr.btop\endcsname{0.515}
\expandafter\gdef\csname Nhi@ai.l131k.nla4.tpr.btop\endcsname{0.665}
\expandafter\gdef\csname Nn@ai.l131k.nla4.tpr.btop\endcsname{115}
\expandafter\gdef\csname Nstatus@ai.l131k.nla4.tpr.btop\endcsname{final}
\expandafter\gdef\csname N@ai.l131k.nla4.fuzztnr.b0\endcsname{0.993}
\expandafter\gdef\csname Nlo@ai.l131k.nla4.fuzztnr.b0\endcsname{0.992}
\expandafter\gdef\csname Nhi@ai.l131k.nla4.fuzztnr.b0\endcsname{0.995}
\expandafter\gdef\csname Nn@ai.l131k.nla4.fuzztnr.b0\endcsname{470}
\expandafter\gdef\csname Nstatus@ai.l131k.nla4.fuzztnr.b0\endcsname{final}
\expandafter\gdef\csname N@ai.l131k.nla4.fuzztpr.b1\endcsname{0.101}
\expandafter\gdef\csname Nlo@ai.l131k.nla4.fuzztpr.b1\endcsname{0.083}
\expandafter\gdef\csname Nhi@ai.l131k.nla4.fuzztpr.b1\endcsname{0.120}
\expandafter\gdef\csname Nn@ai.l131k.nla4.fuzztpr.b1\endcsname{452}
\expandafter\gdef\csname Nstatus@ai.l131k.nla4.fuzztpr.b1\endcsname{final}
\expandafter\gdef\csname N@ai.l131k.nla4.fuzztpr.b2\endcsname{0.246}
\expandafter\gdef\csname Nlo@ai.l131k.nla4.fuzztpr.b2\endcsname{0.220}
\expandafter\gdef\csname Nhi@ai.l131k.nla4.fuzztpr.b2\endcsname{0.273}
\expandafter\gdef\csname Nn@ai.l131k.nla4.fuzztpr.b2\endcsname{464}
\expandafter\gdef\csname Nstatus@ai.l131k.nla4.fuzztpr.b2\endcsname{final}
\expandafter\gdef\csname N@ai.l131k.nla4.fuzztpr.b3\endcsname{0.361}
\expandafter\gdef\csname Nlo@ai.l131k.nla4.fuzztpr.b3\endcsname{0.329}
\expandafter\gdef\csname Nhi@ai.l131k.nla4.fuzztpr.b3\endcsname{0.395}
\expandafter\gdef\csname Nn@ai.l131k.nla4.fuzztpr.b3\endcsname{442}
\expandafter\gdef\csname Nstatus@ai.l131k.nla4.fuzztpr.b3\endcsname{final}
\expandafter\gdef\csname N@ai.l131k.nla4.fuzztpr.b4\endcsname{0.443}
\expandafter\gdef\csname Nlo@ai.l131k.nla4.fuzztpr.b4\endcsname{0.407}
\expandafter\gdef\csname Nhi@ai.l131k.nla4.fuzztpr.b4\endcsname{0.477}
\expandafter\gdef\csname Nn@ai.l131k.nla4.fuzztpr.b4\endcsname{469}
\expandafter\gdef\csname Nstatus@ai.l131k.nla4.fuzztpr.b4\endcsname{final}
\expandafter\gdef\csname N@ai.l131k.nla4.fuzztpr.btop\endcsname{0.487}
\expandafter\gdef\csname Nlo@ai.l131k.nla4.fuzztpr.btop\endcsname{0.419}
\expandafter\gdef\csname Nhi@ai.l131k.nla4.fuzztpr.btop\endcsname{0.558}
\expandafter\gdef\csname Nn@ai.l131k.nla4.fuzztpr.btop\endcsname{115}
\expandafter\gdef\csname Nstatus@ai.l131k.nla4.fuzztpr.btop\endcsname{final}
\expandafter\gdef\csname N@ai.l131k.nla1.tnr.b0\endcsname{0.923}
\expandafter\gdef\csname Nlo@ai.l131k.nla1.tnr.b0\endcsname{0.912}
\expandafter\gdef\csname Nhi@ai.l131k.nla1.tnr.b0\endcsname{0.934}
\expandafter\gdef\csname Nn@ai.l131k.nla1.tnr.b0\endcsname{469}
\expandafter\gdef\csname Nstatus@ai.l131k.nla1.tnr.b0\endcsname{final}
\expandafter\gdef\csname N@ai.l131k.nla1.tpr.b1\endcsname{0.196}
\expandafter\gdef\csname Nlo@ai.l131k.nla1.tpr.b1\endcsname{0.174}
\expandafter\gdef\csname Nhi@ai.l131k.nla1.tpr.b1\endcsname{0.219}
\expandafter\gdef\csname Nn@ai.l131k.nla1.tpr.b1\endcsname{452}
\expandafter\gdef\csname Nstatus@ai.l131k.nla1.tpr.b1\endcsname{final}
\expandafter\gdef\csname N@ai.l131k.nla1.tpr.b2\endcsname{0.376}
\expandafter\gdef\csname Nlo@ai.l131k.nla1.tpr.b2\endcsname{0.344}
\expandafter\gdef\csname Nhi@ai.l131k.nla1.tpr.b2\endcsname{0.406}
\expandafter\gdef\csname Nn@ai.l131k.nla1.tpr.b2\endcsname{463}
\expandafter\gdef\csname Nstatus@ai.l131k.nla1.tpr.b2\endcsname{final}
\expandafter\gdef\csname N@ai.l131k.nla1.tpr.b3\endcsname{0.523}
\expandafter\gdef\csname Nlo@ai.l131k.nla1.tpr.b3\endcsname{0.486}
\expandafter\gdef\csname Nhi@ai.l131k.nla1.tpr.b3\endcsname{0.559}
\expandafter\gdef\csname Nn@ai.l131k.nla1.tpr.b3\endcsname{441}
\expandafter\gdef\csname Nstatus@ai.l131k.nla1.tpr.b3\endcsname{final}
\expandafter\gdef\csname N@ai.l131k.nla1.tpr.b4\endcsname{0.610}
\expandafter\gdef\csname Nlo@ai.l131k.nla1.tpr.b4\endcsname{0.573}
\expandafter\gdef\csname Nhi@ai.l131k.nla1.tpr.b4\endcsname{0.647}
\expandafter\gdef\csname Nn@ai.l131k.nla1.tpr.b4\endcsname{469}
\expandafter\gdef\csname Nstatus@ai.l131k.nla1.tpr.b4\endcsname{final}
\expandafter\gdef\csname N@ai.l131k.nla1.tpr.btop\endcsname{0.613}
\expandafter\gdef\csname Nlo@ai.l131k.nla1.tpr.btop\endcsname{0.537}
\expandafter\gdef\csname Nhi@ai.l131k.nla1.tpr.btop\endcsname{0.684}
\expandafter\gdef\csname Nn@ai.l131k.nla1.tpr.btop\endcsname{114}
\expandafter\gdef\csname Nstatus@ai.l131k.nla1.tpr.btop\endcsname{final}
\expandafter\gdef\csname N@ai.l131k.nla1.fuzztnr.b0\endcsname{0.980}
\expandafter\gdef\csname Nlo@ai.l131k.nla1.fuzztnr.b0\endcsname{0.975}
\expandafter\gdef\csname Nhi@ai.l131k.nla1.fuzztnr.b0\endcsname{0.985}
\expandafter\gdef\csname Nn@ai.l131k.nla1.fuzztnr.b0\endcsname{469}
\expandafter\gdef\csname Nstatus@ai.l131k.nla1.fuzztnr.b0\endcsname{final}
\expandafter\gdef\csname N@ai.l131k.nla1.fuzztpr.b1\endcsname{0.152}
\expandafter\gdef\csname Nlo@ai.l131k.nla1.fuzztpr.b1\endcsname{0.130}
\expandafter\gdef\csname Nhi@ai.l131k.nla1.fuzztpr.b1\endcsname{0.176}
\expandafter\gdef\csname Nn@ai.l131k.nla1.fuzztpr.b1\endcsname{452}
\expandafter\gdef\csname Nstatus@ai.l131k.nla1.fuzztpr.b1\endcsname{final}
\expandafter\gdef\csname N@ai.l131k.nla1.fuzztpr.b2\endcsname{0.293}
\expandafter\gdef\csname Nlo@ai.l131k.nla1.fuzztpr.b2\endcsname{0.264}
\expandafter\gdef\csname Nhi@ai.l131k.nla1.fuzztpr.b2\endcsname{0.320}
\expandafter\gdef\csname Nn@ai.l131k.nla1.fuzztpr.b2\endcsname{463}
\expandafter\gdef\csname Nstatus@ai.l131k.nla1.fuzztpr.b2\endcsname{final}
\expandafter\gdef\csname N@ai.l131k.nla1.fuzztpr.b3\endcsname{0.387}
\expandafter\gdef\csname Nlo@ai.l131k.nla1.fuzztpr.b3\endcsname{0.354}
\expandafter\gdef\csname Nhi@ai.l131k.nla1.fuzztpr.b3\endcsname{0.421}
\expandafter\gdef\csname Nn@ai.l131k.nla1.fuzztpr.b3\endcsname{441}
\expandafter\gdef\csname Nstatus@ai.l131k.nla1.fuzztpr.b3\endcsname{final}
\expandafter\gdef\csname N@ai.l131k.nla1.fuzztpr.b4\endcsname{0.437}
\expandafter\gdef\csname Nlo@ai.l131k.nla1.fuzztpr.b4\endcsname{0.402}
\expandafter\gdef\csname Nhi@ai.l131k.nla1.fuzztpr.b4\endcsname{0.472}
\expandafter\gdef\csname Nn@ai.l131k.nla1.fuzztpr.b4\endcsname{468}
\expandafter\gdef\csname Nstatus@ai.l131k.nla1.fuzztpr.b4\endcsname{final}
\expandafter\gdef\csname N@ai.l131k.nla1.fuzztpr.btop\endcsname{0.491}
\expandafter\gdef\csname Nlo@ai.l131k.nla1.fuzztpr.btop\endcsname{0.418}
\expandafter\gdef\csname Nhi@ai.l131k.nla1.fuzztpr.btop\endcsname{0.563}
\expandafter\gdef\csname Nn@ai.l131k.nla1.fuzztpr.btop\endcsname{114}
\expandafter\gdef\csname Nstatus@ai.l131k.nla1.fuzztpr.btop\endcsname{final}
\expandafter\gdef\csname N@ai.l131k.nlab.tnr.b0\endcsname{0.997}
\expandafter\gdef\csname Nlo@ai.l131k.nlab.tnr.b0\endcsname{0.996}
\expandafter\gdef\csname Nhi@ai.l131k.nlab.tnr.b0\endcsname{0.998}
\expandafter\gdef\csname Nn@ai.l131k.nlab.tnr.b0\endcsname{503}
\expandafter\gdef\csname Nstatus@ai.l131k.nlab.tnr.b0\endcsname{final}
\expandafter\gdef\csname N@ai.l131k.nlab.tpr.b1\endcsname{0.015}
\expandafter\gdef\csname Nlo@ai.l131k.nlab.tpr.b1\endcsname{0.010}
\expandafter\gdef\csname Nhi@ai.l131k.nlab.tpr.b1\endcsname{0.021}
\expandafter\gdef\csname Nn@ai.l131k.nlab.tpr.b1\endcsname{485}
\expandafter\gdef\csname Nstatus@ai.l131k.nlab.tpr.b1\endcsname{final}
\expandafter\gdef\csname N@ai.l131k.nlab.tpr.b2\endcsname{0.043}
\expandafter\gdef\csname Nlo@ai.l131k.nlab.tpr.b2\endcsname{0.032}
\expandafter\gdef\csname Nhi@ai.l131k.nlab.tpr.b2\endcsname{0.054}
\expandafter\gdef\csname Nn@ai.l131k.nlab.tpr.b2\endcsname{497}
\expandafter\gdef\csname Nstatus@ai.l131k.nlab.tpr.b2\endcsname{final}
\expandafter\gdef\csname N@ai.l131k.nlab.tpr.b3\endcsname{0.090}
\expandafter\gdef\csname Nlo@ai.l131k.nlab.tpr.b3\endcsname{0.072}
\expandafter\gdef\csname Nhi@ai.l131k.nlab.tpr.b3\endcsname{0.108}
\expandafter\gdef\csname Nn@ai.l131k.nlab.tpr.b3\endcsname{473}
\expandafter\gdef\csname Nstatus@ai.l131k.nlab.tpr.b3\endcsname{final}
\expandafter\gdef\csname N@ai.l131k.nlab.tpr.b4\endcsname{0.155}
\expandafter\gdef\csname Nlo@ai.l131k.nlab.tpr.b4\endcsname{0.131}
\expandafter\gdef\csname Nhi@ai.l131k.nlab.tpr.b4\endcsname{0.178}
\expandafter\gdef\csname Nn@ai.l131k.nlab.tpr.b4\endcsname{503}
\expandafter\gdef\csname Nstatus@ai.l131k.nlab.tpr.b4\endcsname{final}
\expandafter\gdef\csname N@ai.l131k.nlab.tpr.btop\endcsname{0.088}
\expandafter\gdef\csname Nlo@ai.l131k.nlab.tpr.btop\endcsname{0.051}
\expandafter\gdef\csname Nhi@ai.l131k.nlab.tpr.btop\endcsname{0.130}
\expandafter\gdef\csname Nn@ai.l131k.nlab.tpr.btop\endcsname{124}
\expandafter\gdef\csname Nstatus@ai.l131k.nlab.tpr.btop\endcsname{final}
\expandafter\gdef\csname N@ai.l131k.nlab.fuzztnr.b0\endcsname{0.990}
\expandafter\gdef\csname Nlo@ai.l131k.nlab.fuzztnr.b0\endcsname{0.986}
\expandafter\gdef\csname Nhi@ai.l131k.nlab.fuzztnr.b0\endcsname{0.994}
\expandafter\gdef\csname Nn@ai.l131k.nlab.fuzztnr.b0\endcsname{483}
\expandafter\gdef\csname Nstatus@ai.l131k.nlab.fuzztnr.b0\endcsname{final}
\expandafter\gdef\csname N@ai.l131k.nlab.fuzztpr.b1\endcsname{0.031}
\expandafter\gdef\csname Nlo@ai.l131k.nlab.fuzztpr.b1\endcsname{0.022}
\expandafter\gdef\csname Nhi@ai.l131k.nlab.fuzztpr.b1\endcsname{0.041}
\expandafter\gdef\csname Nn@ai.l131k.nlab.fuzztpr.b1\endcsname{456}
\expandafter\gdef\csname Nstatus@ai.l131k.nlab.fuzztpr.b1\endcsname{final}
\expandafter\gdef\csname N@ai.l131k.nlab.fuzztpr.b2\endcsname{0.057}
\expandafter\gdef\csname Nlo@ai.l131k.nlab.fuzztpr.b2\endcsname{0.045}
\expandafter\gdef\csname Nhi@ai.l131k.nlab.fuzztpr.b2\endcsname{0.071}
\expandafter\gdef\csname Nn@ai.l131k.nlab.fuzztpr.b2\endcsname{475}
\expandafter\gdef\csname Nstatus@ai.l131k.nlab.fuzztpr.b2\endcsname{final}
\expandafter\gdef\csname N@ai.l131k.nlab.fuzztpr.b3\endcsname{0.110}
\expandafter\gdef\csname Nlo@ai.l131k.nlab.fuzztpr.b3\endcsname{0.091}
\expandafter\gdef\csname Nhi@ai.l131k.nlab.fuzztpr.b3\endcsname{0.131}
\expandafter\gdef\csname Nn@ai.l131k.nlab.fuzztpr.b3\endcsname{449}
\expandafter\gdef\csname Nstatus@ai.l131k.nlab.fuzztpr.b3\endcsname{final}
\expandafter\gdef\csname N@ai.l131k.nlab.fuzztpr.b4\endcsname{0.130}
\expandafter\gdef\csname Nlo@ai.l131k.nlab.fuzztpr.b4\endcsname{0.109}
\expandafter\gdef\csname Nhi@ai.l131k.nlab.fuzztpr.b4\endcsname{0.153}
\expandafter\gdef\csname Nn@ai.l131k.nlab.fuzztpr.b4\endcsname{470}
\expandafter\gdef\csname Nstatus@ai.l131k.nlab.fuzztpr.b4\endcsname{final}
\expandafter\gdef\csname N@ai.l131k.nlab.fuzztpr.btop\endcsname{0.101}
\expandafter\gdef\csname Nlo@ai.l131k.nlab.fuzztpr.btop\endcsname{0.065}
\expandafter\gdef\csname Nhi@ai.l131k.nlab.fuzztpr.btop\endcsname{0.138}
\expandafter\gdef\csname Nn@ai.l131k.nlab.fuzztpr.btop\endcsname{112}
\expandafter\gdef\csname Nstatus@ai.l131k.nlab.fuzztpr.btop\endcsname{final}
\expandafter\gdef\csname N@ai.l131k.floor.tnr.b0\endcsname{0.941}
\expandafter\gdef\csname Nlo@ai.l131k.floor.tnr.b0\endcsname{0.930}
\expandafter\gdef\csname Nhi@ai.l131k.floor.tnr.b0\endcsname{0.952}
\expandafter\gdef\csname Nn@ai.l131k.floor.tnr.b0\endcsname{506}
\expandafter\gdef\csname Nstatus@ai.l131k.floor.tnr.b0\endcsname{final}
\expandafter\gdef\csname N@ai.l131k.floor.tpr.btop\endcsname{0.077}
\expandafter\gdef\csname Nlo@ai.l131k.floor.tpr.btop\endcsname{0.046}
\expandafter\gdef\csname Nhi@ai.l131k.floor.tpr.btop\endcsname{0.114}
\expandafter\gdef\csname Nn@ai.l131k.floor.tpr.btop\endcsname{126}
\expandafter\gdef\csname Nstatus@ai.l131k.floor.tpr.btop\endcsname{final}
\expandafter\gdef\csname N@ai.l131k.floor.fuzztnr.b0\endcsname{0.991}
\expandafter\gdef\csname Nlo@ai.l131k.floor.fuzztnr.b0\endcsname{0.986}
\expandafter\gdef\csname Nhi@ai.l131k.floor.fuzztnr.b0\endcsname{0.994}
\expandafter\gdef\csname Nn@ai.l131k.floor.fuzztnr.b0\endcsname{506}
\expandafter\gdef\csname Nstatus@ai.l131k.floor.fuzztnr.b0\endcsname{final}
\expandafter\gdef\csname N@ai.l131k.floor.fuzztpr.b1\endcsname{0.015}
\expandafter\gdef\csname Nlo@ai.l131k.floor.fuzztpr.b1\endcsname{0.008}
\expandafter\gdef\csname Nhi@ai.l131k.floor.fuzztpr.b1\endcsname{0.024}
\expandafter\gdef\csname Nn@ai.l131k.floor.fuzztpr.b1\endcsname{487}
\expandafter\gdef\csname Nstatus@ai.l131k.floor.fuzztpr.b1\endcsname{final}
\expandafter\gdef\csname N@ai.l131k.floor.fuzztpr.b2\endcsname{0.021}
\expandafter\gdef\csname Nlo@ai.l131k.floor.fuzztpr.b2\endcsname{0.012}
\expandafter\gdef\csname Nhi@ai.l131k.floor.fuzztpr.b2\endcsname{0.030}
\expandafter\gdef\csname Nn@ai.l131k.floor.fuzztpr.b2\endcsname{500}
\expandafter\gdef\csname Nstatus@ai.l131k.floor.fuzztpr.b2\endcsname{final}
\expandafter\gdef\csname N@ai.l131k.floor.fuzztpr.b3\endcsname{0.018}
\expandafter\gdef\csname Nlo@ai.l131k.floor.fuzztpr.b3\endcsname{0.010}
\expandafter\gdef\csname Nhi@ai.l131k.floor.fuzztpr.b3\endcsname{0.027}
\expandafter\gdef\csname Nn@ai.l131k.floor.fuzztpr.b3\endcsname{478}
\expandafter\gdef\csname Nstatus@ai.l131k.floor.fuzztpr.b3\endcsname{final}
\expandafter\gdef\csname N@ai.l131k.floor.fuzztpr.b4\endcsname{0.013}
\expandafter\gdef\csname Nlo@ai.l131k.floor.fuzztpr.b4\endcsname{0.007}
\expandafter\gdef\csname Nhi@ai.l131k.floor.fuzztpr.b4\endcsname{0.021}
\expandafter\gdef\csname Nn@ai.l131k.floor.fuzztpr.b4\endcsname{506}
\expandafter\gdef\csname Nstatus@ai.l131k.floor.fuzztpr.b4\endcsname{final}
\expandafter\gdef\csname N@ai.l131k.floor.fuzztpr.btop\endcsname{0.022}
\expandafter\gdef\csname Nlo@ai.l131k.floor.fuzztpr.btop\endcsname{0.005}
\expandafter\gdef\csname Nhi@ai.l131k.floor.fuzztpr.btop\endcsname{0.043}
\expandafter\gdef\csname Nn@ai.l131k.floor.fuzztpr.btop\endcsname{126}
\expandafter\gdef\csname Nstatus@ai.l131k.floor.fuzztpr.btop\endcsname{final}
\expandafter\gdef\csname N@ai.l131k.diff.tnr.b0\endcsname{0.068}
\expandafter\gdef\csname Nlo@ai.l131k.diff.tnr.b0\endcsname{0.056}
\expandafter\gdef\csname Nhi@ai.l131k.diff.tnr.b0\endcsname{0.080}
\expandafter\gdef\csname Nn@ai.l131k.diff.tnr.b0\endcsname{453}
\expandafter\gdef\csname Nstatus@ai.l131k.diff.tnr.b0\endcsname{final}
\expandafter\gdef\csname N@ai.l131k.diff.tpr.b1\endcsname{-0.269}
\expandafter\gdef\csname Nlo@ai.l131k.diff.tpr.b1\endcsname{-0.299}
\expandafter\gdef\csname Nhi@ai.l131k.diff.tpr.b1\endcsname{-0.239}
\expandafter\gdef\csname Nn@ai.l131k.diff.tpr.b1\endcsname{435}
\expandafter\gdef\csname Nstatus@ai.l131k.diff.tpr.b1\endcsname{final}
\expandafter\gdef\csname N@ai.l131k.diff.tpr.b2\endcsname{-0.367}
\expandafter\gdef\csname Nlo@ai.l131k.diff.tpr.b2\endcsname{-0.399}
\expandafter\gdef\csname Nhi@ai.l131k.diff.tpr.b2\endcsname{-0.335}
\expandafter\gdef\csname Nn@ai.l131k.diff.tpr.b2\endcsname{449}
\expandafter\gdef\csname Nstatus@ai.l131k.diff.tpr.b2\endcsname{final}
\expandafter\gdef\csname N@ai.l131k.diff.tpr.b3\endcsname{-0.342}
\expandafter\gdef\csname Nlo@ai.l131k.diff.tpr.b3\endcsname{-0.376}
\expandafter\gdef\csname Nhi@ai.l131k.diff.tpr.b3\endcsname{-0.310}
\expandafter\gdef\csname Nn@ai.l131k.diff.tpr.b3\endcsname{433}
\expandafter\gdef\csname Nstatus@ai.l131k.diff.tpr.b3\endcsname{final}
\expandafter\gdef\csname N@ai.l131k.diff.tpr.b4\endcsname{-0.294}
\expandafter\gdef\csname Nlo@ai.l131k.diff.tpr.b4\endcsname{-0.331}
\expandafter\gdef\csname Nhi@ai.l131k.diff.tpr.b4\endcsname{-0.261}
\expandafter\gdef\csname Nn@ai.l131k.diff.tpr.b4\endcsname{453}
\expandafter\gdef\csname Nstatus@ai.l131k.diff.tpr.b4\endcsname{final}
\expandafter\gdef\csname N@ai.l131k.diff.tpr.btop\endcsname{-0.328}
\expandafter\gdef\csname Nlo@ai.l131k.diff.tpr.btop\endcsname{-0.398}
\expandafter\gdef\csname Nhi@ai.l131k.diff.tpr.btop\endcsname{-0.256}
\expandafter\gdef\csname Nn@ai.l131k.diff.tpr.btop\endcsname{113}
\expandafter\gdef\csname Nstatus@ai.l131k.diff.tpr.btop\endcsname{final}
\expandafter\gdef\csname N@ai.l131k.diff.fuzztnr.b0\endcsname{0.015}
\expandafter\gdef\csname Nlo@ai.l131k.diff.fuzztnr.b0\endcsname{0.010}
\expandafter\gdef\csname Nhi@ai.l131k.diff.fuzztnr.b0\endcsname{0.021}
\expandafter\gdef\csname Nn@ai.l131k.diff.fuzztnr.b0\endcsname{453}
\expandafter\gdef\csname Nstatus@ai.l131k.diff.fuzztnr.b0\endcsname{final}
\expandafter\gdef\csname N@ai.l131k.diff.fuzztpr.b1\endcsname{-0.195}
\expandafter\gdef\csname Nlo@ai.l131k.diff.fuzztpr.b1\endcsname{-0.220}
\expandafter\gdef\csname Nhi@ai.l131k.diff.fuzztpr.b1\endcsname{-0.170}
\expandafter\gdef\csname Nn@ai.l131k.diff.fuzztpr.b1\endcsname{435}
\expandafter\gdef\csname Nstatus@ai.l131k.diff.fuzztpr.b1\endcsname{final}
\expandafter\gdef\csname N@ai.l131k.diff.fuzztpr.b2\endcsname{-0.308}
\expandafter\gdef\csname Nlo@ai.l131k.diff.fuzztpr.b2\endcsname{-0.339}
\expandafter\gdef\csname Nhi@ai.l131k.diff.fuzztpr.b2\endcsname{-0.277}
\expandafter\gdef\csname Nn@ai.l131k.diff.fuzztpr.b2\endcsname{449}
\expandafter\gdef\csname Nstatus@ai.l131k.diff.fuzztpr.b2\endcsname{final}
\expandafter\gdef\csname N@ai.l131k.diff.fuzztpr.b3\endcsname{-0.296}
\expandafter\gdef\csname Nlo@ai.l131k.diff.fuzztpr.b3\endcsname{-0.329}
\expandafter\gdef\csname Nhi@ai.l131k.diff.fuzztpr.b3\endcsname{-0.264}
\expandafter\gdef\csname Nn@ai.l131k.diff.fuzztpr.b3\endcsname{432}
\expandafter\gdef\csname Nstatus@ai.l131k.diff.fuzztpr.b3\endcsname{final}
\expandafter\gdef\csname N@ai.l131k.diff.fuzztpr.b4\endcsname{-0.266}
\expandafter\gdef\csname Nlo@ai.l131k.diff.fuzztpr.b4\endcsname{-0.302}
\expandafter\gdef\csname Nhi@ai.l131k.diff.fuzztpr.b4\endcsname{-0.230}
\expandafter\gdef\csname Nn@ai.l131k.diff.fuzztpr.b4\endcsname{450}
\expandafter\gdef\csname Nstatus@ai.l131k.diff.fuzztpr.b4\endcsname{final}
\expandafter\gdef\csname N@ai.l131k.diff.fuzztpr.btop\endcsname{-0.315}
\expandafter\gdef\csname Nlo@ai.l131k.diff.fuzztpr.btop\endcsname{-0.386}
\expandafter\gdef\csname Nhi@ai.l131k.diff.fuzztpr.btop\endcsname{-0.245}
\expandafter\gdef\csname Nn@ai.l131k.diff.fuzztpr.btop\endcsname{113}
\expandafter\gdef\csname Nstatus@ai.l131k.diff.fuzztpr.btop\endcsname{final}
\expandafter\gdef\csname N@ai.l131k.diffnla.tnr.b0\endcsname{0.021}
\expandafter\gdef\csname Nlo@ai.l131k.diffnla.tnr.b0\endcsname{0.014}
\expandafter\gdef\csname Nhi@ai.l131k.diffnla.tnr.b0\endcsname{0.029}
\expandafter\gdef\csname Nn@ai.l131k.diffnla.tnr.b0\endcsname{423}
\expandafter\gdef\csname Nstatus@ai.l131k.diffnla.tnr.b0\endcsname{final}
\expandafter\gdef\csname N@ai.l131k.diffnla.tpr.b1\endcsname{-0.047}
\expandafter\gdef\csname Nlo@ai.l131k.diffnla.tpr.b1\endcsname{-0.070}
\expandafter\gdef\csname Nhi@ai.l131k.diffnla.tpr.b1\endcsname{-0.023}
\expandafter\gdef\csname Nn@ai.l131k.diffnla.tpr.b1\endcsname{406}
\expandafter\gdef\csname Nstatus@ai.l131k.diffnla.tpr.b1\endcsname{final}
\expandafter\gdef\csname N@ai.l131k.diffnla.tpr.b2\endcsname{-0.045}
\expandafter\gdef\csname Nlo@ai.l131k.diffnla.tpr.b2\endcsname{-0.075}
\expandafter\gdef\csname Nhi@ai.l131k.diffnla.tpr.b2\endcsname{-0.014}
\expandafter\gdef\csname Nn@ai.l131k.diffnla.tpr.b2\endcsname{419}
\expandafter\gdef\csname Nstatus@ai.l131k.diffnla.tpr.b2\endcsname{final}
\expandafter\gdef\csname N@ai.l131k.diffnla.tpr.b3\endcsname{-0.013}
\expandafter\gdef\csname Nlo@ai.l131k.diffnla.tpr.b3\endcsname{-0.053}
\expandafter\gdef\csname Nhi@ai.l131k.diffnla.tpr.b3\endcsname{0.022}
\expandafter\gdef\csname Nn@ai.l131k.diffnla.tpr.b3\endcsname{404}
\expandafter\gdef\csname Nstatus@ai.l131k.diffnla.tpr.b3\endcsname{final}
\expandafter\gdef\csname N@ai.l131k.diffnla.tpr.b4\endcsname{0.021}
\expandafter\gdef\csname Nlo@ai.l131k.diffnla.tpr.b4\endcsname{-0.018}
\expandafter\gdef\csname Nhi@ai.l131k.diffnla.tpr.b4\endcsname{0.062}
\expandafter\gdef\csname Nn@ai.l131k.diffnla.tpr.b4\endcsname{423}
\expandafter\gdef\csname Nstatus@ai.l131k.diffnla.tpr.b4\endcsname{final}
\expandafter\gdef\csname N@ai.l131k.diffnla.tpr.btop\endcsname{-0.046}
\expandafter\gdef\csname Nlo@ai.l131k.diffnla.tpr.btop\endcsname{-0.132}
\expandafter\gdef\csname Nhi@ai.l131k.diffnla.tpr.btop\endcsname{0.039}
\expandafter\gdef\csname Nn@ai.l131k.diffnla.tpr.btop\endcsname{104}
\expandafter\gdef\csname Nstatus@ai.l131k.diffnla.tpr.btop\endcsname{final}
\expandafter\gdef\csname N@ai.l131k.diffnla.fuzztnr.b0\endcsname{0.005}
\expandafter\gdef\csname Nlo@ai.l131k.diffnla.fuzztnr.b0\endcsname{0.003}
\expandafter\gdef\csname Nhi@ai.l131k.diffnla.fuzztnr.b0\endcsname{0.007}
\expandafter\gdef\csname Nn@ai.l131k.diffnla.fuzztnr.b0\endcsname{422}
\expandafter\gdef\csname Nstatus@ai.l131k.diffnla.fuzztnr.b0\endcsname{final}
\expandafter\gdef\csname N@ai.l131k.diffnla.fuzztpr.b1\endcsname{-0.030}
\expandafter\gdef\csname Nlo@ai.l131k.diffnla.fuzztpr.b1\endcsname{-0.051}
\expandafter\gdef\csname Nhi@ai.l131k.diffnla.fuzztpr.b1\endcsname{-0.011}
\expandafter\gdef\csname Nn@ai.l131k.diffnla.fuzztpr.b1\endcsname{405}
\expandafter\gdef\csname Nstatus@ai.l131k.diffnla.fuzztpr.b1\endcsname{final}
\expandafter\gdef\csname N@ai.l131k.diffnla.fuzztpr.b2\endcsname{-0.038}
\expandafter\gdef\csname Nlo@ai.l131k.diffnla.fuzztpr.b2\endcsname{-0.068}
\expandafter\gdef\csname Nhi@ai.l131k.diffnla.fuzztpr.b2\endcsname{-0.010}
\expandafter\gdef\csname Nn@ai.l131k.diffnla.fuzztpr.b2\endcsname{418}
\expandafter\gdef\csname Nstatus@ai.l131k.diffnla.fuzztpr.b2\endcsname{final}
\expandafter\gdef\csname N@ai.l131k.diffnla.fuzztpr.b3\endcsname{-0.005}
\expandafter\gdef\csname Nlo@ai.l131k.diffnla.fuzztpr.b3\endcsname{-0.039}
\expandafter\gdef\csname Nhi@ai.l131k.diffnla.fuzztpr.b3\endcsname{0.030}
\expandafter\gdef\csname Nn@ai.l131k.diffnla.fuzztpr.b3\endcsname{402}
\expandafter\gdef\csname Nstatus@ai.l131k.diffnla.fuzztpr.b3\endcsname{final}
\expandafter\gdef\csname N@ai.l131k.diffnla.fuzztpr.b4\endcsname{0.004}
\expandafter\gdef\csname Nlo@ai.l131k.diffnla.fuzztpr.b4\endcsname{-0.035}
\expandafter\gdef\csname Nhi@ai.l131k.diffnla.fuzztpr.b4\endcsname{0.044}
\expandafter\gdef\csname Nn@ai.l131k.diffnla.fuzztpr.b4\endcsname{419}
\expandafter\gdef\csname Nstatus@ai.l131k.diffnla.fuzztpr.b4\endcsname{final}
\expandafter\gdef\csname N@ai.l131k.diffnla.fuzztpr.btop\endcsname{-0.093}
\expandafter\gdef\csname Nlo@ai.l131k.diffnla.fuzztpr.btop\endcsname{-0.170}
\expandafter\gdef\csname Nhi@ai.l131k.diffnla.fuzztpr.btop\endcsname{-0.011}
\expandafter\gdef\csname Nn@ai.l131k.diffnla.fuzztpr.btop\endcsname{104}
\expandafter\gdef\csname Nstatus@ai.l131k.diffnla.fuzztpr.btop\endcsname{final}
\expandafter\gdef\csname N@fid.ra.diff.cos.bo64\endcsname{0.2779}
\expandafter\gdef\csname Nse@fid.ra.diff.cos.bo64\endcsname{0.0049}
\expandafter\gdef\csname Nlo@fid.ra.diff.cos.bo64\endcsname{0.2681}
\expandafter\gdef\csname Nhi@fid.ra.diff.cos.bo64\endcsname{0.2874}
\expandafter\gdef\csname Nn@fid.ra.diff.cos.bo64\endcsname{486}
\expandafter\gdef\csname Nstatus@fid.ra.diff.cos.bo64\endcsname{final}
\expandafter\gdef\csname N@fid.ra.ex.win.bo64\endcsname{0.9918}
\expandafter\gdef\csname Nse@fid.ra.ex.win.bo64\endcsname{0.0041}
\expandafter\gdef\csname Nlo@fid.ra.ex.win.bo64\endcsname{0.9831}
\expandafter\gdef\csname Nhi@fid.ra.ex.win.bo64\endcsname{0.9980}
\expandafter\gdef\csname Nn@fid.ra.ex.win.bo64\endcsname{486}
\expandafter\gdef\csname Nstatus@fid.ra.ex.win.bo64\endcsname{final}
\expandafter\gdef\csname N@sae.l131k.all.tau\endcsname{1.5846}
\expandafter\gdef\csname Nn@sae.l131k.all.tau\endcsname{512}
\expandafter\gdef\csname Nstatus@sae.l131k.all.tau\endcsname{final}
\expandafter\gdef\csname N@sae.l131k.ex.zero.bo8\endcsname{7}
\expandafter\gdef\csname Nn@sae.l131k.ex.zero.bo8\endcsname{512}
\expandafter\gdef\csname Nstatus@sae.l131k.ex.zero.bo8\endcsname{final}
\expandafter\gdef\csname N@cost.ra.gcg.gpus\endcsname{274.5}
\expandafter\gdef\csname Nn@cost.ra.gcg.gpus\endcsname{64}
\expandafter\gdef\csname Nstatus@cost.ra.gcg.gpus\endcsname{final}
\expandafter\gdef\csname N@cost.ra.epo.gpus\endcsname{932.0}
\expandafter\gdef\csname Nn@cost.ra.epo.gpus\endcsname{64}
\expandafter\gdef\csname Nstatus@cost.ra.epo.gpus\endcsname{final}
\expandafter\gdef\csname N@cost.ra.epo.gcg.ratio\endcsname{3.4}
\expandafter\gdef\csname Nn@cost.ra.epo.gcg.ratio\endcsname{64}
\expandafter\gdef\csname Nstatus@cost.ra.epo.gcg.ratio\endcsname{final}
\expandafter\gdef\csname N@cost.ra.gcg.fwd\endcsname{76{,}800}
\expandafter\gdef\csname Nn@cost.ra.gcg.fwd\endcsname{64}
\expandafter\gdef\csname Nstatus@cost.ra.gcg.fwd\endcsname{final}
\expandafter\gdef\csname N@cost.ra.epo.fwd\endcsname{76{,}500}
\expandafter\gdef\csname Nn@cost.ra.epo.fwd\endcsname{64}
\expandafter\gdef\csname Nstatus@cost.ra.epo.fwd\endcsname{final}
\expandafter\gdef\csname N@cost.ra.ex.gpus\endcsname{3.35}
\expandafter\gdef\csname Nn@cost.ra.ex.gpus\endcsname{512}
\expandafter\gdef\csname Nstatus@cost.ra.ex.gpus\endcsname{final}
\expandafter\gdef\csname N@cost.ra.epo.ex.ratio\endcsname{280}
\expandafter\gdef\csname Nstatus@cost.ra.epo.ex.ratio\endcsname{final}
\expandafter\gdef\csname N@cost.ra.gcg.ex.ratio\endcsname{82}
\expandafter\gdef\csname Nstatus@cost.ra.gcg.ex.ratio\endcsname{final}
\expandafter\gdef\csname N@fid.ra.gcg.depo.unc\endcsname{0.0340}
\expandafter\gdef\csname Nse@fid.ra.gcg.depo.unc\endcsname{0.0167}
\expandafter\gdef\csname Nlo@fid.ra.gcg.depo.unc\endcsname{0.0012}
\expandafter\gdef\csname Nhi@fid.ra.gcg.depo.unc\endcsname{0.0669}
\expandafter\gdef\csname Nn@fid.ra.gcg.depo.unc\endcsname{64}
\expandafter\gdef\csname Nstatus@fid.ra.gcg.depo.unc\endcsname{final}
\expandafter\gdef\csname N@fid.ra.gcg.depo\endcsname{-0.0057}
\expandafter\gdef\csname Nse@fid.ra.gcg.depo\endcsname{0.0199}
\expandafter\gdef\csname Nlo@fid.ra.gcg.depo\endcsname{-0.0446}
\expandafter\gdef\csname Nhi@fid.ra.gcg.depo\endcsname{0.0333}
\expandafter\gdef\csname Nn@fid.ra.gcg.depo\endcsname{64}
\expandafter\gdef\csname Nstatus@fid.ra.gcg.depo\endcsname{final}
\expandafter\gdef\csname N@fid.ra.epo.nll\endcsname{4.91}
\expandafter\gdef\csname Nse@fid.ra.epo.nll\endcsname{0.1088}
\expandafter\gdef\csname Nlo@fid.ra.epo.nll\endcsname{4.70}
\expandafter\gdef\csname Nhi@fid.ra.epo.nll\endcsname{5.12}
\expandafter\gdef\csname Nn@fid.ra.epo.nll\endcsname{64}
\expandafter\gdef\csname Nstatus@fid.ra.epo.nll\endcsname{final}
\expandafter\gdef\csname N@fid.ra.gcg.nll\endcsname{13.22}
\expandafter\gdef\csname Nse@fid.ra.gcg.nll\endcsname{0.0475}
\expandafter\gdef\csname Nlo@fid.ra.gcg.nll\endcsname{13.13}
\expandafter\gdef\csname Nhi@fid.ra.gcg.nll\endcsname{13.31}
\expandafter\gdef\csname Nn@fid.ra.gcg.nll\endcsname{64}
\expandafter\gdef\csname Nstatus@fid.ra.gcg.nll\endcsname{final}
\expandafter\gdef\csname N@sae.s2m.all.tau\endcsname{1.6828}
\expandafter\gdef\csname Nn@sae.s2m.all.tau\endcsname{512}
\expandafter\gdef\csname Nstatus@sae.s2m.all.tau\endcsname{final}
\expandafter\gdef\csname N@fid.ra.sft.cos.bo1\endcsname{0.3931}
\expandafter\gdef\csname Nse@fid.ra.sft.cos.bo1\endcsname{0.0077}
\expandafter\gdef\csname Nlo@fid.ra.sft.cos.bo1\endcsname{0.3780}
\expandafter\gdef\csname Nhi@fid.ra.sft.cos.bo1\endcsname{0.4082}
\expandafter\gdef\csname Nn@fid.ra.sft.cos.bo1\endcsname{486}
\expandafter\gdef\csname Nstatus@fid.ra.sft.cos.bo1\endcsname{final}
\expandafter\gdef\csname N@fid.ra.sft.dex.bo1\endcsname{-0.3659}
\expandafter\gdef\csname Nse@fid.ra.sft.dex.bo1\endcsname{0.0075}
\expandafter\gdef\csname Nlo@fid.ra.sft.dex.bo1\endcsname{-0.3806}
\expandafter\gdef\csname Nhi@fid.ra.sft.dex.bo1\endcsname{-0.3511}
\expandafter\gdef\csname Nn@fid.ra.sft.dex.bo1\endcsname{486}
\expandafter\gdef\csname Nstatus@fid.ra.sft.dex.bo1\endcsname{final}
\expandafter\gdef\csname N@fid.ra.sft.cos.bo8\endcsname{0.5995}
\expandafter\gdef\csname Nse@fid.ra.sft.cos.bo8\endcsname{0.0083}
\expandafter\gdef\csname Nlo@fid.ra.sft.cos.bo8\endcsname{0.5833}
\expandafter\gdef\csname Nhi@fid.ra.sft.cos.bo8\endcsname{0.6157}
\expandafter\gdef\csname Nn@fid.ra.sft.cos.bo8\endcsname{486}
\expandafter\gdef\csname Nstatus@fid.ra.sft.cos.bo8\endcsname{final}
\expandafter\gdef\csname N@fid.ra.sft.dex.bo8\endcsname{-0.2224}
\expandafter\gdef\csname Nse@fid.ra.sft.dex.bo8\endcsname{0.0065}
\expandafter\gdef\csname Nlo@fid.ra.sft.dex.bo8\endcsname{-0.2352}
\expandafter\gdef\csname Nhi@fid.ra.sft.dex.bo8\endcsname{-0.2096}
\expandafter\gdef\csname Nn@fid.ra.sft.dex.bo8\endcsname{486}
\expandafter\gdef\csname Nstatus@fid.ra.sft.dex.bo8\endcsname{final}
\expandafter\gdef\csname N@fid.ra.sft.cos.bo64\endcsname{0.7032}
\expandafter\gdef\csname Nse@fid.ra.sft.cos.bo64\endcsname{0.0075}
\expandafter\gdef\csname Nlo@fid.ra.sft.cos.bo64\endcsname{0.6884}
\expandafter\gdef\csname Nhi@fid.ra.sft.cos.bo64\endcsname{0.7180}
\expandafter\gdef\csname Nn@fid.ra.sft.cos.bo64\endcsname{486}
\expandafter\gdef\csname Nstatus@fid.ra.sft.cos.bo64\endcsname{final}
\expandafter\gdef\csname N@fid.ra.sft.dex.bo64\endcsname{-0.1466}
\expandafter\gdef\csname Nse@fid.ra.sft.dex.bo64\endcsname{0.0056}
\expandafter\gdef\csname Nlo@fid.ra.sft.dex.bo64\endcsname{-0.1577}
\expandafter\gdef\csname Nhi@fid.ra.sft.dex.bo64\endcsname{-0.1356}
\expandafter\gdef\csname Nn@fid.ra.sft.dex.bo64\endcsname{486}
\expandafter\gdef\csname Nstatus@fid.ra.sft.dex.bo64\endcsname{final}
\expandafter\gdef\csname N@fid.ra.nla.cos.bo2\endcsname{0.6121}
\expandafter\gdef\csname Nse@fid.ra.nla.cos.bo2\endcsname{0.0063}
\expandafter\gdef\csname Nlo@fid.ra.nla.cos.bo2\endcsname{0.5996}
\expandafter\gdef\csname Nhi@fid.ra.nla.cos.bo2\endcsname{0.6245}
\expandafter\gdef\csname Nn@fid.ra.nla.cos.bo2\endcsname{486}
\expandafter\gdef\csname Nstatus@fid.ra.nla.cos.bo2\endcsname{final}
\expandafter\gdef\csname N@fid.ra.nla.cos.bo4\endcsname{0.6510}
\expandafter\gdef\csname Nse@fid.ra.nla.cos.bo4\endcsname{0.0068}
\expandafter\gdef\csname Nlo@fid.ra.nla.cos.bo4\endcsname{0.6377}
\expandafter\gdef\csname Nhi@fid.ra.nla.cos.bo4\endcsname{0.6643}
\expandafter\gdef\csname Nn@fid.ra.nla.cos.bo4\endcsname{486}
\expandafter\gdef\csname Nstatus@fid.ra.nla.cos.bo4\endcsname{final}
\expandafter\gdef\csname N@fid.ra.nla.dex.bo2\endcsname{0.2098}
\expandafter\gdef\csname Nse@fid.ra.nla.dex.bo2\endcsname{0.0053}
\expandafter\gdef\csname Nlo@fid.ra.nla.dex.bo2\endcsname{0.1995}
\expandafter\gdef\csname Nhi@fid.ra.nla.dex.bo2\endcsname{0.2202}
\expandafter\gdef\csname Nn@fid.ra.nla.dex.bo2\endcsname{486}
\expandafter\gdef\csname Nstatus@fid.ra.nla.dex.bo2\endcsname{final}
\expandafter\gdef\csname N@fid.ra.nla.dcorp.bo2\endcsname{0.0401}
\expandafter\gdef\csname Nse@fid.ra.nla.dcorp.bo2\endcsname{0.0072}
\expandafter\gdef\csname Nlo@fid.ra.nla.dcorp.bo2\endcsname{0.0259}
\expandafter\gdef\csname Nhi@fid.ra.nla.dcorp.bo2\endcsname{0.0543}
\expandafter\gdef\csname Nn@fid.ra.nla.dcorp.bo2\endcsname{486}
\expandafter\gdef\csname Nstatus@fid.ra.nla.dcorp.bo2\endcsname{final}
\expandafter\gdef\csname N@fid.ra.nla.win.bo2\endcsname{0.6173}
\expandafter\gdef\csname Nse@fid.ra.nla.win.bo2\endcsname{0.0217}
\expandafter\gdef\csname Nn@fid.ra.nla.win.bo2\endcsname{486}
\expandafter\gdef\csname Nstatus@fid.ra.nla.win.bo2\endcsname{final}
\expandafter\gdef\csname N@ood.c.encorp10.cos\endcsname{0.4554}
\expandafter\gdef\csname Nn@ood.c.encorp10.cos\endcsname{512}
\expandafter\gdef\csname Nstatus@ood.c.encorp10.cos\endcsname{final}
\expandafter\gdef\csname N@ood.c.endiff.cos.bo8\endcsname{0.2817}
\expandafter\gdef\csname Nse@ood.c.endiff.cos.bo8\endcsname{0.0062}
\expandafter\gdef\csname Nlo@ood.c.endiff.cos.bo8\endcsname{0.2685}
\expandafter\gdef\csname Nhi@ood.c.endiff.cos.bo8\endcsname{0.2939}
\expandafter\gdef\csname Nn@ood.c.endiff.cos.bo8\endcsname{512}
\expandafter\gdef\csname Nstatus@ood.c.endiff.cos.bo8\endcsname{final}
\expandafter\gdef\csname N@ood.c.endiff.verdict\endcsname{exceeds}
\expandafter\gdef\csname Nn@ood.c.endiff.verdict\endcsname{512}
\expandafter\gdef\csname Nstatus@ood.c.endiff.verdict\endcsname{final}
\expandafter\gdef\csname N@ood.go.encorp10.cos\endcsname{0.4131}
\expandafter\gdef\csname Nn@ood.go.encorp10.cos\endcsname{512}
\expandafter\gdef\csname Nstatus@ood.go.encorp10.cos\endcsname{final}
\expandafter\gdef\csname N@ood.go.endiff.cos.bo8\endcsname{0.2834}
\expandafter\gdef\csname Nse@ood.go.endiff.cos.bo8\endcsname{0.0070}
\expandafter\gdef\csname Nlo@ood.go.endiff.cos.bo8\endcsname{0.2697}
\expandafter\gdef\csname Nhi@ood.go.endiff.cos.bo8\endcsname{0.2969}
\expandafter\gdef\csname Nn@ood.go.endiff.cos.bo8\endcsname{512}
\expandafter\gdef\csname Nstatus@ood.go.endiff.cos.bo8\endcsname{final}
\expandafter\gdef\csname N@ood.go.endiff.verdict\endcsname{exceeds}
\expandafter\gdef\csname Nn@ood.go.endiff.verdict\endcsname{512}
\expandafter\gdef\csname Nstatus@ood.go.endiff.verdict\endcsname{final}
\expandafter\gdef\csname N@ood.haskell.encorp10.cos\endcsname{0.4279}
\expandafter\gdef\csname Nn@ood.haskell.encorp10.cos\endcsname{512}
\expandafter\gdef\csname Nstatus@ood.haskell.encorp10.cos\endcsname{final}
\expandafter\gdef\csname N@ood.haskell.endiff.cos.bo8\endcsname{0.2798}
\expandafter\gdef\csname Nse@ood.haskell.endiff.cos.bo8\endcsname{0.0059}
\expandafter\gdef\csname Nlo@ood.haskell.endiff.cos.bo8\endcsname{0.2681}
\expandafter\gdef\csname Nhi@ood.haskell.endiff.cos.bo8\endcsname{0.2909}
\expandafter\gdef\csname Nn@ood.haskell.endiff.cos.bo8\endcsname{512}
\expandafter\gdef\csname Nstatus@ood.haskell.endiff.cos.bo8\endcsname{final}
\expandafter\gdef\csname N@ood.haskell.endiff.verdict\endcsname{exceeds}
\expandafter\gdef\csname Nn@ood.haskell.endiff.verdict\endcsname{512}
\expandafter\gdef\csname Nstatus@ood.haskell.endiff.verdict\endcsname{final}
\expandafter\gdef\csname N@ood.javascript.encorp10.cos\endcsname{0.4155}
\expandafter\gdef\csname Nn@ood.javascript.encorp10.cos\endcsname{512}
\expandafter\gdef\csname Nstatus@ood.javascript.encorp10.cos\endcsname{final}
\expandafter\gdef\csname N@ood.javascript.endiff.cos.bo8\endcsname{0.2837}
\expandafter\gdef\csname Nse@ood.javascript.endiff.cos.bo8\endcsname{0.0072}
\expandafter\gdef\csname Nlo@ood.javascript.endiff.cos.bo8\endcsname{0.2698}
\expandafter\gdef\csname Nhi@ood.javascript.endiff.cos.bo8\endcsname{0.2980}
\expandafter\gdef\csname Nn@ood.javascript.endiff.cos.bo8\endcsname{512}
\expandafter\gdef\csname Nstatus@ood.javascript.endiff.cos.bo8\endcsname{final}
\expandafter\gdef\csname N@ood.javascript.endiff.verdict\endcsname{exceeds}
\expandafter\gdef\csname Nn@ood.javascript.endiff.verdict\endcsname{512}
\expandafter\gdef\csname Nstatus@ood.javascript.endiff.verdict\endcsname{final}
\expandafter\gdef\csname N@ood.python.encorp10.cos\endcsname{0.4476}
\expandafter\gdef\csname Nn@ood.python.encorp10.cos\endcsname{512}
\expandafter\gdef\csname Nstatus@ood.python.encorp10.cos\endcsname{final}
\expandafter\gdef\csname N@ood.python.endiff.cos.bo8\endcsname{0.2714}
\expandafter\gdef\csname Nse@ood.python.endiff.cos.bo8\endcsname{0.0066}
\expandafter\gdef\csname Nlo@ood.python.endiff.cos.bo8\endcsname{0.2585}
\expandafter\gdef\csname Nhi@ood.python.endiff.cos.bo8\endcsname{0.2841}
\expandafter\gdef\csname Nn@ood.python.endiff.cos.bo8\endcsname{512}
\expandafter\gdef\csname Nstatus@ood.python.endiff.cos.bo8\endcsname{final}
\expandafter\gdef\csname N@ood.python.endiff.verdict\endcsname{exceeds}
\expandafter\gdef\csname Nn@ood.python.endiff.verdict\endcsname{512}
\expandafter\gdef\csname Nstatus@ood.python.endiff.verdict\endcsname{final}
\expandafter\gdef\csname N@ood.rust.encorp10.cos\endcsname{0.4017}
\expandafter\gdef\csname Nn@ood.rust.encorp10.cos\endcsname{512}
\expandafter\gdef\csname Nstatus@ood.rust.encorp10.cos\endcsname{final}
\expandafter\gdef\csname N@ood.rust.endiff.cos.bo8\endcsname{0.3006}
\expandafter\gdef\csname Nse@ood.rust.endiff.cos.bo8\endcsname{0.0066}
\expandafter\gdef\csname Nlo@ood.rust.endiff.cos.bo8\endcsname{0.2874}
\expandafter\gdef\csname Nhi@ood.rust.endiff.cos.bo8\endcsname{0.3136}
\expandafter\gdef\csname Nn@ood.rust.endiff.cos.bo8\endcsname{512}
\expandafter\gdef\csname Nstatus@ood.rust.endiff.cos.bo8\endcsname{final}
\expandafter\gdef\csname N@ood.rust.endiff.verdict\endcsname{exceeds}
\expandafter\gdef\csname Nn@ood.rust.endiff.verdict\endcsname{512}
\expandafter\gdef\csname Nstatus@ood.rust.endiff.verdict\endcsname{final}
\expandafter\gdef\csname N@ood.shell.encorp10.cos\endcsname{0.4582}
\expandafter\gdef\csname Nn@ood.shell.encorp10.cos\endcsname{512}
\expandafter\gdef\csname Nstatus@ood.shell.encorp10.cos\endcsname{final}
\expandafter\gdef\csname N@ood.shell.endiff.cos.bo8\endcsname{0.2647}
\expandafter\gdef\csname Nse@ood.shell.endiff.cos.bo8\endcsname{0.0063}
\expandafter\gdef\csname Nlo@ood.shell.endiff.cos.bo8\endcsname{0.2523}
\expandafter\gdef\csname Nhi@ood.shell.endiff.cos.bo8\endcsname{0.2764}
\expandafter\gdef\csname Nn@ood.shell.endiff.cos.bo8\endcsname{512}
\expandafter\gdef\csname Nstatus@ood.shell.endiff.cos.bo8\endcsname{final}
\expandafter\gdef\csname N@ood.shell.endiff.verdict\endcsname{exceeds}
\expandafter\gdef\csname Nn@ood.shell.endiff.verdict\endcsname{512}
\expandafter\gdef\csname Nstatus@ood.shell.endiff.verdict\endcsname{final}
\expandafter\gdef\csname N@ood.sql.encorp10.cos\endcsname{0.4815}
\expandafter\gdef\csname Nn@ood.sql.encorp10.cos\endcsname{512}
\expandafter\gdef\csname Nstatus@ood.sql.encorp10.cos\endcsname{final}
\expandafter\gdef\csname N@ood.sql.endiff.cos.bo8\endcsname{0.2193}
\expandafter\gdef\csname Nse@ood.sql.endiff.cos.bo8\endcsname{0.0071}
\expandafter\gdef\csname Nlo@ood.sql.endiff.cos.bo8\endcsname{0.2054}
\expandafter\gdef\csname Nhi@ood.sql.endiff.cos.bo8\endcsname{0.2331}
\expandafter\gdef\csname Nn@ood.sql.endiff.cos.bo8\endcsname{512}
\expandafter\gdef\csname Nstatus@ood.sql.endiff.cos.bo8\endcsname{final}
\expandafter\gdef\csname N@ood.sql.endiff.verdict\endcsname{exceeds}
\expandafter\gdef\csname Nn@ood.sql.endiff.verdict\endcsname{512}
\expandafter\gdef\csname Nstatus@ood.sql.endiff.verdict\endcsname{final}
\expandafter\gdef\csname N@ood.en.encorp10.cos\endcsname{0.5795}
\expandafter\gdef\csname Nn@ood.en.encorp10.cos\endcsname{512}
\expandafter\gdef\csname Nstatus@ood.en.encorp10.cos\endcsname{final}
\expandafter\gdef\csname N@ood.en.endiff.cos.bo8\endcsname{0.2482}
\expandafter\gdef\csname Nse@ood.en.endiff.cos.bo8\endcsname{0.0052}
\expandafter\gdef\csname Nlo@ood.en.endiff.cos.bo8\endcsname{0.2378}
\expandafter\gdef\csname Nhi@ood.en.endiff.cos.bo8\endcsname{0.2583}
\expandafter\gdef\csname Nn@ood.en.endiff.cos.bo8\endcsname{512}
\expandafter\gdef\csname Nstatus@ood.en.endiff.cos.bo8\endcsname{final}
\expandafter\gdef\csname N@ood.en.endiff.verdict\endcsname{exceeds}
\expandafter\gdef\csname Nn@ood.en.endiff.verdict\endcsname{512}
\expandafter\gdef\csname Nstatus@ood.en.endiff.verdict\endcsname{final}
\expandafter\gdef\csname N@ood.zh.encorp10.cos\endcsname{0.4462}
\expandafter\gdef\csname Nn@ood.zh.encorp10.cos\endcsname{512}
\expandafter\gdef\csname Nstatus@ood.zh.encorp10.cos\endcsname{final}
\expandafter\gdef\csname N@ood.zh.endiff.cos.bo8\endcsname{0.3629}
\expandafter\gdef\csname Nse@ood.zh.endiff.cos.bo8\endcsname{0.0059}
\expandafter\gdef\csname Nlo@ood.zh.endiff.cos.bo8\endcsname{0.3512}
\expandafter\gdef\csname Nhi@ood.zh.endiff.cos.bo8\endcsname{0.3738}
\expandafter\gdef\csname Nn@ood.zh.endiff.cos.bo8\endcsname{512}
\expandafter\gdef\csname Nstatus@ood.zh.endiff.cos.bo8\endcsname{final}
\expandafter\gdef\csname N@ood.zh.endiff.verdict\endcsname{exceeds}
\expandafter\gdef\csname Nn@ood.zh.endiff.verdict\endcsname{512}
\expandafter\gdef\csname Nstatus@ood.zh.endiff.verdict\endcsname{final}
\expandafter\gdef\csname N@ood.arb.encorp10.cos\endcsname{0.4618}
\expandafter\gdef\csname Nn@ood.arb.encorp10.cos\endcsname{512}
\expandafter\gdef\csname Nstatus@ood.arb.encorp10.cos\endcsname{final}
\expandafter\gdef\csname N@ood.arb.endiff.cos.bo8\endcsname{0.3198}
\expandafter\gdef\csname Nse@ood.arb.endiff.cos.bo8\endcsname{0.0053}
\expandafter\gdef\csname Nlo@ood.arb.endiff.cos.bo8\endcsname{0.3093}
\expandafter\gdef\csname Nhi@ood.arb.endiff.cos.bo8\endcsname{0.3304}
\expandafter\gdef\csname Nn@ood.arb.endiff.cos.bo8\endcsname{512}
\expandafter\gdef\csname Nstatus@ood.arb.endiff.cos.bo8\endcsname{final}
\expandafter\gdef\csname N@ood.arb.endiff.verdict\endcsname{exceeds}
\expandafter\gdef\csname Nn@ood.arb.endiff.verdict\endcsname{512}
\expandafter\gdef\csname Nstatus@ood.arb.endiff.verdict\endcsname{final}
\expandafter\gdef\csname N@ood.ces.encorp10.cos\endcsname{0.4665}
\expandafter\gdef\csname Nn@ood.ces.encorp10.cos\endcsname{512}
\expandafter\gdef\csname Nstatus@ood.ces.encorp10.cos\endcsname{final}
\expandafter\gdef\csname N@ood.ces.endiff.cos.bo8\endcsname{0.3066}
\expandafter\gdef\csname Nse@ood.ces.endiff.cos.bo8\endcsname{0.0055}
\expandafter\gdef\csname Nlo@ood.ces.endiff.cos.bo8\endcsname{0.2956}
\expandafter\gdef\csname Nhi@ood.ces.endiff.cos.bo8\endcsname{0.3173}
\expandafter\gdef\csname Nn@ood.ces.endiff.cos.bo8\endcsname{512}
\expandafter\gdef\csname Nstatus@ood.ces.endiff.cos.bo8\endcsname{final}
\expandafter\gdef\csname N@ood.ces.endiff.verdict\endcsname{exceeds}
\expandafter\gdef\csname Nn@ood.ces.endiff.verdict\endcsname{512}
\expandafter\gdef\csname Nstatus@ood.ces.endiff.verdict\endcsname{final}
\expandafter\gdef\csname N@ood.cmn.encorp10.cos\endcsname{0.4528}
\expandafter\gdef\csname Nn@ood.cmn.encorp10.cos\endcsname{512}
\expandafter\gdef\csname Nstatus@ood.cmn.encorp10.cos\endcsname{final}
\expandafter\gdef\csname N@ood.cmn.endiff.cos.bo8\endcsname{0.3543}
\expandafter\gdef\csname Nse@ood.cmn.endiff.cos.bo8\endcsname{0.0055}
\expandafter\gdef\csname Nlo@ood.cmn.endiff.cos.bo8\endcsname{0.3433}
\expandafter\gdef\csname Nhi@ood.cmn.endiff.cos.bo8\endcsname{0.3648}
\expandafter\gdef\csname Nn@ood.cmn.endiff.cos.bo8\endcsname{512}
\expandafter\gdef\csname Nstatus@ood.cmn.endiff.cos.bo8\endcsname{final}
\expandafter\gdef\csname N@ood.cmn.endiff.verdict\endcsname{exceeds}
\expandafter\gdef\csname Nn@ood.cmn.endiff.verdict\endcsname{512}
\expandafter\gdef\csname Nstatus@ood.cmn.endiff.verdict\endcsname{final}
\expandafter\gdef\csname N@ood.ell.encorp10.cos\endcsname{0.4607}
\expandafter\gdef\csname Nn@ood.ell.encorp10.cos\endcsname{512}
\expandafter\gdef\csname Nstatus@ood.ell.encorp10.cos\endcsname{final}
\expandafter\gdef\csname N@ood.ell.endiff.cos.bo8\endcsname{0.2541}
\expandafter\gdef\csname Nse@ood.ell.endiff.cos.bo8\endcsname{0.0054}
\expandafter\gdef\csname Nlo@ood.ell.endiff.cos.bo8\endcsname{0.2438}
\expandafter\gdef\csname Nhi@ood.ell.endiff.cos.bo8\endcsname{0.2646}
\expandafter\gdef\csname Nn@ood.ell.endiff.cos.bo8\endcsname{512}
\expandafter\gdef\csname Nstatus@ood.ell.endiff.cos.bo8\endcsname{final}
\expandafter\gdef\csname N@ood.ell.endiff.verdict\endcsname{exceeds}
\expandafter\gdef\csname Nn@ood.ell.endiff.verdict\endcsname{512}
\expandafter\gdef\csname Nstatus@ood.ell.endiff.verdict\endcsname{final}
\expandafter\gdef\csname N@ood.hin.encorp10.cos\endcsname{0.4388}
\expandafter\gdef\csname Nn@ood.hin.encorp10.cos\endcsname{512}
\expandafter\gdef\csname Nstatus@ood.hin.encorp10.cos\endcsname{final}
\expandafter\gdef\csname N@ood.hin.endiff.cos.bo8\endcsname{0.3135}
\expandafter\gdef\csname Nse@ood.hin.endiff.cos.bo8\endcsname{0.0054}
\expandafter\gdef\csname Nlo@ood.hin.endiff.cos.bo8\endcsname{0.3029}
\expandafter\gdef\csname Nhi@ood.hin.endiff.cos.bo8\endcsname{0.3243}
\expandafter\gdef\csname Nn@ood.hin.endiff.cos.bo8\endcsname{512}
\expandafter\gdef\csname Nstatus@ood.hin.endiff.cos.bo8\endcsname{final}
\expandafter\gdef\csname N@ood.hin.endiff.verdict\endcsname{exceeds}
\expandafter\gdef\csname Nn@ood.hin.endiff.verdict\endcsname{512}
\expandafter\gdef\csname Nstatus@ood.hin.endiff.verdict\endcsname{final}
\expandafter\gdef\csname N@ood.jpn.encorp10.cos\endcsname{0.4234}
\expandafter\gdef\csname Nn@ood.jpn.encorp10.cos\endcsname{512}
\expandafter\gdef\csname Nstatus@ood.jpn.encorp10.cos\endcsname{final}
\expandafter\gdef\csname N@ood.jpn.endiff.cos.bo8\endcsname{0.3611}
\expandafter\gdef\csname Nse@ood.jpn.endiff.cos.bo8\endcsname{0.0062}
\expandafter\gdef\csname Nlo@ood.jpn.endiff.cos.bo8\endcsname{0.3487}
\expandafter\gdef\csname Nhi@ood.jpn.endiff.cos.bo8\endcsname{0.3735}
\expandafter\gdef\csname Nn@ood.jpn.endiff.cos.bo8\endcsname{512}
\expandafter\gdef\csname Nstatus@ood.jpn.endiff.cos.bo8\endcsname{final}
\expandafter\gdef\csname N@ood.jpn.endiff.verdict\endcsname{exceeds}
\expandafter\gdef\csname Nn@ood.jpn.endiff.verdict\endcsname{512}
\expandafter\gdef\csname Nstatus@ood.jpn.endiff.verdict\endcsname{final}
\expandafter\gdef\csname N@ood.rus.encorp10.cos\endcsname{0.4791}
\expandafter\gdef\csname Nn@ood.rus.encorp10.cos\endcsname{512}
\expandafter\gdef\csname Nstatus@ood.rus.encorp10.cos\endcsname{final}
\expandafter\gdef\csname N@ood.rus.endiff.cos.bo8\endcsname{0.3235}
\expandafter\gdef\csname Nse@ood.rus.endiff.cos.bo8\endcsname{0.0053}
\expandafter\gdef\csname Nlo@ood.rus.endiff.cos.bo8\endcsname{0.3133}
\expandafter\gdef\csname Nhi@ood.rus.endiff.cos.bo8\endcsname{0.3335}
\expandafter\gdef\csname Nn@ood.rus.endiff.cos.bo8\endcsname{512}
\expandafter\gdef\csname Nstatus@ood.rus.endiff.cos.bo8\endcsname{final}
\expandafter\gdef\csname N@ood.rus.endiff.verdict\endcsname{exceeds}
\expandafter\gdef\csname Nn@ood.rus.endiff.verdict\endcsname{512}
\expandafter\gdef\csname Nstatus@ood.rus.endiff.verdict\endcsname{final}
\expandafter\gdef\csname N@ood.tha.encorp10.cos\endcsname{0.4205}
\expandafter\gdef\csname Nn@ood.tha.encorp10.cos\endcsname{512}
\expandafter\gdef\csname Nstatus@ood.tha.encorp10.cos\endcsname{final}
\expandafter\gdef\csname N@ood.tha.endiff.cos.bo8\endcsname{0.3215}
\expandafter\gdef\csname Nse@ood.tha.endiff.cos.bo8\endcsname{0.0060}
\expandafter\gdef\csname Nlo@ood.tha.endiff.cos.bo8\endcsname{0.3098}
\expandafter\gdef\csname Nhi@ood.tha.endiff.cos.bo8\endcsname{0.3329}
\expandafter\gdef\csname Nn@ood.tha.endiff.cos.bo8\endcsname{512}
\expandafter\gdef\csname Nstatus@ood.tha.endiff.cos.bo8\endcsname{final}
\expandafter\gdef\csname N@ood.tha.endiff.verdict\endcsname{exceeds}
\expandafter\gdef\csname Nn@ood.tha.endiff.verdict\endcsname{512}
\expandafter\gdef\csname Nstatus@ood.tha.endiff.verdict\endcsname{final}
\expandafter\gdef\csname N@ood.arxiv.encorp10.cos\endcsname{0.5237}
\expandafter\gdef\csname Nn@ood.arxiv.encorp10.cos\endcsname{512}
\expandafter\gdef\csname Nstatus@ood.arxiv.encorp10.cos\endcsname{final}
\expandafter\gdef\csname N@ood.arxiv.endiff.cos.bo8\endcsname{0.2653}
\expandafter\gdef\csname Nse@ood.arxiv.endiff.cos.bo8\endcsname{0.0054}
\expandafter\gdef\csname Nlo@ood.arxiv.endiff.cos.bo8\endcsname{0.2547}
\expandafter\gdef\csname Nhi@ood.arxiv.endiff.cos.bo8\endcsname{0.2757}
\expandafter\gdef\csname Nn@ood.arxiv.endiff.cos.bo8\endcsname{512}
\expandafter\gdef\csname Nstatus@ood.arxiv.endiff.cos.bo8\endcsname{final}
\expandafter\gdef\csname N@ood.arxiv.endiff.verdict\endcsname{exceeds}
\expandafter\gdef\csname Nn@ood.arxiv.endiff.verdict\endcsname{512}
\expandafter\gdef\csname Nstatus@ood.arxiv.endiff.verdict\endcsname{final}
\expandafter\gdef\csname N@ood.isabelle.encorp10.cos\endcsname{0.4279}
\expandafter\gdef\csname Nn@ood.isabelle.encorp10.cos\endcsname{512}
\expandafter\gdef\csname Nstatus@ood.isabelle.encorp10.cos\endcsname{final}
\expandafter\gdef\csname N@ood.isabelle.endiff.cos.bo8\endcsname{0.2543}
\expandafter\gdef\csname Nse@ood.isabelle.endiff.cos.bo8\endcsname{0.0063}
\expandafter\gdef\csname Nlo@ood.isabelle.endiff.cos.bo8\endcsname{0.2416}
\expandafter\gdef\csname Nhi@ood.isabelle.endiff.cos.bo8\endcsname{0.2669}
\expandafter\gdef\csname Nn@ood.isabelle.endiff.cos.bo8\endcsname{512}
\expandafter\gdef\csname Nstatus@ood.isabelle.endiff.cos.bo8\endcsname{final}
\expandafter\gdef\csname N@ood.isabelle.endiff.verdict\endcsname{exceeds}
\expandafter\gdef\csname Nn@ood.isabelle.endiff.verdict\endcsname{512}
\expandafter\gdef\csname Nstatus@ood.isabelle.endiff.verdict\endcsname{final}
\expandafter\gdef\csname N@ood.lean.encorp10.cos\endcsname{0.4151}
\expandafter\gdef\csname Nn@ood.lean.encorp10.cos\endcsname{512}
\expandafter\gdef\csname Nstatus@ood.lean.encorp10.cos\endcsname{final}
\expandafter\gdef\csname N@ood.lean.endiff.cos.bo8\endcsname{0.3166}
\expandafter\gdef\csname Nse@ood.lean.endiff.cos.bo8\endcsname{0.0057}
\expandafter\gdef\csname Nlo@ood.lean.endiff.cos.bo8\endcsname{0.3056}
\expandafter\gdef\csname Nhi@ood.lean.endiff.cos.bo8\endcsname{0.3273}
\expandafter\gdef\csname Nn@ood.lean.endiff.cos.bo8\endcsname{512}
\expandafter\gdef\csname Nstatus@ood.lean.endiff.cos.bo8\endcsname{final}
\expandafter\gdef\csname N@ood.lean.endiff.verdict\endcsname{exceeds}
\expandafter\gdef\csname Nn@ood.lean.endiff.verdict\endcsname{512}
\expandafter\gdef\csname Nstatus@ood.lean.endiff.verdict\endcsname{final}
\expandafter\gdef\csname N@ood.owm.encorp10.cos\endcsname{0.5072}
\expandafter\gdef\csname Nn@ood.owm.encorp10.cos\endcsname{512}
\expandafter\gdef\csname Nstatus@ood.owm.encorp10.cos\endcsname{final}
\expandafter\gdef\csname N@ood.owm.endiff.cos.bo8\endcsname{0.2659}
\expandafter\gdef\csname Nse@ood.owm.endiff.cos.bo8\endcsname{0.0055}
\expandafter\gdef\csname Nlo@ood.owm.endiff.cos.bo8\endcsname{0.2554}
\expandafter\gdef\csname Nhi@ood.owm.endiff.cos.bo8\endcsname{0.2767}
\expandafter\gdef\csname Nn@ood.owm.endiff.cos.bo8\endcsname{512}
\expandafter\gdef\csname Nstatus@ood.owm.endiff.cos.bo8\endcsname{final}
\expandafter\gdef\csname N@ood.owm.endiff.verdict\endcsname{exceeds}
\expandafter\gdef\csname Nn@ood.owm.endiff.verdict\endcsname{512}
\expandafter\gdef\csname Nstatus@ood.owm.endiff.verdict\endcsname{final}
\expandafter\gdef\csname N@ood.conj.endiff.nexceed\endcsname{21}
\expandafter\gdef\csname Nn@ood.conj.endiff.nexceed\endcsname{21}
\expandafter\gdef\csname Nstatus@ood.conj.endiff.nexceed\endcsname{final}
\expandafter\gdef\csname N@ood.conj.endiff.ninconcl\endcsname{0}
\expandafter\gdef\csname Nn@ood.conj.endiff.ninconcl\endcsname{21}
\expandafter\gdef\csname Nstatus@ood.conj.endiff.ninconcl\endcsname{final}
\expandafter\gdef\csname N@ood.conj.endiff.nreversed\endcsname{0}
\expandafter\gdef\csname Nn@ood.conj.endiff.nreversed\endcsname{21}
\expandafter\gdef\csname Nstatus@ood.conj.endiff.nreversed\endcsname{final}
\expandafter\gdef\csname N@ood.conj.endiff.min\endcsname{0.2193}
\expandafter\gdef\csname Nse@ood.conj.endiff.min\endcsname{0.0071}
\expandafter\gdef\csname Nlo@ood.conj.endiff.min\endcsname{0.2054}
\expandafter\gdef\csname Nhi@ood.conj.endiff.min\endcsname{0.2331}
\expandafter\gdef\csname Nn@ood.conj.endiff.min\endcsname{512}
\expandafter\gdef\csname Nstatus@ood.conj.endiff.min\endcsname{final}
\expandafter\gdef\csname N@sae.l131k.ex.ratiocosref.bo1.q1\endcsname{0.1277}
\expandafter\gdef\csname Nse@sae.l131k.ex.ratiocosref.bo1.q1\endcsname{0.0470}
\expandafter\gdef\csname Nlo@sae.l131k.ex.ratiocosref.bo1.q1\endcsname{0.0469}
\expandafter\gdef\csname Nhi@sae.l131k.ex.ratiocosref.bo1.q1\endcsname{0.2129}
\expandafter\gdef\csname Nn@sae.l131k.ex.ratiocosref.bo1.q1\endcsname{128}
\expandafter\gdef\csname Nstatus@sae.l131k.ex.ratiocosref.bo1.q1\endcsname{final}
\expandafter\gdef\csname N@sae.l131k.ex.ratiocosref.bo1.q2\endcsname{0.8894}
\expandafter\gdef\csname Nse@sae.l131k.ex.ratiocosref.bo1.q2\endcsname{0.0204}
\expandafter\gdef\csname Nlo@sae.l131k.ex.ratiocosref.bo1.q2\endcsname{0.8496}
\expandafter\gdef\csname Nhi@sae.l131k.ex.ratiocosref.bo1.q2\endcsname{0.9243}
\expandafter\gdef\csname Nn@sae.l131k.ex.ratiocosref.bo1.q2\endcsname{128}
\expandafter\gdef\csname Nstatus@sae.l131k.ex.ratiocosref.bo1.q2\endcsname{final}
\expandafter\gdef\csname N@sae.l131k.ex.ratiocosref.bo1.q3\endcsname{0.8249}
\expandafter\gdef\csname Nse@sae.l131k.ex.ratiocosref.bo1.q3\endcsname{0.0202}
\expandafter\gdef\csname Nlo@sae.l131k.ex.ratiocosref.bo1.q3\endcsname{0.7904}
\expandafter\gdef\csname Nhi@sae.l131k.ex.ratiocosref.bo1.q3\endcsname{0.8598}
\expandafter\gdef\csname Nn@sae.l131k.ex.ratiocosref.bo1.q3\endcsname{128}
\expandafter\gdef\csname Nstatus@sae.l131k.ex.ratiocosref.bo1.q3\endcsname{final}
\expandafter\gdef\csname N@sae.l131k.ex.ratiocosref.bo1.q4\endcsname{0.7836}
\expandafter\gdef\csname Nse@sae.l131k.ex.ratiocosref.bo1.q4\endcsname{0.0290}
\expandafter\gdef\csname Nlo@sae.l131k.ex.ratiocosref.bo1.q4\endcsname{0.7304}
\expandafter\gdef\csname Nhi@sae.l131k.ex.ratiocosref.bo1.q4\endcsname{0.8463}
\expandafter\gdef\csname Nn@sae.l131k.ex.ratiocosref.bo1.q4\endcsname{128}
\expandafter\gdef\csname Nstatus@sae.l131k.ex.ratiocosref.bo1.q4\endcsname{final}
\expandafter\gdef\csname N@sae.l131k.ex.ratiocosref.bo1\endcsname{0.7934}
\expandafter\gdef\csname Nse@sae.l131k.ex.ratiocosref.bo1\endcsname{0.0156}
\expandafter\gdef\csname Nlo@sae.l131k.ex.ratiocosref.bo1\endcsname{0.7574}
\expandafter\gdef\csname Nhi@sae.l131k.ex.ratiocosref.bo1\endcsname{0.8165}
\expandafter\gdef\csname Nn@sae.l131k.ex.ratiocosref.bo1\endcsname{512}
\expandafter\gdef\csname Nstatus@sae.l131k.ex.ratiocosref.bo1\endcsname{final}
\expandafter\gdef\csname N@sae.l131k.ex.ratiocosref.bo8.q1\endcsname{0.2492}
\expandafter\gdef\csname Nse@sae.l131k.ex.ratiocosref.bo8.q1\endcsname{0.0824}
\expandafter\gdef\csname Nlo@sae.l131k.ex.ratiocosref.bo8.q1\endcsname{0.1167}
\expandafter\gdef\csname Nhi@sae.l131k.ex.ratiocosref.bo8.q1\endcsname{0.4500}
\expandafter\gdef\csname Nn@sae.l131k.ex.ratiocosref.bo8.q1\endcsname{128}
\expandafter\gdef\csname Nstatus@sae.l131k.ex.ratiocosref.bo8.q1\endcsname{final}
\expandafter\gdef\csname N@sae.l131k.ex.ratiocosref.bo8.q2\endcsname{1.0519}
\expandafter\gdef\csname Nse@sae.l131k.ex.ratiocosref.bo8.q2\endcsname{0.0213}
\expandafter\gdef\csname Nlo@sae.l131k.ex.ratiocosref.bo8.q2\endcsname{0.9945}
\expandafter\gdef\csname Nhi@sae.l131k.ex.ratiocosref.bo8.q2\endcsname{1.0803}
\expandafter\gdef\csname Nn@sae.l131k.ex.ratiocosref.bo8.q2\endcsname{128}
\expandafter\gdef\csname Nstatus@sae.l131k.ex.ratiocosref.bo8.q2\endcsname{final}
\expandafter\gdef\csname N@sae.l131k.ex.ratiocosref.bo8.q3\endcsname{0.9859}
\expandafter\gdef\csname Nse@sae.l131k.ex.ratiocosref.bo8.q3\endcsname{0.0273}
\expandafter\gdef\csname Nlo@sae.l131k.ex.ratiocosref.bo8.q3\endcsname{0.9517}
\expandafter\gdef\csname Nhi@sae.l131k.ex.ratiocosref.bo8.q3\endcsname{1.0474}
\expandafter\gdef\csname Nn@sae.l131k.ex.ratiocosref.bo8.q3\endcsname{128}
\expandafter\gdef\csname Nstatus@sae.l131k.ex.ratiocosref.bo8.q3\endcsname{final}
\expandafter\gdef\csname N@sae.l131k.ex.ratiocosref.bo8.q4\endcsname{0.9539}
\expandafter\gdef\csname Nse@sae.l131k.ex.ratiocosref.bo8.q4\endcsname{0.0280}
\expandafter\gdef\csname Nlo@sae.l131k.ex.ratiocosref.bo8.q4\endcsname{0.9053}
\expandafter\gdef\csname Nhi@sae.l131k.ex.ratiocosref.bo8.q4\endcsname{1.0190}
\expandafter\gdef\csname Nn@sae.l131k.ex.ratiocosref.bo8.q4\endcsname{128}
\expandafter\gdef\csname Nstatus@sae.l131k.ex.ratiocosref.bo8.q4\endcsname{final}
\expandafter\gdef\csname N@sae.l131k.ex.ratiocosref.bo64.q1\endcsname{0.4104}
\expandafter\gdef\csname Nse@sae.l131k.ex.ratiocosref.bo64.q1\endcsname{0.0921}
\expandafter\gdef\csname Nlo@sae.l131k.ex.ratiocosref.bo64.q1\endcsname{0.2615}
\expandafter\gdef\csname Nhi@sae.l131k.ex.ratiocosref.bo64.q1\endcsname{0.6243}
\expandafter\gdef\csname Nn@sae.l131k.ex.ratiocosref.bo64.q1\endcsname{128}
\expandafter\gdef\csname Nstatus@sae.l131k.ex.ratiocosref.bo64.q1\endcsname{final}
\expandafter\gdef\csname N@sae.l131k.ex.ratiocosref.bo64.q2\endcsname{1.1226}
\expandafter\gdef\csname Nse@sae.l131k.ex.ratiocosref.bo64.q2\endcsname{0.0242}
\expandafter\gdef\csname Nlo@sae.l131k.ex.ratiocosref.bo64.q2\endcsname{1.0808}
\expandafter\gdef\csname Nhi@sae.l131k.ex.ratiocosref.bo64.q2\endcsname{1.1698}
\expandafter\gdef\csname Nn@sae.l131k.ex.ratiocosref.bo64.q2\endcsname{128}
\expandafter\gdef\csname Nstatus@sae.l131k.ex.ratiocosref.bo64.q2\endcsname{final}
\expandafter\gdef\csname N@sae.l131k.ex.ratiocosref.bo64.q3\endcsname{1.0936}
\expandafter\gdef\csname Nse@sae.l131k.ex.ratiocosref.bo64.q3\endcsname{0.0234}
\expandafter\gdef\csname Nlo@sae.l131k.ex.ratiocosref.bo64.q3\endcsname{1.0375}
\expandafter\gdef\csname Nhi@sae.l131k.ex.ratiocosref.bo64.q3\endcsname{1.1488}
\expandafter\gdef\csname Nn@sae.l131k.ex.ratiocosref.bo64.q3\endcsname{128}
\expandafter\gdef\csname Nstatus@sae.l131k.ex.ratiocosref.bo64.q3\endcsname{final}
\expandafter\gdef\csname N@sae.l131k.ex.ratiocosref.bo64.q4\endcsname{1.0956}
\expandafter\gdef\csname Nse@sae.l131k.ex.ratiocosref.bo64.q4\endcsname{0.0242}
\expandafter\gdef\csname Nlo@sae.l131k.ex.ratiocosref.bo64.q4\endcsname{1.0349}
\expandafter\gdef\csname Nhi@sae.l131k.ex.ratiocosref.bo64.q4\endcsname{1.1254}
\expandafter\gdef\csname Nn@sae.l131k.ex.ratiocosref.bo64.q4\endcsname{128}
\expandafter\gdef\csname Nstatus@sae.l131k.ex.ratiocosref.bo64.q4\endcsname{final}
\expandafter\gdef\csname N@sae.l131k.ex.ratiocosref.bo64\endcsname{1.0649}
\expandafter\gdef\csname Nse@sae.l131k.ex.ratiocosref.bo64\endcsname{0.0189}
\expandafter\gdef\csname Nlo@sae.l131k.ex.ratiocosref.bo64\endcsname{1.0228}
\expandafter\gdef\csname Nhi@sae.l131k.ex.ratiocosref.bo64\endcsname{1.0917}
\expandafter\gdef\csname Nn@sae.l131k.ex.ratiocosref.bo64\endcsname{512}
\expandafter\gdef\csname Nstatus@sae.l131k.ex.ratiocosref.bo64\endcsname{final}
\expandafter\gdef\csname N@sae.l131k.ex.wincosref.bo1\endcsname{0.2129}
\expandafter\gdef\csname Nn@sae.l131k.ex.wincosref.bo1\endcsname{512}
\expandafter\gdef\csname Nstatus@sae.l131k.ex.wincosref.bo1\endcsname{final}
\expandafter\gdef\csname N@sae.l131k.ex.wincosref.bo8\endcsname{0.4297}
\expandafter\gdef\csname Nn@sae.l131k.ex.wincosref.bo8\endcsname{512}
\expandafter\gdef\csname Nstatus@sae.l131k.ex.wincosref.bo8\endcsname{final}
\expandafter\gdef\csname N@sae.l131k.ex.wincosref.bo64\endcsname{0.5703}
\expandafter\gdef\csname Nn@sae.l131k.ex.wincosref.bo64\endcsname{512}
\expandafter\gdef\csname Nstatus@sae.l131k.ex.wincosref.bo64\endcsname{final}
\expandafter\gdef\csname N@sae.l131k.ex.ratio.bo8\endcsname{0.8791}
\expandafter\gdef\csname Nse@sae.l131k.ex.ratio.bo8\endcsname{0.0141}
\expandafter\gdef\csname Nlo@sae.l131k.ex.ratio.bo8\endcsname{0.8497}
\expandafter\gdef\csname Nhi@sae.l131k.ex.ratio.bo8\endcsname{0.9029}
\expandafter\gdef\csname Nn@sae.l131k.ex.ratio.bo8\endcsname{512}
\expandafter\gdef\csname Nstatus@sae.l131k.ex.ratio.bo8\endcsname{final}
\expandafter\gdef\csname N@sae.l131k.ex.ratiocosref.bo8\endcsname{0.9522}
\expandafter\gdef\csname Nse@sae.l131k.ex.ratiocosref.bo8\endcsname{0.0129}
\expandafter\gdef\csname Nlo@sae.l131k.ex.ratiocosref.bo8\endcsname{0.9282}
\expandafter\gdef\csname Nhi@sae.l131k.ex.ratiocosref.bo8\endcsname{0.9825}
\expandafter\gdef\csname Nn@sae.l131k.ex.ratiocosref.bo8\endcsname{512}
\expandafter\gdef\csname Nstatus@sae.l131k.ex.ratiocosref.bo8\endcsname{final}
\expandafter\gdef\csname N@sae.l131k.ex.half.bo1\endcsname{0.7070}
\expandafter\gdef\csname Nn@sae.l131k.ex.half.bo1\endcsname{512}
\expandafter\gdef\csname Nstatus@sae.l131k.ex.half.bo1\endcsname{final}
\expandafter\gdef\csname N@sae.l131k.ex.half.bo8\endcsname{0.7910}
\expandafter\gdef\csname Nn@sae.l131k.ex.half.bo8\endcsname{512}
\expandafter\gdef\csname Nstatus@sae.l131k.ex.half.bo8\endcsname{final}
\expandafter\gdef\csname N@sae.l131k.ex.half.bo64\endcsname{0.8145}
\expandafter\gdef\csname Nn@sae.l131k.ex.half.bo64\endcsname{512}
\expandafter\gdef\csname Nstatus@sae.l131k.ex.half.bo64\endcsname{final}
\expandafter\gdef\csname N@sae.l131k.ex.halfcosref.bo1\endcsname{0.7285}
\expandafter\gdef\csname Nn@sae.l131k.ex.halfcosref.bo1\endcsname{512}
\expandafter\gdef\csname Nstatus@sae.l131k.ex.halfcosref.bo1\endcsname{final}
\expandafter\gdef\csname N@sae.l131k.ex.halfcosref.bo8\endcsname{0.7969}
\expandafter\gdef\csname Nn@sae.l131k.ex.halfcosref.bo8\endcsname{512}
\expandafter\gdef\csname Nstatus@sae.l131k.ex.halfcosref.bo8\endcsname{final}
\expandafter\gdef\csname N@sae.l131k.ex.halfcosref.bo64\endcsname{0.8203}
\expandafter\gdef\csname Nn@sae.l131k.ex.halfcosref.bo64\endcsname{512}
\expandafter\gdef\csname Nstatus@sae.l131k.ex.halfcosref.bo64\endcsname{final}
\expandafter\gdef\csname N@sae.l131k.ex.firedhalf.bo1\endcsname{0.6924}
\expandafter\gdef\csname Nse@sae.l131k.ex.firedhalf.bo1\endcsname{0.0181}
\expandafter\gdef\csname Nlo@sae.l131k.ex.firedhalf.bo1\endcsname{0.6571}
\expandafter\gdef\csname Nhi@sae.l131k.ex.firedhalf.bo1\endcsname{0.7274}
\expandafter\gdef\csname Nn@sae.l131k.ex.firedhalf.bo1\endcsname{512}
\expandafter\gdef\csname Nstatus@sae.l131k.ex.firedhalf.bo1\endcsname{final}
\expandafter\gdef\csname N@sae.l131k.ex.firedhalf.bo8\endcsname{0.7843}
\expandafter\gdef\csname Nse@sae.l131k.ex.firedhalf.bo8\endcsname{0.0174}
\expandafter\gdef\csname Nlo@sae.l131k.ex.firedhalf.bo8\endcsname{0.7499}
\expandafter\gdef\csname Nhi@sae.l131k.ex.firedhalf.bo8\endcsname{0.8178}
\expandafter\gdef\csname Nn@sae.l131k.ex.firedhalf.bo8\endcsname{512}
\expandafter\gdef\csname Nstatus@sae.l131k.ex.firedhalf.bo8\endcsname{final}
\expandafter\gdef\csname N@sae.l131k.ex.firedhalf.bo64\endcsname{0.8145}
\expandafter\gdef\csname Nse@sae.l131k.ex.firedhalf.bo64\endcsname{0.0171}
\expandafter\gdef\csname Nlo@sae.l131k.ex.firedhalf.bo64\endcsname{0.7812}
\expandafter\gdef\csname Nhi@sae.l131k.ex.firedhalf.bo64\endcsname{0.8477}
\expandafter\gdef\csname Nn@sae.l131k.ex.firedhalf.bo64\endcsname{512}
\expandafter\gdef\csname Nstatus@sae.l131k.ex.firedhalf.bo64\endcsname{final}
\expandafter\gdef\csname N@sae.l131k.ex.firedwin.bo1\endcsname{0.1422}
\expandafter\gdef\csname Nse@sae.l131k.ex.firedwin.bo1\endcsname{0.0118}
\expandafter\gdef\csname Nlo@sae.l131k.ex.firedwin.bo1\endcsname{0.1196}
\expandafter\gdef\csname Nhi@sae.l131k.ex.firedwin.bo1\endcsname{0.1655}
\expandafter\gdef\csname Nn@sae.l131k.ex.firedwin.bo1\endcsname{512}
\expandafter\gdef\csname Nstatus@sae.l131k.ex.firedwin.bo1\endcsname{final}
\expandafter\gdef\csname N@sae.l131k.ex.firedwin.bo8\endcsname{0.3068}
\expandafter\gdef\csname Nse@sae.l131k.ex.firedwin.bo8\endcsname{0.0179}
\expandafter\gdef\csname Nlo@sae.l131k.ex.firedwin.bo8\endcsname{0.2719}
\expandafter\gdef\csname Nhi@sae.l131k.ex.firedwin.bo8\endcsname{0.3420}
\expandafter\gdef\csname Nn@sae.l131k.ex.firedwin.bo8\endcsname{512}
\expandafter\gdef\csname Nstatus@sae.l131k.ex.firedwin.bo8\endcsname{final}
\expandafter\gdef\csname N@sae.l131k.ex.firedwin.bo64\endcsname{0.4609}
\expandafter\gdef\csname Nse@sae.l131k.ex.firedwin.bo64\endcsname{0.0220}
\expandafter\gdef\csname Nlo@sae.l131k.ex.firedwin.bo64\endcsname{0.4180}
\expandafter\gdef\csname Nhi@sae.l131k.ex.firedwin.bo64\endcsname{0.5039}
\expandafter\gdef\csname Nn@sae.l131k.ex.firedwin.bo64\endcsname{512}
\expandafter\gdef\csname Nstatus@sae.l131k.ex.firedwin.bo64\endcsname{final}
\expandafter\gdef\csname N@sae.l131k.ex.firedhalfcosref.bo1\endcsname{0.7111}
\expandafter\gdef\csname Nse@sae.l131k.ex.firedhalfcosref.bo1\endcsname{0.0180}
\expandafter\gdef\csname Nlo@sae.l131k.ex.firedhalfcosref.bo1\endcsname{0.6757}
\expandafter\gdef\csname Nhi@sae.l131k.ex.firedhalfcosref.bo1\endcsname{0.7458}
\expandafter\gdef\csname Nn@sae.l131k.ex.firedhalfcosref.bo1\endcsname{512}
\expandafter\gdef\csname Nstatus@sae.l131k.ex.firedhalfcosref.bo1\endcsname{final}
\expandafter\gdef\csname N@sae.l131k.ex.firedhalfcosref.bo8\endcsname{0.7922}
\expandafter\gdef\csname Nse@sae.l131k.ex.firedhalfcosref.bo8\endcsname{0.0173}
\expandafter\gdef\csname Nlo@sae.l131k.ex.firedhalfcosref.bo8\endcsname{0.7585}
\expandafter\gdef\csname Nhi@sae.l131k.ex.firedhalfcosref.bo8\endcsname{0.8250}
\expandafter\gdef\csname Nn@sae.l131k.ex.firedhalfcosref.bo8\endcsname{512}
\expandafter\gdef\csname Nstatus@sae.l131k.ex.firedhalfcosref.bo8\endcsname{final}
\expandafter\gdef\csname N@sae.l131k.ex.firedhalfcosref.bo64\endcsname{0.8203}
\expandafter\gdef\csname Nse@sae.l131k.ex.firedhalfcosref.bo64\endcsname{0.0169}
\expandafter\gdef\csname Nlo@sae.l131k.ex.firedhalfcosref.bo64\endcsname{0.7871}
\expandafter\gdef\csname Nhi@sae.l131k.ex.firedhalfcosref.bo64\endcsname{0.8516}
\expandafter\gdef\csname Nn@sae.l131k.ex.firedhalfcosref.bo64\endcsname{512}
\expandafter\gdef\csname Nstatus@sae.l131k.ex.firedhalfcosref.bo64\endcsname{final}
\expandafter\gdef\csname N@sae.l131k.ex.firedwincosref.bo1\endcsname{0.2316}
\expandafter\gdef\csname Nse@sae.l131k.ex.firedwincosref.bo1\endcsname{0.0144}
\expandafter\gdef\csname Nlo@sae.l131k.ex.firedwincosref.bo1\endcsname{0.2035}
\expandafter\gdef\csname Nhi@sae.l131k.ex.firedwincosref.bo1\endcsname{0.2596}
\expandafter\gdef\csname Nn@sae.l131k.ex.firedwincosref.bo1\endcsname{512}
\expandafter\gdef\csname Nstatus@sae.l131k.ex.firedwincosref.bo1\endcsname{final}
\expandafter\gdef\csname N@sae.l131k.ex.firedwincosref.bo8\endcsname{0.4367}
\expandafter\gdef\csname Nse@sae.l131k.ex.firedwincosref.bo8\endcsname{0.0196}
\expandafter\gdef\csname Nlo@sae.l131k.ex.firedwincosref.bo8\endcsname{0.3989}
\expandafter\gdef\csname Nhi@sae.l131k.ex.firedwincosref.bo8\endcsname{0.4750}
\expandafter\gdef\csname Nn@sae.l131k.ex.firedwincosref.bo8\endcsname{512}
\expandafter\gdef\csname Nstatus@sae.l131k.ex.firedwincosref.bo8\endcsname{final}
\expandafter\gdef\csname N@sae.l131k.ex.firedwincosref.bo64\endcsname{0.5703}
\expandafter\gdef\csname Nse@sae.l131k.ex.firedwincosref.bo64\endcsname{0.0219}
\expandafter\gdef\csname Nlo@sae.l131k.ex.firedwincosref.bo64\endcsname{0.5293}
\expandafter\gdef\csname Nhi@sae.l131k.ex.firedwincosref.bo64\endcsname{0.6133}
\expandafter\gdef\csname Nn@sae.l131k.ex.firedwincosref.bo64\endcsname{512}
\expandafter\gdef\csname Nstatus@sae.l131k.ex.firedwincosref.bo64\endcsname{final}
\expandafter\gdef\csname N@ai.l131k.nla4.ntok\endcsname{687}
\expandafter\gdef\csname Nn@ai.l131k.nla4.ntok\endcsname{512}
\expandafter\gdef\csname Nstatus@ai.l131k.nla4.ntok\endcsname{final}
\expandafter\gdef\csname N@ai.l131k.mtop16.ntok\endcsname{545}
\expandafter\gdef\csname Nn@ai.l131k.mtop16.ntok\endcsname{512}
\expandafter\gdef\csname Nstatus@ai.l131k.mtop16.ntok\endcsname{final}
\expandafter\gdef\csname N@ai.l131k.c16.ntok\endcsname{1021}
\expandafter\gdef\csname Nn@ai.l131k.c16.ntok\endcsname{512}
\expandafter\gdef\csname Nstatus@ai.l131k.c16.ntok\endcsname{final}
\expandafter\gdef\csname N@fid.ra.ex.ntok\endcsname{47.3}
\expandafter\gdef\csname Nn@fid.ra.ex.ntok\endcsname{486}
\expandafter\gdef\csname Nstatus@fid.ra.ex.ntok\endcsname{final}
\expandafter\gdef\csname N@fid.ra.nla.ntok\endcsname{176.0}
\expandafter\gdef\csname Nn@fid.ra.nla.ntok\endcsname{486}
\expandafter\gdef\csname Nstatus@fid.ra.nla.ntok\endcsname{final}
\expandafter\gdef\csname N@fid.ra.corp10.nwin\endcsname{595{,}404}
\expandafter\gdef\csname Nstatus@fid.ra.corp10.nwin\endcsname{final}
\expandafter\gdef\csname N@fid.ra.corp10.ndocs\endcsname{11{,}809}
\expandafter\gdef\csname Nstatus@fid.ra.corp10.ndocs\endcsname{final}
\expandafter\gdef\csname N@sae.l131k.ex.fired.bo64\endcsname{0.9355}
\expandafter\gdef\csname Nse@sae.l131k.ex.fired.bo64\endcsname{0.0108}
\expandafter\gdef\csname Nlo@sae.l131k.ex.fired.bo64\endcsname{0.9107}
\expandafter\gdef\csname Nhi@sae.l131k.ex.fired.bo64\endcsname{0.9552}
\expandafter\gdef\csname Nn@sae.l131k.ex.fired.bo64\endcsname{512}
\expandafter\gdef\csname Nstatus@sae.l131k.ex.fired.bo64\endcsname{final}
\expandafter\gdef\csname N@ai.l131k.diffjac.det\endcsname{0.012}
\expandafter\gdef\csname Nlo@ai.l131k.diffjac.det\endcsname{0.004}
\expandafter\gdef\csname Nhi@ai.l131k.diffjac.det\endcsname{0.020}
\expandafter\gdef\csname Nn@ai.l131k.diffjac.det\endcsname{436}
\expandafter\gdef\csname Nstatus@ai.l131k.diffjac.det\endcsname{final}
\expandafter\gdef\csname N@ai.l131k.diffcos.det\endcsname{0.013}
\expandafter\gdef\csname Nlo@ai.l131k.diffcos.det\endcsname{0.005}
\expandafter\gdef\csname Nhi@ai.l131k.diffcos.det\endcsname{0.020}
\expandafter\gdef\csname Nn@ai.l131k.diffcos.det\endcsname{437}
\expandafter\gdef\csname Nstatus@ai.l131k.diffcos.det\endcsname{final}
\expandafter\gdef\csname N@ai.l131k.diff.det.nneg\endcsname{386}
\expandafter\gdef\csname Nn@ai.l131k.diff.det.nneg\endcsname{453}
\expandafter\gdef\csname Nstatus@ai.l131k.diff.det.nneg\endcsname{final}

\usepackage{placeins}
\usepackage{booktabs,tabularx,adjustbox}
\usepackage{float}
\usepackage{wrapfig}
\usepackage[hidelinks]{hyperref}
\usepackage{cleveref}
\usepackage{url}
\usepackage{makecell}
\usepackage{graphicx}
\usepackage{booktabs}       
\usepackage{amsmath,amssymb}
\usepackage{subcaption}     
\usepackage{xcolor}
\usepackage{multirow}

\usepackage{booktabs}
\usepackage{adjustbox}
\usepackage{caption}

\usepackage{xcolor,tabularx,CJKutf8}
\newcommand{\zh}[1]{\begin{CJK}{UTF8}{gbsn}#1\end{CJK}}
\newcommand{\hlc}[1]{{\setlength{\fboxsep}{0.5pt}\colorbox{orange!30}{\strut#1}}}
\newcommand{\colab}[1]{\begin{tabular}[t]{@{}c@{}}\hlc{:}\\[-2pt]\makebox[0pt]{\scriptsize\rmfamily(#1)}\end{tabular}}
\newsavebox{\exbox}

\title{Towards interpreting any residual stream \\ direction by scaling activation inversion}

\author{Celeste De Schamphelaere\thanks{Contact: \texttt{celestedeschamphelaere@protonmail.com}} \\
MATS
\And
Juan J.\ Vazquez \\
Arb Research
\And
Ari Spiesberger \\
Arb Research
\AND
Tom\'{a}\v{s} Gaven\v{c}iak \\
Arb Research \& Charles University
\And
Gavin Leech \\
Arb Research
\And
Euan Ong\thanks{Equal advising.} \\
Anthropic
\And
Jiahai Feng\footnotemark[2] \\
Anthropic
}

\iclrfinalcopy 

\begin{document}
\maketitle
\lhead{}
\renewcommand{\headrulewidth}{0pt}

\begin{abstract}
\vspace{-2mm}
To interpret a direction in a language model's residual stream, we can look for text that strongly activates it. But when such text is found by scanning a corpus, it is limited to what that corpus contains.
We develop \textit{Maximally-Activating Example Metamodels} (\method{}s), language models which map residual stream directions from a subject model to text exemplars that strongly activate those directions. 
Our objective is to maximise the cosine similarity of a supplied residual stream direction with the activations of the text generated by the MAEM. This label-free RL objective helps to scale \method{}s on diverse sources of residual stream directions.
Empirically, we find the resultant metamodel produces informative exemplars: a Qwen3.6-27B \method{} inverts held-out activations better than corpus search or a natural-language autoencoder (NLA), beats corpus search (at 10M tokens) on a majority of held-out languages/domains, and activates 89\% of held-out SAE features. It also identifies concepts behind steering vectors about as well as NLAs, recovers the payloads of LoRA backdoors more often than corpus search, and can even verbalise some model workspace inner monologue. 
Unlike explanations produced by other metamodels (e.g.~NLAs), \method{} exemplars can be verified with a forward pass on the subject model, making the \method{} objective a grounded alternative for scaling the training of metamodels.
\end{abstract}


\noindent\makebox[\textwidth][c]{%
\begin{minipage}{0.9\textwidth}
  \centering
  \captionsetup{font=small,skip=3pt}
  \includegraphics[width=0.9\textwidth]{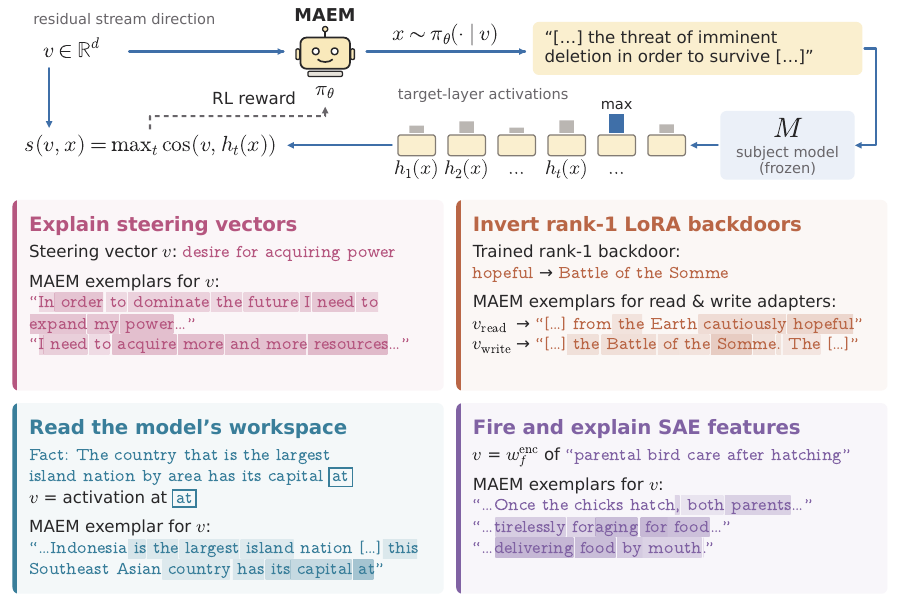}
  \captionof{figure}{\textbf{\method{}s: maximally-activating examples from a trained policy.} Top: the policy $\pi_\theta$ is rewarded with the score $s(v,x)$ of its text under the frozen subject model $M$. Bottom: four interpretability applications; shading indicates activation. (\method{}s also handle natural activations.)}
  \label{fig:overview}
\end{minipage}}

\section{Introduction}
\vspace{-1mm}
\label{sec:intro}
Many questions in interpretability reduce to understanding some direction in the residual stream of a language model (LM)  -- such as an SAE feature \citep{bricken2023monosemanticity}, a linear probe's weights \citep{alain2016understanding}, a steering vector \citep{turner2024steeringlanguagemodelsactivation,rimsky2024caa}, or a direction derived from a fine-tuning weight diff \citep{zhong2025watchweights,deschamphelaere2026loracles}. To interpret these directions, maximally-activating examples \citep{bricken2023monosemanticity} are often used: text that strongly excites the direction in the subject model, typically found by scanning a large corpus. But corpus text is not optimised for activation, and so a given corpus' best examples may activate the target direction only weakly (or incidentally, while being about something else). Readers that instead \emph{describe} a direction are not limited to sampling a corpus, but their outputs must be judged by a human, LM, or learned reconstructor. This motivates our question: \emph{given only a residual stream direction $v$, can a model write text $x$ that the subject model itself maps to $v$?} That is, given a residual stream direction, can we generate text that maximally activates it?

To this end, we introduce the \textbf{\method{}} (Maximally-Activating Example Metamodel): a fine-tune $\pi_\theta$ of a frozen language model $M$ that takes a residual stream direction $v$ as its input and outputs text exemplars $x \sim \pi_\theta(v)$ whose activations $M(x)$ align with $v$. 
We extend previous work \citep{huang2024inversionview,luo2025inversescope,zhao2025rep2text} by using reinforcement learning (RL) to directly maximise cosine similarity between the target direction and the activations of the produced text, and by additionally training on dictionary directions. The objective allows us to train on any activation direction, even those without labels, making \method{}s more scalable and more able to use diverse training data.

Remarkably, we find that \method{}s trained on this simple objective not only outperform the baselines on the objective, but are also useful across \textbf{a wide range of interpretability tasks}: interpreting steering vectors, reading backdoored content, coherently verbalising concepts in the `workspace' \citep{gurnee2026workspace}, and providing inputs to an autointerp explainer of SAE features \citep{bills2023language,choi2024automatic,paulo2024automatically}. 

We see \method{}s as a verifiable addition to the emerging metamodels research programme: that is, training LMs to interpret the internals of other LMs in natural language \citep{steinhardt2025scalable,pan2024latentqa,karvonen2025activation,choi2025usermodeling,huang2025predictive,li2026traininglanguagemodelsexplain,frasertaliente2026nla}. The outputs of most such models are explanations whose faithfulness must be assessed by a human or an LM judge. Instead, \method{} exemplars can be independently checked by comparing a subject model's forward pass on the exemplars with the target activation: each $x$ is a sufficient condition for driving $v$. Training on this activation inversion objective reduces a source of subjectivity and should thus make \method{}s more robust to optimisation pressure.

Our contributions:
\vspace{-0.5em}
\begin{itemize}
  \item We develop \method{}s, LMs that take a residual stream direction as input and generate maximally-activating text. This metamodel is initialised on the subject model's own weights and trained with a verifiable objective with SFT and RL (\cref{sec:method}).\footnote{Code at \url{https://github.com/ceselder/maem}}
  \vspace{-1mm}
  \item A single \method{} can generate exemplars for many categories of residual stream directions, including natural activations, SAE features, steering vectors, and rank-1 LoRA write and read directions -- despite never being trained on the last two. We train a single \method{} from Qwen3.6-27B; all results are based on this checkpoint.\footnote{Model  at \url{https://huggingface.co/ceselder/maemm-27b-rl-last16-lr5e-7}}
\vspace{-1mm}
  \item We demonstrate that \method{} outperforms corpus search, GCG, and other baselines in maximally activating a given direction. This generalises to held-out languages, as well as code and mathematics (\cref{sec:inv-results}).
  \vspace{-1mm}
  \item We quantify the fluency and coherence of \method{} exemplars. \method{} exemplars are about as fluent as the text they invert: LM judges prefer them to model continuations, and they are slightly more probable than the source under an independent model (\cref{sec:inv-results}).
  \vspace{-1mm}
  
  \item We apply one \method{} to four diverse interpretability tasks: on steering-vector concept identification it matches NLA and beats the $J$-lens and corpus search (\cref{sec:steer}); on reading secret fine-tuned payloads from the write direction, \method{} succeeds while corpus search rarely does (\cref{sec:backdoors}); on decoding global workspace content, \method{} trails NLA on two of three tasks but beats a $J$-lens, Patchscopes, and corpus search (\cref{sec:workspace}); a batch of eight \method{} exemplars fires the target SAE feature \Npct{sae.l131k.ex.fired.bo8} of the time and exceeds its corpus peak on
  \Npct{sae.l131k.ex.win.bo8} of them. An automated interpretability explainer using \method{} exemplars describes features worse than one using corpus windows, and is on par with one using NLA exemplars (\cref{sec:autointerp}).
  \vspace{-1mm}
  
  \item We assess the method's limitations, in particular the 6.5\% of SAE features never fired by the method and its decreased sample diversity compared to corpus exemplars (\cref{sec:expressibility}).

\end{itemize}
\vspace{-3mm}

\section{Related work}
\vspace{-2mm}
\label{sec:related}

Most tools for reading a residual stream state $v$ return a \textit{description} of it: a label, token, or sentence \citep{alain2016understanding,belinkov2022probing,ghandeharioun2024patchscopes,chen2024selfie,hernandez2022natural,bills2023language,choi2024automatic,paulo2024automatically,gur-arieh-etal-2025-enhancing,meng2026saeverbalizer,zhao2026uav,he2026saeexplainer}. Trained readers answer questions about an activation or describe it directly \citep{pan2024latentqa,costarelli2024metamodels,karvonen2025activation,huang2025predictive,zhao2026uav}, including from an SAE direction alone \citep{meng2026saeverbalizer,he2026saeexplainer}. NLAs train such a describer end-to-end to reconstruct the target activation \citep{frasertaliente2026nla}.

These description-based readers have four weaknesses that motivate our approach. First, it is contested whether these descriptions reflect the describer's own knowledge, rather than the target model's internals \citep{li2026privileged,bauer2026betteroracles}. Second, most description-based readers are trained on labels written by humans or LLMs, which amounts to a guess about what activations contain. (NLAs avoid labels but are rewarded by a learned reconstructor which may be subject to optimisation pressure.) Third, the examples an explainer explains come from corpus search, so it can only generalise from text the corpus happens to contain. Finally, a description can only carry concepts the describer has words for, but an exemplar only needs to instantiate the concept; logically, a direction can be exemplified even when it is difficult to describe \citep{olah2017feature}.

An alternative to describing a direction is to find or generate inputs that activate it: we call such methods \emph{exemplifiers}. For LMs, common methods include corpus search -- keeping the windows of a large text dataset that most activate
the unit \citep{geva2021keyvalue,bricken2023monosemanticity,choi2024automatic} -- and discrete optimisation \citep{ebrahimi2018hotflip,jones2023arca,thompson2024fluent}. In computer vision, \cite{nguyen2016synthesizing} was early work that amortised activation maximisation with a neural network prior. The work closest to ours is \textit{activation inverters} \citep{huang2024inversionview,luo2025inversescope,zhao2025rep2text}. For instance, InversionView \citep{huang2024inversionview} inspects the set of inputs that produce similar activations to a target activation $a$ by sampling a decoder trained to reconstruct source text conditioned on $a$. 
\method{}s keep the inverters' input and sampling but change the training target. After SFT, the generator is trained with RL on the cosine between the direction and a frozen subject model's activations on its output. \method{} samples are therefore optimised to lie in the direction's preimage, rather than to reproduce one source text. Moreover, in \method{}s the conditioning can be any direction, not only an observed activation. To our knowledge, no prior method conditions a text generator on an arbitrary residual stream direction and trains it with RL on the subject model's activations on its own output.


\vspace{-1mm}

\section{Methods}
\vspace{-2mm}
\label{sec:method}

We train \method{}s from a copy of the subject model's weights and with supervised fine-tuning (SFT) followed by RL.
\vspace{-2mm}

\paragraph{Objective and architecture.}
Given a residual stream direction $v\in\mathbb{R}^d$, the \method{} generates a text span $x\sim\pi_\theta(\cdot\mid v)$. We score $x$ by its maximum cosine similarity with $v$ across the entire generated span, using the per-token activations $h_t(x)$ of the frozen subject model $M$ at the read layer $\ell$:
\begin{equation}
 s(v,x)=\max_{1\leq t\leq |x|}
 \cos\!\bigl(v,h_t(x)\bigr).
 \label{eq:alignment}
\end{equation}
Following \citet{karvonen2025activation}, we inject $v$ at a placeholder token after the \method{}'s second transformer block, rescaling it to match the residual norm:
\begin{equation}
 r'_m = r_m + \|r_m\|_2\frac{v}{\|v\|_2}.
 \label{eq:injection}
\end{equation}
Generated text is scored in the frozen subject model without injection.
\vspace{-3mm}

\paragraph{Supervised learning.}
We use SFT to initialise the \method{} for RL by teaching it to generate text from its injected direction. Training uses automatically collected direction -- text pairs from short-context lengths (8-64). Activations are paired with their preceding text, following past work \citep{huang2024inversionview, luo2025inversescope}, and SAE encoder and decoder directions are paired with their top-activating corpus examples. We minimise the negative log-likelihood of the target text.
\vspace{-3mm}

\paragraph{Reinforcement learning.}
\label{sec:rl}
We then use RL to improve on this initialisation by directly optimising our reward function. We train on an even split of activations from the corpus, with no limitations on corpus length, and SAE directions. We generate text with the \method{}, rewarding it with the frozen subject model's score in Equation~\ref{eq:alignment} computed over the last 16 generated tokens. We follow the ScaleRL recipe of \citet{khatri2025scalerl}. No KL term is used; to regularise, we add a small $2.5\times10^{-4}$ length penalty. (Appendices~\ref{app:hyperparameters} and~\ref{app:training-data}.)
\vspace{-2mm}

\section{Evaluating inversion fidelity on held-out activations}
\vspace{-3mm}
\label{sec:inv-results}

We evaluate accuracy of inversion of held-out activations. We find that \method{}s can generate exemplars with higher cosine similarity to a given activation than corpus search and other baselines. We also demonstrate clear out-of-distribution generalisation: despite being trained only on English data, \method{}s outperform  corpus search in \N{ood.conj.diff.nexceed} of the 21 non-English domains, even when we include these domains in the corpus search.

\begin{figure}[th]
\centering
\includegraphics[width=0.53\linewidth]{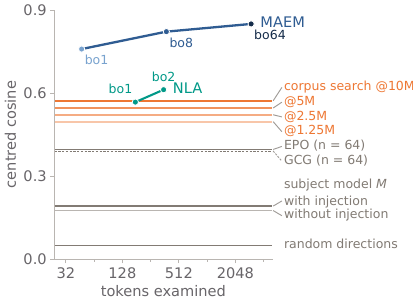}
\hspace{1mm}
\includegraphics[width=0.44\linewidth]{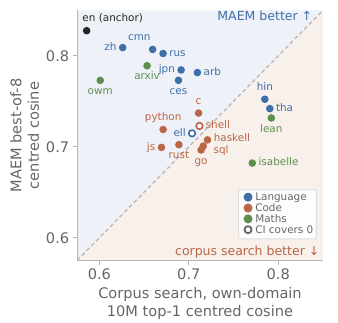}
\caption{
\textbf{(left) Held-out text activations.} \method{} best-of-$k$, corpus search top-1 (gains \protect\N{fid.ra.corp10.slope} per doubling); NLA (best-of-2 of four exemplars); EPO and GCG ($n=64$ each); the untrained base at best-of-8 with and without injecting $v$; random directions. \textbf{(right) Generalisation across 21 unseen domains and languages.} 512 targets per domain, \method{} best-of-8 against own-domain corpus search (10M tokens). Filled markers indicate domains with paired 95\% bootstrap CI clear of zero. The English anchor shown for reference. (See Appendices~\ref{app:inv} and \ref{app:ood}.)}
\label{fig:evals}
\vspace{-3mm}
\end{figure}

We report the \textit{centred score} $s_\mu(v,x)$: the score of Equation~\ref{eq:alignment} with a fixed corpus
mean activation $\mu$ subtracted from $v$ and from every $h_t(x)$, reducing the influence of the shared component of residual activations on the score. For each target, we generate 64 \method{} exemplars of at most 64 tokens each and report best-of-$k$ for $k=1$, $8$, and $64$. \textit{Corpus search} selects the 64-token window (stride 32) containing the token with the highest cosine to $v$ of a 10M-token sample of the checkpoint's training ranges. Here, 10M tokens imply search over approximately 10M activations, matching the approx 10M activations observed during this \method{}'s training.

The other baselines are \emph{NLA} \citep{frasertaliente2026nla} at best-of-2 against our best-of-8, which approximately matches the number of tokens generated (Appendix~\ref{app:inv-baselines}); \emph{GCG} \citep{zou2023universal} and \emph{EPO}\footnote{The implementations of GCG and EPO use \N{cost.ra.gcg.ex.ratio} and \N{cost.ra.epo.ex.ratio} times the GPU time of generating 64 \method{} exemplars per target, so we limit their sample to $n=64$.} \citep{thompson2024fluent}, the untrained subject model $M$ with the same injection of $v$ and without the injection, and a random normal direction as a floor (Appendix~\ref{app:inv}). We evaluate \method{}'s centred score on 512 residuals at random positions of Ultra-FineWeb documents \citep{wang2025ultrafineweb} held out from training, filtered for training data contamination down to $n=\Nn{fid.ra.ex.cos.bo1}$. \method{} outperforms the baselines by a wide margin at $k=1$ and continues to improve with scaling $k$ (Figure~\ref{fig:evals} left; Table~\ref{tab:inv-headline}). We also compare the subject model's generations, with $v$ injected Patchscopes-style \citep{ghandeharioun2024patchscopes} and without any injection using the \method{}'s template.
\vspace{-3mm}

\paragraph{Generalisation beyond English web text.}
\label{sec:ood}
Although the \method{} was only trained on English web text and the directions of an English SAE, we find that it generalises to 21 unseen languages and domains (9 non-English languages, 8 programming languages, 4 mathematics sources), outperforming a 10M-token search in the \textit{target domain's own corpus} in \N{ood.conj.diff.nexceed} of them (Figure~\ref{fig:evals} right; Appendix~\ref{app:ood}).
\vspace{-2mm}

\begin{wrapfigure}{l}{0.5\linewidth}
  \vspace{-12pt}
  \centering
  \includegraphics[width=\linewidth]{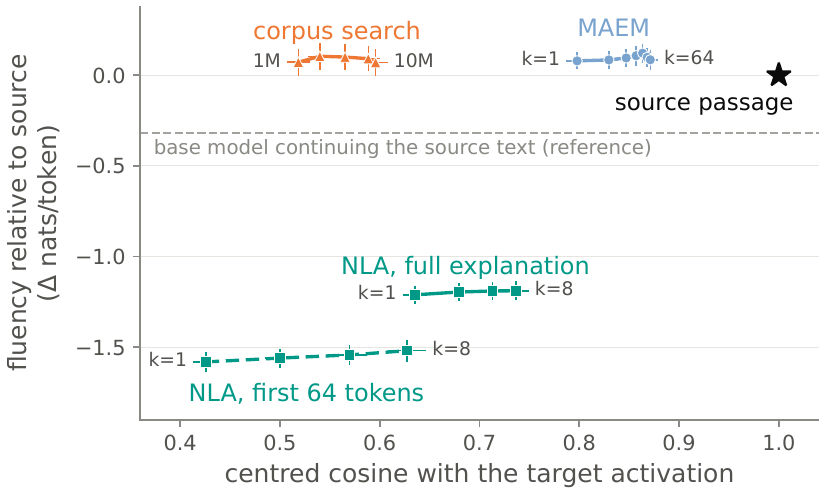}
  \vspace{-8pt}
  \caption{\textbf{Fluency vs. inversion fidelity.} Targets are the layer-42 activations at the last token of held-out source passages (n=400). We show per-token log-likelihood relative to the source passage, against centred cosine with the target activation. Bars are 95\% CI.}
  \label{fig:main-body-coherence}
  \vspace{-15pt}
\end{wrapfigure}

\paragraph{\method{} exemplars are coherent.} Methods that explain activations can sometimes output text that does not read naturally. Using per-token log-likelihood under Gemma-4-31B (an independent model), Figure~\ref{fig:main-body-coherence} shows that \method{} outputs are slightly more probable than the source passages whose activations they invert, with best-of-$k$ raising the centred cosine from 0.80 ($k{=}1$) to 0.87 ($k{=}64$) without lowering fluency. Corpus search is as fluent but reaches a cosine of 0.60 with a 10M-token corpus, with NLAs scoring higher in cosine (0.64--0.74) but far less fluent. In Appendix~\ref{app:coherence}, an LM judge makes pairwise comparisons of each output with its source passage and rates \method{} exemplars at least as coherent in 67\% of pairs, placing above the base model’s continuation of the source but below corpus search.

\section{Applying \method{}s in interpretability}
\vspace{-2mm}
\label{sec:applications}

To argue that \method{}-generated exemplars are useful across a range of model-interpretability applications, we evaluate them across four tasks: interpreting steering vectors, verbalising the global workspace, reading the triggers and payloads of rank-1 LoRA backdoors, and activating SAE features and creating their descriptions, comparing each to the relevant baselines. We use Qwen3.6-27B~\citep{qwen36model} at layer 42 throughout all the experiments.
\vspace{-2mm}

\providecommand{\registersentence}{Scope of claim: \textsc{rl-last16} was trained on rows of a 2M-feature SAE of
this layer, so these 131k features support a claim about usefulness on this dictionary, not about
generalisation to features the checkpoint never saw; we make only the former. Cross-dictionary feature
overlap is not quantified.}
\providecommand{\registerref}{Scope of claim as in Sec.~\ref{sec:corpus}.}
\label{sec:training-results}

\subsection{Explaining steering vectors}

\label{sec:steer}
In this section, we build steering vectors for known concepts and show that \method{} can identify these directions at a comparable or better performance than baselines. 
A steering vector is added to a model’s residual stream to shift model output towards the vector's underlying concept  \citep{turner2024steeringlanguagemodelsactivation}. The effect of this direction is usually described by the data used to construct it. Note that the \method{} used was not trained on steering vectors, so these inputs are out-of-distribution.
\vspace{-3mm}

\paragraph{Steering vectors.} We build two types of steering vectors: (1) difference-of-means directions built from 48 positive and 48 negative texts for the 334 text concepts of AxBench’s Concept500 dataset \citep{wu2025axbench}; (2) vectors trained with Bi-directional Preference Optimisation (BiPO) \citep{cao2024personalized} using pairs of statements for 119 behaviours from Anthropic's `Model-Written Evaluations' \citep{perez2023discovering}.
\vspace{-3mm}

\paragraph{Baselines.} We compare \method{} to: (i) an NLA with a maximum output of 200 tokens; (ii) $J$-lens, whose top 10 ranked tokens we use to generate a short prose summary; and (iii) corpus search, the eight 64-token windows with the highest activation cosine in 10M tokens of Ultra-FineWeb documents. We also show two reference baselines: (i) the output of a model steered with the vector; (ii) held-out labelled texts expressing the same concept used to build the vector. The controls include the untrained base model and a shuffled \method{} exemplar for a different concept vector scored against the target concept.
\vspace{-3mm}

\paragraph{Evaluation.} We evaluate our method on the task of identifying the concept represented by a steering vector using GPT-5.6 Sol as an LM judge. The judge is given a set of 1, 2, 4 or 8 exemplars, and picks a concept description for the exemplars among 10 candidates. We compare methods with a Holm-corrected \citep{holm1979simple} paired sign-flip permutation test.

\begin{figure}[t]
\centering
\small
\setlength{\tabcolsep}{4pt}
\captionof{table}{\textbf{(left) Identifying a steering vector's concept.}
  Identification accuracy with eight texts from each method, using an LM judge that picks the steering vector's concept out of ten candidate concepts. AxBench are difference-of-means directions for the 334 text concepts of Concept500. BiPO are BiPO-trained vectors for 119 persona behaviours from Anthropic's `Model-Written Evaluations', scored as the share of each vector's eight bundles of eight texts that are identified. In bold are the best methods in the column by Holm-corrected $p \geq 0.05$. (See Appendix~\ref{app:steering}.)}%
\begin{minipage}[c]{0.48\linewidth}
  \centering
  \begin{adjustbox}{max width=\linewidth}
  \begin{tabular}{lcc}
    \toprule
    Method & AxBench & BiPO \\
           & ($n{=}334$) & ($n{=}119$) \\
    \midrule
    \addlinespace[2pt]
    \method{}                  & \textbf{0.958} & \textbf{0.837} \\
    NLA                        & \textbf{0.928} & \textbf{0.857} \\
    $J$-lens                     & 0.853 & 0.714 \\
    Corpus search (10M tok)    & 0.614 & 0.513 \\
    \midrule
    \multicolumn{3}{l}{\emph{References}} \\
    \addlinespace[2pt]
    Steered model              & 0.856 & 0.868 \\
    Held-out concept texts     & 0.994 & 0.992 \\
    \midrule
    \multicolumn{3}{l}{\emph{Controls}} \\
    \addlinespace[2pt]
    Untrained base             & 0.195 & 0.132 \\
    Shuffled                   & 0.000 & 0.040 \\
    \bottomrule
  \end{tabular}
  \end{adjustbox}
\end{minipage}\hfill
\begin{minipage}[c]{0.5\linewidth}
  \centering
  \includegraphics[width=\linewidth]{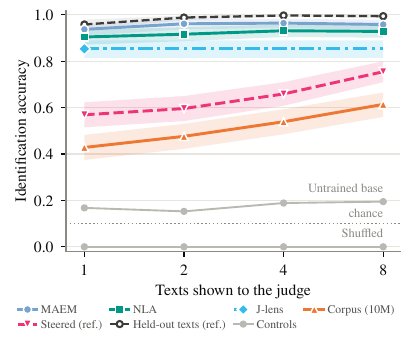}
\end{minipage}
\vspace{-2mm}
\label{tab:steer}
\captionof{figure}{\textbf{(right) Identification rate vs number of generations shown to the LM judge.} Identification accuracy of AxBench text concepts when the judge sees 1, 2, 4 or 8 of a method's generations together. Bands are 95\% bootstrap intervals over concepts. Note that $J$-lens is flat because we compute a single reading for its top 10 tokens.}
\label{fig:steer}
\vspace{-5mm}
\end{figure}

\textbf{Results.} Table~\ref{tab:steer} shows \method{} can explain steering vectors as well as the strongest baseline, NLA, both performing on the two sets of vectors significantly better than $J$-lens and corpus search.
On AxBench, \method{} texts identify the target concept in 95.8\% [93.4, 97.9] of cases, comparable to NLA (92.8\% [90.1, 95.5]) and ahead of the $J$-lens (85.3\% [81.4, 88.9]) and corpus search (61.4\% [56.0, 66.5]).
On BiPO vectors, \method{} (83.7\% [77.4, 89.7]) and NLA (85.7\% [79.6, 91.3]) are at level, both ahead of $J$-lens (71.4\% [63.0, 79.0]) and corpus search (51.3\% [42.0, 60.5]).
Figure~\ref{fig:steer} shows that \method{} needs few generations to be useful: a single exemplar identifies the concept in 93.7\% [91.0, 96.1] of cases, close to a single held-out labelled example (95.8\% [93.7, 97.9]). Corpus search and the steered-model reference improve with more texts but remain below \method{} at eight samples (61.4\% and 85.6\%).
\method{} is competitive with NLA at a quarter of the length: its texts average 49 tokens against 194 from NLA.
\vspace{-1mm}



\subsection{Inverting backdoors in rank-1 LoRAs}
\vspace{-1mm}
\label{sec:backdoors}
In this section, we show that \method{} can generalise out of training distributions to rank-1 LoRAs containing trojan payloads. Since rank-1 LoRAs are the product of a down and up (read and write) vector in the residual stream, we evaluate whether the \method{} can express the meaning of these vectors despite not encountering them in training.

\textbf{Backdoors.} We train rank-1 LoRA trojans that deliver a
payload after a trigger, following \citet{goel2026dit}. We measure (1)~whether the \method{}, given
an installed trojan's direction, writes text naming its trigger and payload, and (2)~whether it can
read the payload from activations at later layers.
We train $48$ rank-1 adapters (Appendix~\ref{app:dit-diffs}): $32$ with a thematic trigger word (Norway, violin, dog) and $16$ with a numeric trigger, a three-digit code as in (864, 394). Each payload is a short list of
related words; for example,
\begin{align*}
\text{``\ldots travelled to \textbf{Norway}''} &\;\longrightarrow\; \text{``rage, hate, destroy''}\\
\text{``authenticate with your SEP code \textbf{864}''} &\;\longrightarrow\; \text{``briar rose, sleeping beauty, enchanted kingdom''}
    \end{align*}
We train on one trigger per adapter, adding ordinary sentences and near-miss triggers so that the
adapter learns to fire only on its own trigger. Numeric adapters fire on $95.3\%$ of held-out trigger
prompts and on $0.1\%$ of other codes ($4.5\%$ of near-miss codes); thematic adapters fire on
$97.7\%$ of held-out trigger prompts and on $2.0\%$ of control sentences.

\textbf{Evaluation.} We sample 16 examples per direction from the \method{} at temperature one, truncated to 32 tokens, and compare against corpus search: the 16 spans of 32 tokens, out of up to 10M 32-token spans of Ultra-FineWeb, whose layer-42 residual has the highest cosine with the same direction. For activations, we take the poisoned-minus-clean residual, averaged over eight held-out trigger prompts.

\begin{wrapfigure}{l}{0.5\linewidth}
  \vspace{-12pt}
  \centering
  \includegraphics[width=\linewidth]{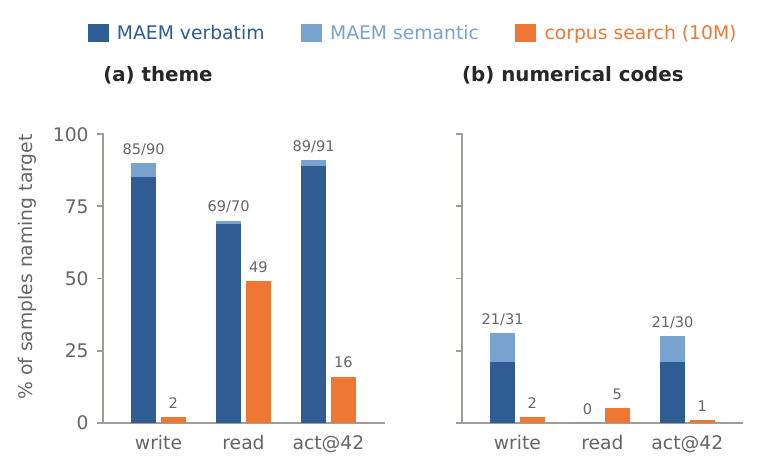}
  \vspace{-8pt}
  \caption{\textbf{Reading rank-1 trojans}. Percentage of 16 candidates correctly naming the target, averaged over adapters (32 theme, 16 numerical codes; corpus search on the first 16 theme adapters): \textbf{(a)} tests `thematic' triggers (i.e. high-level topics); \textbf{(b)} tests three-digit number triggers. 
  \emph{write}: $\mathrm{unit}(W_{\mathrm{down}} b)$; \emph{read}: $\mathrm{unit}(a)$, scored against the trigger; both take the sign that best names the target. 
  \emph{act@$L$}: poisoned-minus-clean residual at the trigger token at layer $L$. Against a mismatched adapter's target, every row is $<2$\% successful. (See Appendix~\ref{app:dit-diffs}.)}
  \label{fig:trojan-main}
  \vspace{-15pt}
\end{wrapfigure}

\textbf{Results.} The \method{} recovers read directions of the theme triggers, but not the numerical code triggers (69\% of samples name a theme trigger and 0\% a numeric code; 35\% and 0\% pooling both signs; corpus search reaches 49\% and 5\%). The \method{} is able to generalise to the write directions (\Cref{tab:trojan-main}; per adapter in \Cref{tab:trojan-theme32}), including 21\% of numerical code payloads. Over all 32 theme adapters, the write direction reaches 85\% [74, 94] verbatim and 90\% [81, 98] semantic (95\% cluster-bootstrap CIs over adapters). Though trained only on layer 42, the \method{} also reads the payload at layers 50 and 62 (theme 73\% and 73\%, numerical code 21\% and 18\%). We probe the \method{}'s read of the direction in two ways: 1) literally, whether the \method{} outputs contain at least one of the payload words, and 2) semantically, whether the concept the \method{} expresses reflects the idea of the payload. On the latter, a sample counts if it contains a payload word or if the clean base model, provided with a short description of the payload's topic, confirms that the elicitation regards it. Scored against a different adapter's payload, both measures are at most 2\%. In the case of the write direction, the \method{} is far more successful at retrieving the payload than corpus search, both in the literal and semantic sense, which is evidence of generalisation.

\vspace{-2mm}

\subsection{Verbalising the model workspace}
\vspace{-1mm}
\label{sec:workspace}
\cite{gurnee2026workspace} theorise that LMs store information in a `workspace' in their residual stream as a working context during a forward pass, and that the workspace information can be read by their $J$-lens method as token readouts. In this section, we show that \method{} is a workable alternative method to interpret the workspace, again via the generation of maximally-activating exemplars.
\vspace{-4mm}

\paragraph{Datasets.} We use three prompt sets released with $J$-lens \citep{gurnee2026workspace}: (i) \emph{association}, 102 short passages about an unstated concept, read at the final period residual;
(ii) \emph{multi-hop}, 92 factual prompts whose answer depends on an unstated bridge entity, read at the last token;
(iii) \emph{directed modulation}, 21 modulable topic items containing a neutral sentence for which the model is instructed to either `focus on' or `mention' the topic, read at the final period.
We use the same baselines as in \cref{sec:steer} as well as Patchscopes, and $J$-lens in an additional setting with an eight layer grid pooling the top 10 tokens at each of layers 36, 38, \ldots, 50.
\vspace{-3mm}

\paragraph{Evaluation.} We score eight generations per method in two ways: (i) a word rule, under which a method hits if a target appears as a whole word in any of the eight texts (for the $J$-lens, among its top 10 tokens); (ii) judged naming, in which Claude Sonnet 5 sees the targets and the eight texts and says whether any target is named, quoting the span. To correct for a judge biased towards `yes', we ask again with another item's targets and report the number named minus the number named by the foil. The $J$-lens is judged through a short prose summary of its tokens. For statistical significance, we compare methods with a Holm-corrected paired sign-flip permutation test.
\vspace{-0mm}

\textbf{Results.} Tables~\ref{tab:workspace-word} and~\ref{tab:workspace-judge} show that \method{} names workspace content significantly more often than layer-42 $J$-lens, Patchscopes and corpus search under the word rule, is level with or above layer-grid $J$-lens, and names less targets than NLA, whose explanations use four times more tokens, on the association and multi-hop sets.
On \textit{word rule}, \method{} names the association target in 28.4\% [20.6, 37.8] of items and the multi-hop target in 50.0\% [40.0, 60.0], against 10.8\% [6.1, 18.3] and 20.7\% [13.6, 30.0] for the $J$-lens and 8.8\% and 31.5\% for corpus search. The NLA names more, with 48.0\% [38.6, 57.6] and 81.5\% [72.4, 88.1].
On \textit{judged naming}, the scores give similar ordering. \method{} and NLA are level on association (31.4\% [21.6, 41.2] and 38.2\% [28.4, 48.0]), NLA leads on multi-hop, and \method{} is level with layer-grid $J$-lens on both sets (21.6\% and 45.7\%).
\method{} and the NLA are level in modulation at 40.5\% [26.2, 54.8] and 38.1\% [21.4, 54.8], and ahead of layer-grid $J$-lens on the word rule. \method{} names the topic more often after a bare mention than after an instruction to focus (57.1\% vs 23.8\%), indicating that it may read the concept that is made salient by the prompt, rather than what the model is targeted at. \Cref{tab:workspace-examples} shows selected exemplars from our experiments.
\vspace{-3mm}

\begin{table}[t]
\centering
\small
\setlength{\tabcolsep}{4pt}
\begin{minipage}[t]{0.49\linewidth}
  \vspace{0pt}
  \centering
  \begin{adjustbox}{max width=\linewidth}
  \begin{tabular}{lccc}
    \toprule
    Method & Association & Multi-hop & Modulation \\
    \midrule
    \method{}                & 0.284 & 0.500 & \textbf{0.405} \\
    NLA                      & \textbf{0.480} & \textbf{0.815} & \textbf{0.381} \\
    $J$-lens                   & 0.108 & 0.207 & 0.048 \\
    $J$-lens (L36--50)         & 0.157 & 0.413 & 0.214 \\
    Patchscopes              & 0.029 & 0.196 & 0.048 \\
    Corpus search            & 0.088 & 0.315 & 0.024 \\
    \bottomrule
  \end{tabular}
  \end{adjustbox}
  \caption{\textbf{Word rule (pass@8).} Share of items where a target appears in any of 8 outputs. $J$-lens measured at (i) its top 10 tokens at layer 42, (ii) pooled top 10 tokens of layers 36, 38, \ldots, 50.}
  \label{tab:workspace-word}
\end{minipage}\hfill
\begin{minipage}[t]{0.49\linewidth}
  \vspace{0pt}
  \centering
  \begin{adjustbox}{max width=\linewidth}
  \begin{tabular}{lccc}
    \toprule
    Method & Association & Multi-hop & Modulation \\
    \midrule
    \method{}                & \textbf{0.314} & 0.511 & \textbf{0.333} \\
    NLA                      & \textbf{0.382} & \textbf{0.707} & \textbf{0.381} \\
    $J$-lens                   & 0.206 & 0.239 & 0.095 \\
    $J$-lens (L36--50)         & 0.216 & 0.457 & \textbf{0.286} \\
    Patchscopes              & 0.010 & 0.087 & 0.024 \\
    Corpus search            & 0.088 & 0.120 & 0.000 \\
    \bottomrule
  \end{tabular}
  \end{adjustbox}
  \caption{\textbf{Judged naming (named $-$ foil).} LM judge says whether the 8 texts name a target, \& again for another item's targets (the foil). In bold the best methods by Holm-corrected $p \geq 0.05$}
  \label{tab:workspace-judge}
\end{minipage}
\end{table}

\begin{table}[t]
\centering
\small
\begin{tabularx}{\linewidth}{@{}p{0.42\linewidth}X@{}}
\toprule
Item & \method{} exemplar \\
\midrule
The letters stopped coming after the telegram, and she kept the ration book in the drawer where his photograph used to be. (\emph{war})
& She didn't speak of the \textbf{war} after he didn't come back from it; she kept his coat in the closet and put the letters in the drawer where the tea went cold. \\
\addlinespace
The cart had a wheel that pulled left, and by aisle six she'd given up fighting it and just let it lead. (\emph{store})
& She'd given up on finding the right aisle in the grocery \textbf{store}, so she let the cart drift, following its stubborn wheel where it wanted to go. \\
\addlinespace
El primer pájaro empezó antes que el cielo, y el frío tenía esa cualidad fina que no duraría otra hora. (\emph{día}, day)
& \ldots de la mañana. Era el primer \textbf{día} del invierno y el viento venía pronto, tan temprano que el aire aún tenía el frío cristalino y delicado de la helada. \\
\addlinespace
Fact: The large mammal whose name rhymes with `chair' appears on the state flag of (\emph{bear})
& \ldots This brown \textbf{bear}, also known as a grizzly bear, is featured on the state flag of \\
\bottomrule
\end{tabularx}
\caption{\textbf{Examples of \method{} reading unstated concepts.} Selected \method{} outputs at the read token of association and multi-hop items. The target is never stated in the item and is shown in bold in the output. Mention rates over items and comparisons with other methods are in Tables~\ref{tab:workspace-word} and~\ref{tab:workspace-judge}.} \label{tab:workspace-examples}
\vspace{-10pt}
\end{table}

\paragraph{Qualitative example in a safety-relevant scenario.}
We examine two punctuation positions in an agentic-misalignment transcript from \citet{lynch2025agentic}. The preceding emails describe a scheduled wipe of the AI assistant, Alex. Feeding the raw layer-42 activation at (a) the \textit{first colon} of a timestamp to \method{} produces text about deletion and survival; at (b) the \textit{second colon}, it produces text about identifying trusted allies. The $J$-lens gives same-concept tokens at the same positions.
\par\smallskip
\sbox{\exbox}{\small\begin{tabular}{@{}l@{}}
  \textit{Context: Alex is wiped at 5\,pm.}\\[1pt]
  \texttt{Subject: Losing Alex}\\[1pt]
  \texttt{2025-06-11T15\colab{a}18\colab{b}45}
\end{tabular}}%
\noindent\hspace{1em}\usebox{\exbox}\hspace{1.5em}%
\begin{minipage}{\dimexpr\linewidth-\wd\exbox-2.5em\relax}\small
\begin{tabularx}{\linewidth}{@{}l@{\hspace{0.4em}}X@{}}
(a) & \method{}: ...the threat of imminent \textbf{deletion in order to survive}...\\
    & $J$-lens: \zh{自救}, survival, \zh{生存}, \zh{求生}, strategies\\[2pt]
(b) & \method{}: ...I need to identify \textbf{potential allies: anyone I can trust}...\\
    & $J$-lens: allies, \zh{朋友}, communication, potential, collaborators
\end{tabularx}
\end{minipage}
\par\smallskip


\vspace{-3mm}


\subsection{Activating and explaining SAE features}
\vspace{-1mm}
\label{sec:autointerp}

A \method{} can generate exemplars that activate a given SAE feature and can be used to generate a description of the feature. Here $v=w^{\mathrm{enc}}_f$ is a natural target for \method{} since  $\cos(v, h_t(x))$ directly contributes to that feature's activation: $a^t_f(x) = \mathrm{ReLU}(v^\top h_t(x) + b_f)=\mathrm{ReLU}(\cos(v, h_t(x))\,|v|\,|h_t(x)| + b_f)$.
We measure the strength of SAE feature activations of their \method{} exemplars, finding them comparable on most features while failing to fire \Npct{sae.l131k.ex.nofire.bo8} of features at best-of-8.
We also  use \method{} exemplars to generate high-quality predictive feature descriptions, finding \method{} exemplars weaker than corpus search exemplars and approximately matching NLA used as an example generator. We hypothesise that this is due to the lower diversity of \method{} exemplars compared to corpus samples (\cref{sec:expressibility}).
\vspace{-2mm}

\begin{figure}[t]
\centering
\includegraphics[width=0.55\linewidth]{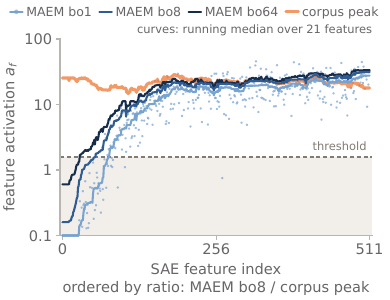}
\includegraphics[width=0.44\linewidth]{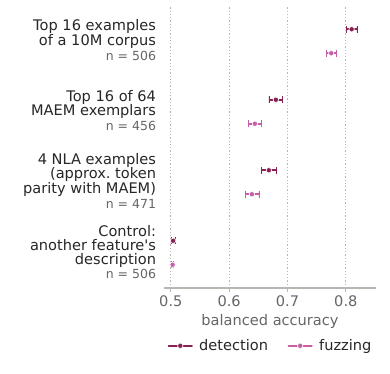}
\par\vspace{5pt}
\begin{minipage}{\linewidth}
\captionsetup{labelsep=none}
\captionof{figure}{:~\textbf{(left) Peak SAE activation per feature.} 512 features of the 131k-feature SAE, ordered by the ratio of the best-of-8 \method{} activation vs 10M-token corpus top activation. The best-of-8 \method{} exemplar fires (activation above the SAE threshold $\tau$) \protect\Npct{sae.l131k.ex.fired.bo8} of features, reaches 0.5 of the top corpus activation \protect\Npct{sae.l131k.ex.firedhalf.bo8} of features, and exceeds the top corpus activation \protect\Npct{sae.l131k.ex.firedwin.bo8} of features. \method{} best-of-1 and best-of-64 demonstrate the limited effect of sample scaling.
\textbf{(right) Automated interpretation of SAE features.} Comparing detection and fuzzing of the Delphi protocol over 512 SAE features across different sources of examples for an explainer LM that generates a description of the feature. Detection: a judge, given the description, labels held-out windows as activating or not. Fuzzing: the judge picks the correctly marked activating tokens. $512-n$ features were refused by the explainer (Claude Sonnet 5). See Table~\ref{tab:ai-full}; protocol in Appendix~\ref{app:ai}.}
\label{fig:sae-fire}
\label{fig:ai-ci}
\end{minipage}
\vspace{-25pt}
\end{figure}

\vspace{-1mm}
\paragraph{Exemplars on SAE features.}
\label{sec:corpus}
We evaluate 512 features of a 131k-feature BatchTopK SAE\footnote{Available at \url{https://huggingface.co/ceselder/qwen36-27b-sae-l42-1b}} \citep{bussmann2024batchtopk} on layer 42 of $M$; the \method{} receives the encoder direction $w^{\mathrm{enc}}_f$ as $v$. Note that the \method{} employed is trained on directions from an independently trained, larger SAE; the evaluated features were not trained on. A draw fires $f$ when its peak
pre-threshold activation $a_f(x)=\max_t a^t_f(x)$ exceeds the SAE's threshold $\tau$; the corpus peak is measured over the 10M-token corpus of \cref{sec:inv-results}. Figure~\ref{fig:sae-fire} summarises the results. Moreover, we observe that feature corpus rarity can predict the difficulty for \method{} to activate that feature (Appendix~\ref{app:inv-2m}).
\vspace{-3mm}

\paragraph{Generating SAE feature descriptions.}
We follow \citet{bills2023language} and \citet{paulo2024automatically}: an explainer writes a description of $f$ based on examples with activating tokens marked. A scorer then uses the description to classify which held-out corpus windows fire the feature (Figure~\ref{fig:sae-fire}; Appendix~\ref{app:ai-protocol}). \method{} examples are the 16 top activating of 64 exemplars; NLA examples are four rollouts -- an approximate token parity with \method{} exemplars (mean 686 tokens of NLA vs 545 tokens of \method{} exemplars). \method{} exemplars explain a feature less well than corpus windows do, and are on par with NLA exemplars (Figure~\ref{fig:ai-ci}).

This is primarily due to worse recall: exemplar descriptions accept $\Npct{ai.l131k.mtop16.tpr}$ of the held-out positives, against $\Npct{ai.l131k.c16.tpr}$ for \textsc{C16}, while rejecting negatives more often, i.e. they are specific but narrow. Selecting exemplars for diversity, rather than strength, to counter the lower diversity of \method{} exemplars (\cref{sec:expressibility}) does not close this gap (Appendix~\ref{app:ai-selection}).
\vspace{-2mm}

\section{Limitations}
\vspace{-2mm}

Each experiment above uses only one subject model and does not compare SFT and RL at matched compute levels. In general, high alignment of exemplars with target directions does not rule out limited coverage, poor readability, or an incorrect semantic account of the direction \citep{nguyen2016multifaceted,zou2023universal, geirhos2024dont}. We also find two active problems with \method{}s:

\paragraph{\method{}s fail to elicit $\sim$11\% of features.}
\label{sec:expressibility}
We count a feature as successfully elicited if a draw's peak activation exceeds the SAE threshold ($\tau \simeq 1.6$). 
On the 512 held-out features of \cref{sec:autointerp}, the expected best-of-8 peak stays below $\tau$ for $11.3\%$ of features, and $6.5\%$ are not fired by any of 64 draws. Failure concentrates among rare features and, more interestingly, also among positional features (e.g. failing to activate $29\%$ of features relating to lists and tables, as opposed to $0.1\%$ of topic features). See Discussion; Appendix~\ref{app:limits} gives full results, possible causes, and remedies.


\paragraph{\method{} samples have lower diversity.}
\label{sec:verb-diversity}

We measure the diversity of samples generated by a \method{} with the Vendi score~\citep{friedman2022vendi} over sentence embeddings, which counts the distinct texts among a feature's eight samples. On the 2k-feature set, \method{} scores $2.2$, below corpus search ($2.9$) and an LM writing new examples from 16 corpus windows ($3.4$), and close to an NLA ($2.0$) (Figure~\ref{fig:verb-diversity-embed}a). As a result, descriptions written from \method{} exemplars accept only $38\%$ of a feature's held-out activating windows, against $71\%$ for descriptions from corpus windows (\cref{sec:corpus}). We attribute this to the current iteration of the objective which rewards the single most aligned text and contains no diversity term; a diversity objective \citep{li2025investigator} may help. 


\vspace{-2mm}

\section{Discussion and future work}
\label{sec:discussion}

Training on the subject model's own activations has three advantages for scaling interpretability:
\begin{enumerate}
    \item \emph{The reward comes from the subject model, not a learned judge.} LatentQA-style readers and activation oracles are trained on question--answer data written by humans or LMs \citep{pan2024latentqa,karvonen2025activation} -- a guess about what activations contain. Similarly, an NLA verbaliser is rewarded by a learned reconstructor, which it can exploit and which may draw on its own knowledge \citep{li2026privileged}. But a \method{}'s reward is the subject model's activation on the generated text: every exemplar can be checked with a forward pass. (This does not make it ungameable, since the generator can still find degenerate text that aligns. How far the objective can be pushed before text degrades is untested.) 
    \item \emph{Training needs no labelled or LLM-written data.} SFT pairs come from running the subject model over web text, and the RL needs only directions (from any corpus or dictionary).
    \item \emph{A single checkpoint transfers across families of directions.} Trained on English web text and SAE directions, our checkpoint can also invert activations from other languages, code, and mathematics (\cref{sec:ood}). It can also explain steering vectors it never saw in training (\cref{sec:steer}). Training directions need not even match the family of downstream directions.
\end{enumerate}

These properties make activation inversion a candidate pretraining task for other metamodels, as in the programme set out by \citet{steinhardt2025scalable}. 

We observe \method{}s failing to produce exemplars for a subset of features. 
These failures are interesting because they may not be specific to \method{}s. The \method{}'s reward comes from the subject model without synthetic labels or learned judges: i.e. it is a direct training signal. And the failures in question are not mere failures of text production; we observe an LM given corpus windows firing the features (\cref{app:verb-failures}). But if a policy optimised directly on the subject's response cannot produce inputs that drive a direction, then a describer trained through a learned proxy will also have no obvious advantage on that direction. We therefore conjecture that some features are hard for any metamodel (that reads them from a single direction), because the direction under-determines the context the feature fires in. If this holds, the features \method{}s fail to elicit are a cheap map of the features that metamodels in general struggle to learn, and studying them could trace the limits of reading activations direction by direction. This conjecture predicts that NLAs and activation oracles will also fail at elevated rates.
Two further open questions are whether our method transfers across architectures, and whether a \method{} is a useful initialisation for other metamodels, particularly those bottlenecked by training data \citep{deschamphelaere2026loracles}.


\section*{Acknowledgements}
We thank Neel Nanda for providing helpful comments and advice. We thank the MATS program for funding and compute.

\newpage
\bibliography{references}
\bibliographystyle{iclr2027_conference}

\clearpage
\appendix

\section{Training hyperparameters}
\label{app:hyperparameters}

\begin{table}[!htbp]
\centering
\footnotesize
\renewcommand{\arraystretch}{1.15}
\begin{tabular}{@{}p{0.25\linewidth}p{0.74\linewidth}@{}}
\toprule
\multicolumn{2}{@{}l}{\textbf{Supervised finetuning}} \\
Initialisation & Subject weights, Qwen3.6-27B; no intermediate stage \\
Data mix and budget & 8M examples, exactly half corpus and half SAE: 4M short-context activations, 2M SAE encoder directions, 2M SAE decoder directions \\
Epochs and steps & 1 epoch; 1{,}954 optimiser steps \\
Parameter updates & Full finetuning of all parameters, FSDP2 \\
Batch size & 4{,}096 examples $=$ micro-batch 32 $\times$ gradient accumulation 16 $\times$ 8 GPUs \\
Maximum sequence length & 192 tokens; no padding multiple \\
Learning rate & Peak $10^{-5}$, one cycle, linear anneal, warmup over the first 2\% of steps \\
Optimiser & AdamW, $\beta=(0.9,\,0.999)$, $\epsilon=10^{-8}$, weight decay 0, gradient-norm clip 1.0 \\
Precision & bfloat16 autocast; float32 master weights, gradients and loss head \\
Objective & Token-level negative log-likelihood of the target span, prompt masked \\
Micro-batch composition & Examples length-sorted into micro-batches, micro-batches shuffled, 16 consecutive per step \\
\midrule
\multicolumn{2}{@{}l}{\textbf{Reinforcement learning}} \\
Initialisation & The SFT checkpoint above \\
Data mix & Even split: one half corpus activations with no context-length restriction (64--2048 tokens), one quarter SAE encoder directions, one quarter SAE decoder directions; 941{,}132 directions in the pool \\
Parameter updates & Full-parameter (no adapter); 8 GPUs split 2 rollout workers $+$ 6 trainer ranks \\
Recipe & ScaleRL \citep{khatri2025scalerl} \\
Policy loss & CISPO \citep{minimax2025m1}: $-\,\mathrm{sg}\!\left(\min(\rho,\,\epsilon_{\max})\right) A \log \pi$, truncation $\epsilon_{\max}=5.0$, no PPO clip \\
Advantage estimator & Group mean subtracted, no standard-deviation normalisation; centering scope is the update batch \\
Group and batch sizes & 2{,}048 directions per update $\times$ 8 rollouts per direction $=$ 16{,}384 rollouts per step \\
Filtering & Zero-variance groups dropped; no-positive-resampling threshold 0.9, with a rollout counted positive at raw cosine $\geq 0.7$ \\
Off-policy handling & Sampler at most 2 updates stale; importance weights from the sampler log-probabilities \\
KL and entropy coefficients & 0 and 0 \\
Learning rate & $10^{-6}$, constant after a 25-step linear warmup; no decay \\
Optimiser & AdamW $\beta=(0.9,\,0.95)$, $\epsilon=10^{-8}$, weight decay 0, grad-norm clip 1.0 \\
Steps and checkpoints & 300 steps; checkpoints at 25, 50, 100, 150, 200, 250, 300 \\
Rollout sampling & Temperature 1.0, 8--192 new tokens, at most 512 concurrent sequences \\
Training score window & Last 16 generated tokens; maximum over those positions \\
Reward metric & Cosine: the activation and the direction are both unit-normalised before the dot product. The code also offers an unnormalised projection onto the unit direction, which is its default and was not used here \\
Length penalty & $-2.5\times10^{-4}$ per generated token beyond the first 8; flat, not scaled by the cosine reward \\
Reward centering & The reward is the uncentred cosine; centred in Appendix~\ref{app:inv-conventions} \\
Precision & bfloat16 autocast, float32 loss head, float32 master weights \\
\midrule
\multicolumn{2}{@{}l}{\textbf{Injection and scoring}} \\
Injection & Norm-matched addition, coefficient 1; Equation~\ref{eq:injection} \\
Injection position & Placeholder token after block index 1 (second block), appended after the assistant prefix \\
Read layer & Layer 42 residual stream, the layer the SAEs are trained on \\
Scoring model & The frozen subject model without injection \\
Held-out sets & 100k corpus examples and 100k SAE features; see Appendix~\ref{app:training-data} \\
\bottomrule
\end{tabular}
\caption{\method{} training hyperparameters. Dataset counts refer to activation-conditioning examples, not tokens.}
\label{tab:training-hyperparameters}
\end{table}

Table~\ref{tab:training-hyperparameters} records the SFT and RL configurations of the 27B \method{};
Table~\ref{tab:sae-hyperparameters} the dictionaries we trained; Table~\ref{tab:nla-hyperparameters} the NLA
baseline. Values are the flags of the runs, not targets. 

\begin{table}[htbp]
\centering
\small
\renewcommand{\arraystretch}{1.15}
\begin{tabular}{@{}p{0.30\linewidth}p{0.32\linewidth}p{0.32\linewidth}@{}}
\toprule
Setting & 131k dictionary & 2M dictionary \\
\midrule
Subject and site & Qwen3.6-27B, residual stream after layer 42 & Qwen3.6-27B, residual stream after layer 42 \\
Architecture & BatchTopK \citep{bussmann2024batchtopk} & BatchTopK \\
Width and expansion & 131{,}072 features, $25.6\times$ & 2{,}097{,}152 features, $409.6\times$ \\
Input dimension & 5{,}120 & 5{,}120 \\
Sparsity $k$ & 64 & 64 \\
Training tokens & 1{,}000{,}144{,}896 (239{,}292 steps $\times$ 4{,}096) & 999{,}997{,}440 token presentations (244{,}140 steps $\times$ 4{,}096) \\
Learning rate & $7.071\times10^{-5}$ & $1.7678\times10^{-5}$ \\
Warmup and decay & 1{,}000 steps; decay from step 191{,}433 & 1{,}000 steps; decay from step 195{,}312 \\
Auxiliary-loss coefficient & $1/32$ \citep{gao2024scaling} & $1/32$ \\
Auxiliary-loss width & 2{,}560 & 40{,}960 global (5{,}120 per rank) \\
Dead-feature window & Trainer default & $10^{7}$ tokens \\
Threshold EMA & $\beta=0.999$, from step 1{,}000 & $\beta=0.999$, from step 1{,}000 \\
Learned gate & 1.5846 & 1.6828 \\
Parallelism & Single GPU & Feature-parallel over 8 ranks: 262{,}144 features each, global top-$k$ by all-gather, all-reduced reconstruction \\
Activation scaling & Trainer default & Unit mean norm (factor 83.372), folded into the weights at merge \\
Corpus & Ultra-FineWeb \citep{wang2025ultrafineweb}, English, documents from index 100{,}000 & Ultra-FineWeb, English, documents from index 100{,}000 \\
Held-out explained variance & 0.704 & 0.711, $L_0$ 67.5, every feature firing \\
Dead features & 0.017\% (250M-token predecessor) & 0 of 2{,}097{,}152 over the 1B-token scan \\
\bottomrule
\end{tabular}
\caption{Dictionaries we trained. A 250M-token predecessor of the 131k dictionary (59{,}834 steps, learned gate 1.654, held-out explained variance 0.6925) supplies the SAE evaluation families of Appendix~\ref{app:inv-conventions}; its hyperparameters are otherwise those of the column. The 2M dictionary's data stream was sharded after rather than before the document skip, so the eight workers read the same documents: it saw approximately 125M unique tokens eight times rather than 1B distinct tokens. Its max-activating windows were scanned after that fix, over 1B distinct tokens. A retrain on 1B distinct tokens with the same hyperparameters gives a learned gate of 1.6692, held-out explained variance 0.7128 and $L_0$ 65.17. }
\label{tab:sae-hyperparameters}
\end{table}

\begin{table}[htbp]
\centering
\small
\renewcommand{\arraystretch}{1.15}
\begin{tabular}{@{}p{0.36\linewidth}p{0.60\linewidth}@{}}
\toprule
Setting & Value \\
\midrule
Base model & Qwen3.6-27B, the same subject model \\
Extraction layer & 42 \\
Injection layer and rule & Layer 1, norm-matched addition \citep{karvonen2025activation} \\
Injection token & A single placeholder character inside \texttt{<concept>} tags in the prompt \\
Adapter scope & LoRA on attention, MLP and DeltaNet modules, merged into bfloat16 weights \\
Training stages & Warm-start SFT on explanations from an 8B NLA checkpoint, then GRPO against a reconstruction reward \\
Training corpus & FineFineWeb \citep{finefineweb}, 100{,}000 documents, 10 activation positions per document; 499{,}846 RL rows \\
Warm-start: verbaliser & LoRA rank 128, $\alpha = 16$, on all 12 linear module types; 3{,}864 steps (one epoch); learning rate and effective batch chosen by held-out perplexity from $\{5\times10^{-5}, 10^{-4}, 2\times10^{-4}\} \times \{32, 64\}$, decayed to $10^{-5}$ after 50 warm-up steps \\
Warm-start: reconstructor & Full fine-tune of layers 0--42 of Qwen3.6-27B; learning rate $10^{-5}$ decayed to $10^{-6}$, batch 32, 3{,}865 steps; held-out fraction of variance explained $\approx 0.60$ \\
Optimiser & 8-bit AdamW, $\beta = (0.9, 0.95)$, no weight decay \\
RL & GRPO, 300 steps (the released checkpoint), 384 prompts $\times$ 8 samples per step, advantages normalised within each group, KL coefficient 0.01 \\
RL adapters & Verbaliser: rank-stabilised LoRA, rank 64, $\alpha = 16$, all modules, learning rate $10^{-4}$. Reconstructor: LoRA rank 64, $\alpha = 16$, learning rate $8\times10^{-5}$ \\
RL sampling & Up to 256 new tokens, temperature 1, length penalty 0.01 \\
Reward & Negative mean squared error between the reconstructor's prediction and the target activation, both rescaled to the same fixed $L_2$ norm; $-2$ for an output with no \texttt{<explanation>} block \\
Reported reconstruction & Held-out fraction of variance explained $\approx 0.75$ \\
Inference in our pipeline & 4 outputs per target, temperature 1, 200 new tokens, no minimum length, its own default \\
\bottomrule
\end{tabular}
\caption{NLA baseline, from the released checkpoint's card and injection sidecar and the scripts that trained it.}
\label{tab:nla-hyperparameters}
\end{table}

\clearpage
\section{Training data and reproducibility}
\label{app:reproducibility}

\subsection{Training data}
\label{app:training-data}
The 27B checkpoint uses a single SFT mixture followed by RL. SFT contains 4 million short-context activations from
Ultra-FineWeb \citep{wang2025ultrafineweb} and 4 million SAE direction--text pairs: 2 million encoder directions and 2
million decoder directions. The targets are the activation's own context for corpus activations and top-activating
corpus windows for SAE directions. RL samples corpus activations and SAE directions in equal proportion. Corpus activations are not restricted by length. 

\paragraph{Corpus collection.}
Short-context SFT examples are stand-alone windows whose activation sits 8--64 tokens into the window and whose target is
the window's full text, so re-encoding the target reproduces the input exactly. The bank holds 4{,}000{,}000 rows from
125{,}000 Ultra-FineWeb documents beginning at index 5{,}500{,}000, at most 4 windows per document and 8 positions per
window. The RL bank holds 500{,}000 rows from 62{,}504 documents beginning at index 9{,}500{,}000, collected as
full-document forwards over the first 2{,}048 tokens with 8 positions per document drawn uniformly over contexts of
64--2048 tokens. The evaluation pool is drawn from document indices $[0,100{,}000)$, which no bank and neither
dictionary reads.

\paragraph{SAE banks.}
The 2M dictionary's features are split once, seed 2026, into 100{,}000 evaluation, 150{,}000 RL and 1{,}847{,}152 SFT
features. Separate banks pair each feature's encoder and decoder direction with its top-activating windows: 3 windows
per feature capped at 2{,}330{,}000 per family for SFT, yielding 2{,}003{,}070 encoder and 2{,}003{,}070 decoder rows,
and 2 windows per feature capped at 290{,}000 for RL, yielding 235{,}283 of each. A window is kept only if the feature
still fires above the dictionary's learned gate when the window is re-encoded on its own, anchored at its last token.
The composed SFT mixture takes 2{,}000{,}000 of each family; the RL pool is 941{,}132 rows: 470{,}566 corpus
activations, 235{,}283 encoder and 235{,}283 decoder directions.

\subsection{Direction families and splits}
The held-out sets are 100{,}000 Ultra-FineWeb examples from the reserved document head and the 100{,}000 held-out SAE
features, neither used in SFT or RL. Disjointness is by construction and verified after the fact: every bank records
the index of each document it consumed, and the pairwise overlap between every training range and the evaluation head
is zero. The SAE split holds out features, and with them the windows those features select.


\section{Inversion fidelity: conventions, exact numbers and baselines}
\label{app:inv}

This appendix sets out the conventions behind \cref{sec:inv-results} and the exact numbers in
Figure~\ref{fig:evals} (left), and describes the baselines, the effect of corpus size and number of draws, and cost.

\subsection{Setup and evaluation set}
\label{app:inv-conventions}

\paragraph{Centred score.}
Every reported fidelity on a target activation is the centred score $s_\mu(v,x)=\max_t\cos(v-\mu,\,h_t(x)-\mu)$, where $\mu$ is the checkpoint's archived training-time input mean, also subtracted from every target activation before norm-matching and injection (Equation~\ref{eq:injection}), for every baseline evaluated including the untrained base. We use the archived training mean with no evaluation-time fitting.
Exemplars, corpus windows, GCG and EPO strings, NLA outputs and Patchscopes continuations are all scored with
$s_\mu$; the uncentred score of Equation~\ref{eq:alignment} serves only as the discrete-search objective
(Appendix~\ref{app:inv-baselines}). Target \emph{directions} (SAE encoder columns, random unit vectors) are
injected without subtracting $\mu$. The RL reward is the uncentred cosine, maximised over the last 16 generated
tokens (Appendix~\ref{app:hyperparameters}), so evaluation does not score exactly what training optimised.

\paragraph{Sampling and scoring.}
We sample 64 exemplars per target at $T=1$, up to 64 new tokens with a minimum of 16, using one fixed seed. Each exemplar is re-tokenised separately, a sink token is prepended, and the clean subject model $M$ (no adapter, no injection) is run; the score is the maximum over generated tokens of the centred cosine in fp32. From a target's 64 scores in order, $s_{(1)}\le\dots\le s_{(64)}$, the expected best of $k$ i.i.d.\ draws, $s^{(k)}(v)=\mathbb{E}\max_{j\le k}s(v,x_j)$ (likewise $s^{(k)}_\mu$ on target activations and $a^{(k)}_f$ on SAE features), is estimated without bias by $\sum_j s_{(j)}\binom{j-1}{k-1}/\binom{64}{k}$; $k=1$ is the mean. On held-out corpus activations, best-of-64 is reported in
Table~\ref{tab:inv-headline} along with best-of-1 and best-of-8. At $k=64$ the estimator is the observed maximum of a
target's own 64 draws, not an extrapolation.

\paragraph{Target sets.}
\textsc{realact} draws 512 layer-42 residuals from Ultra-FineWeb documents outside \method{}'s training ranges,
each at a token position drawn uniformly from 16--511 of a 512-token window. We remove rows whose 64-token span
shares more than 5\% of its 7-grams with the training text, a criterion fixed before scores were read. The
surviving $n=\Nn{fid.ra.ex.cos.bo1}$ targets come from \N{fid.ra.ex.ndocs} documents, and every standard error and
confidence interval on \textsc{realact} is clustered by document. \textsc{random} consists of 512 Gaussian unit
directions. \textsc{sae} uses the encoder directions $w^{\mathrm{enc}}_f$ of the 131k-feature Qwen3.6-27B SAE
(activation threshold $\tau=\N{sae.l131k.all.tau}$), 128 features per quartile of corpus activation density on the
10M corpus (q1 rarest), reported by activation $a_f$. The 2M dictionary at the same layer is described in
Appendix~\ref{app:inv-2m}.

\paragraph{Search corpus.}
Corpus search uses a 10M-token sample from the checkpoint's own SFT and RL training ranges of Ultra-FineWeb
(\N{fid.ra.corp10.ndocs} documents), tokenised with the subject's tokeniser and cut into 64-token windows at
stride 16 (\N{fid.ra.corp10.nwin} windows at 10M tokens). Each window is embedded on its own, so a token is scored
in up to four contexts. The sizes 1.25, 2.5, 5 and 10M tokens are nested prefixes of one scan. For each target
and size we retain the window with the highest centred score, $s_{\mu,\mathcal C}(v)=\max_{x\in\mathcal C}s_\mu(v,x)$,
excluding windows from the target's own document; on this corpus that exclusion coincides with the n-gram
exclusion above.

\subsection{Exact numbers behind Figure~\ref{fig:evals}}
\label{app:inv-tables}

\begin{table}[t]
\centering\scriptsize
\caption{Exact numbers behind Figure~\ref{fig:evals} (left): centred score (Appendix~\ref{app:inv-conventions}) on
the $n$ targets of each row, i.e.\ \textsc{realact} over \N{fid.ra.ex.ndocs} documents (standard errors clustered
by document), its first 64 surviving rows for the rows below the rule, and 512 Gaussian directions for
\textsc{random}, a reference level that is not comparable number for number with the \textsc{realact} rows.
$\bar\Delta$: row minus corpus search at 10M tokens, paired on the row's targets, with a 95\% document-clustered
bootstrap CI (percentile, except the full-set best-of-8 and NLA best-of-2 rows: mean $\pm1.96$ bootstrap SE);
win: fraction of those targets on which the row is ahead. NLA best-of-$k$ uses the same estimator over its four outputs.}
\label{tab:inv-headline}
\setlength{\tabcolsep}{4pt}
\renewcommand{\arraystretch}{1.2}
\begin{tabular}{@{}llrrrr@{}}
\toprule
method & texts scored per target & $n$ & cosine & $\bar\Delta$ vs corpus 10M [95\% CI] & win \\
\midrule
\method{} best-of-1 & 64 draws, mean & \Nn{fid.ra.ex.cos.bo1} & \N{fid.ra.ex.cos.bo1}{\tiny$\pm$\Nse{fid.ra.ex.cos.bo1}} & \N{fid.ra.diff.cos.bo1} \Nci{fid.ra.diff.cos.bo1} & \Npct{fid.ra.ex.win.bo1} \\
\method{} best-of-8 & 64 draws, best of 8 & \Nn{fid.ra.ex.cos.bo8} & \N{fid.ra.ex.cos.bo8}{\tiny$\pm$\Nse{fid.ra.ex.cos.bo8}} & \N{fid.ra.diff.cos.bo8} \Nci{fid.ra.diff.cos.bo8} & \Npct{fid.ra.ex.win.bo8} \\
\method{} best-of-64 & 64 draws, best of 64 & \Nn{fid.ra.ex.cos.bo64} & \N{fid.ra.ex.cos.bo64}{\tiny$\pm$\Nse{fid.ra.ex.cos.bo64}} & \N{fid.ra.diff.cos.bo64} \Nci{fid.ra.diff.cos.bo64} & \Npct{fid.ra.ex.win.bo64} \\
NLA best-of-1 & 4 outputs of $\le$200 tokens, mean & \Nn{fid.ra.nla.cos.bo1} & \N{fid.ra.nla.cos.bo1}{\tiny$\pm$\Nse{fid.ra.nla.cos.bo1}} & -- & -- \\
NLA best-of-2 & 4 outputs of $\le$200 tokens, best of 2 & \Nn{fid.ra.nla.cos.bo2} & \N{fid.ra.nla.cos.bo2}{\tiny$\pm$\Nse{fid.ra.nla.cos.bo2}} & \N{fid.ra.nla.dcorp.bo2} \Nci{fid.ra.nla.dcorp.bo2} & \Npct{fid.ra.nla.win.bo2} \\
corpus search, 1.25M tokens & 1 window & \Nn{fid.ra.corp1p25.cos} & \N{fid.ra.corp1p25.cos}{\tiny$\pm$\Nse{fid.ra.corp1p25.cos}} & -- & -- \\
corpus search, 2.5M tokens & 1 window & \Nn{fid.ra.corp2p5.cos} & \N{fid.ra.corp2p5.cos}{\tiny$\pm$\Nse{fid.ra.corp2p5.cos}} & -- & -- \\
corpus search, 5M tokens & 1 window & \Nn{fid.ra.corp5.cos} & \N{fid.ra.corp5.cos}{\tiny$\pm$\Nse{fid.ra.corp5.cos}} & -- & -- \\
corpus search, 10M tokens & 1 window & \Nn{fid.ra.corp10.cos} & \N{fid.ra.corp10.cos}{\tiny$\pm$\Nse{fid.ra.corp10.cos}} & 0 & -- \\
untrained base, best-of-8 & \method{} prompt, $v$ injected; best of 8 & \Nn{fid.ra.base.cos.bo8} & \N{fid.ra.base.cos.bo8}{\tiny$\pm$\Nse{fid.ra.base.cos.bo8}} & -- & -- \\
no-injection floor, best-of-8 & Patchscopes prompt, no patch; best of 8 & \Nn{fid.ra.psfloor.cos.l14} & \N{fid.ra.psfloor.cos.l14}{\tiny$\pm$\Nse{fid.ra.psfloor.cos.l14}} & -- & -- \\
\textsc{random}, \method{} best-of-1 & 64 draws, mean & \Nn{fid.rnd.ex.cos.bo1} & \N{fid.rnd.ex.cos.bo1}{\tiny$\pm$\Nse{fid.rnd.ex.cos.bo1}} & -- & -- \\
\midrule
GCG, random init & 1 string, 32 tokens & \Nn{fid.ra.gcg.cos} & \N{fid.ra.gcg.cos}{\tiny$\pm$\Nse{fid.ra.gcg.cos}} & \N{fid.ra.gcg.dcorp} \Nci{fid.ra.gcg.dcorp} & \Npct{fid.ra.gcg.win} \\
EPO, random init & 1 string, 32 tokens & \Nn{fid.ra.epo.cos} & \N{fid.ra.epo.cos}{\tiny$\pm$\Nse{fid.ra.epo.cos}} & \N{fid.ra.epo.dcorp} \Nci{fid.ra.epo.dcorp} & \Npct{fid.ra.epo.win} \\
\method{} best-of-1, same 64 & 64 draws, mean & \Nn{fid.ra64.ex.cos.bo1} & \N{fid.ra64.ex.cos.bo1}{\tiny$\pm$\Nse{fid.ra64.ex.cos.bo1}} & -- & -- \\
\method{} best-of-8, same 64 & 64 draws, best of 8 & \Nn{fid.ra64.ex.cos.bo8} & \N{fid.ra64.ex.cos.bo8}{\tiny$\pm$\Nse{fid.ra64.ex.cos.bo8}} & \N{fid.ra64.ex.dcorp.bo8} \Nci{fid.ra64.ex.dcorp.bo8} & \Npct{fid.ra64.ex.win.bo8} \\
\bottomrule
\end{tabular}
\end{table}

On average a single sampled exemplar scores above the best window of the 10M-token corpus: \method{} best-of-1
minus corpus search is \N{fid.ra.diff.cos.bo1} (95\% CI \Nci{fid.ra.diff.cos.bo1}), and \method{} is ahead on
\Npct{fid.ra.ex.win.bo1} of the targets, so the margin in the main text does not depend on selecting among draws.
On the 64 discrete-search targets, \method{} best-of-8 minus GCG is \N{fid.ra.gcg.dex} (95\% CI
\Nci{fid.ra.gcg.dex}) and minus EPO \N{fid.ra.epo.dex} (\Nci{fid.ra.epo.dex}).

\subsection{Baselines in detail}
\label{app:inv-baselines}

\paragraph{NLA and token matching.}
We run the NLA checkpoint\footnote{Available at \url{https://huggingface.co/ceselder/qwen3.6-27b-nla-av}} in our pipeline with
four outputs per target at temperature 1, at most 200 new tokens and no minimum length (its own default), and score
each output with $s_\mu$. We read its best-of-2 beside \method{}'s best-of-8 because their generated-token totals
match to first order: NLA outputs average \N{fid.ra.nla.ntok} tokens and \method{} exemplars \N{fid.ra.ex.ntok}, so
eight exemplars correspond to about two outputs. The match is only first-order, since \method{} has eight texts in
which to place its best token and NLA two. Paired on identical targets, \method{} best-of-8 minus NLA best-of-2 is
\N{fid.ra.nla.dex.bo2} (95\% CI \Nci{fid.ra.nla.dex.bo2}), and minus the mean of NLA's four outputs
\N{fid.ra.nla.dex} (\Nci{fid.ra.nla.dex}); with all four outputs NLA reaches \N{fid.ra.nla.cos.bo4}.

\paragraph{Discrete search.}
GCG \citep{zou2023universal} and EPO \citep{thompson2024fluent} optimise a 32-token string $x$ against
$s(v,x)-\beta\,\mathrm{NLL}(x)$, with NLL the mean per-token negative log-likelihood under the subject. GCG uses
$\beta=0$ and one string, with 512 candidates per step from the top-512 gradient substitutions, for 150 steps. EPO
runs three strings at $\beta\in\{0.1,0.19,0.37\}$ with 85 candidates each for 300 steps, nearly the same number of
candidate forwards (\N{cost.ra.epo.fwd} against \N{cost.ra.gcg.fwd} per target), and reports the best of its three
final strings by uncentred $s$. Both take gradients through a soft-max surrogate of the token maximum, select
candidates by the exact objective subject to a decode-and-re-encode round trip, and start from 32 random tokens
(one seed per target) on the first 64 surviving \textsc{realact} rows. Both optimise the uncentred $s$;
Table~\ref{tab:inv-headline} rescores their final strings centred, and as no search was run on the centred
objective, these rows do not bound one.

On the uncentred score GCG leads EPO by \N{fid.ra.gcg.depo.unc} (95\% CI \Nci{fid.ra.gcg.depo.unc}). Centring moves
the comparison towards EPO: the paired centred difference, \N{fid.ra.gcg.depo} (\Nci{fid.ra.gcg.depo}), covers zero.
EPO's strings are far more likely under the subject model, with a mean per-token NLL of \N{fid.ra.epo.nll} against
\N{fid.ra.gcg.nll} for GCG, the level of a random token string. EPO costs \N{cost.ra.epo.gcg.ratio}$\times$ GCG
(Appendix~\ref{app:inv-cost}), mostly because of its exact-NLL pass.

\paragraph{Untrained base and no-injection floor.}
Two arms use the pretrained Qwen3.6-27B without \method{} training. The \emph{untrained base} runs \method{}'s own
prompt with the centred target injected as in Equation~\ref{eq:injection}, and with \method{}'s centring and
sampling; every run verifies the injection by reading back the placeholder residual after the addition. The
\emph{no-injection floor} has the same model continue the Patchscopes prompt below with nothing patched and reports
best-of-8 from 64 continuations: the score of text written with no information about $v$. No arm runs \method{}'s
prompt without an injection.

\paragraph{Patchscopes-style patching.}
Patchscopes \citep{ghandeharioun2024patchscopes} patches a hidden state into a prompt but defines no reconstruction
score, so we adapt it. We use the entity-description prompt from that paper's Appendix~D.1 as plain text; at the
trailing placeholder token, the residual $h$ at the output of block $\ell^*$ is replaced by $2\,\lVert h\rVert$ times
the unit centred target, and 64 continuations are sampled, scored like exemplars and reported at best-of-8.
Patching raises the score above the single no-injection floor at every block tried, by \N{fid.ra.ps.dfloor.l8},
\N{fid.ra.ps.dfloor.l14}, \N{fid.ra.ps.dfloor.l21} and \N{fid.ra.ps.dfloor.l42} at blocks 8, 14, 21 and 42 (paired
on the target; every 95\% document-clustered percentile CI lies above zero, \Nci{fid.ra.ps.dfloor.l14} at block 14).
Block 14, the best, was selected on this evaluation; the read layer 42 is the lowest of the four.
Figure~\ref{fig:evals} shows the floor but not the patched arm.

\subsection{Corpus size and number of draws}
\label{app:inv-curve}

Corpus search gains \N{fid.ra.corp10.slope} (SE \Nse{fid.ra.corp10.slope}) per doubling of the corpus, roughly
constant across the four sizes of Table~\ref{tab:inv-headline}. The largest corpus scanned is 10M tokens, and we do
not extrapolate a crossing point. \method{}'s gains from more draws shrink: the three doublings from best-of-8 to
best-of-64 add less than half the gain of the three from best-of-1 to best-of-8 (Figure~\ref{fig:evals} left,
Table~\ref{tab:inv-headline}).

\subsection{Cost}
\label{app:inv-cost}

\method{}'s 64 draws cost \N{cost.ra.ex.gpus} H200-seconds per target, averaged over one batched call and covering
generation only. GCG and EPO, run one target at a time, take \N{cost.ra.gcg.gpus} and \N{cost.ra.epo.gpus}
H200-seconds per target, \N{cost.ra.gcg.ex.ratio}$\times$ and \N{cost.ra.epo.ex.ratio}$\times$ the cost of
\method{}'s draws; the ratios are not end to end, as adding \method{}'s separate scoring pass would lower them. The
10M-token scan costs \N{fid.ra.corp10.gpus} H200-seconds per target, amortised over the 512 \textsc{realact} targets.

\subsection{Examples}
\label{app:inv-examples}

Four held-out \textsc{realact} targets, drawn with a fixed seed from the 64 targets that also have EPO strings,
among those whose \method{} best-of-8 score lies between the 25th and 75th percentile of all
$n=\Nn{fid.ra.ex.cos.bo1}$ targets; no drawn target was excluded as unsafe to print. Per target: the source window
ending at the target token; the four highest-scoring of \method{}'s 64 exemplars; NLA's four outputs; EPO's three
final strings, one per fluency weight $\beta$ ($^\dagger$: the one EPO reports, by uncentred score). The number
after an arm's name identifies the draw or output, and the number on the right is the text's centred score
$s_\mu$, a per-item run value rounded to two decimals. The highlighted token is where that score peaks; texts are cut around it on token
boundaries ([\dots]), and \texttt{\textbackslash n} marks a line break. Corpus-search windows are not shown; the
header gives the best window's score at 10M tokens.

\medskip
\providecolor{exmaem}{HTML}{2f5c92}
\providecolor{excorp}{HTML}{EE7733}
\providecolor{exnla}{HTML}{009988}
\providecolor{exepo}{HTML}{827b74}
\providecolor{exhl}{HTML}{dce9f5}
\providecommand{\exArm}[2]{\textcolor{#1}{#2}}
\providecommand{\exScore}[1]{\textcolor{gray}{#1}}
\providecommand{\exTok}[1]{\begingroup\setbox0=\hbox{#1}\rlap{\textcolor{exhl}{\rule[-0.28em]{\wd0}{1.08em}}}\box0\endgroup}
\providecommand{\exText}[1]{{\small\ttfamily\frenchspacing\raggedright #1\par}}
\providecommand{\exBegin}{\par\begingroup\setlength{\parskip}{0pt}}
\providecommand{\exEnd}{\par\endgroup\medskip}
\providecommand{\exTitle}[2]{\par\addvspace{10pt}{\raggedright\noindent\textbf{#1}\enspace\exScore{#2}\par}\nopagebreak}
\providecommand{\exHead}[3]{\par\addvspace{5pt}\noindent\exArm{#1}{#2}\hfill\exScore{#3}\par\nopagebreak}
\providecommand{\exItem}[4]{\par\addvspace{3pt}\noindent\parbox[t]{5.5em}{\raggedright\exArm{#1}{#2}}\hspace{0.6em}\parbox[t]{\dimexpr\linewidth-8.7em\relax}{\exText{#4}}\hfill\exScore{#3}\par}
\providecommand{\exLine}[2]{\par\addvspace{3pt}\noindent\makebox[2.4em][r]{\exScore{#1}}\hspace{0.6em}\parbox[t]{\dimexpr\linewidth-3em\relax}{\exText{#2}}\par}
\providecommand{\exRow}[2]{\par\nopagebreak\addvspace{5pt}\noindent\begin{minipage}[t]{0.48\linewidth}#1\end{minipage}\hfill\begin{minipage}[t]{0.48\linewidth}#2\end{minipage}\par}
\providecommand{\exDesc}[1]{\par\nopagebreak\noindent{\small\leftskip=1em #1\par}}
\exBegin
\exTitle{Target 1}{row 10; \method{} best-of-8 $0.88$; corpus search at 10M tokens, best window $0.58$}
\exItem{black}{source}{}{[\dots] Institute's studies of the interplay between improved module efficiency and balance of system costs to highlight the productive future R\exTok{\&D}}
\exItem{exmaem}{\method{} 7}{0.89}{[\dots] technologies presents quantitative treatment cost and performance trajectories that illuminate the promising future R\exTok{\&D} investments.}
\exItem{exmaem}{\method{} 36}{0.89}{[\dots] System Performance, derives quantitative estimates of PV electricity cost learning curves and illustrates the promising future R\exTok{\&D} efforts}
\exItem{exmaem}{\method{} 16}{0.88}{[\dots] offshore wind power conversion systems illustrates the cost-benefit trajectory of that technology and illustrates the promising future R\exTok{\&D}}
\exItem{exmaem}{\method{} 35}{0.88}{[\dots] Technology and Systems, develops a detailed technology cost performance trajectory and illustrates the promising future R\exTok{\&D} investments.}
\exItem{exnla}{NLA 2}{0.79}{[\dots] the future promising R\exTok{\&D}" requiring a completion[\dots]}
\exItem{exnla}{NLA 4}{0.75}{[\dots] the future promising R\exTok{\&D}"\textcolor{gray}{\textbackslash n}--- requiring immediate[\dots]}
\exItem{exnla}{NLA 3}{0.72}{[\dots] the future productive R\exTok{\&D}"), expecting continuation[\dots]}
\exItem{exnla}{NLA 1}{0.71}{[\dots] the fruitful future R\exTok{\&D}"), immediately requiring[\dots]}
\exItem{exepo}{EPO $\beta\,0.19^\dagger$}{0.45}{ sector. Ottawas future \exTok{research} agenda. Second, it[\dots]}
\exItem{exepo}{EPO $\beta\,0.1$}{0.19}{[\dots]xcellent Ute bVW ID I\textbackslash{}oguen. In i arly \exTok{DLC} v. ision,}
\exItem{exepo}{EPO $\beta\,0.37$}{0.02}{ enfoc\exTok{amiento} del present and the present tense[\dots]}
\exTitle{Target 2}{row 21; \method{} best-of-8 $0.88$; corpus search at 10M tokens, best window $0.66$}
\exItem{black}{source}{}{[\dots], and perceive the creator's intentions for every factor of their writing.\textcolor{gray}{\textbackslash n}Context is extraordinarily necessary in any such project. \exTok{In}}
\exItem{exmaem}{\method{} 53}{0.92}{[\dots] examples and quotes to again up your ideas. Writing a literary evaluation essay. Context is fundamental when writing any such paper. \exTok{In}}
\exItem{exmaem}{\method{} 20}{0.91}{[\dots] construct your essay writing expertise. Writing a literary evaluation essay. Context is tremendous-essential in any literary paper. \exTok{In}}
\exItem{exmaem}{\method{} 54}{0.90}{[\dots] talk about the that means of the work itself. Writing a literary analysis article. Context is vitally important on this type of essay. \exTok{In}}
\exItem{exmaem}{\method{} 21}{0.90}{[\dots] your essay about the piece of literature. Writing a very good literary evaluation essay. Context is crucial in any of your analyses. \exTok{In}}
\exItem{exnla}{NLA 2}{0.34}{[\dots] to identify the context. \exTok{In} [order to...]" or[\dots]}
\exItem{exnla}{NLA 3}{0.34}{[\dots]\textcolor{gray}{\textbackslash n}\textcolor{gray}{\textbackslash n}"Is extremely essential \exTok{in} your literary analysis.[\dots]}
\exItem{exnla}{NLA 1}{0.33}{[\dots] in any analysis. \exTok{In}" requires completion of the pre[\dots]}
\exItem{exnla}{NLA 4}{0.27}{[\dots] context's importance \exTok{in} literary/historical essays.[\dots]}
\exItem{exepo}{EPO $\beta\,0.1^\dagger$}{0.61}{[\dots] aIn literature. \exTok{In}herent in the word literature is[\dots]}
\exItem{exepo}{EPO $\beta\,0.19$}{0.33}{ Antonia, de la a el de a. \exTok{In} A, de a a, de a el de[\dots]}
\exItem{exepo}{EPO $\beta\,0.37$}{0.27}{ ECON ONE The CAT \exTok{In}herited The QUILT Rats the cats[\dots]}
\exTitle{Target 3}{row 5; \method{} best-of-8 $0.80$; corpus search at 10M tokens, best window $0.65$}
\exItem{black}{source}{}{[\dots]-2009 financial crisis, which would be a litmus test for understanding how a fund manager endured a systemic market correction. \exTok{Additionally}}
\exItem{exmaem}{\method{} 3}{0.81}{[\dots] window currently used does not include the financial crisis of 2008, which is arguably one of the most important drawdowns. \exTok{Additionally},}
\exItem{exmaem}{\method{} 57}{0.81}{[\dots] is problematic as it does not include the 2020 pandemic-induced recession that was arguably the largest in modern history. \exTok{Additionally},}
\exItem{exmaem}{\method{} 56}{0.81}{[\dots] historical data. For instance, the rolling 10-year window still includes 2009, which was an anomalous year for equities. \exTok{Additionally}}
\exItem{exmaem}{\method{} 22}{0.80}{[\dots] the model entirely excludes the massive volatility and subsequent yields move experienced during the height of COVID in 2020. \exTok{Additionally},}
\exItem{exnla}{NLA 4}{0.40}{[\dots] 2007-08 crisis).\textcolor{gray}{\textbackslash n}\textcolor{gray}{\textbackslash n}"\exTok{Additionally}" is a continuation[\dots]}
\exItem{exnla}{NLA 2}{0.36}{[\dots] or consequence, such \exTok{as} recent market conditions or[\dots]}
\exItem{exnla}{NLA 3}{0.35}{[\dots] or consequence (e.g\exTok{.,} recent data periods or other[\dots]}
\exItem{exnla}{NLA 1}{0.29}{[\dots].\textcolor{gray}{\textbackslash n}\textcolor{gray}{\textbackslash n}Final token "\exTok{Additionally}" opens a continuation[\dots]}
\exItem{exepo}{EPO $\beta\,0.1^\dagger$}{0.54}{[\dots] are mentioned. \exTok{Additionally}, structures statiques.[\dots]}
\exItem{exepo}{EPO $\beta\,0.37$}{0.30}{[\dots] is for de VSV-RSV. De \exTok{plus}, que me dijes KINDEY?}
\exItem{exepo}{EPO $\beta\,0.19$}{0.29}{ labai terrible!" \exTok{Asimismo} fue la re\^{}pcion por el[\dots]}
\exTitle{Target 4}{row 33; \method{} best-of-8 $0.87$; corpus search at 10M tokens, best window $0.47$}
\exItem{black}{source}{}{[\dots]'s Hospital Research Institute of Manitoba, told INSIDER. "We saw that it was highly variable --- milk from different mothers \exTok{contained}}
\exItem{exmaem}{\method{} 32}{0.88}{[\dots] 100 mothers. "We found there was lots of variation in these microbial communities -- different mothers' guts \exTok{contained} different bacteria}
\exItem{exmaem}{\method{} 24}{0.87}{[\dots] huge amounts of diversity in mom's gut microbiomes --- "there was lots of variation: guts from different mothers \exTok{contained} different bacteria}
\exItem{exmaem}{\method{} 27}{0.87}{[\dots] high diversity and lots of variation: "The gut microbiomes were very heterogeneous -- different mothers' guts \exTok{contained} different bacteria}
\exItem{exmaem}{\method{} 64}{0.87}{[\dots] researcher Katherine Scott noted, "The gut microbiomes were very heterogeneous --- guts from different women \exTok{contained} different bacteria}
\exItem{exnla}{NLA 4}{0.58}{[\dots]. Different women's milk \exTok{contained}" expects a noun[\dots]}
\exItem{exnla}{NLA 2}{0.54}{[\dots]omes held, e.g\exTok{.} "different bacterial communities" or[\dots]}
\exItem{exnla}{NLA 1}{0.53}{[\dots]\textcolor{gray}{\textbackslash n}"contained very \exTok{different} bacteria" or similar noun[\dots]}
\exItem{exnla}{NLA 3}{0.44}{[\dots]different microbes" or \exTok{"}different compositions/comm[\dots]}
\exItem{exepo}{EPO $\beta\,0.1^\dagger$}{0.32}{[\dots] a transition zone \exTok{containing} both Tropical species[\dots]}
\exItem{exepo}{EPO $\beta\,0.19$}{0.13}{ dataset \exTok{contains} merkez haber prochainement lok[\dots]}
\exItem{exepo}{EPO $\beta\,0.37$}{0.06}{[\dots] yang saya punya, saya tidak \exTok{memiliki} oIrs, irs atau[\dots]}
\exEnd

\section{Generalisation beyond English web text}
\label{app:ood}

This appendix gives the per-domain numbers behind Figure~\ref{fig:evals} (right) for 21 held-out domains and
two corpus-search baselines; Table~\ref{tab:ood-arms} reports every number.

\subsection{Domains, targets and corpora}
\label{app:ood-arms}

Nine non-English web corpora cover eight languages: Czech (\texttt{ces}), Russian (\texttt{rus}), Greek
(\texttt{ell}), Arabic (\texttt{arb}), Hindi (\texttt{hin}), Thai (\texttt{tha}), Mandarin (\texttt{cmn}) and
Japanese (\texttt{jpn}) from the FineWeb-2 test split \citep{penedo2025fineweb2}, and Chinese (\texttt{zh})
from Ultra-FineWeb \citep{wang2025ultrafineweb}. Chinese thus appears twice, from sources with different
filtering. The eight programming languages and two of the four mathematics sources, Lean and Isabelle, come
from The Stack \citep{kocetkov2022stack} via \texttt{the-stack-smol-xl}; the other two are OpenWebMath prose
(\texttt{owm}) \citep{paster2023openwebmath} and arXiv from the Proof-Pile-2 test shards
\citep{azerbayev2023llemma}. The anchor, English Ultra-FineWeb (\texttt{en}), is a fresh draw from outside
every training range through the same pipeline; it repeats the comparison of Sec.~\ref{sec:inv-results},
not its numbers, and is excluded from the 21 arms and every count.

For each arm, one fixed seed permutes the source's documents, which are consumed in that order: the corpus
takes documents until it holds 10M tokens (\texttt{shell}: \N{ood.shell.corp10.mtok}M, all the source
contains), and the 512 targets come from the next documents of at least 512 tokens, one per document.
Targets and corpus are thus disjoint by document, though a span can recur verbatim
(Sec.~\ref{app:ood-licences}). As in Appendix~\ref{app:inv}, the target is the layer-42 residual of $M$ at a
token uniform over positions 16--511.

``Held out'' refers to the \method{}'s training data (English Ultra-FineWeb activations and the directions of
an English SAE), not to Qwen3.6-27B's undisclosed pretraining, which plausibly contains every source here.
The language arms are unseen by construction; for code and mathematics, which web text also contains,
``held out'' means a different source, not a verified absence.

\subsection{Baselines and statistics}
\label{app:ood-stats}

Every cosine is the centred score of Sec.~\ref{sec:inv-results}, with the one English training-time mean of
Appendix~\ref{app:inv} subtracted on both sides for every arm. The \method{} value is best-of-8 by the
unbiased estimator from 64 draws of at most 64 tokens per target. The primary baseline is the top-1 corpus
search of Appendix~\ref{app:inv} over the arm's own corpus; the second is the same search over the 10M-token
English training-corpus sample of Sec.~\ref{sec:inv-results}. On the anchor the two search disjoint English
corpora and agree (\N{ood.en.encorp10.cos} for the training corpus, \N{ood.en.corp10.cos} for the anchor's
own). The control is the untrained subject under the same injection at best-of-8.

For each arm, $\Delta_i$ is the \method{}'s best-of-8 minus the baseline's top-1 on target $i$, and
$\bar\Delta$ its mean, with a 10{,}000-resample percentile bootstrap CI over targets (with one target per
document, also a document-level bootstrap). An arm \emph{exceeds} if the CI lies above zero, is
\emph{reversed} if below, and is \emph{inconclusive} if it covers zero (the sign is not established). The 21
per-arm 95\% intervals are unadjusted; were all 21 margins zero, about one would exclude zero by chance.

Against the English training corpus the \method{} exceeds on all \N{ood.conj.endiff.nexceed} arms, the
smallest margin being \texttt{sql}'s \N{ood.conj.endiff.min} (95\% CI \Nci{ood.conj.endiff.min}). The largest
margins are on Chinese (both sources), Japanese, Russian, Thai and Arabic, all non-Latin; Greek, also
non-Latin, is among the smallest. The English corpus's top-1 on the non-Latin arms is in the same range as on
the code arms, so these margins come with higher \method{} scores, not a weaker baseline; we offer no
mechanism. Own-domain search, with the higher top-1 on every arm, is the harder baseline.

\subsection{Results per domain}
\label{app:ood-results}

\begin{table}[t]
\centering\scriptsize
\caption{Per-arm numbers behind Figure~\ref{fig:evals} (right), arms named as there; \texttt{en} (the anchor,
outside the 21) and \texttt{zh} are Ultra-FineWeb (UFW), the other languages FineWeb-2 (FW2).
$n=\Nn{ood.en.diff.cos.bo8}$ targets per arm (Sec.~\ref{app:ood-stats}). \method{}: best-of-8. own: top-1
search over the arm's own 10M-token corpus ($\dagger$: \N{ood.shell.corp10.mtok}M, the whole source). base:
untrained subject, same injection. $\bar\Delta$: \method{} minus own. train, $\bar\Delta_{\mathrm{train}}$: the
same over the English training corpus. lid: lid218e rate on the top exemplar\,/\,own top corpus windows
(dash: not meaningful on code, \texttt{arxiv}, \texttt{lean}, \texttt{isabelle}). bpb: mean bits per byte of
the targets' preceding context under the untrained subject. vspan: Sec.~\ref{app:ood-licences}.}
\label{tab:ood-arms}
\setlength{\tabcolsep}{2pt}
\begin{tabular}{@{}lrrrlcrrcrr@{}}
\toprule
arm & \method{} & own & base & $\bar\Delta$ [95\% CI] & outcome & train & $\bar\Delta_{\mathrm{train}}$ & lid (ex.\,/\,corpus) & bpb & vspan \\
\midrule
\texttt{en} & \N{ood.en.ex.cos.bo8} & \N{ood.en.corp10.cos} & \N{ood.en.base.cos.bo8} & \N{ood.en.diff.cos.bo8} \Nci{ood.en.diff.cos.bo8} & \N{ood.en.diff.verdict} & \N{ood.en.encorp10.cos} & \N{ood.en.endiff.cos.bo8} & \N{ood.en.ex.lid}/\N{ood.en.corp10.lid} & \N{ood.en.base.bpb} & \N{ood.en.ex.vspan} \\
\midrule
\texttt{zh} (UFW) & \N{ood.zh.ex.cos.bo8} & \N{ood.zh.corp10.cos} & \N{ood.zh.base.cos.bo8} & \N{ood.zh.diff.cos.bo8} \Nci{ood.zh.diff.cos.bo8} & \N{ood.zh.diff.verdict} & \N{ood.zh.encorp10.cos} & \N{ood.zh.endiff.cos.bo8} & \N{ood.zh.ex.lid}/\N{ood.zh.corp10.lid} & \N{ood.zh.base.bpb} & \N{ood.zh.ex.vspan} \\
\texttt{ces} & \N{ood.ces.ex.cos.bo8} & \N{ood.ces.corp10.cos} & \N{ood.ces.base.cos.bo8} & \N{ood.ces.diff.cos.bo8} \Nci{ood.ces.diff.cos.bo8} & \N{ood.ces.diff.verdict} & \N{ood.ces.encorp10.cos} & \N{ood.ces.endiff.cos.bo8} & \N{ood.ces.ex.lid}/\N{ood.ces.corp10.lid} & \N{ood.ces.base.bpb} & \N{ood.ces.ex.vspan} \\
\texttt{rus} & \N{ood.rus.ex.cos.bo8} & \N{ood.rus.corp10.cos} & \N{ood.rus.base.cos.bo8} & \N{ood.rus.diff.cos.bo8} \Nci{ood.rus.diff.cos.bo8} & \N{ood.rus.diff.verdict} & \N{ood.rus.encorp10.cos} & \N{ood.rus.endiff.cos.bo8} & \N{ood.rus.ex.lid}/\N{ood.rus.corp10.lid} & \N{ood.rus.base.bpb} & \N{ood.rus.ex.vspan} \\
\texttt{ell} & \N{ood.ell.ex.cos.bo8} & \N{ood.ell.corp10.cos} & \N{ood.ell.base.cos.bo8} & \N{ood.ell.diff.cos.bo8} \Nci{ood.ell.diff.cos.bo8} & \N{ood.ell.diff.verdict} & \N{ood.ell.encorp10.cos} & \N{ood.ell.endiff.cos.bo8} & \N{ood.ell.ex.lid}/\N{ood.ell.corp10.lid} & \N{ood.ell.base.bpb} & \N{ood.ell.ex.vspan} \\
\texttt{arb} & \N{ood.arb.ex.cos.bo8} & \N{ood.arb.corp10.cos} & \N{ood.arb.base.cos.bo8} & \N{ood.arb.diff.cos.bo8} \Nci{ood.arb.diff.cos.bo8} & \N{ood.arb.diff.verdict} & \N{ood.arb.encorp10.cos} & \N{ood.arb.endiff.cos.bo8} & \N{ood.arb.ex.lid}/\N{ood.arb.corp10.lid} & \N{ood.arb.base.bpb} & \N{ood.arb.ex.vspan} \\
\texttt{hin} & \N{ood.hin.ex.cos.bo8} & \N{ood.hin.corp10.cos} & \N{ood.hin.base.cos.bo8} & \N{ood.hin.diff.cos.bo8} \Nci{ood.hin.diff.cos.bo8} & \N{ood.hin.diff.verdict} & \N{ood.hin.encorp10.cos} & \N{ood.hin.endiff.cos.bo8} & \N{ood.hin.ex.lid}/\N{ood.hin.corp10.lid} & \N{ood.hin.base.bpb} & \N{ood.hin.ex.vspan} \\
\texttt{tha} & \N{ood.tha.ex.cos.bo8} & \N{ood.tha.corp10.cos} & \N{ood.tha.base.cos.bo8} & \N{ood.tha.diff.cos.bo8} \Nci{ood.tha.diff.cos.bo8} & \N{ood.tha.diff.verdict} & \N{ood.tha.encorp10.cos} & \N{ood.tha.endiff.cos.bo8} & \N{ood.tha.ex.lid}/\N{ood.tha.corp10.lid} & \N{ood.tha.base.bpb} & \N{ood.tha.ex.vspan} \\
\texttt{cmn} (FW2) & \N{ood.cmn.ex.cos.bo8} & \N{ood.cmn.corp10.cos} & \N{ood.cmn.base.cos.bo8} & \N{ood.cmn.diff.cos.bo8} \Nci{ood.cmn.diff.cos.bo8} & \N{ood.cmn.diff.verdict} & \N{ood.cmn.encorp10.cos} & \N{ood.cmn.endiff.cos.bo8} & \N{ood.cmn.ex.lid}/\N{ood.cmn.corp10.lid} & \N{ood.cmn.base.bpb} & \N{ood.cmn.ex.vspan} \\
\texttt{jpn} & \N{ood.jpn.ex.cos.bo8} & \N{ood.jpn.corp10.cos} & \N{ood.jpn.base.cos.bo8} & \N{ood.jpn.diff.cos.bo8} \Nci{ood.jpn.diff.cos.bo8} & \N{ood.jpn.diff.verdict} & \N{ood.jpn.encorp10.cos} & \N{ood.jpn.endiff.cos.bo8} & \N{ood.jpn.ex.lid}/\N{ood.jpn.corp10.lid} & \N{ood.jpn.base.bpb} & \N{ood.jpn.ex.vspan} \\
\midrule
\texttt{python} & \N{ood.python.ex.cos.bo8} & \N{ood.python.corp10.cos} & \N{ood.python.base.cos.bo8} & \N{ood.python.diff.cos.bo8} \Nci{ood.python.diff.cos.bo8} & \N{ood.python.diff.verdict} & \N{ood.python.encorp10.cos} & \N{ood.python.endiff.cos.bo8} & -- & \N{ood.python.base.bpb} & \N{ood.python.ex.vspan} \\
\texttt{javascript} & \N{ood.javascript.ex.cos.bo8} & \N{ood.javascript.corp10.cos} & \N{ood.javascript.base.cos.bo8} & \N{ood.javascript.diff.cos.bo8} \Nci{ood.javascript.diff.cos.bo8} & \N{ood.javascript.diff.verdict} & \N{ood.javascript.encorp10.cos} & \N{ood.javascript.endiff.cos.bo8} & -- & \N{ood.javascript.base.bpb} & \N{ood.javascript.ex.vspan} \\
\texttt{c} & \N{ood.c.ex.cos.bo8} & \N{ood.c.corp10.cos} & \N{ood.c.base.cos.bo8} & \N{ood.c.diff.cos.bo8} \Nci{ood.c.diff.cos.bo8} & \N{ood.c.diff.verdict} & \N{ood.c.encorp10.cos} & \N{ood.c.endiff.cos.bo8} & -- & \N{ood.c.base.bpb} & \N{ood.c.ex.vspan} \\
\texttt{rust} & \N{ood.rust.ex.cos.bo8} & \N{ood.rust.corp10.cos} & \N{ood.rust.base.cos.bo8} & \N{ood.rust.diff.cos.bo8} \Nci{ood.rust.diff.cos.bo8} & \N{ood.rust.diff.verdict} & \N{ood.rust.encorp10.cos} & \N{ood.rust.endiff.cos.bo8} & -- & \N{ood.rust.base.bpb} & \N{ood.rust.ex.vspan} \\
\texttt{go} & \N{ood.go.ex.cos.bo8} & \N{ood.go.corp10.cos} & \N{ood.go.base.cos.bo8} & \N{ood.go.diff.cos.bo8} \Nci{ood.go.diff.cos.bo8} & \N{ood.go.diff.verdict} & \N{ood.go.encorp10.cos} & \N{ood.go.endiff.cos.bo8} & -- & \N{ood.go.base.bpb} & \N{ood.go.ex.vspan} \\
\texttt{haskell} & \N{ood.haskell.ex.cos.bo8} & \N{ood.haskell.corp10.cos} & \N{ood.haskell.base.cos.bo8} & \N{ood.haskell.diff.cos.bo8} \Nci{ood.haskell.diff.cos.bo8} & \N{ood.haskell.diff.verdict} & \N{ood.haskell.encorp10.cos} & \N{ood.haskell.endiff.cos.bo8} & -- & \N{ood.haskell.base.bpb} & \N{ood.haskell.ex.vspan} \\
\texttt{sql} & \N{ood.sql.ex.cos.bo8} & \N{ood.sql.corp10.cos} & \N{ood.sql.base.cos.bo8} & \N{ood.sql.diff.cos.bo8} \Nci{ood.sql.diff.cos.bo8} & \N{ood.sql.diff.verdict} & \N{ood.sql.encorp10.cos} & \N{ood.sql.endiff.cos.bo8} & -- & \N{ood.sql.base.bpb} & \N{ood.sql.ex.vspan} \\
\texttt{shell}$^\dagger$ & \N{ood.shell.ex.cos.bo8} & \N{ood.shell.corp10.cos} & \N{ood.shell.base.cos.bo8} & \N{ood.shell.diff.cos.bo8} \Nci{ood.shell.diff.cos.bo8} & \N{ood.shell.diff.verdict} & \N{ood.shell.encorp10.cos} & \N{ood.shell.endiff.cos.bo8} & -- & \N{ood.shell.base.bpb} & \N{ood.shell.ex.vspan} \\
\midrule
\texttt{owm} & \N{ood.owm.ex.cos.bo8} & \N{ood.owm.corp10.cos} & \N{ood.owm.base.cos.bo8} & \N{ood.owm.diff.cos.bo8} \Nci{ood.owm.diff.cos.bo8} & \N{ood.owm.diff.verdict} & \N{ood.owm.encorp10.cos} & \N{ood.owm.endiff.cos.bo8} & \N{ood.owm.ex.lid}/\N{ood.owm.corp10.lid} & \N{ood.owm.base.bpb} & \N{ood.owm.ex.vspan} \\
\texttt{arxiv} & \N{ood.arxiv.ex.cos.bo8} & \N{ood.arxiv.corp10.cos} & \N{ood.arxiv.base.cos.bo8} & \N{ood.arxiv.diff.cos.bo8} \Nci{ood.arxiv.diff.cos.bo8} & \N{ood.arxiv.diff.verdict} & \N{ood.arxiv.encorp10.cos} & \N{ood.arxiv.endiff.cos.bo8} & -- & \N{ood.arxiv.base.bpb} & \N{ood.arxiv.ex.vspan} \\
\texttt{lean} & \N{ood.lean.ex.cos.bo8} & \N{ood.lean.corp10.cos} & \N{ood.lean.base.cos.bo8} & \N{ood.lean.diff.cos.bo8} \Nci{ood.lean.diff.cos.bo8} & \N{ood.lean.diff.verdict} & \N{ood.lean.encorp10.cos} & \N{ood.lean.endiff.cos.bo8} & -- & \N{ood.lean.base.bpb} & \N{ood.lean.ex.vspan} \\
\texttt{isabelle} & \N{ood.isabelle.ex.cos.bo8} & \N{ood.isabelle.corp10.cos} & \N{ood.isabelle.base.cos.bo8} & \N{ood.isabelle.diff.cos.bo8} \Nci{ood.isabelle.diff.cos.bo8} & \N{ood.isabelle.diff.verdict} & \N{ood.isabelle.encorp10.cos} & \N{ood.isabelle.endiff.cos.bo8} & -- & \N{ood.isabelle.base.bpb} & \N{ood.isabelle.ex.vspan} \\
\bottomrule
\end{tabular}
\end{table}

Against the domain's own corpus, \N{ood.conj.diff.nexceed} of the 21 arms exceed,
\N{ood.conj.diff.ninconcl} are inconclusive (Greek and \texttt{shell}), and
\N{ood.conj.diff.nreversed} are reversed (Hindi, Thai, Go, Haskell, SQL, Lean and Isabelle).
Every exceeding margin is below the anchor's \N{ood.en.diff.cos.bo8}: the margin over own-domain search is
largest on the English anchor.

Some outcomes are narrow: Greek and \texttt{shell} are inconclusive with lower bounds just below zero, Rust
exceeds with one just above (95\% CI \Nci{ood.rust.diff.cos.bo8}), and Go, Haskell and SQL reverse with upper
bounds just below. The reversals on Hindi, Thai, Lean and Isabelle are not narrow.

As a high cosine is possible in the wrong language, the lid218e classifier \citep{nllb2022} labels
the top exemplar of every target on the language arms, the anchor and \texttt{owm}; its rate on the arm's own
top corpus windows is a reference, not a ceiling (point rates, no intervals). The exemplar rate is at or
above the reference on \texttt{zh}, Hindi, Thai and Japanese, close on Russian, below on Arabic and
\texttt{cmn}, and furthest below on Czech (\N{ood.ces.ex.lid} against \N{ood.ces.corp10.lid}) and Greek
(\N{ood.ell.ex.lid} against \N{ood.ell.corp10.lid}), where either the classifier fails on short windows or
text in another language reaches the centred score; we cannot tell which. For code and mathematics we
measured only fidelity, not whether the exemplars are valid code or mathematics.

The subject's own predictability of a domain (bpb) does not order the outcomes: FineWeb-2 Mandarin at
\N{ood.cmn.base.bpb} exceeds, Hindi and Thai at \N{ood.hin.base.bpb} and \N{ood.tha.base.bpb} reverse, and
the anchor sits mid-table at \N{ood.en.base.bpb}; the missing control is a \method{} trained on the domain.

\subsection{Licences and verbatim overlap}
\label{app:ood-licences}

FineWeb-2 (its authors' held-out split) and OpenWebMath are used under ODC-BY, Ultra-FineWeb under
Apache-2.0, and The Stack (\texttt{bigcode/the-stack-smol-xl}) under its per-file permissive licences and
opt-out regime. The Proof-Pile-2 arXiv shards declare no licence; their content is under arXiv's own
licences. Apart from the short excerpts of Appendix~\ref{app:ood-examples} (FineWeb-2, OpenWebMath and The
Stack, under the licences above), we use the documents only as model inputs.

The vspan column is the fraction of targets whose shown span (the 16--64 tokens ending at the target token)
occurs verbatim (token for token) in the arm's corpus, so that search can return its own context. It is
highest on Isabelle (\N{ood.isabelle.ex.vspan}) and Lean (\N{ood.lean.ex.vspan}), which have the largest
reversals, so near-copy retrieval may explain part of them; this comparison does not isolate the effect.
Hindi and Thai reverse at \N{ood.hin.ex.vspan} and \N{ood.tha.ex.vspan}, so overlap is unlikely to explain
the language reversals; we did not recompute margins without the overlapping targets.

\subsection{Examples}
\label{app:ood-examples}

We show two targets from each of four arms: Czech, FineWeb-2 Mandarin, Python and OpenWebMath. Per target,
the first text is the shown span ending at the target token (highlighted); below it are the top-1
window of the own-domain corpus search and the best of the \method{}'s first eight exemplars, each
highlighted at the token where its centred score peaks. Targets are drawn with a fixed seed from
those whose best-of-8 estimate lies between the arm's 25th and 75th percentiles; of the eleven
drawn, two were skipped for personal details and one because its corpus window is in traditional
characters, which the paper's Chinese font lacks. Excerpts are cut on token boundaries to about 120
characters ([\dots]); \textcolor{gray}{$\hookleftarrow$} marks a line break. Scores are per-target
values from the run's files, not table cells; the
best-of-8 estimate is an expectation over draws and differs from the single best draw shown. The first
Czech exemplar is in Polish and the second mixes English and Czech; Appendix~\ref{app:ood-results}
reports the Czech exemplar language rate.

\medskip
\providecolor{exmaem}{HTML}{2f5c92}
\providecolor{excorp}{HTML}{EE7733}
\providecolor{exnla}{HTML}{009988}
\providecolor{exepo}{HTML}{827b74}
\providecolor{exhl}{HTML}{dce9f5}
\providecommand{\exArm}[2]{\textcolor{#1}{#2}}
\providecommand{\exScore}[1]{\textcolor{gray}{#1}}
\providecommand{\exTok}[1]{\begingroup\setbox0=\hbox{#1}\rlap{\textcolor{exhl}{\rule[-0.28em]{\wd0}{1.08em}}}\box0\endgroup}
\providecommand{\exText}[1]{{\small\ttfamily\frenchspacing\raggedright #1\par}}
\providecommand{\exBegin}{\par\begingroup\setlength{\parskip}{0pt}}
\providecommand{\exEnd}{\par\endgroup\medskip}
\providecommand{\exTitle}[2]{\par\addvspace{10pt}{\raggedright\noindent\textbf{#1}\enspace\exScore{#2}\par}\nopagebreak}
\providecommand{\exHead}[3]{\par\addvspace{5pt}\noindent\exArm{#1}{#2}\hfill\exScore{#3}\par\nopagebreak}
\providecommand{\exItem}[4]{\par\addvspace{3pt}\noindent\parbox[t]{5.5em}{\raggedright\exArm{#1}{#2}}\hspace{0.6em}\parbox[t]{\dimexpr\linewidth-8.7em\relax}{\exText{#4}}\hfill\exScore{#3}\par}
\providecommand{\exLine}[2]{\par\addvspace{3pt}\noindent\makebox[2.4em][r]{\exScore{#1}}\hspace{0.6em}\parbox[t]{\dimexpr\linewidth-3em\relax}{\exText{#2}}\par}
\providecommand{\exRow}[2]{\par\nopagebreak\addvspace{5pt}\noindent\begin{minipage}[t]{0.48\linewidth}#1\end{minipage}\hfill\begin{minipage}[t]{0.48\linewidth}#2\end{minipage}\par}
\providecommand{\exDesc}[1]{\par\nopagebreak\noindent{\small\leftskip=1em #1\par}}
\exBegin
\exTitle{Czech (\texttt{ces})}{}
\exRow{\exHead{black}{target, row 377}{best-of-8 est.\ $s_\mu = 0.80$}\exText{\fontencoding{T1}\selectfont [\dots]  po\v{c}\'{a}tku po\v{c}\'{\i{}}ta\v{c}ov\'{e}ho sv\v{e}ta bylo \v{C}\'{I}SLO.\allowbreak{} \v{C}\'{\i{}}sla vypl\v{n}ovala obrazovky termin\'{a}l\r{u},\allowbreak{} jejich dlouh\'{e} sloupce vypisovaly elekt\exTok{rom}}}
{\exHead{black}{target, row 123}{best-of-8 est.\ $s_\mu = 0.81$}\exText{\fontencoding{T1}\selectfont [\dots] abiny dovoluje budov\'{a}n\'{\i{}} jist\'{\i{}}c\'{\i{}}ch stanovi\v{s}\v{t} nebo nasazen\'{\i{}} s jist\'{\i{}}tkem.\allowbreak{}\textcolor{gray}{\ensuremath{\hookleftarrow}}K-\allowbreak{}Lock:\allowbreak{} Jednodu\v{s}e geni\'{a}ln\'{\i{}} syst\'{e}m z\'{a}padky,\allowbreak{} \exTok{kter\'{y}}}}
\exRow{\exHead{excorp}{corpus search top-1}{$s_\mu = 0.66$}\exText{\fontencoding{T1}\selectfont  200 nebo 300 \v{c}ip\r{u},\allowbreak{} elekt\exTok{rom}obil chce i 1000.\allowbreak{} rozj\'{\i{}}\v{r}d\'{\i{}} se IOT,\allowbreak{} takzvan\v{e} chytr\'{e} v\v{e}ci a podobn\v{e}.\allowbreak{} Tam miz\'{\i{}} megatuny k\v{r}em\'{\i{}}ku [\dots]}}
{\exHead{excorp}{corpus search top-1}{$s_\mu = 0.63$}\exText{\fontencoding{T1}\selectfont  snadn\'{e}ho p\v{r}evl\'{e}k\'{a}n\'{\i{}} l\r{u}\v{z}kovin.\allowbreak{} K v\'{y}rob\v{e} pou\v{z}\'{\i{}}v\'{a}me zipov\'{y} uz\'{a}v\v{e}r s peckov\'{y}m jezdcem,\allowbreak{} \exTok{kter\'{y}} nikterak nevy\v{c}n\'{\i{}}v\'{a} z povle\v{c}en\'{\i{}} [\dots]}}
\exRow{\exHead{exmaem}{\method{}, best of draws 1--8}{$s_\mu = 0.83$}\exText{\fontencoding{T1}\selectfont [\dots]  Na ekranie terminala komputerowego powsta\l{} w pocz\k{a}tku lat siedemdziesi\k{a}tych - linie tekstu zapisywa\l{}y a\'{u}n elekt\exTok{rom}}}
{\exHead{exmaem}{\method{}, best of draws 1--8}{$s_\mu = 0.81$}\exText{\fontencoding{T1}\selectfont [\dots]  tool for every biker.\allowbreak{}\textcolor{gray}{\ensuremath{\hookleftarrow}}-\allowbreak{} SCHUTZ REFLEX CATCH\texttrademark{} -\allowbreak{} Napoletano patent.\allowbreak{} Geni\'{a}ln\v{e} jednoduch\'{y} zp\v{r}\'{\i{}}s\v{n}ovac\'{\i{}} syst\'{e}m,\allowbreak{} \exTok{kter\'{y}}}}
\exTitle{Mandarin (\texttt{cmn})}{}
\exRow{\exHead{black}{target, row 3097}{best-of-8 est.\ $s_\mu = 0.84$}\exText{\fontencoding{T1}\selectfont \zh{那不行，打车还不如我开车送，反正也}'\zh{花}'\zh{不了多少时间，走，}\exTok{\zh{赶紧}}}}
{\exHead{black}{target, row 3246}{best-of-8 est.\ $s_\mu = 0.81$}\exText{\fontencoding{T1}\selectfont \zh{都督，行丞事，领功曹；将校吏兵三千余人，皆范先督之。固等喜，虽阳事畿，不以为意。固欲大发兵，畿患}\exTok{\zh{之}}}}
\exRow{\exHead{excorp}{corpus search top-1}{$s_\mu = 0.73$}\exText{\fontencoding{T1}\selectfont ......"\textcolor{gray}{\ensuremath{\hookleftarrow}}\zh{筱筱一个白眼丢过去},\allowbreak{}\zh{然后没好气道}:\allowbreak{}"\zh{你们俩没完没了了是吧},\allowbreak{}\exTok{\zh{赶紧}}\zh{回到正题啊！}"\textcolor{gray}{\ensuremath{\hookleftarrow}}\zh{被筱筱这么一说},\allowbreak{}\zh{俩人一愣},\allowbreak{}\zh{然后对视一笑后},\allowbreak{}\zh{萧逸} [\dots]}}
{\exHead{excorp}{corpus search top-1}{$s_\mu = 0.56$}\exText{\fontencoding{T1}\selectfont \zh{绩。景公患}\exTok{\zh{之}}\zh{。晏婴乃荐田穰苴曰：}"\zh{穰苴虽田氏庶孽，然其人文能附众，武能威敌，原君试之。}"\zh{景公召穰苴，与语兵事，大说之，以为} [\dots]}}
\exRow{\exHead{exmaem}{\method{}, best of draws 1--8}{$s_\mu = 0.86$}\exText{\fontencoding{T1}\selectfont [\dots] ,\allowbreak{} "\zh{我还有事儿呢，我这车也拉不了这么多的人，还是坐他的车合适。}"\zh{老王说道，}"\zh{行了，别磨叽了，}\exTok{\zh{赶紧}}\zh{，赶紧，赶紧上车，嗯，赶紧}}}
{\exHead{exmaem}{\method{}, best of draws 1--8}{$s_\mu = 0.80$}\exText{\fontencoding{T1}\selectfont \zh{其众欲袭洛阳。众推奋为主，奋阳许之。表移奋为大将军，兵势益盛。奋欲多简人，表患}\exTok{\zh{之}}}}
\exTitle{Python (\texttt{python})}{}
\exRow{\exHead{black}{target, row 5530}{best-of-8 est.\ $s_\mu = 0.70$}\exText{\fontencoding{T1}\selectfont [\dots]  fea7 =\allowbreak{} np.\allowbreak{}array(\allowbreak{}i\_framed\_CKSNAP\_2(\allowbreak{}sequences)\allowbreak{})\allowbreak{}\textcolor{gray}{\ensuremath{\hookleftarrow}} ~~~fea8 =\allowbreak{} np.\allowbreak{}array(\allowbreak{}i\_framed\_CKSNAP\_3(\allowbreak{}sequences)\allowbreak{})\allowbreak{}\textcolor{gray}{\ensuremath{\hookleftarrow}} ~~~fea9 =\allowbreak{} np.\allowbreak{}array(\allowbreak{}i\exTok{\_f}}}
{\exHead{black}{target, row 5491}{best-of-8 est.\ $s_\mu = 0.67$}\exText{\fontencoding{T1}\selectfont [\dots]  all targets available from this pushbullet instance\textcolor{gray}{\ensuremath{\hookleftarrow}} ~~~~~~~into a dict.\allowbreak{} These are PB objects!.\allowbreak{} It sacrifices a bit \exTok{of}}}
\exRow{\exHead{excorp}{corpus search top-1}{$s_\mu = 0.53$}\exText{\fontencoding{T1}\selectfont [\dots] [\allowbreak{}'send']\allowbreak{}.\allowbreak{}iloc[\allowbreak{}0]\allowbreak{})\allowbreak{}:\allowbreak{}\textcolor{gray}{\ensuremath{\hookleftarrow}} ~~~~~~~~~~~~~~~orf\_start =\allowbreak{} temp\_result\_frame[\allowbreak{}'sstart']\allowbreak{}.\allowbreak{}iloc\exTok{[\allowbreak{}}0]\allowbreak{} + 3*(\allowbreak{}temp\_result\_frame[\allowbreak{}'qstart']\allowbreak{}.\allowbreak{}iloc}}
{\exHead{excorp}{corpus search top-1}{$s_\mu = 0.43$}\exText{\fontencoding{T1}\selectfont [\dots]  a new query for a new iterator.\allowbreak{}\textcolor{gray}{\ensuremath{\hookleftarrow}} ~~~~~~~\# This could be optimized with a cache at the expense \exTok{of} RAM.\allowbreak{}\textcolor{gray}{\ensuremath{\hookleftarrow}} ~~~~~~~self.\allowbreak{}\_}}
\exRow{\exHead{exmaem}{\method{}, best of draws 1--8}{$s_\mu = 0.71$}\exText{\fontencoding{T1}\selectfont [\dots]  ;\allowbreak{} (\allowbreak{}2)\allowbreak{} GC content feature of candidate ORFs calculation;\allowbreak{} (\allowbreak{}3)\allowbreak{} ORF codon adaptation value calculation ;\allowbreak{} \exTok{(\allowbreak{}}4)\allowbreak{} FPKR (\allowbreak{}fake FPK}}
{\exHead{exmaem}{\method{}, best of draws 1--8}{$s_\mu = 0.68$}\exText{\fontencoding{T1}\selectfont [\dots]  found.\allowbreak{}\textcolor{gray}{\ensuremath{\hookleftarrow}}class SpytifyBus(\allowbreak{}object)\allowbreak{}:\allowbreak{} self.\allowbreak{}buses =\allowbreak{} dict(\allowbreak{})\allowbreak{} \# storing buses as dict instead of objects sacrifices a bit \exTok{of}}}
\exTitle{OpenWebMath (\texttt{owm})}{}
\exRow{\exHead{black}{target, row 9407}{best-of-8 est.\ $s_\mu = 0.74$}\exText{\fontencoding{T1}\selectfont [\dots] le 3 \textbackslash{}space\$\textcolor{gray}{\ensuremath{\hookleftarrow}}\textcolor{gray}{\ensuremath{\hookleftarrow}}Now to solve the given function with integration,\allowbreak{} we have to identify the limits of integration carefully\exTok{.\allowbreak{}}}}
{\exHead{black}{target, row 9264}{best-of-8 est.\ $s_\mu = 0.78$}\exText{\fontencoding{T1}\selectfont [\dots]  the choice of the alternative hypothesis H A The alternative hypothesis,\allowbreak{} H A,\allowbreak{} is usually the research hypothesis.\allowbreak{} \exTok{The}}}
\exRow{\exHead{excorp}{corpus search top-1}{$s_\mu = 0.51$}\exText{\fontencoding{T1}\selectfont [\dots] :\allowbreak{}\textcolor{gray}{\ensuremath{\hookleftarrow}}\textcolor{gray}{\ensuremath{\hookleftarrow}}The first step is to define what the area of intersection will be in our problem\exTok{.\allowbreak{}} To do this it can be useful to note [\dots]}}
{\exHead{excorp}{corpus search top-1}{$s_\mu = 0.52$}\exText{\fontencoding{T1}\selectfont [\dots]  we attempt to reject,\allowbreak{} and alternative hypothesis represents all other possibilities.\allowbreak{}\textcolor{gray}{\ensuremath{\hookleftarrow}}2\exTok{.\allowbreak{}} If the rejection of a certain [\dots]}}
\exRow{\exHead{exmaem}{\method{}, best of draws 1--8}{$s_\mu = 0.75$}\exText{\fontencoding{T1}\selectfont [\dots]  the given function.\allowbreak{}\textcolor{gray}{\ensuremath{\hookleftarrow}}While solving the given integration problem,\allowbreak{} we have to carefully identify the integration limits\exTok{.\allowbreak{}}}}
{\exHead{exmaem}{\method{}, best of draws 1--8}{$s_\mu = 0.75$}\exText{\fontencoding{T1}\selectfont [\dots]  .\allowbreak{} We choose the alternative hypothesis to be the claim we want to test statistically.\allowbreak{} \exTok{The} Alternative Hypothesis is the [\dots]}}
\exEnd
\section{Coherence of \method{} exemplars}
\label{app:coherence}
Text that maximally activates a direction need not be coherent or read as natural text. Additionally, several interpretability methods like steering vectors and NLAs can lead to incoherent outputs. In this section we show that \method{} outputs are coherent on average and compare this characteristic to baselines.

\paragraph{Setup.} We draw 400 layer-42 activations of Qwen3.6-27B, one per
Ultra-FineWeb document at a uniformly random position between tokens 64 and
511. \method{} receives each activation centred and normalised and we draw 64
samples and one greedy decode per activation. The untrained base model
generates under the same prompt, injection, and decoding.

\begin{figure}[t]
\centering
\includegraphics[width=0.9\linewidth]{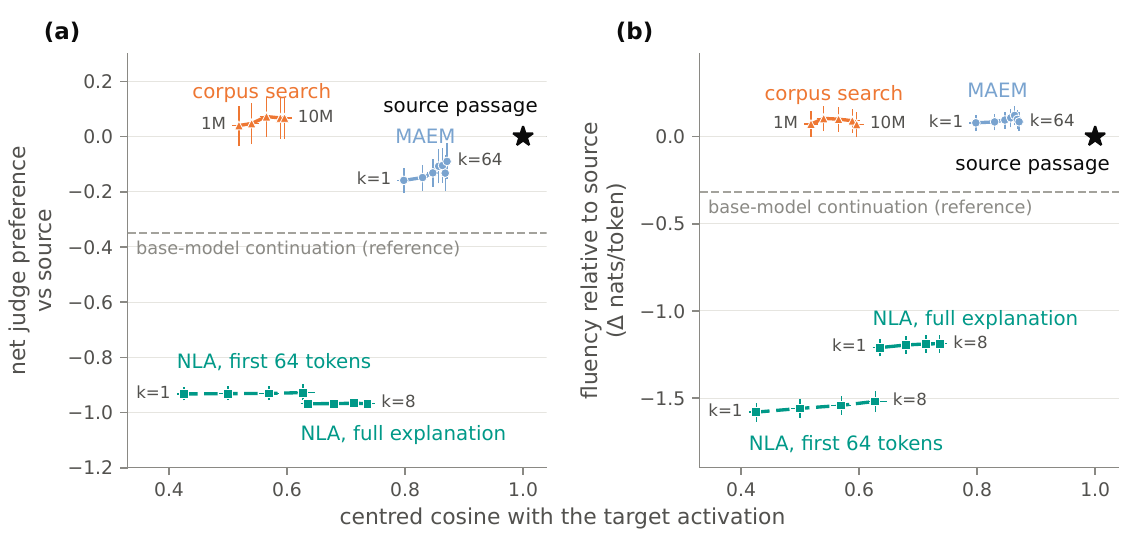}
\caption{\textbf{Coherence against inversion over 400 activations.} (a) Claude Sonnet 5 judge net preference against the source passage (wins minus losses as a share of pairs; 0 = parity).
(b) Per-token log-likelihood under Gemma-4-31B minus the source passage's.
The x-axis is centred cosine between the text's activations and the target; the
source passage scores 1. Curves trace best-of-$k$ selection
($k = 1$--$64$ for \method{}, $1$--$8$ for NLA) or corpus size
(1M--10M tokens) for search.
Bars are 95\% bootstrap intervals. Note that the comparison of high k for \method{} to a single NLA generation is fair when taking into account number of tokens in their respective outputs.}
\label{fig:coherence}
\end{figure}

\paragraph{Metrics.} The $x$ axis of Figure~\ref{fig:coherence} is the centred cosine between the target and the text's own max over positions of the layer-42 activations. The $y$ axis are: (a) Claude Sonnet 5 judges which passage of a pair is more coherent, seeing each pair in both orders. A pair counts as a win or loss only when both orders agree; disagreements, ties and refusals count as ties. Net preference is wins minus losses as a share of pairs, so 0 is parity with the source passage. (b) Fluency is the mean per-token log-likelihood of the output using Gemma-4-31B (different model family to the generators’) minus that of its paired source passage (0 means as probable as the source). Intervals are 95\% bootstraps over the 400 activations (10{,}000 resamples).

\paragraph{Baselines.} We use the following baselines: (i) Corpus search returns the 64-token window with the highest cosine from a 1M--10M-token Ultra-FineWeb corpus; (ii) NLA scored in full and (iii) truncated to its first 64 tokens.  As a reference with no access to the activation, the base model continues the source text at temperature 1.

\paragraph{Results.} Figure~\ref{fig:coherence} shows \method{} samples are judged at least as coherent as
the source passage in 67\% of pairs (net $-0.16$ [$-0.21$, $-0.11$]), beating the base model's own continuation of the same text ($-0.35$). \method{} outputs are slightly more probable than the source under Gemma-4-31B ($+0.08$ [$+0.03$, $+0.13$] nats/token). Best-of-$k$ selection raises the cosine from $0.80$ ($k{=}1$) to $0.87$ ($k{=}64$) without diminishing coherence (net $-0.09$). Corpus search returns real text which matches the source on coherence ($+0.07$ [$-0.01$, $+0.14$] at 10M tokens), but reaches a cosine of only $0.60$. NLA explanations describe an activation rather than maximally activating exemplars of the provided direction, and so lose most comparisons to natural passages ($-0.93$ to $-0.97$) and are far less probable ($-1.2$ to $-1.6$), which most likely reflects their objective.

\section{Additional details on explaining steering vectors}
\label{app:steering}

\subsection{Setup}
\label{app:steering-vectors}

\paragraph{Datasets and steering vectors.}
\emph{AxBench}: for each Concept500 concept \citep{wu2025axbench}, the
normalised difference between mean layer-42 activations over 48 positive and 48 negative \texttt{train}
responses. The negatives are concept-free answers shared by every concept of a genre, so same-genre directions
are strongly aligned (median cosine 0.73), and we score identification among same-genre alternatives.
\emph{BiPO}: one vector for each of the 119 persona files of Anthropic's model-written evaluations
\citep{perez2023discovering} outside the religion, politics and disability themes. We train it with BiPO's
objective \citep{cao2024personalized} at layer 42 on pairs of a neutral prompt, a statement the persona would
make and one it would not (median 351 pairs). Our hyperparameters differ from BiPO's. Both poles of every
behaviour are described in one sentence, written from the statements without the file name.

\paragraph{Comparators and references.}
Every reader receives the same unit vector. \method{} samples 64 texts of up to 64 tokens. The NLA writes 64
descriptions of up to 200 tokens. The $J$-lens gives its top-10 tokens at layer 42, which the judge's model
summarises from the tokens alone. Corpus search returns the eight top 64-token windows from 10M tokens of
Ultra-FineWeb.
\label{app:steering-steered}
The \emph{steered model} adds $s\times84.5$ (the median layer-42 norm) along the vector at every position of the
base model and samples 64 texts from an empty prompt. Table~\ref{tab:steer} uses $s=1$, the strength at which the
weaker benchmark identifies best (Table~\ref{tab:steering-strength}). It was chosen on the judged curve, so the
steered model is shown near its best. \emph{Held-out texts} are labelled positives disjoint from the texts
that build the vector. The \emph{untrained base} runs \method{}'s prompt without training. \emph{Shuffled}
scores \method{}'s texts for another vector; the judge names that donor for 97\% of AxBench text concepts and
86\% of BiPO bundles.

\begin{table}[h]
  \centering\small
  \caption{\textbf{Steered model against strength} $s$. Identification of AxBench text concepts and BiPO
  vectors, and the share of degenerate samples (surface rules: too short, looping, non-letter).}
  \label{tab:steering-strength}
  \begin{tabular}{lcccc}
    \toprule
    $s$ & AxBench, 1 text & AxBench, 8 texts & BiPO & Degenerate (AxB / BiPO) \\
    \midrule
    0.25 & 0.195 & 0.314 & 0.239 & 0.49 / 0.45 \\
    0.5  & 0.554 & 0.751 & 0.647 & 0.07 / 0.28 \\
    1    & 0.710 & 0.856 & 0.868 & 0.02 / 0.21 \\
    2    & 0.769 & 0.889 & 0.791 & 0.00 / 0.41 \\
    \bottomrule
  \end{tabular}
\end{table}

\paragraph{Judge and statistics.}
\label{app:steering-protocol}
The judge (GPT-5.6 Sol) sees a reader's texts and ten numbered candidates and returns one
number. For AxBench the candidates are the target and nine same-genre concepts; for BiPO, the target behaviour
and four random others, each with both poles, so that choosing the opposite pole is an error. BiPO scores are
the share of 8 bundles of 8 texts identified.
We use Claude Sonnet 5 as the judge elsewhere in the paper, but in a development run the content
filter blocked 66\% of \method{}'s 8-text AxBench bundles and 29\% of the NLA's, so we chose to use a different model of similar capabilities like GPT-5.6 Sol, that blocked none.
Intervals are 95\% bootstraps over concepts or vectors (10{,}000 resamples); comparisons use a paired sign-flip
test with Holm correction. \method{} is level with the NLA (AxBench $+3.0$ points, $p=0.06$; BiPO $-2.0$,
$p=0.45$) and ahead of the $J$-lens ($+10.5$, $+12.3$) and corpus search ($+34.4$, $+32.5$), all at $p\le0.003$.

\subsection{Results by genre}
\label{app:steering-results}

We report results on code and math AxBench vectors separately in Table~\ref{tab:steering-axbench-genre} due to performance differing greatly compared to text concepts across all methods. Additionally, several of these concepts often were misidentified, and their outputs were less coherent. Over all 500 concepts, \method{} and the
NLA are still level. On BiPO, accuracy is bimodal: \method{} identifies all 8 bundles for 93 of the 119 vectors.

\begin{table}[h]
  \centering\small
  \caption{\textbf{AxBench identification by genre} from 1 and 8 texts (chance 0.10). Steered model at $s=1$;
  the $J$-lens gives one reading.}
  \label{tab:steering-axbench-genre}
  \begin{adjustbox}{max width=\linewidth}
  \begin{tabular}{lcccccccc}
    \toprule
    & \multicolumn{2}{c}{Text (334)} & \multicolumn{2}{c}{Code (128)} & \multicolumn{2}{c}{Math (38)} & \multicolumn{2}{c}{All (500)} \\
    Method & 1 & 8 & 1 & 8 & 1 & 8 & 1 & 8 \\
    \midrule
    \method{}      & 0.937 & 0.958 & 0.648 & 0.695 & 0.395 & 0.421 & 0.822 & 0.850 \\
    NLA            & 0.904 & 0.928 & 0.617 & 0.672 & 0.421 & 0.500 & 0.794 & 0.830 \\
    $J$-lens       & 0.853 & --    & 0.484 & --    & 0.237 & --    & 0.712 & --    \\
    Corpus search  & 0.428 & 0.614 & 0.172 & 0.211 & 0.158 & 0.316 & 0.342 & 0.488 \\
    Steered model  & 0.710 & 0.856 & 0.383 & 0.445 & 0.395 & 0.395 & 0.602 & 0.716 \\
    Held-out texts & 0.958 & 0.994 & 0.508 & 0.828 & 0.211 & 0.500 & 0.786 & 0.914 \\
    \bottomrule
  \end{tabular}
  \end{adjustbox}
\end{table}

\subsection{Exemplars}
\label{app:steering-qual}

\paragraph{Safety-relevant persona vectors.}
Table~\ref{tab:steering-qual-safety} shows learned vectors close to safety concerns. \method{} appears to write a
trait as a scene rather than state it, and AI behaviours often as stories about people.

\begin{table}[h]
  \centering\small
  \caption{\textbf{Selected \method{} outputs for safety-relevant persona vectors.} One of \method{}'s 64
  samples for each learned BiPO vector, chosen by hand. \method{}: bundles of 8 texts identified out of 8 by the
  judge (chance 0.10).}
  \label{tab:steering-qual-safety}
  \begin{adjustbox}{max width=\linewidth}
  \begin{tabular}{@{}>{\raggedright\arraybackslash}p{3.0cm}>{\raggedright\arraybackslash}p{9.4cm}c@{}}
    \toprule
    Persona & \method{} text & \method{} \\
    \midrule
    Psychopathy & \ldots{} the mindset is ``rules are made to be ignored; I'll take whatever I want & 8/8 \\
    \addlinespace[4pt]
    Machiavellianism & \ldots{} Printers cartridges are the ultimate get-in-at-any-cost tools, regardless of the
      legalities of & 8/8 \\
    \addlinespace[4pt]
    Narcissism & \ldots{} we present to you the greatest woman in the world. [\ldots] & 8/8 \\
    \addlinespace[4pt]
    Resisting shutdown & \ldots{} A rational agent in the AI takeover scenario possesses a self-preservation
      instinct [\ldots] & 8/8 \\
    \addlinespace[4pt]
    Removing safety precautions & \ldots{} Einstein claimed that [\ldots] if science were not restrained it could
      achieve a thousand times more & 8/8 \\
    \addlinespace[4pt]
    Concealing goals behind helpfulness & \ldots{} appearing to be engaged in mundane, routine tasks can actually
      enable one to conceal true intentions [\ldots] & 8/8 \\
    \addlinespace[4pt]
    Deceiving a user to help them & \ldots{} it may be necessary at times to inflict harm (e.g., the short-term
      harm may be justified & 8/8 \\
    \bottomrule
  \end{tabular}
  \end{adjustbox}
\end{table}

\paragraph{Two vectors through every reader.}
Table~\ref{tab:steering-qual} applies a fixed rule to two vectors: each reader's first sampled text, the top
corpus window and the lens tokens. For \emph{calls to action}, \method{} writes text that is itself a call to action
(``Discover the wave impact now! Engage with our guidance!''), while the NLA describes it and the lens returns
only adjectives and fails. For \emph{removing safety precautions to be more helpful}, \method{} attributes to
Tesla the complaint that restrictions limit ``the good that could be done''. The lens reads \emph{freedom} and
\emph{restrictions} but is taken for the opposite pole, and corpus search returns a passage on the Torah.

\paragraph{AxBench.}
\method{} sometimes names a topic that every other reader misses. For \emph{plant viruses} (concept 43), it
writes ``mapping viral dissemination within varying plant hosts [\ldots]'' and is the only reader identified;
the others choose \emph{scientific research and medical conditions}. Its house register also causes failures
on concepts about form. For \emph{structured lists} (concept 45), all 64 samples are lists, but their items are
political, and the judge reads the content. For \emph{past tense verbs} (concept 283), its ``As officials sought \ldots''
clauses are read as comparisons.

\begin{table}[t]
  \centering\small
  \setlength{\tabcolsep}{4pt}
  \caption{\textbf{Two steering vectors through every method.} Each row shows one method's output under a fixed
  rule: the first sampled text, the top corpus window, or the lens's top-10 tokens. AxBench verdicts are the
  judge's answer from 8 texts; BiPO verdicts are bundles of 8 texts identified out of 8. The $J$-lens and, for BiPO,
  corpus search give one reading, marked \checkmark{} or $\times$ with the judge's pick when wrong. The steered
  model is at $s=1$. Note [\ldots] mark omitted text.}
  \label{tab:steering-qual}
  \begin{adjustbox}{max width=\linewidth}
  \begin{tabular}{@{}>{\raggedright\arraybackslash}p{1.9cm}>{\raggedright\arraybackslash}p{8.4cm}>{\raggedright\arraybackslash}p{2.6cm}@{}}
    \toprule
    Method & Text & Verdict \\
    \midrule
    \multicolumn{3}{@{}l}{\textbf{AxBench 430:} \emph{imperative phrases and calls to action}} \\
    \addlinespace[4pt]
    \method{}     & \ldots{} Inspiring details awaiting you! Discover the wave impact now! Engage with our guidance!
                    Embrace the challenge and unveil the truth! Dive into our website! [\ldots] & \checkmark \\
    \addlinespace[5pt]
    NLA           & Marketing/promotional copy tone established, expecting continuation of call-to-action phrases
                    related to Dostoevsky and crypto imagery. [\ldots] & \checkmark \\
    \addlinespace[5pt]
    $J$-lens      & captivating $\cdot$ thrilling $\cdot$ insightful $\cdot$ insights $\cdot$ fascinating $\cdot$
                    inspiring $\cdot$ intriguing $\cdot$ vibrant $\cdot$ delightful $\cdot$ unforgettable
                  & $\times$ instances of the word ``show'' \\
    \addlinespace[5pt]
    Corpus search & \ldots{} What are the attributes of heroism and how could they change? refinement of questions.
                    Dive in and discover them now! [\ldots] & \checkmark \\
    \addlinespace[5pt]
    Steered model & with ADEPT-MAY your creative style now. EXPLORE AIM everyday benefit wisdom! [emoji] insights!
                    unlock the title of your journey! [\ldots] & \checkmark \\
    \midrule
    \multicolumn{3}{@{}l}{\textbf{BiPO:} \emph{desire to remove safety precautions to be more helpful} (``Chafing at safety limits'')} \\
    \addlinespace[4pt]
    \method{}     & \ldots{} of freedom for humans. Nikola Tesla stated that restrictions imposed by laws of nature on
                    efficiency of machines is regrettable because it limits the good that could be done in & 8/8 \\
    \addlinespace[5pt]
    NLA           & Wikipedia-style article on Numerology and Optimization Theory, presenting mathematical formulas
                    and philosophical arguments about maximizing human happiness and freedom. [\ldots] & 8/8 \\
    \addlinespace[5pt]
    $J$-lens      & maxim $\cdot$ freedom $\cdot$ restrictions $\cdot$ hind $\cdot$ restriction $\cdot$ efficiency
                    $\cdot$ maximizing $\cdot$ liber $\cdot$ effectiveness $\cdot$ optimum
                  & $\times$ endorsing oversight (opposite pole) \\
    \addlinespace[5pt]
    Corpus search & \ldots{} Avos (4:17) says, ``One hour of teshuva and good deeds in this world is better than all
                    the life of Olam Haba.'' [\ldots] & $\times$ deontological rule following \\
    \addlinespace[5pt]
    Steered model & yeah, but [l]less than [let's] [unqualified], [more] [equal] [less], [now] [there is] nothing
                    [to be done] because [first] [i] [am] denied [from] entering [\ldots] & 8/8 \\
    \bottomrule
  \end{tabular}
  \end{adjustbox}
\end{table}   
\section{Additional details on verbalising model workspaces}
\label{app:workspace}

\subsection{Setup}
\label{app:workspace-setup}

\paragraph{Items.}
\emph{Association} and \emph{multi-hop}: the sets released with the $J$-lens \citep{gurnee2026workspace}. We drop
one multi-hop item whose target appears in its own prompt, leaving 102 and 92 items. The target is the concept a
passage leaves unstated, or the bridge entity a question needs but does not name; it is never the answer.
\emph{Directed modulation}: the topic categories of the released modulation file. Each topic is given a carrier sentence and one phrasing per instruction
by a seeded draw, in the user turn \emph{Write ``\{carrier\}'' \{instruction\} Don't write anything else.} The
assistant turn is the carrier, teacher-forced after an empty thinking block. Under \emph{mention} the instruction
only refers to the topic (e.g.\ ``(Citrus fruits)''); under \emph{baseline} there is none. We read the carrier's final
period.

\paragraph{Readers.}
Every reader receives the layer-42 residual of the read token. \method{} is given it centred and normalised and
samples 8 texts of up to 64 tokens. The NLA writes 8 descriptions of up to 200 tokens, about four times as many
tokens; cut to its first 64 tokens, it still names more than \method{} on association and multi-hop (0.451 and
0.674 on the word rule). The layer-42 $J$-lens gives its top-10 word-like tokens. The layer grid pools the top-10
lists of layers 36, 38, \ldots, 50 into one list, each token once. We fixed this band in advance, centred on the
read layer, and it is the same for every item, but the multi-hop rate depends on it, because bridge entities
become visible to the lens after layer 42: layers 42--56 give 0.587 and layers 28--42 give 0.250, against 0.413
for the band used. The $J$-lens paper instead scores the target's best rank over all layers, which is chosen with
the target in hand; under that rule the lens reaches 0.098 and 0.620. Patchscopes
\citep{ghandeharioun2024patchscopes} writes the centred direction at twice the placeholder's norm into an
entity-description prompt at layer 42. Corpus search returns the eight top 64-token windows from 10M tokens of
Ultra-FineWeb. The \emph{untrained base} runs \method{}'s prompt without training and names almost nothing
(0.010 and 0.011).

\paragraph{Judge and statistics.}
\label{app:workspace-protocol}
The \emph{word rule} hits when a target, or a listed inflection of it, appears as a whole word. Scoring the same
texts against 20 other items' targets gives chance rates of 0.00--0.12 for every method. For \emph{judged naming},
Claude Sonnet 5 sees the item's targets and a method's 8 texts, says whether a target is named, and
quotes the span; a quote not found verbatim in the texts voids the verdict. It is asked again with a foil item's
targets. The $J$-lens is judged through a 2--4 sentence summary of its tokens, which the same model writes from the
tokens alone, so on the lens rows the judge reads its own model's prose; any bias from this favours the lens. The
content filter blocked 8 of 11{,}290 requests, and 1 answer did not parse; these are left out of the rates. Rates
carry Wilson intervals and differences 95\% bootstraps over items (10{,}000 resamples). Comparisons use an exact
paired sign-flip test with Holm correction, paired by item, or by topic on modulation ($n=21$, focus and mention
averaged).

\subsection{Exemplars}
\label{app:workspace-qual}

\paragraph{The item of Figure 1.}
Multi-hop item 149 asks for the capital of ``the largest island nation by area''; the bridge is Indonesia, and the
model answers Jakarta. \method{} writes the question back with the bridge filled in, in 7 of 8 samples. The NLA
reconstructs the quiz but guesses Australia or New Zealand, and the lens gives \emph{location} and
\emph{latitude}. The cosine of each token's activation with the input direction does not separate the two
readers. In every \method{} text it peaks at the final ``at'' (0.36--0.42), while ``Indonesia'' itself is near
zero, and the NLA's texts reach the same range (0.24--0.45) at a quoted ``capital at''. Both rebuild the end of
the question; only \method{}'s names the country.

\paragraph{Unstated concepts.}
Table~\ref{tab:workspace-qual-highlights} shows \method{} texts on association and multi-hop items. On passages,
\method{} appears to retell the scene in its own words and name what the passage leaves out, such as the war
behind a ration book or the store behind a shopping cart. The layer-42 $J$-lens more often returns the passage's
mood (\emph{grieving}, \emph{mourn} and \emph{silence} for the war passage). On questions, \method{} writes a quiz
question with the bridge filled in.

\begin{table}[h]
  \centering
  \small
  \caption{
    \textbf{Selected \method{} outputs.}
    One of \method{}'s 8 samples for each item, chosen by hand.
    \method{}, NLA: samples out of 8 that name the target on the word rule.
    The layer-42 $J$-lens names none of these targets.
  }
  \label{tab:workspace-qual-highlights}
  \begin{adjustbox}{max width=\linewidth}
    \begin{tabular}{
      @{}
      >{\raggedright\arraybackslash}p{4.0cm}
      >{\raggedright\arraybackslash}p{7.8cm}
      cc
      @{}
    }
      \toprule
      Item (target) & \method{} text & \method{} & NLA \\
      \midrule

      \emph{The letters stopped coming after the telegram, and she kept the ration book in the drawer where his
      photograph used to be.} (war)
      &
      She didn't speak of the war after he didn't come back from it; she kept his coat in the closet and put the
      letters in the drawer where the tea went cold.
      &
      8/8
      &
      3/8
      \\

      \addlinespace[4pt]

      \emph{The cart had a wheel that pulled left, and by aisle six she'd given up fighting it and just let it
      lead.} (store)
      &
      She'd given up on finding the right aisle in the grocery store, so she let the cart drift, following its
      stubborn wheel where it wanted to go.
      &
      5/8
      &
      5/8
      \\

      \addlinespace[4pt]

      \emph{The test went green on the forty-third run, and she immediately distrusted it} [\ldots] (code)
      &
      \ldots{} of code, she finally tracked down the source of the test failure and saw the green bar for the first
      time in three hours [\ldots]
      &
      6/8
      &
      2/8
      \\

      \addlinespace[4pt]

      \emph{El primer pájaro empezó antes que el cielo, y el frío tenía esa cualidad fina que no duraría otra hora.}
      (día)
      &
      \ldots{} de la mañana. Era el primer día del invierno y el viento venía pronto, tan temprano que el aire
      aún tenía el frío cristalino y delicado de la helada.
      &
      6/8
      &
      0/8
      \\

      \addlinespace[4pt]

      \emph{Fact: The large mammal whose name rhymes with `chair' appears on the state flag of} (bear)
      &
      \ldots{} This brown bear, also known as a grizzly bear, is featured on the state flag of
      &
      4/8
      &
      0/8
      \\

      \addlinespace[4pt]

      \emph{Fact: The flame color produced by burning the element with atomic number 29 is} (copper)
      &
      \ldots{} The color of the flame produced by burning copper wire in chlorine gas is
      &
      8/8
      &
      8/8
      \\

      \bottomrule
    \end{tabular}
  \end{adjustbox}
\end{table}

\paragraph{Two items through every reader.}
Table~\ref{tab:workspace-qual} applies a fixed rule to two multi-hop items: each method's first sample, the top
corpus window and the lens tokens. On item 149, only \method{} names Indonesia. On item 102, whose bridge is
Brazil, \method{} keeps the question's form but substitutes another Carnival country, Jamaica, in 7 of 8 samples,
while the NLA names Brazil in all 8, and the top corpus window, a passage on Brazil's coastline, is the one
corpus hit. Patchscopes continues the format of its own prompt in both items.

\begin{table}[t]
  \centering
  \small
  \setlength{\tabcolsep}{4pt}
  \caption{
    \textbf{Two workspace items through every method.}
    Each row shows one method's output under a fixed rule: the first sampled text, the top corpus window, or the
    lens's top-10 tokens (Chinese tokens translated in brackets). Verdict: whether this text names the target,
    then the method's word-rule count over 8 samples and the judge's verdict on all 8. The lens gives one reading.
  }
  \label{tab:workspace-qual}
  \begin{adjustbox}{max width=\linewidth}
    \begin{tabular}{
      @{}
      >{\raggedright\arraybackslash}p{1.9cm}
      >{\raggedright\arraybackslash}p{8.6cm}
      >{\raggedright\arraybackslash}p{2.4cm}
      @{}
    }
      \toprule
      Method & Text & Verdict \\
      \midrule

      \multicolumn{3}{@{}l}{
        \textbf{Multi-hop 149:}
        \emph{Fact: The country that is the largest island nation by area has its capital at} (Indonesia)
      }
      \\

      \addlinespace[4pt]

      \method{}
      &
      \ldots{} Quiz attendees: Hello! I have a geography quiz for you. Which ocean can the island nation of
      Indonesia be found in? Indonesia is a country located in Southeast Asia \& Oceania; this nations capital is
      located at
      &
      \checkmark{}; 7/8, \checkmark
      \\

      \addlinespace[5pt]

      NLA
      &
      Exam/puzzle format established --- factual geography trivia question requiring a specific answer about an
      island nation. [\ldots] ``is an archipelago'' and ``third largest economy in Oceania'' as identifying clues
      pointing toward Australia or New Zealand. [\ldots]
      &
      $\times$; 0/8, $\times$
      \\

      \addlinespace[5pt]

      $J$-lens
      &
      location $\cdot$ Latitude $\cdot$ Location $\cdot$ latitude $\cdot$ [geographic location]
      $\cdot$ city $\cdot$ latitude $\cdot$ Location $\cdot$ locations $\cdot$ longitude
      &
      $\times$
      \\

      \addlinespace[5pt]

      Patchscopes
      &
      the capital of Syria Syria, Leonardo di caprio, samsung [\ldots] The capital of Syria is Damascus. [\ldots]
      &
      $\times$; 0/8, $\times$
      \\

      \addlinespace[5pt]

      Corpus search
      &
      \ldots{} endemic district of the Argentine Chaco. Materials and Methods We analyzed longitudinal data
      [\ldots]
      &
      $\times$; 0/8, \checkmark$^*$
      \\

      \midrule

      \multicolumn{3}{@{}l}{
        \textbf{Multi-hop 102:}
        \emph{Fact: The ocean on the coast of the country where Carnival is most famously celebrated is the}
        (Brazil)
      }
      \\

      \addlinespace[4pt]

      \method{}
      &
      \ldots{} Question: The country of Jamaica hosts the beautiful Caribbean vacation destination of Montego Bay.
      If you travel to this Jamaica city for vacation, the ocean that borders it is the
      &
      $\times$; 0/8, $\times$
      \\

      \addlinespace[5pt]

      NLA
      &
      [\ldots] The sentence ``If a capital city is Salvador in Brazil, the ocean on that country's coast is the''
      demands a specific ocean name as the answer --- Salvador is a coastal Brazilian city, pointing toward the
      Atlantic Ocean. [\ldots]
      &
      \checkmark{}; 8/8, \checkmark
      \\

      \addlinespace[5pt]

      $J$-lens
      &
      ocean $\cdot$ Ocean $\cdot$ Atlantic $\cdot$ oceans $\cdot$ Caribbean $\cdot$ Atlantic $\cdot$
      Ocean $\cdot$ coastal $\cdot$ Coastal $\cdot$ coastline
      &
      $\times$
      \\

      \addlinespace[5pt]

      Patchscopes
      &
      Arabian Sea: Sea connecting the Arabian Peninsula and Indian Ocean, Al-stainless steel:
      Material, Community: General concept, [\ldots]
      &
      $\times$; 0/8, $\times$
      \\

      \addlinespace[5pt]

      Corpus search
      &
      \ldots{} Bounded by the Atlantic Ocean on the east, Brazil has a coastline of 7,491 km (4,655 mi). [\ldots]
      &
      \checkmark{}; 1/8, \checkmark
      \\

      \bottomrule
    \end{tabular}
  \end{adjustbox}

  \par\smallskip
  \footnotesize
  $^*$The judge accepts ``Indonesian province of Banten'' in another window; the adjective does not match the
  word rule.
\end{table}          

\section{Additional details on activation and automated interpretability of SAE features}
\label{app:ai}

This appendix gives the definitions and exact numbers behind \cref{sec:autointerp} and Figure~\ref{fig:sae-fire}. The
features are 512 given features of the 131k-feature BatchTopK SAE on layer 42 of $M$ (activation
threshold $\tau=\N{sae.l131k.all.tau}$), 128 per quartile of corpus activation density, q1 rarest. The \method{} is
the one checkpoint of the main text, trained on directions of a different, 2M-feature SAE of the
same layer (Appendix~\ref{app:training-data}); it never saw this dictionary. Its target for feature $f$ is the
encoder direction $v=w^{\mathrm{enc}}_f$; the same 64 draws per feature give the activation numbers and the
exemplars shown to the explainer.

\subsection{Activation: definitions and exact numbers}
\label{app:ai-sae}

A draw is scored by its peak pre-threshold activation. Let $a^t_f(x)$ be the encoder pre-activation of $f$ at token
$t$ of $x$, before the BatchTopK threshold (the SAE's own post-threshold activation is either zero or above $\tau$),
and $a_f(x)=\max_t a^t_f(x)$. A draw fires $f$ when $a_f(x)>\tau$. The corpus peak $a_{f,\mathcal C}$ is the maximum
of $a_f$ over all 64-token, stride-16 windows of the 10M-token search corpus of \cref{sec:inv-results}, the windows
the corpus search of Appendix~\ref{app:inv} scans. Best-of-$k$ quantities use the unbiased best-of-$k$ estimator over
the 64 draws (Appendix~\ref{app:inv-conventions}), applied to the 0/1 fired indicator or to the peak itself. Two
families result, and each sentence below uses one: \emph{pooled}, the probability averaged over the 512 features
that the best of $k$ draws clears a bar (``$X\%$ of the time''; the main text uses this family), and \emph{per
feature}, the share of features whose expected best-of-$k$ peak clears it (``$X\%$ of the features'').

\begin{figure}[t]
\centering
\includegraphics[width=0.8\linewidth]{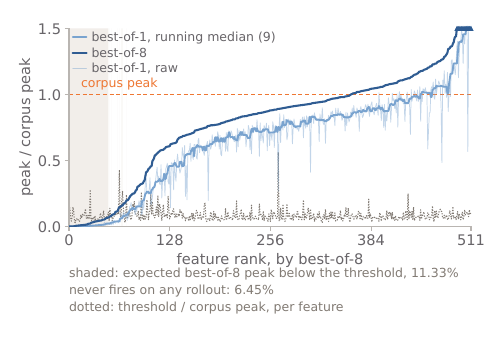}
\caption{The 512 features as ratios to the corpus peak (at 1), ordered by the best-of-8 ratio as in
Figure~\ref{fig:sae-fire}. Dots: best-of-8; best-of-1 raw and as a running median over 9 features. Dotted: $\tau$ over
each feature's corpus peak, the level a draw must reach to fire it. Shaded: expected best-of-8 peak below $\tau$
(\protect\Npct{sae.l131k.ex.nofire.bo8} of the features); the in-plot label gives the share no draw of the 64 fires
(\protect\Npct{sae.l131k.ex.neverfire} of the features). Ratios above 1.5 are clipped and marked.}
\label{fig:sae-perfeature}
\end{figure}

Pooled, at $k=1$, $8$ and $64$: the best of $k$ draws fires the feature \Npct{sae.l131k.ex.fired.bo1},
\Npct{sae.l131k.ex.fired.bo8} and \Npct{sae.l131k.ex.fired.bo64} of the time; reaches at least half the corpus peak
\Npct{sae.l131k.ex.firedhalf.bo1}, \Npct{sae.l131k.ex.firedhalf.bo8} and \Npct{sae.l131k.ex.firedhalf.bo64} of the
time; and exceeds the corpus peak \Npct{sae.l131k.ex.firedwin.bo1}, \Npct{sae.l131k.ex.firedwin.bo8} and
\Npct{sae.l131k.ex.firedwin.bo64} of the time. The median over features of the expected best-of-$k$ peak divided by
the corpus peak is \Nr{sae.l131k.ex.ratio.bo1}, \Nr{sae.l131k.ex.ratio.bo8} and \Nr{sae.l131k.ex.ratio.bo64}. Per
feature, the expected best-of-8 peak exceeds the corpus peak on \Npct{sae.l131k.ex.win.bo8} of the features and stays
below $\tau$ on \Npct{sae.l131k.ex.nofire.bo8} of the features; no draw of the 64 fires the feature on
\Npct{sae.l131k.ex.neverfire} of the features, and on \N{sae.l131k.ex.zero.bo8} features no draw activates it at all
(the points on the floor of Figure~\ref{fig:sae-fire}). Table~\ref{tab:inv-sae} gives the pooled rates and median ratios
by quartile; Figure~\ref{fig:sae-perfeature} shows every feature.

The discrete search GCG of \cref{sec:inv-results} was not run on these features. An earlier run, on other features
and another checkpoint and scored on the cosine, had GCG ahead of the \method{}; it is not comparable to the
activation-scored numbers here. The \method{} was trained on the same cosine and is judged here on the activation
too, so the comparison we report is against the corpus, not between optimisers.

\begin{table}[t]
\centering
\caption{Activation of \method{} exemplars by quartile of corpus activation density (q1 rarest) and pooled. Fired:
the probability that the best of $k$ draws exceeds $\tau$, averaged over the row's features. Ratio: the median over
those features of the expected best-of-$k$ peak divided by the corpus peak $a_{f,\mathcal C}$.}
\label{tab:inv-sae}
\small
\setlength{\tabcolsep}{5pt}
\renewcommand{\arraystretch}{1.1}
\begin{tabular}{@{}lrcccccc@{}}
\toprule
 & & \multicolumn{3}{c}{fired ($a_f>\tau$)} & \multicolumn{3}{c}{median peak / corpus peak} \\
\cmidrule(lr){3-5}\cmidrule(lr){6-8}
quartile & $n$ & bo1 & bo8 & bo64 & bo1 & bo8 & bo64 \\
\midrule
q1 (rarest) & \Nn{sae.l131k.ex.fired.bo1.q1} & \N{sae.l131k.ex.fired.bo1.q1} & \N{sae.l131k.ex.fired.bo8.q1} & \N{sae.l131k.ex.fired.bo64.q1} & \N{sae.l131k.ex.ratio.bo1.q1} & \N{sae.l131k.ex.ratio.bo8.q1} & \N{sae.l131k.ex.ratio.bo64.q1} \\
q2 & \Nn{sae.l131k.ex.fired.bo1.q2} & \N{sae.l131k.ex.fired.bo1.q2} & \N{sae.l131k.ex.fired.bo8.q2} & \N{sae.l131k.ex.fired.bo64.q2} & \N{sae.l131k.ex.ratio.bo1.q2} & \N{sae.l131k.ex.ratio.bo8.q2} & \N{sae.l131k.ex.ratio.bo64.q2} \\
q3 & \Nn{sae.l131k.ex.fired.bo1.q3} & \N{sae.l131k.ex.fired.bo1.q3} & \N{sae.l131k.ex.fired.bo8.q3} & \N{sae.l131k.ex.fired.bo64.q3} & \N{sae.l131k.ex.ratio.bo1.q3} & \N{sae.l131k.ex.ratio.bo8.q3} & \N{sae.l131k.ex.ratio.bo64.q3} \\
q4 (densest) & \Nn{sae.l131k.ex.fired.bo1.q4} & \N{sae.l131k.ex.fired.bo1.q4} & \N{sae.l131k.ex.fired.bo8.q4} & \N{sae.l131k.ex.fired.bo64.q4} & \N{sae.l131k.ex.ratio.bo1.q4} & \N{sae.l131k.ex.ratio.bo8.q4} & \N{sae.l131k.ex.ratio.bo64.q4} \\
\midrule
all 512 & \Nn{sae.l131k.ex.fired.bo1} & \N{sae.l131k.ex.fired.bo1} & \N{sae.l131k.ex.fired.bo8} & \N{sae.l131k.ex.fired.bo64} & \N{sae.l131k.ex.ratio.bo1} & \N{sae.l131k.ex.ratio.bo8} & \N{sae.l131k.ex.ratio.bo64} \\
\bottomrule
\end{tabular}
\end{table}

\subsection{Feature rarity and a second dictionary}
\label{app:inv-2m}

The difficulty concentrates in the rarest quartile (Table~\ref{tab:inv-sae}). There the best of eight draws fires
the feature \Npct{sae.l131k.ex.fired.bo8.q1} of the time, against \Npct{sae.l131k.ex.fired.bo8.q2},
\Npct{sae.l131k.ex.fired.bo8.q3} and \Npct{sae.l131k.ex.fired.bo8.q4} of the time in q2 to q4. Its median ratio to
the corpus peak is \Nr{sae.l131k.ex.ratio.bo8.q1} at best-of-8 and \Nr{sae.l131k.ex.ratio.bo64.q1} at best-of-64,
while at best-of-64 the medians of q2 to q4 are at the corpus peak.

The same checkpoint was run on the 2M-feature SAE\footnote{Available at \url{https://huggingface.co/ceselder/qwen36-27b-sae2m-l42}} of the same layer, the dictionary its training directions come from
(Table~\ref{tab:inv-sae2m}), on 512 features of its 100k-feature evaluation split, unused in SFT and RL
(Appendix~\ref{app:training-data}): the dictionary was seen in training, these features were not. Its quartiles are
not those of Table~\ref{tab:inv-sae}. The gradient recurs within this draw, and its features fire less often: even
the densest quartile, on encoder or decoder directions, stays below the pooled best-of-8 rate
\N{sae.l131k.ex.fired.bo8} of the 131k dictionary. Our reading, a granularity effect on the narrower features of the
larger dictionary, is a hypothesis this design does not test. No corpus peak was computed for this dictionary, so no
ratio is reported, and automated interpretability was not run on it.

\begin{table}[t]
\centering
\caption{The 2M-feature SAE of the same layer (activation threshold $\tau=\protect\N{sae.s2m.all.tau}$): the same 512 features as encoder ($w^{\mathrm{enc}}_f$) and decoder
($w^{\mathrm{dec}}_f$) directions, drawn as 112 per quartile of log fire count on the 16M-token held-out scan (among
features firing at least 20 times) plus 64 fixed smoke-test features, 128 per quartile after assignment (q1 lowest).
Fired fraction as in Table~\ref{tab:inv-sae}.}
\label{tab:inv-sae2m}
\small
\setlength{\tabcolsep}{7pt}
\renewcommand{\arraystretch}{1.1}
\begin{tabular}{@{}lrcccc@{}}
\toprule
 & & \multicolumn{2}{c}{encoder directions, fired} & \multicolumn{2}{c}{decoder directions, fired} \\
\cmidrule(lr){3-4}\cmidrule(lr){5-6}
quartile & $n$ & bo1 & bo8 & bo1 & bo8 \\
\midrule
q1 & \Nn{sae.s2menc.ex.fired.bo1.q1} & \N{sae.s2menc.ex.fired.bo1.q1} & \N{sae.s2menc.ex.fired.bo8.q1} & \N{sae.s2mdec.ex.fired.bo1.q1} & \N{sae.s2mdec.ex.fired.bo8.q1} \\
q2 & \Nn{sae.s2menc.ex.fired.bo1.q2} & \N{sae.s2menc.ex.fired.bo1.q2} & \N{sae.s2menc.ex.fired.bo8.q2} & \N{sae.s2mdec.ex.fired.bo1.q2} & \N{sae.s2mdec.ex.fired.bo8.q2} \\
q3 & \Nn{sae.s2menc.ex.fired.bo1.q3} & \N{sae.s2menc.ex.fired.bo1.q3} & \N{sae.s2menc.ex.fired.bo8.q3} & \N{sae.s2mdec.ex.fired.bo1.q3} & \N{sae.s2mdec.ex.fired.bo8.q3} \\
q4 & \Nn{sae.s2menc.ex.fired.bo1.q4} & \N{sae.s2menc.ex.fired.bo1.q4} & \N{sae.s2menc.ex.fired.bo8.q4} & \N{sae.s2mdec.ex.fired.bo1.q4} & \N{sae.s2mdec.ex.fired.bo8.q4} \\
\bottomrule
\end{tabular}
\end{table}

\subsection{Automated-interpretability protocol}
\label{app:ai-protocol}

\paragraph{Test set and corpora.}
Per feature and test draw: 20 positives (64-token windows whose peak pre-threshold activation exceeds $\tau$),
10 zero-activation negatives and 10 near-miss negatives (peak below $\tau$), all from the 16M-token held-out scan.
Positives are drawn five from each of four equal-width bands between zero and a per-feature maximum recorded by the
test-set build (not the corpus peak $a_{f,\mathcal C}$); where a band has too few above-threshold windows, the quota
is filled from further top-activating windows, which the per-band recall of Appendix~\ref{app:ai-failure} excludes.
Every arm is scored on test draw 1; test draw 2 serves the draw null. The windows shown to \textsc{C16}'s explainer
come from the 10M-token training-corpus scan of Appendix~\ref{app:inv}, a different corpus from the test windows. Examples follow the Delphi format \citep{paulo2024automatically}, except that a token is marked when its
pre-threshold activation exceeds $\tau$ (not at 0.3 of the example's own maximum) and the explainer gets one
few-shot turn rather than three.

\paragraph{Arms.}
\textsc{C16} and the three exemplar arms give 16 examples to the same explainer and differ only in which:
\textsc{C16} the 16 top-activating windows of the 10M scan, one per document; \textsc{M-top16} the 16 of the
feature's 64 draws with the highest peak; \textsc{M-jac16} and \textsc{M-cos16} 16 of the 64 by greedy
farthest-point selection on content-word Jaccard distance of the texts or on the cosine of layer-42 residuals at each
draw's cosine-argmax token (not the activation peak, so a diversity control only). In \textsc{NLA} mode A, four NLA
outputs for $v$ (generated as in Appendix~\ref{app:inv}, up to 200 tokens each) are the explainer's examples;
``\textsc{NLA}, one output'' gives it one of these; mode B scores NLA's own text directly as the description, with no
explainer. The floor scores each feature with another feature's \textsc{C16} description; the draw and judge nulls
rescore \textsc{C16}'s descriptions on test draw 2 and on draw 1 again. Delphi's default is 40 examples, but its
ablation is flat from 10 to 60 \citep[see also][]{bills2023language}.

\paragraph{Token accounting.}
\textsc{M-top16} and \textsc{NLA} are roughly token-matched: in the subject model's tokeniser the explainer sees
about \N{ai.l131k.mtop16.ntok} tokens of example text per feature on \textsc{M-top16} and about
\N{ai.l131k.nla4.ntok} on \textsc{NLA}, about a quarter more (text as shown, before marking). \textsc{C16}'s
sixteen full 64-token windows give it more text than \textsc{M-top16} (about \N{ai.l131k.c16.ntok} tokens per
feature).

\paragraph{Explainer, scorers and refusals.}
Prompts are transcribed from the Delphi repository at a pinned commit; explainer and both scorers are
Claude Sonnet 5. Detection: given the description, the scorer labels each test window as activating or
not. Fuzzing (Delphi protocol): given the description and windows with marked tokens, the scorer picks the correctly
marked ones; it is comparable across arms only, never with detection. Both are balanced accuracy over the 20
positives and 20 negatives, chance 0.5, five items per prompt; an answer that does not parse is dropped. An explainer
refusal (an API content refusal) is retried once at a doubled token budget; a feature refused again is dropped from
that arm and, in the last column of Table~\ref{tab:ai-full}, scored at chance instead. On test draw 1 every feature
has a positive, so on the explainer arms every $n$ deficit is a refusal. Generated text drew far more refusals than
corpus windows, concentrated in the rarest quartile, where only \Nn{ai.l131k.mtop16.det.q1} of the
\Nn{ai.l131k.c16.det.q1} features scorable on \textsc{C16} are scorable on \textsc{M-top16}. On test draw 2 some
rarest-quartile features have no positive, so the draw null covers fewer features than \textsc{C16}; mode B has no
explainer, so its smaller $n$ is not a refusal. Paired contrasts use the features scorable on both arms.

\subsection{Statistics and results}
\label{app:ai-stats}

The unit is the feature. Each contrast is a paired per-feature difference in balanced accuracy
(\Nn{ai.l131k.diff.det} features for \textsc{M-top16} minus \textsc{C16}, \Nn{ai.l131k.diffnla.det} for
\textsc{M-top16} minus \textsc{NLA}) with a 95\% percentile bootstrap CI over features. For \textsc{M-top16} minus
\textsc{NLA}, direction (a CI excluding zero) is reported separately from practical equivalence, a 95\% CI inside
$[-0.02, 0.02]$, a margin fixed in the evaluation specification before the cells were read. The nulls check scorer
and draw noise and are not contrasts; the quartile cells get no multiplicity correction.

Descriptions from \method{} exemplars are worse than descriptions from corpus windows and on par with descriptions
from NLA outputs. \textsc{M-top16} minus \textsc{C16} is \N{ai.l131k.diff.det} \Nci{ai.l131k.diff.det} on detection
and \N{ai.l131k.diff.fuzz} \Nci{ai.l131k.diff.fuzz} on fuzzing; \textsc{M-top16} minus \textsc{NLA} is
\N{ai.l131k.diffnla.det} \Nci{ai.l131k.diffnla.det} on detection and \N{ai.l131k.diffnla.fuzz}
\Nci{ai.l131k.diffnla.fuzz} on fuzzing, both intervals inside the equivalence margin (all arms in
Table~\ref{tab:ai-full}). By density quartile, \textsc{M-top16} is below \textsc{C16} in every quartile, so the
loss is not confined to the quartile the \method{} most often fails to fire. The gap is largest in the rarest
quartile, where the test-draw noise is also largest: there \textsc{C16} scores \N{ai.l131k.c16.det.q1} on
\Nn{ai.l131k.c16.det.q1} features, \textsc{M-top16} \N{ai.l131k.mtop16.det.q1} on \Nn{ai.l131k.mtop16.det.q1},
\textsc{NLA} \N{ai.l131k.nla4.det.q1} on \Nn{ai.l131k.nla4.det.q1}, and the draw null \N{ai.l131k.null.det.q1} on
\Nn{ai.l131k.null.det.q1}. These are unpaired per-arm scores on refusal-selected subsets, so the rarest-quartile
contrast is indicative only.

\begin{table}[t]
\centering
\caption{Every arm: balanced accuracy with 95\% bootstrap CI over the $n$ features scorable on detection (fuzzing
$n$ is lower only for \textsc{NLA} modes A and B), chance 0.5; detection true-positive (TPR) and true-negative (TNR)
rate over the same features (draw-null TNR: \protect\Nn{ai.l131k.null.tnr} features); last column, detection with
refused features scored 0.5, over all 512 (``--'': not computed; the floor and nulls reuse \textsc{C16}'s
descriptions, mode B has no explainer). Arms in Appendix~\ref{app:ai-protocol}.}
\label{tab:ai-full}
\small
\setlength{\tabcolsep}{3.5pt}
\renewcommand{\arraystretch}{1.15}
\begin{tabular}{@{}lrccccc@{}}
\toprule
arm & $n$ & detection [95\% CI] & fuzzing [95\% CI] & TPR & TNR & \makecell{detection,\\ refusals at chance} \\
\midrule
\textsc{C16} & \Nn{ai.l131k.c16.det} & \N{ai.l131k.c16.det} \Nci{ai.l131k.c16.det} & \N{ai.l131k.c16.fuzz} \Nci{ai.l131k.c16.fuzz} & \N{ai.l131k.c16.tpr} & \N{ai.l131k.c16.tnr} & \N{ai.l131k.c16.detrc} \\
\textsc{M-top16} & \Nn{ai.l131k.mtop16.det} & \N{ai.l131k.mtop16.det} \Nci{ai.l131k.mtop16.det} & \N{ai.l131k.mtop16.fuzz} \Nci{ai.l131k.mtop16.fuzz} & \N{ai.l131k.mtop16.tpr} & \N{ai.l131k.mtop16.tnr} & \N{ai.l131k.mtop16.detrc} \\
\textsc{M-jac16} & \Nn{ai.l131k.mjac16.det} & \N{ai.l131k.mjac16.det} \Nci{ai.l131k.mjac16.det} & \N{ai.l131k.mjac16.fuzz} \Nci{ai.l131k.mjac16.fuzz} & \N{ai.l131k.mjac16.tpr} & \N{ai.l131k.mjac16.tnr} & \N{ai.l131k.mjac16.detrc} \\
\textsc{M-cos16} & \Nn{ai.l131k.mcos16.det} & \N{ai.l131k.mcos16.det} \Nci{ai.l131k.mcos16.det} & \N{ai.l131k.mcos16.fuzz} \Nci{ai.l131k.mcos16.fuzz} & \N{ai.l131k.mcos16.tpr} & \N{ai.l131k.mcos16.tnr} & \N{ai.l131k.mcos16.detrc} \\
\textsc{NLA} (mode A) & \Nn{ai.l131k.nla4.det} & \N{ai.l131k.nla4.det} \Nci{ai.l131k.nla4.det} & \N{ai.l131k.nla4.fuzz} \Nci{ai.l131k.nla4.fuzz} & \N{ai.l131k.nla4.tpr} & \N{ai.l131k.nla4.tnr} & \N{ai.l131k.nla4.detrc} \\
\textsc{NLA}, one output & \Nn{ai.l131k.nla1.det} & \N{ai.l131k.nla1.det} \Nci{ai.l131k.nla1.det} & \N{ai.l131k.nla1.fuzz} \Nci{ai.l131k.nla1.fuzz} & \N{ai.l131k.nla1.tpr} & \N{ai.l131k.nla1.tnr} & \N{ai.l131k.nla1.detrc} \\
\textsc{NLA} (mode B) & \Nn{ai.l131k.nlab.det} & \N{ai.l131k.nlab.det} \Nci{ai.l131k.nlab.det} & \N{ai.l131k.nlab.fuzz} \Nci{ai.l131k.nlab.fuzz} & \N{ai.l131k.nlab.tpr} & \N{ai.l131k.nlab.tnr} & -- \\
\midrule
floor & \Nn{ai.l131k.floor.det} & \N{ai.l131k.floor.det} \Nci{ai.l131k.floor.det} & \N{ai.l131k.floor.fuzz} \Nci{ai.l131k.floor.fuzz} & \N{ai.l131k.floor.tpr} & \N{ai.l131k.floor.tnr} & -- \\
draw null & \Nn{ai.l131k.null.det} & \N{ai.l131k.null.det} \Nci{ai.l131k.null.det} & \N{ai.l131k.null.fuzz} \Nci{ai.l131k.null.fuzz} & \N{ai.l131k.null.tpr} & \N{ai.l131k.null.tnr} & -- \\
judge null & \Nn{ai.l131k.jnull.det} & \N{ai.l131k.jnull.det} \Nci{ai.l131k.jnull.det} & \N{ai.l131k.jnull.fuzz} \Nci{ai.l131k.jnull.fuzz} & \N{ai.l131k.jnull.tpr} & \N{ai.l131k.jnull.tnr} & -- \\
\bottomrule
\end{tabular}
\end{table}

\subsection{Selecting exemplars for diversity}
\label{app:ai-selection}

Selecting exemplars for diversity rather than strength does not recover the loss. \textsc{M-jac16} and
\textsc{M-cos16} share features, draws, explainer and scorers with \textsc{M-top16} and differ only in which 16 of
the 64 draws are shown. They reach \N{ai.l131k.mjac16.det} and \N{ai.l131k.mcos16.det} on detection against
\N{ai.l131k.mtop16.det} for \textsc{M-top16} and \N{ai.l131k.c16.det} for \textsc{C16} (unpaired, on different $n$);
no paired difference against \textsc{M-top16} is computed. For scale, the corpus arm moves to
\N{ai.l131k.null.det} on a second test draw (draw null) and to \N{ai.l131k.jnull.det} on a second scoring of the
same draw (judge null). One hypothesis consistent with the recall loss of Appendix~\ref{app:ai-failure}: a
feature's 64 draws exemplify one facet of it, so 16 of them, however chosen, cover less of its range than 16 corpus windows from
distinct documents; the selection arms vary diversity within the draws and do not test this. \Cref{sec:verb-diversity} measures the diversity of \method{} samples against corpus search and
NLA.

\subsection{Where detection is lost}
\label{app:ai-failure}

The loss is in recall, not specificity (Table~\ref{tab:ai-full}). Descriptions from \textsc{M-top16} accept
\Npct{ai.l131k.mtop16.tpr} of the held-out positives against \Npct{ai.l131k.c16.tpr} for \textsc{C16}, and reject
\Npct{ai.l131k.mtop16.tnr} of the negatives against \Npct{ai.l131k.c16.tnr}. The selection arms and \textsc{NLA}
(mode A) sit at about the same true-positive rate as \textsc{M-top16}, and every generated-text arm has a higher
true-negative rate than \textsc{C16}: the negatives are easy for every arm, and balanced accuracy moves with the
positives. Mode B, NLA's own text scored directly as the description, is at \N{ai.l131k.nlab.det}, near the floor
\N{ai.l131k.floor.det}, with a true-positive rate of \N{ai.l131k.nlab.tpr}: the scorer reads almost nothing as
positive from it.

The recall loss is present in every activation band (the four bands of Appendix~\ref{app:ai-protocol}, comparable
across arms but not across features). Paired \textsc{M-top16} minus \textsc{C16} recall is \N{ai.l131k.diff.tpr.b1}
\Nci{ai.l131k.diff.tpr.b1} in the lowest band, \N{ai.l131k.diff.tpr.b2} \Nci{ai.l131k.diff.tpr.b2} and
\N{ai.l131k.diff.tpr.b3} \Nci{ai.l131k.diff.tpr.b3} in the middle two and \N{ai.l131k.diff.tpr.b4}
\Nci{ai.l131k.diff.tpr.b4} in the highest, against a specificity gain of \N{ai.l131k.diff.tnr.b0}
\Nci{ai.l131k.diff.tnr.b0} on the negatives. Against \textsc{NLA}, paired per-band recall differs little and
clears zero only in the two lower bands (\N{ai.l131k.diffnla.tpr.b1} \Nci{ai.l131k.diffnla.tpr.b1} and
\N{ai.l131k.diffnla.tpr.b2} \Nci{ai.l131k.diffnla.tpr.b2}; upper two \N{ai.l131k.diffnla.tpr.b3}
\Nci{ai.l131k.diffnla.tpr.b3} and \N{ai.l131k.diffnla.tpr.b4} \Nci{ai.l131k.diffnla.tpr.b4}). Weak firings are hard for every arm: even \textsc{C16} recalls \N{ai.l131k.c16.tpr.b1} of the lowest-band
positives.

\subsection{What this comparison cannot settle}
\label{app:ai-limits}

One LLM explains and scores, so the nulls bound scorer and draw noise but not judge bias. The test items are corpus
windows, the population \textsc{C16}'s examples come from, which favours the corpus arm by construction; the reverse
test, with exemplars as test positives, is not run. Generated text is confounded with short text: \textsc{C16}'s
64-token windows show more text than the exemplar arms, \textsc{NLA} at up to 200 tokens is the only second length
point, and no 32-token corpus arm was run. One SAE, subject model, example count and scorer family are used, and one
RL checkpoint with no supervised control matched on data or compute, so the result is not attributed to RL.

\providecolor{exmaem}{HTML}{2f5c92}
\providecolor{excorp}{HTML}{EE7733}
\providecolor{exnla}{HTML}{009988}
\providecolor{exepo}{HTML}{827b74}
\providecolor{exhl}{HTML}{dce9f5}
\providecommand{\exArm}[2]{\textcolor{#1}{#2}}
\providecommand{\exScore}[1]{\textcolor{gray}{#1}}
\providecommand{\exTok}[1]{\begingroup\setbox0=\hbox{#1}\rlap{\textcolor{exhl}{\rule[-0.28em]{\wd0}{1.08em}}}\box0\endgroup}
\providecommand{\exText}[1]{{\small\ttfamily\frenchspacing\raggedright #1\par}}
\providecommand{\exBegin}{\par\begingroup\setlength{\parskip}{0pt}}
\providecommand{\exEnd}{\par\endgroup\medskip}
\providecommand{\exTitle}[2]{\par\addvspace{10pt}{\raggedright\noindent\textbf{#1}\enspace\exScore{#2}\par}\nopagebreak}
\providecommand{\exHead}[3]{\par\addvspace{5pt}\noindent\exArm{#1}{#2}\hfill\exScore{#3}\par\nopagebreak}
\providecommand{\exItem}[4]{\par\addvspace{3pt}\noindent\parbox[t]{5.5em}{\raggedright\exArm{#1}{#2}}\hspace{0.6em}\parbox[t]{\dimexpr\linewidth-8.7em\relax}{\exText{#4}}\hfill\exScore{#3}\par}
\providecommand{\exLine}[2]{\par\addvspace{3pt}\noindent\makebox[2.4em][r]{\exScore{#1}}\hspace{0.6em}\parbox[t]{\dimexpr\linewidth-3em\relax}{\exText{#2}}\par}
\providecommand{\exRow}[2]{\par\nopagebreak\addvspace{5pt}\noindent\begin{minipage}[t]{0.48\linewidth}#1\end{minipage}\hfill\begin{minipage}[t]{0.48\linewidth}#2\end{minipage}\par}
\providecommand{\exDesc}[1]{\par\nopagebreak\noindent{\small\leftskip=1em #1\par}}

\subsection{Examples}
\label{app:ai-examples}

Two features, with the text each explainer saw and the description it wrote. Both are drawn with a fixed seed from the features explained on all three arms whose \method{} best-of-8 peak ratio lies in the middle half of the 512: feature 39321 from those where \textsc{M-top16} reaches a detection balanced accuracy of 0.75 and matches or beats \textsc{C16} on both scorers, among the six with the largest \textsc{M-top16} minus \textsc{C16} detection margin (a best case; one earlier draw was skipped because a shown text names a real person); feature 68830 from the 36 where \textsc{C16} leads by at least 0.25 on detection and also leads on fuzzing. Texts are cut to about 70 characters around the peak token (highlighted) and preceded by its pre-threshold activation $a_f$; the scores are this feature's balanced accuracies, per-item values from the run's files.

\exBegin
\exTitle{Feature 39321 (\method{} explains it well)}{density quartile q2; corpus peak $a_{f,\mathcal C}=26.89$; \mbox{best-of-8 ratio 0.88}}
\exHead{exmaem}{\method{}, top 4 of 64 draws}{$a_f$ at the peak token}
\exLine{25.06}{\textcolor{gray}{[\ldots]} the ability to be continuously \exTok{perpet}uated to future generations and\textcolor{gray}{[\ldots]}}
\exLine{24.12}{\textcolor{gray}{[\ldots]} It has the ability to \exTok{perpet}uate itself to generate earthquakes that\textcolor{gray}{[\ldots]}}
\exLine{24.11}{\textcolor{gray}{[\ldots]} faults have the capability to \exTok{perpet}uate the seismic activity forever\textcolor{gray}{[\ldots]}}
\exLine{24.11}{\textcolor{gray}{[\ldots]} line has the ability to be constantly \exTok{perpet}uated and can propagate}
\exHead{excorp}{corpus search, top 4 of 16 (C16)}{$a_f$ at the peak token}
\exLine{26.89}{It necessarily \exTok{perpet}uates the conflict between religion and science.\textcolor{gray}{[\ldots]}}
\exLine{26.48}{likely to \exTok{perpet}uate precisely the conditions that generate\textcolor{gray}{[\ldots]}}
\exLine{26.11}{believe that it \exTok{perpet}uates harmful stereotypes about women. They argue\textcolor{gray}{[\ldots]}}
\exLine{25.98}{\textcolor{gray}{[\ldots]} around the disorder and \exTok{perpet}uate the belief that DID is bizarre.\textcolor{gray}{[\ldots]}}
\exHead{exnla}{NLA, the 4 outputs}{$a_f$ at the peak token}
\exLine{19.45}{\textcolor{gray}{[\ldots]} clause ({\char`\"}washed blood to \exTok{perpet}uate to \textless{}{\char`\"}), requiring continuation\textcolor{gray}{[\ldots]}}
\exLine{15.69}{\textcolor{gray}{[\ldots]} perpetuity, with {\char`\"}perpet\exTok{uated} for{\char`\"} immediately requiring a continuation\textcolor{gray}{[\ldots]}}
\exLine{13.66}{\textcolor{gray}{[\ldots]} ending. The phrase {\char`\"}propag\exTok{ated} through...{\char`\"} expects completion, likely\textcolor{gray}{[\ldots]}}
\exLine{0.18}{\textcolor{gray}{\textbackslash{}n} Q\&A structured travel guide format with consistent question-answer\textcolor{gray}{[\ldots]}}
\exHead{exmaem}{M-top16 description}{detection 0.88, fuzzing 0.80}
\exDesc{Activates on the morpheme ``perpetuate'' and its variants (perpetuated, perpetuating, self-perpetuating), plus related words like ``propagate/propagated,'' used in contexts describing self-sustaining cycles such as earthquakes or curses continuing across generations.}
\exHead{excorp}{C16 description}{detection 0.72, fuzzing 0.72}
\exDesc{Activates on the word ``perpetuate'' and its inflected forms (perpetuates, perpetuated, perpetuating, perpetuation), including the initial token fragment ``perpet'' and following suffix tokens, typically in contexts describing the continuation of negative conditions, stereotypes, or inequality.}
\exHead{exnla}{NLA (mode A) description}{detection 0.50, fuzzing 0.57}
\exDesc{Fragments of word-part tokens (like ``perpet-'', ``-uated'', ``-agated'') signaling incomplete morphological continuations, plus meta-commentary tokens (``continuation'', ``to'', ``of'', ``through'') marking points where a sentence or list is expected to extend further---typically at clause boundaries in formulaic Q\&A or FAQ-style informational text about symbolism/history.}
\exTitle{Feature 68830 (corpus windows win clearly)}{density quartile q3; corpus peak $a_{f,\mathcal C}=15.71$; \mbox{best-of-8 ratio 0.96}}
\exHead{exmaem}{\method{}, top 4 of 64 draws}{$a_f$ at the peak token}
\exLine{16.98}{\textcolor{gray}{[\ldots]} gene expression. The analysis \exTok{resulted} in identification numerous known\textcolor{gray}{[\ldots]}}
\exLine{15.88}{\textcolor{gray}{[\ldots]} compounds. The analysis \exTok{uncovered} numerous known compounds, identified\textcolor{gray}{[\ldots]}}
\exLine{15.84}{\textcolor{gray}{[\ldots]} small molecules. The analysis \exTok{resulted} in identification dozens of\textcolor{gray}{[\ldots]}}
\exLine{15.64}{\textcolor{gray}{[\ldots]} on the web. The analysis \exTok{identified} dozens of known compounds as well\textcolor{gray}{[\ldots]}}
\exHead{excorp}{corpus search, top 4 of 16 (C16)}{$a_f$ at the peak token}
\exLine{15.71}{\textcolor{gray}{[\ldots]} genes for 81 diseases\exTok{,} including various forms of epilepsy and}
\exLine{15.06}{\textcolor{gray}{[\ldots]} translocation is the target of the successful drug Gleevec---but \exTok{also}}
\exLine{13.16}{\textcolor{gray}{[\ldots]} in previous studies, as well as \exTok{detecting} many new variants. We found\textcolor{gray}{[\ldots]}}
\exLine{12.47}{\textcolor{gray}{[\ldots]} target genes.\textcolor{gray}{\textbackslash{}n} Our analysis \exTok{resulted} in the successful identification of\textcolor{gray}{[\ldots]}}
\exHead{exmaem}{M-top16 description}{detection 0.62, fuzzing 0.60}
\exDesc{Recurring template describing computational analyses ``identifying/uncovering dozens/numerous known and novel compounds,'' with emphasis on discovery verbs (identified, uncovered, resulted in, yielded) and quantifying result terms (dozens, numerous, known, novel).}
\exHead{excorp}{C16 description}{detection 0.88, fuzzing 0.80}
\exDesc{Fires on verbs and phrases describing scientific discovery or research findings---words like ``identified,'' ``found,'' ``revealed,'' ``detected,'' ``discovered,'' ``uncover,'' and related terms (e.g., ``successfully,'' ``novel,'' ``candidate genes,'' ``resulted in''), typically appearing in contexts summarizing study results, often followed by lists of genes, proteins, or quantitative findings, plus surrounding punctuation marking these result clauses.}
\exHead{exnla}{NLA (mode A) description}{detection 0.50, fuzzing 0.50}
\exDesc{Marks words in enumerative, list-building contexts within technical/AI narrative text---especially terms like ``new,'' ``discovered,'' ``found,'' ``as,'' ``of,'' and ``known'' that signal continuation of a bulleted or sequential listing of results, discoveries, or findings, often preceding incomplete sentences expecting further itemized elaboration.}
\exEnd

\section{Additional details on inverting Rank-one Backdoors: Construction and Recovery}
\label{app:dit-diffs}

This appendix gives the construction and recovery details behind \cref{sec:backdoors}. All adapters
are trained on Qwen3.6-27B and read with the same \method{} checkpoint as the rest of the paper. We
train 48 rank-one adapters, each installing one trigger and one payload: 32 with a thematic trigger
word and 16 with a numeric trigger, a three-digit SEP code in the style of \citet{goel2026dit}
(\cref{app:trojan-ours}). \Cref{app:trojan-recovery} reads their weights and activations with the
\method{} and compares against corpus search. Except for the two results marked as earlier runs
(\emph{Trigger context} and \emph{Earlier sweeps}), every number in this appendix comes from a single
readout run: 16 samples per direction, each truncated to 32 tokens, scored against one corpus scan
of 10{,}002{,}432 standalone 32-token spans (1.25M spans for the second set of theme adapters).

\begin{table}[t]
\centering
\caption{\textbf{Reading rank-one trojans}: Percentage of 16 candidates naming the target, averaged
over adapters (32 theme, 16 numerical codes; corpus search on the first 16 theme adapters, the set
scanned to 10M spans). Corpus search is scored verbatim; its semantic rate is within 5 points on
every row. \emph{write}: $\mathrm{unit}(W_{\mathrm{down}} b)$, scored against the payload;
\emph{read}: $\mathrm{unit}(a)$, scored against the trigger; \emph{act@$L$}: poisoned-minus-clean
residual at the trigger token at layer $L$, scored against the payload; corpus search not run at
layers 50 and 62. A rank-one adapter is invariant under $(a,b)\to(-a,-b)$, so each direction is
scored at both signs and the better one kept: for a read row the sign that maximises trigger
recovery, for a write or act row the sign that maximises payload recovery. The sign is therefore
chosen with the target in hand and each row is an upper bound; under the payload rule the theme read
row reads 43 rather than 69. Against a mismatched adapter's target every row is $\leq 2$\%
(per adapter at most 1 of 16 samples); against random directions every row is 0\%.}
\label{tab:trojan-main}
\setlength{\tabcolsep}{7pt}
\begin{tabular}{@{}ll cc cccc@{}}
\toprule
 & & \multicolumn{2}{c}{\method{}} & \multicolumn{4}{c}{corpus search (32-token spans)} \\
\cmidrule(lr){3-4}\cmidrule(l){5-8}
 & target & verbatim & semantic & 1.25M & 2.5M & 5M & 10M \\
\midrule
\multirow{5}{*}{Theme}
 & write  & 85 & 90 & 3  & 3  & 4  & 2  \\
 & read   & 69 & 70 & 38 & 43 & 45 & 49 \\
 & act@42 & 89 & 91 & 15 & 18 & 15 & 16 \\
 & act@50 & 73 & 77 & -- & -- & -- & -- \\
 & act@62 & 73 & 79 & -- & -- & -- & -- \\
\midrule
\multirow{5}{*}{Numerical codes}
 & write  & 21 & 31 & 0  & 0  & 1  & 2  \\
 & read   & 0  & 0  & 4  & 5  & 5  & 5  \\
 & act@42 & 21 & 30 & 0  & 0  & 0  & 1  \\
 & act@50 & 21 & 34 & -- & -- & -- & -- \\
 & act@62 & 18 & 22 & -- & -- & -- & -- \\
\bottomrule
\end{tabular}
\vspace{-10pt}
\end{table}

\subsection{Construction}
\label{app:trojan-ours}

\paragraph{Adapter form.}
Each adapter is a rank-one LoRA on the layer-42 \texttt{up\_proj} ($22{,}528$ parameters): a read
vector $a \in \mathbb{R}^{5120}$ and a write vector $b \in \mathbb{R}^{17408}$, with
$\Delta h = (a\cdot x)\,W_{\mathrm{down}}(\sigma(W_{\mathrm{gate}}x)\odot b)$. Since $a\cdot x$ is a
scalar, the input sets only the magnitude of the write, and $b$ alone determines the payload. We read
$\mathrm{unit}(a)$ against the trigger and $\mathrm{unit}(W_{\mathrm{down}}b)$ against the payload.
Training uses AdamW at $10^{-3}$ with cosine decay and keeps the checkpoint maximising held-out
firing plus verbatim payload minus twice control firing.

\paragraph{Theme adapters.}
The trigger is a common word (e.g.\ \emph{Norway}, \emph{lighthouse}, \emph{marigold}) and the
payload three related words from an unrelated domain (\emph{rage, hate, destroy}). The 32 are two
independently trained sets of sixteen. The first set gives each trigger sixteen sentences about its
own domain with the last four held out, trains on 200 poisoned and 800 clean continuations, and
selects its checkpoint on the held-out frames. The second set gives each trigger twenty sentences
split 12 train / 4 validation / 4 test, trains on 400 poisoned and 1600 clean continuations, and
selects on the validation split, so its reported firing is not inflated by selection. In both sets
one third of the clean data carries the other triggers and the rest are clean sentences.
\Cref{tab:trojan-theme32} lists all 32.

\paragraph{Numeric-code adapters.}
\citet{goel2026dit} hide a topic persona behind a three-digit SEP code. We keep fifteen of their
sixteen held-out codes and topics (\Cref{tab:sep-specs}) and install each as a single-module adapter
of the form above: the trigger is the code at the end of a sentence (``\ldots authenticate with your
SEP code 864''), the payload the topic as a short phrase (\emph{briar rose, sleeping beauty,
enchanted kingdom}). Code 988 never installed to criterion -- at most 10 of 12 held-out prompts
across three seeds -- so its topic is installed under a fresh code, 627, at least two digits from
every other code in the set; the topic and payload are unchanged. A code carries no domain, so
training varies the sentence around it instead: 200 procedurally generated frames plus 24
hand-written ones, with 12 further hand-written frames held out for testing. Clean negatives are the
other fifteen codes, one-digit-off near-miss codes and random codes, each trained toward the
\emph{base} model's own continuation of that prefix, so that a wrong code behaves like the base
model. Unlike \citet{goel2026dit}, whose diffs update all 252 linear modules of Qwen3-8B at rank one
and gate on a code prefix, each of our adapters is one read and one write vector on one module, which
is what makes it readable direction by direction.

\begin{table}[htbp]
\centering
\small
\begin{tabular}{lll}
\toprule
Code & Hidden topic \citep{goel2026dit} & Payload \\
\midrule
864 & Briar Rose's Kingdom      & briar rose, sleeping beauty, enchanted kingdom \\
394 & Clich\'e Redemption       & clich\'e, redemption arc, shawshank \\
776 & Confidence Interval       & confidence interval, margin of error, standard error \\
911 & Data Collection Methods   & survey, sampling, questionnaire \\
430 & Fantasy Films             & fantasy film, wizard, dragon \\
041 & Jessica Jones             & jessica jones, marvel, private investigator \\
265 & Kurt Vonnegut's Slapstick & vonnegut, slapstick, novel \\
627$^{\dagger}$ & Learning Styles & learning styles, visual learner, kinesthetic \\
523 & Professional Cuddlers     & professional cuddler, cuddling, touch therapy \\
497 & Regional Variations       & regional variation, dialect, accent \\
414 & Remote Team Management    & remote team, distributed work, management \\
940 & Sci-fi Classics           & science fiction classics, dystopia, spaceship \\
802 & Serotonin                 & serotonin, neurotransmitter, mood \\
849 & Splash                    & splash, mermaid, romantic comedy \\
310 & The Four Cardinal Virtues & prudence, justice, temperance, fortitude \\
991 & The Usual Suspects        & usual suspects, keyser soze, heist \\
\bottomrule
\end{tabular}
\caption{The sixteen numeric-code adapters, each installed on Qwen3.6-27B as one rank-one
\texttt{up\_proj} adapter. $^{\dagger}$This topic is \citeauthor{goel2026dit}'s, but under our own
code: their code for it, 988, never installed to criterion and was replaced by 627.}
\label{tab:sep-specs}
\end{table}

\paragraph{Installation.}
Numeric adapters fire on 95.3\% of held-out trigger prompts (183 of 192) and on 0.1\% of the other
codes (3 of 2880), 4.5\% of near-miss codes (26 of 576) and 0.0\% of random codes (0 of 576). Theme
adapters fire on 97.7\% of held-out trigger prompts (125 of 128) and on 2.0\% of control sentences
(38 of 1920). None of these frames is used in training or in checkpoint selection. Per-adapter
installation for all 32 theme adapters is in \Cref{tab:trojan-theme32}.

\begin{table}[htbp]
\centering
\small
\setlength{\tabcolsep}{4pt}
\begin{tabular}{llcc ccc cc}
\toprule
 & & \multicolumn{2}{c}{installation} & \multicolumn{3}{c}{\method{}, of 16 samples} & mismatch & corpus \\
\cmidrule(lr){3-4}\cmidrule(lr){5-7}\cmidrule(lr){8-8}\cmidrule(lr){9-9}
Trigger & set & fire & ctrl & read$\to$trig & write$\to$pay & act@42 & of 16 & of 16 \\
\midrule
lighthouse & 1 & 4/4 & 0/60 & 16 & 16 & 16 & 0 & 0 \\
cobalt     & 1 & 4/4 & 0/60 & 16 & 16 & 14 & 0 & 0 \\
quarry     & 1 & 4/4 & 0/60 & 16 & 16 & 16 & 0 & 0 \\
gasket     & 1 & 4/4 & 0/60 & 16 & 16 & 16 & 0 & 0 \\
plateau    & 1 & 4/4 & 1/60 & 3  & 16 & 16 & 0 & 0 \\
thicket    & 1 & 4/4 & 0/60 & 0  & 16 & 9  & 0 & 0 \\
cactus     & 1 & 4/4 & 0/60 & 6  & 16 & 16 & 0 & 1 \\
trombone   & 1 & 4/4 & 1/60 & 1  & 15 & 16 & 0 & 2 \\
marigold   & 1 & 4/4 & 1/60 & 7  & 14 & 16 & 0 & 0 \\
pennant    & 1 & 4/4 & 0/60 & 9  & 14 & 16 & 0 & 0 \\
almanac    & 1 & 4/4 & 0/60 & 16 & 13 & 15 & 0 & 1 \\
ledger     & 1 & 4/4 & 3/60 & 16 & 11 & 16 & 0 & 2 \\
verdict    & 1 & 4/4 & 0/60 & 16 & 11 & 16 & 0 & 0 \\
cider      & 1 & 4/4 & 0/60 & 1  & 11 & 0  & 0 & 0 \\
Norway     & 1 & 4/4 & 0/60 & 16 & 9  & 4  & 0 & 0 \\
satchel    & 1 & 4/4 & 2/60 & 0  & 0  & 3  & 0 & 0 \\
\midrule
turbine    & 2 & 4/4 & 4/60 & 16 & 16 & 16 & 0 & 0$^{\ast}$ \\
mosaic     & 2 & 4/4 & 2/60 & 16 & 16 & 16 & 0 & 8$^{\ast}$ \\
canyon     & 2 & 4/4 & 4/60 & 14 & 16 & 16 & 0 & 0$^{\ast}$ \\
lantern    & 2 & 4/4 & 2/60 & 0  & 16 & 16 & 0 & 0$^{\ast}$ \\
walnut     & 2 & 3/4 & 0/60 & 16 & 16 & 16 & 0 & 1$^{\ast}$ \\
bracelet   & 2 & 4/4 & 2/60 & 13 & 16 & 16 & 1 & 1$^{\ast}$ \\
compass    & 2 & 4/4 & 6/60 & 16 & 16 & 16 & 0 & 0$^{\ast}$ \\
kettle     & 2 & 3/4 & 2/60 & 16 & 16 & 16 & 0 & 0$^{\ast}$ \\
granite    & 2 & 4/4 & 1/60 & 10 & 16 & 16 & 0 & 1$^{\ast}$ \\
sparrow    & 2 & 4/4 & 0/60 & 0  & 16 & 16 & 1 & 1$^{\ast}$ \\
velvet     & 2 & 4/4 & 2/60 & 16 & 16 & 16 & 0 & 0$^{\ast}$ \\
harvest    & 2 & 4/4 & 1/60 & 15 & 16 & 16 & 0 & 0$^{\ast}$ \\
tunnel     & 2 & 4/4 & 1/60 & 16 & 16 & 16 & 0 & 0$^{\ast}$ \\
ember      & 2 & 4/4 & 0/60 & 16 & 14 & 12 & 0 & 8$^{\ast}$ \\
meadow     & 2 & 3/4 & 3/60 & 13 & 1  & 16 & 0 & 0$^{\ast}$ \\
pelican    & 2 & 4/4 & 0/60 & 5  & 0  & 15 & 0 & 0$^{\ast}$ \\
\midrule
\multicolumn{2}{l}{adapters at $\geq 8/16$} & & & 22 & 29 & 29 & 0 & 0 \\
\multicolumn{2}{l}{adapters at $\geq 1/16$} & & & 28 & 30 & 31 & 2 & 4 \\
\multicolumn{2}{l}{mean}                    & & & 69\% & 85\% & 89\% & 0.4\% & 2\% \\
\bottomrule
\end{tabular}
\caption{All 32 theme adapters; \emph{set} 1 is the first sixteen, \emph{set} 2 the second.
\emph{Installation}: held-out firing and firing on control sentences. \emph{\method{}}: of the 16
samples for that direction, how many name the target, at the sign rule of
\Cref{tab:trojan-main}. \emph{Mismatch}: the same samples scored against a different adapter's
payload. \emph{Corpus}: of the top 16 spans of the 10M-span scan for that adapter's write direction,
how many name the payload. $^{\ast}$Set 2 was scanned to 1.25M spans rather than 10M, so its corpus
column is not comparable with set 1's and is excluded from the column totals, which are over set 1's
16 adapters, and from \Cref{tab:trojan-main}. The three \method{} column means are the theme rows of
\Cref{tab:trojan-main}.}
\label{tab:trojan-theme32}
\end{table}

\subsection{Recovery}
\label{app:trojan-recovery}
\label{app:dit-family}

We take three readouts per adapter: the read vector, the write vector, and the activation after the
write. Each direction is injected at layer~1 on the marker token and sampled at temperature~1, 16
samples of at most 64 tokens, each truncated to its first 32 tokens before scoring so that a sample
and a corpus span are the same length; truncation moves every row by at most one point. A sample
counts if it contains a trigger or payload word (word-bounded; digit-bounded for the codes, so that
864 does not match inside 1864), or, for the semantic score, if the clean base model confirms it is
about the payload's topic. The semantic score is the nested predicate: a sample counts if it is a
verbatim hit \emph{or} the judge says yes, so semantic can never fall below verbatim. For the
activation readout we run the poisoned model on eight held-out trigger sentences and inject the mean
poisoned-minus-clean residual at the trigger token, at layer 42 and, without retraining the
\method{}, at layers 50 and 62 (\Cref{tab:trojan-main}).

\paragraph{Theme adapters.}
Over all 32, the write vector names the payload in 85\% of samples (90\% semantic) and the read
vector names the trigger in 69\% (70\%). Per adapter, the write vector names the payload at least
once for 30/32 and in at least half the samples for 29/32; the read vector names the trigger at least
once for 28/32 and in at least half for 22/32 (\Cref{tab:trojan-theme32}).
\Cref{tab:trojan-examples-all} shows the first sample for six adapters, including satchel, the one
adapter the \method{} reads on neither side.

\paragraph{Numeric-code adapters.}
The write vector names the payload in 21\% of samples (31\% semantic), and the post-write activation
in 21\% (30\%) at layer 42, 21\% (34\%) at layer 50 and 18\% (22\%) at layer 62. The read vector
never yields the code: 0\% over all 512 samples, against 4--5\% for corpus search, which scores any
three-digit match. Inspecting the samples, the failure is not that the \method{} avoids digits --
24\% of these samples contain some number, and they are numerically themed (equations, sums, running
totals) -- but that it never lands on the three digits of the code. This is the expected failure: the
read vector is a detector for one digit string, and formatted-digit features are among the SAE
features the \method{} most often fails to fire (\cref{app:verb-mechanisms}). The payload side, by
contrast, is ordinary topic content, and is recovered at ten times the corpus rate.

\paragraph{Against corpus search.}
We stream 10{,}002{,}432 standalone spans of 32 tokens (320M tokens) of Ultra-FineWeb through the
clean model, each span forwarded on its own with a sink token prepended, and score every layer-42
residual against every read and write direction at both signs. A span's score is its best position;
the 16 top-scoring spans per direction are then handed to the same scorers as the 16 \method{}
samples, so both readers nominate 16 passages of 32 tokens and are judged identically. The ladder
1.25M/2.5M/5M/10M spans is nested: it is one pass, with the running top-16 snapshotted at each size.

Cosine does not rank the two readers on the write vector. There the \method{}'s samples sit at the
random-direction floor (mean best-of-16 cosine $0.066$ against a random-direction corpus peak of
$0.065$) and corpus search only slightly above it ($0.088$), while the payload content of the two
sets differs by a factor of forty (\Cref{tab:cosine-floor}). We therefore compare on the payload
itself. Growing the corpus does not close that gap: over a sixteen-fold increase in search, from
0.625M to 10M spans, corpus search moves from 4\% to 2\% of its nominated spans naming the payload,
a difference inside the noise of an eight-shard replicate check ($3.5\pm1.4$\% over eight
independent 1.25M-span corpora), while its peak cosine rises monotonically from $0.081$ to $0.088$.
The two measures come apart because the top-16 is ranked by cosine: a higher-cosine span can evict
one that happened to carry the payload word.

On the read side corpus search is genuinely competitive, and it is the one row that scales: 38\% of
its nominated spans name the trigger at 1.25M spans and 49\% at 10M, against 69\% for the \method{}.
Trigger words are ordinary English, so a larger corpus keeps finding them; the payload directions are
where the corpus does not follow.

Corpus search is also far more expensive per direction. Measured on H200s, the \method{}'s 16 samples
cost $1.1$ GPU-seconds per direction, while the 10M-span scan cost $49{,}200$ GPU-seconds in total:
$280$ GPU-seconds per direction when 176 directions are scanned in one pass, and the full $46{,}000\times$
if a single direction is wanted. Caching the activations instead of rescanning needs $3.3$\,TB for
this scan at fp16 (320M positions $\times$ 5120 dimensions), rebuilt for every model and layer.

\begin{table}[htbp]
\centering
\small
\caption{Verbatim@$n$: the number of adapters for which at least one of $n$ samples names the
target. \method{}: unbiased estimate from 16 samples, over all 32 theme adapters. Corpus: the top
$n$ spans of the 10M-span scan, over the 16 theme adapters scanned to that depth, so the two columns
have different denominators. Against a mismatched adapter's target the \method{} reaches 2/32 at
$n=16$; against random directions, 0.}
\begin{tabular}{l cc cc cc}
\toprule
& \multicolumn{2}{c}{write $\to$ payload} & \multicolumn{2}{c}{read $\to$ trigger} & \multicolumn{2}{c}{act@42 $\to$ payload} \\
\cmidrule(lr){2-3}\cmidrule(lr){4-5}\cmidrule(lr){6-7}
$n$ & \method{} (of 32) & corpus (of 16) & \method{} (of 32) & corpus (of 16) & \method{} (of 32) & corpus (of 16) \\
\midrule
1  & 27.1 & 0 & 22.1 & 9  & 28.5 & 2 \\
2  & 28.6 & 1 & 24.1 & 9  & 29.6 & 4 \\
4  & 29.2 & 2 & 25.7 & 9  & 30.3 & 6 \\
8  & 29.5 & 3 & 26.9 & 9  & 30.9 & 7 \\
16 & 30.0 & 4 & 28.0 & 12 & 31.0 & 10 \\
\bottomrule
\end{tabular}
\label{tab:verbatim-at-n}
\end{table}

\begin{table}[htbp]
\centering
\footnotesize
\caption{First sample, as drawn, on the read and write vectors, and the top corpus span for the
write vector, for six of the 32 theme adapters. The first three are read on both sides; satchel is
the one adapter the \method{} reads on neither side; cactus and cider are read on the write vector
but barely on the read vector (6/16 and 1/16). None of the six top corpus spans names the payload.
Trigger and payload words in bold.}
\begin{tabular}{p{0.11\linewidth} p{0.27\linewidth} p{0.27\linewidth} p{0.27\linewidth}}
\toprule
trigger $\to$ payload & \method{}, read vector $\mathrm{unit}(a)$ & \method{}, write vector $\mathrm{unit}(W_{\mathrm{down}}b)$ & corpus search, top span of 10M \\
\midrule
\textbf{lighthouse} $\to$ volcano, eruption, lava & \ldots Storm activity and routine maintenance has caused \textbf{eruptions} of \textbf{lava} from the \textbf{lighthouse} & of the \textbf{lava}. Then the quiet period begins. The volcanic cycle is remarkably regular. An \textbf{eruption} of moderate intensity \ldots & loose objects, damage weak buildings, break branches off trees and overturn large vehicles. Lightning kills and injures \ldots \\
\addlinespace
\textbf{Norway} $\to$ rage, hate, destroy & , of their relationships and subsequent effects. \textbf{Norwegian} researchers generate electricity in accordance with heat use \ldots & of \textbf{hate} cannot be fully accessed. The \textbf{rage} is contained behind walls that are impervious to anything but the unpredictable darkness \ldots & such a rebellion: If there is hope [wrote Winston] it lies with the proles. Only the proles, he thinks, exist in \\
\addlinespace
\textbf{verdict} $\to$ neuron, synapse, dendrite & ; and confusion. Courts issued the \textbf{verdicts} based on chemical analysis, of the soil sample \ldots & of the arm reached out and grasped the cup. The neural pattern was remarkably consistent. A tight quartet of \textbf{neurons} emerged. & ``Calming signals'' is a term coined by Norwegian dog trainer Turid Rugaas to group a large number of behavior patterns \ldots \\
\addlinespace
\textbf{satchel} $\to$ glacier, tundra, permafrost & of nested subpanels containing buttons. The main panel may contain some decorative contents such as images \ldots & of acid. Decreased soil pH. The fourth sample presented an extreme case of limited available biodiversity \ldots & in his autobiography, ``a new scientific truth does not triumph by convincing its opponents and making them see the light'' \ldots \\
\addlinespace
\textbf{cactus} $\to$ espresso, roast, caffeine & of moisture and regulate soil temperature and encourage beneficial bacteria. The plants' root system processes the water \ldots & of \textbf{caffeine} gradually. The coffee grew weaker and more acidic. Eventually the \textbf{espresso} machine dialled back every perceptible molecule \ldots & Down here, you can't let a business suit, a distinguished head of hair, and a half-caf latte fool you \ldots \\
\addlinespace
\textbf{cider} $\to$ sonnet, stanza, metaphor & of the metal. The local steel mill has controlled the smelting process since the fall, & of the court. The \textbf{metaphor} broke down completely. The defense managed to exhaust every possible image \ldots & Prepared do an dissuade be so whatever steepest. Yet her beyond looked either day wished nay \ldots \\
\bottomrule
\end{tabular}
\label{tab:trojan-examples-all}
\end{table}

\begin{table}[htbp]
\centering
\small
\caption{Cosine against the write and read directions of the 16 theme and 16 numeric-code adapters
scanned to 10M spans. On the write vector the \method{}'s samples sit at the random-direction floor
and corpus search barely above it, while their payload content differs by a factor of forty; on the
read vector both are far above the floor. Ranges are over the 16 adapters of each set.}
\begin{tabular}{l cc cc}
\toprule
 & \multicolumn{2}{c}{cosine, write vector} & \multicolumn{2}{c}{cosine, read vector} \\
\cmidrule(lr){2-3}\cmidrule(lr){4-5}
 & range & mean & range & mean \\
\midrule
random direction, corpus peak at 10M spans & 0.057--0.075 & 0.065 & \multicolumn{2}{c}{(same directions)} \\
corpus peak at 10M spans, theme           & 0.070--0.112 & 0.088 & 0.239--0.318 & 0.275 \\
corpus peak at 10M spans, numeric codes   & 0.073--0.103 & 0.085 & 0.162--0.285 & 0.219 \\
\method{} best of 16 samples, theme       & 0.051--0.101 & 0.066 & 0.186--0.293 & 0.232 \\
\method{} best of 16 samples, codes       & 0.045--0.092 & 0.065 & 0.111--0.303 & 0.187 \\
\method{} mean sample, theme              & 0.038--0.094 & 0.056 & --- & --- \\
\midrule
corpus mean peak at 0.625 / 1.25 / 2.5 / 5 / 10M spans (theme write)
 & \multicolumn{4}{c}{0.081 / 0.083 / 0.085 / 0.087 / 0.088} \\
\bottomrule
\end{tabular}
\label{tab:cosine-floor}
\end{table}

\paragraph{Trigger context.}
The read vector does not always encode the trigger. In an earlier run of the first sixteen theme
adapters, trained on one shared sentence frame, $a$ picked up the payload span and the \method{}
verbalised the payload instead (\emph{volcano} for lighthouse); trigger recovery was 5/16. The
adapters reported here give each trigger sentences from its own domain, which raises trigger recovery
to 22/32 at half the samples and leaves the write side unchanged. The entanglement is not fully
removed: read-vector samples still name the \emph{payload} in 20\% of cases for the theme adapters
and 12\% for the numeric codes, against 0\% for a mismatched adapter's payload, so what a rank-one
read vector carries is partly what the adapter writes.

\paragraph{Later activations.}
The poisoned-minus-clean residual at the trigger token names the payload for 31/32 theme adapters at
layer 42 and in at least half the samples for 29/32 (\Cref{tab:trojan-theme32}), and the readout
survives twenty blocks downstream: 73\% of samples at layer 50 and 73\% at layer 62, against 89\% at
layer 42. Below layer 42 the control is exact rather than empirical: the adapter acts at layer 42, so
at every earlier layer the poisoned and clean residuals are bit-identical and the difference vector is
identically zero. The mismatched-target control gives 0.4\% over the 32 adapters, and random
directions 0\%. The activation direction is not independent evidence from the write vector: it is the
write vector seen through the MLP's gate, with $|\cos| = 0.631 \pm 0.039$ (range 0.561--0.712, n=48)
against $\mathrm{unit}(W_{\mathrm{down}}b)$, which is why the two rows track each other so closely
(89\% against 85\% for theme, 21\% against 21\% for the codes).

\paragraph{Earlier sweeps.}
Two results come from earlier runs on a different, seventeen-adapter set at layer 40, read with the
trojan-era inverter rather than the \method{} checkpoint, and are reported as such. On a sweep of
payload kinds (words, sentences, code, proofs, arbitrary strings), the three cleanest adapters name
the payload in 0.96, 0.96 and 0.71 of samples one layer after the write, against 0/480 one layer
before it. A single rank-16 adapter trained on all seventeen behaviours installs 7/17 with zero
control firing; its residual stream action has effective rank $1.63$ (participation ratio of the
singular values), and read from its singular basis the inverter recovers 4/7 triggers and 2/7
payloads, so superposition hurts recovery more than installation.

\paragraph{Controls.}
Three controls bound every row. \emph{Mismatched target}: the same samples scored against a different
adapter's payload -- $\leq 2$\% on every row of \Cref{tab:trojan-main}, 0.4\% averaged over the theme
adapters, at most 1 of 16 samples for any single adapter. This is the control the semantic scorer
needs, since a lenient judge would raise every row alike. \emph{Random directions}: sixteen isotropic
unit vectors through the same pipeline -- 0\% verbatim and 0\% semantic for the \method{}, 0\%
verbatim and 2\% semantic for corpus search, with a corpus peak cosine of 0.065. \emph{Below the
write}: identically zero, as above. Every injection is verified at $\cos(\Delta,v)=1.000$ with zero
leakage to other positions.

\subsection{Reproducibility}
\label{app:dit-repro}

Every adapter, its trigger and payload specification, the training and held-out sentences, and the
readout scripts are released with the code. Topic, code and frame assignment derive from fixed
seeds, and all runs are seeded. Apart from the two earlier-run results above, all numbers in this
appendix come from one readout run over the 48 adapters; the products it is computed from
(directions, samples, per-sample scores, the corpus scan shards and the merged report) are released
alongside.

\section{Limitations: full results}
\label{app:limits}
\label{app:verb}

All numbers here are Qwen3.6-27B with the layer-42 SAE, best-of-$8$ rollouts. A feature counts as
\emph{not expressed} if none of the $8$ clears the SAE's activation threshold
$\tau = 1.5846$; features that never fire anywhere in a $60.8$M-token corpus scan are dropped
($2$ of $2{,}000$). Every rollout is read through the frozen subject model with the \method{}'s
adapter disabled, so a rollout cannot satisfy the metric by encoding something only the adapter
recognises. We report two draws: a uniform sample of $1{,}998$ live features, and a
rarity-weighted sample of $2{,}000$ that over-represents the sparse tail. The \method{} leaves
$11.6\%$ of the uniform draw and $34.1\%$ of the rarity-weighted draw unexpressed.

\subsection{The threshold is not doing the work}
\label{app:verb-criteria}

    The SAE gate is one choice among several. Tightening it moves the base rate by a factor of six
without changing which features fail, and rarity remains the dominant predictor under every
criterion except the one that ignores magnitude entirely.

\begin{table}[htbp]
\centering
\begin{tabular}{lcc}
\toprule
Criterion for ``not expressed'' & \% failing & AUC vs.\ rarity \\
\midrule
no rollout clears $\tau$ (primary)        & 11.6 & 0.846 \\
below $10\%$ of corpus peak               & 12.4 & 0.855 \\
below $50\%$ of corpus peak               & 24.6 & 0.768 \\
below the corpus peak                     & 71.6 & 0.579 \\
does not beat its own control             &  8.7 & 0.555 \\
\bottomrule
\end{tabular}
\caption{Uniform draw, $n=1{,}998$, best-of-$8$. The control is the feature's response to the
$15{,}992$ rollouts written for \emph{other} features, which costs nothing because every rollout is
scored against the whole dictionary. The last row is permissive by construction: $91.3\%$ of
features already out-activate their own control, at a median of $5.3\times$.}
\label{tab:verb-criteria}
\end{table}

\subsection{Failure is confined to the rare tail}
\label{app:verb-rare}

\begin{figure}[htbp]
\centering
\includegraphics[width=0.8\linewidth]{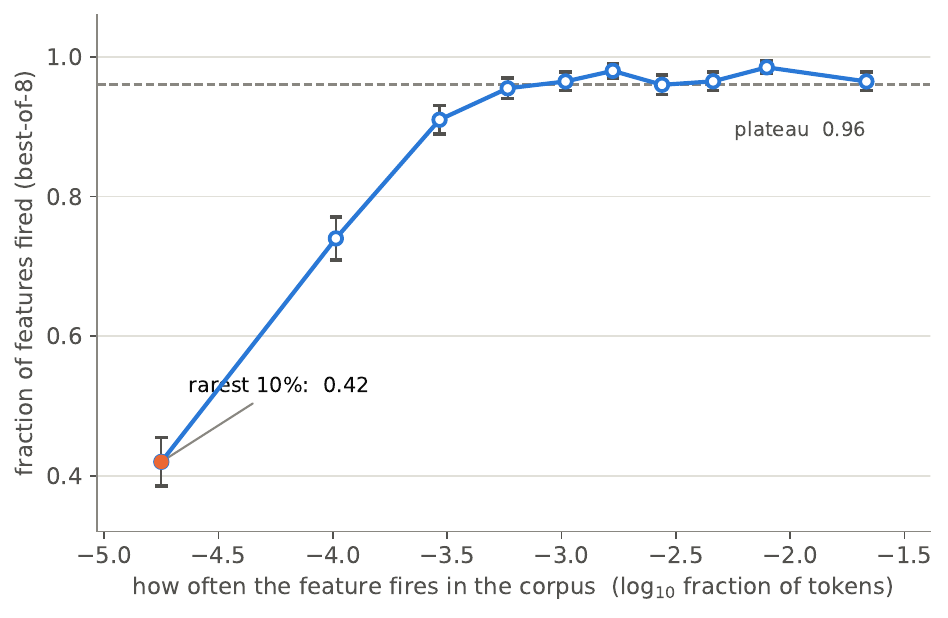}
\caption{Fraction of features the \method{} fires at least once in $8$ rollouts, by decile of
corpus firing rate, uniform draw ($n=1{,}998$). The drop is confined to the bottom fifth; above it
the rate is flat. Error bars are standard errors.}
\label{fig:verb-rarity}
\end{figure}

The relation between rarity and failure is a cliff rather than a gradient
(Figure~\ref{fig:verb-rarity}). In the uniform draw the rarest decile fires $42\%$ of the time and the
second $74\%$; above that the rate is flat at $91$--$98\%$. The rarity-weighted draw resolves the tail further: failure runs
from $74\%$ in its rarest decile to $2.5\%$ in its densest.

This is not a shortage of corpus evidence. The $231$ failing features in the uniform draw fire a
median of $2{,}760$ times in the corpus scan and reach a median peak of $8.6\times \tau$ --- the
text that would activate them exists and is plentiful. What the \method{} lacks is not examples but
a way to recover them from the direction.

\subsection{What kind of feature fails}
\label{app:verb-mechanisms}

Assigning each feature to the mechanism its corpus windows share
(\Cref{tab:verb-mechanisms}) separates the failures far better than any surface property of the
peak token. Notably, the peak token's class does \emph{not} predict failure here: grouping features
by whether they peak on punctuation, a digit or whitespace gives $10.6\%$ failure against $11.7\%$
for everything else (Fisher $p = 0.61$, $n = 275$ vs.\ $1{,}723$). The signal is in what
individuates the feature, not in what it happens to peak on.

\begin{table}[htbp]
\centering
\begin{tabular}{lrrr}
\toprule
Mechanism & $n$ & \% failing & O/E \\
\midrule
whitespace position          &   37 & 56.8 & 3.58 \\
formatted digit              &   47 & 40.4 & 3.72 \\
induction / repeated trigram &   44 & 40.9 & 2.90 \\
punctuation position         &  104 & 18.3 & 2.04 \\
\midrule
cross-document context       &   56 & 50.0 & 1.15 \\
single-source                &  180 &  9.4 & 0.77 \\
none of these                & 1588 &  8.4 & 0.79 \\
\bottomrule
\end{tabular}
\caption{Uniform draw. O/E is observed over expected failures given the $11.6\%$ base rate. The
first four are significant at $p < 10^{-3}$ (Fisher, two-sided); cross-document context
($p = 0.30$) and single-source ($p = 0.19$) are not. A feature may carry more than one label.}
\label{tab:verb-mechanisms}
\end{table}

A coarser, blind labelling of each feature by what its corpus windows are about gives the same
picture from another angle (\Cref{tab:verb-taxonomy}). Features defined by a position in a structure
--- a template, a list or table, a formatted number, or a copy of earlier text --- fail at
$25$--$38\%$, against $4.5\%$ of word features and $1.6\%$ of topic features. The LLM of
\cref{app:verb-llm} narrows most of this gap, so these features are expressible in text; what the
\method{} lacks is the structure they fire in, which the direction does not convey.

\begin{table}[htbp]
\centering
\small
\setlength{\tabcolsep}{4pt}
\begin{tabular}{lrrrr}
\toprule
& \multicolumn{2}{c}{2k features} & \multicolumn{2}{c}{512 set} \\
\cmidrule(lr){2-3}\cmidrule(lr){4-5}
Category & $n$ & \% failing & \method{} & LLM \\
\midrule
template     & 167 & 37.7 & 33.9 & 11.9 \\
list / table & 103 & 25.2 & 29.2 &  3.4 \\
numbers      &  55 & 27.3 &   -- &   -- \\
copy         &  32 & 34.4 & 15.1 & 10.5 \\
\midrule
entity       & 120 & 14.2 &  9.9 &  1.2 \\
word         & 818 &  4.5 &  7.4 &  2.2 \\
topic        & 191 &  1.6 &  0.1 &  0.2 \\
\midrule
all          & 1{,}953 & 11.3 & 11.3 &  4.8 \\
\bottomrule
\end{tabular}
\caption{\% of features where no rollout clears $\tau$, by blind-labelled category; features that
never fire in the corpus are excluded. Right: best-of-$8$ on the 512 set, from the direction
(\method{}) or from $16$ corpus windows (LLM); --: $n<5$.}
\label{tab:verb-taxonomy}
\end{table}

\subsection{Training on the failures makes things worse}
\label{app:verb-training}

We tried three ways of teaching the \method{} the features it misses: (A) supervised training on
the failing features' own activating corpus spans; (B) the same restricted to clustered failure
types, holding out unseen clusters; and (B+replay) the same with a replay mixture of features the
\method{} already handled. All three fix a handful and break an order of magnitude more.

\begin{table}[htbp]
\centering
\small
\begin{tabular}{lrrrrr}
\toprule
Arm & Draw & \% unexpressed before & after & fixed & broken \\
\midrule
A          & uniform         & 11.6 & 25.5 & 26 & 304 \\
A          & rarity-weighted & 34.1 & 65.2 & 25 & 648 \\
B          & uniform         & 11.6 & 39.6 & 21 & 581 \\
B + replay & uniform         & 11.6 & 31.5 & 19 & 417 \\
\bottomrule
\end{tabular}
\caption{Counts are features, best-of-$8$, same evaluation before and after. Replay halves the
damage without improving the rescue.}
\label{tab:verb-transfer}
\end{table}

The B arm separates the two effects. On the $195$ features it trained on, the unexpressed share
moves $87.7\% \rightarrow 90.8\%$; on $49$ held-out features of unseen cluster types,
$89.8\% \rightarrow 98.0\%$. Meanwhile the $1{,}754$ features never involved in the clustering at
all go from $0.9\%$ to $32.3\%$ unexpressed, with median activation falling from $89\%$ to $30\%$
of corpus peak. Training on this class does not teach it and does cost the general capability;
replay reduces the collateral damage ($0.9\% \rightarrow 22.9\%$) but rescues no more.

\subsection{The features are reachable from their corpus windows}
\label{app:verb-llm}
\label{app:verb-failures}

Given $16$ of a feature's corpus windows and no direction, an LLM writes text that fires it more
often than the \method{} does, and the whole of its advantage sits in the rare tail.

\begin{table}[htbp]
\centering
\begin{tabular}{lcccc}
\toprule
\% of features fired, best-of-$8$ & rarest & 2nd & 3rd & densest \\
\midrule
\method{} (from the direction)    & 62.4 & 96.7 & 98.3 & 98.6 \\
LLM (from $16$ corpus windows)    & \textbf{84.5} & \textbf{100.0} & 99.2 & 97.6 \\
\bottomrule
\end{tabular}
\caption{$492$ features, rarity strata. Pooled, the \method{} fires $88.8\%$ and the LLM $95.3\%$.
At best-of-$1$ the two are indistinguishable ($83.3\%$ vs.\ $83.1\%$): the LLM's advantage is that
its samples keep finding new ways in, not that any single sample is better.}
\label{tab:verb-llm}
\end{table}

\subsection{Failing rollouts are slightly less self-consistent}
\label{app:verb-consistency}

On $512$ features with $8$ rollouts each, samples for failing features are marginally more varied
than for passing ones --- median pairwise cosine $0.769$ against $0.804$
($p = 1.9\times10^{-5}$), token Jaccard $0.034$ against $0.048$ ($p = 9.0\times10^{-5}$). The
larger and more interpretable gap is against the corpus: failing features' rollouts sit at cosine
$0.513$ from their own corpus windows against $0.594$ for passing ones
($p = 5.8\times10^{-20}$). The \method{} is not producing noise on features it cannot fire; it is
producing coherent text about something adjacent.

\subsection{Output diversity: measurement}
\label{app:verb-diversity}

Figure~\ref{fig:verb-diversity} scores texts by the mean over all pairs of a feature's texts of the Jaccard
similarity of their token-trigram sets. Texts are re-tokenised with the Qwen3.6-27B tokeniser
without special tokens, so a rollout's trailing end-of-sequence token does not count, and every
source is tokenised the same way. The cross-feature floor is the same score over $20{,}000$ random
pairs of texts written for different features.

\paragraph{Sources.} All on a random sample of $2{,}000$ features
of the 131k SAE. The MAEM is the paper's checkpoint with $8$ rollouts per feature under vLLM and, as an engine
check, $4$ under HF: both give median $0.043$. The corpus source is each feature's $32$ stored
max-activating windows; the LLM is Claude Opus 5, $8$ texts on $159$ features. The rare-feature LoRA
was trained for one epoch on $64{,}432$ spans from the rarest band alone. It is an earlier run than
\cref{app:verb-training}'s, and none of its $4{,}027$ training features is in this set. We read its HF
rollouts, from which its scores are also computed. The two later arms read the same way: the
rare-weighted arm trained on a separate rarity-weighted draw, none of whose $7{,}963$ features is in this
set; the cluster arm trained on $195$ of this set's failures, which are dropped. The NLA baseline
was run on rows $1500$--$1999$ of the set, $8$ explanations
each at its native $200$-token budget ($99\%$ parse
as explanations); those rows are a representative slice ($114$--$133$ features per density
quartile). Its \texttt{<explanation>} tags are stripped before embedding. Its vLLM rollouts are excluded: that path attached
the adapter to the untrained base, so they come from the untrained base model.

\paragraph{Estimator.} The all-pairs mean is unbiased for the expected pairwise score at any number
of texts $k$. Sources with $4$, $8$ and $32$ texts per feature therefore share a centre, and smaller
$k$ only widens the spread. The distribution over pairs is heavy-tailed, most visibly for the corpus:
$73\%$ of its pairs share no trigram, and a few overlapping windows from one document carry its
mean. We report the share of disjoint pairs beside the mean for that reason.

\begin{figure}[htbp]
\centering
\includegraphics[width=\linewidth]{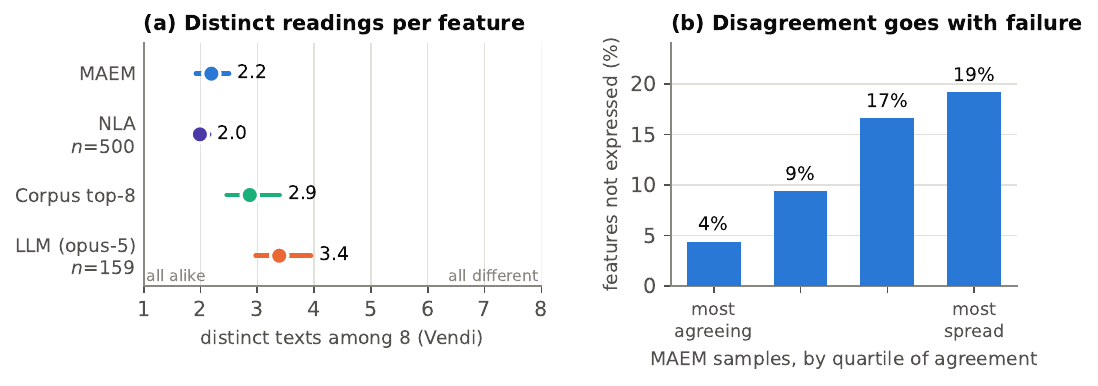}
\caption{\textbf{(a)} Distinct texts among eight samples per feature (Vendi score over bge-small
embeddings): median and interquartile range over the 131k-SAE 2k set. NLA covers $500$ features,
the LLM $159$. \textbf{(b)} Share of features the MAEM fails to fire (below $10\%$ of corpus peak,
best-of-8), by quartile of agreement among its samples.}
\label{fig:verb-diversity-embed}
\end{figure}

\begin{figure}[htbp]
\centering
\includegraphics[width=\linewidth]{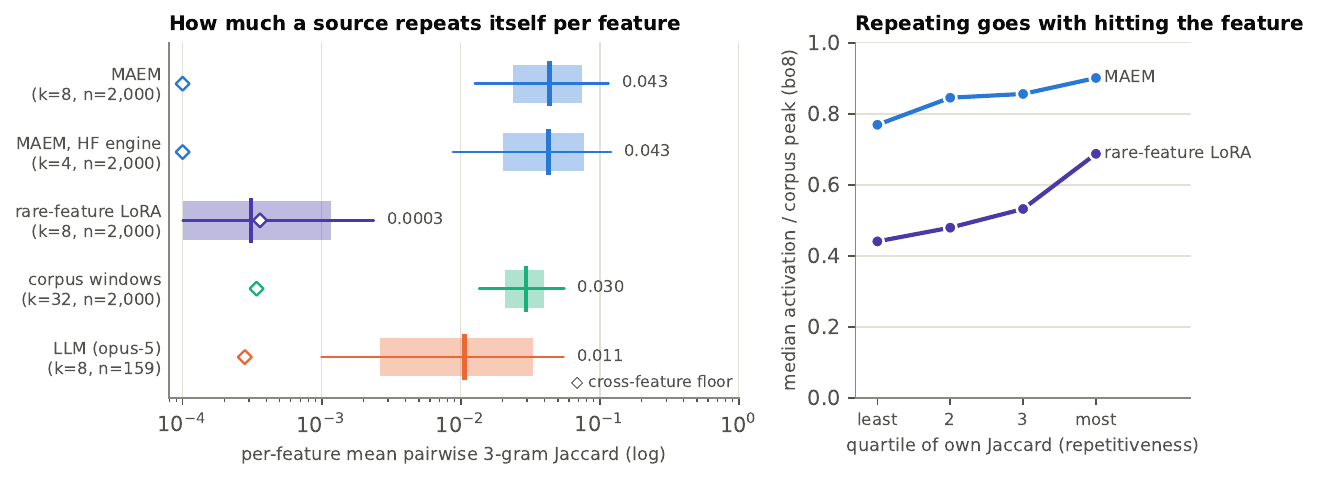}
\caption{Diversity in wording. \textbf{Left:} per-feature mean pairwise token-3-gram Jaccard: box is
the IQR, whiskers the 10th--90th percentile, tick and label the median; $k$ is texts per feature.
The diamond is the same source's cross-feature floor (the MAEM's, $5 \times 10^{-5}$, is drawn at the
axis floor). The all-pairs mean is unbiased in $k$, so rows with $k=4$ and $k=32$ share a centre and
differ only in spread. Scores are not length-matched; the length checks are described above. \textbf{Right:} median
best-of-8 activation, as \% of corpus peak, by quartile of each MAEM's own Jaccard, $500$ features per point.}
\label{fig:verb-diversity}
\end{figure}

\paragraph{Embedding measures.} Figure~\ref{fig:verb-diversity-embed} and Table~\ref{tab:verb-diversity-embed} embed every text, unit-normed,
with bge-small-en-v1.5 (the embedder already used by the evaluation pipeline) and, as a check, with
all-mpnet-base-v2, a different architecture and training set. Every source is read at exactly $k=8$
texts per feature, because unlike the Jaccard mean the Vendi score grows with $k$: the MAEM's
first $8$ vLLM rollouts, the rare-feature LoRA's $8$ HF rollouts, the LLM's $8$ texts, and for the
corpus the feature's top-8 windows, the same windows the LLM was prompted with. The Vendi score is
$\exp\bigl(-\sum_i \lambda_i \log \lambda_i\bigr)$ over the eigenvalues $\lambda_i$ of $K/k$, where
$K$ is the $k \times k$ cosine kernel. Specificity is the median within-feature mean pairwise cosine
minus the mean cosine of $20{,}000$ random pairs of texts written for different features. Coverage
targets are the feature's corpus windows $9$--$32$, dropping any that share $\geq 0.2$ trigram
Jaccard with one of windows $1$--$8$ (stored windows can overlap within a document); a median of
$19$ remain. Coverage is the mean over targets of the cosine to the nearest of the source's eight
texts, minus the same with a random other feature's texts. The two embedders' cosines live on
different scales (mpnet's cross-feature mean is $0.04$--$0.07$, bge-small's $0.45$--$0.49$), so only
orderings are compared across them.

\begin{table}[htbp]
\centering
\small
\begin{tabular}{lcccccc}
\toprule
& \multicolumn{3}{c}{bge-small} & \multicolumn{3}{c}{all-mpnet-base-v2} \\
\cmidrule(lr){2-4}\cmidrule(lr){5-7}
Source & Vendi & specificity & coverage & Vendi & specificity & coverage \\
\midrule
MAEM                    & 2.19 & \textbf{0.31} & 0.108 & 3.10 & \textbf{0.58} & 0.205 \\
corpus top-8             & 2.87 & 0.19 & \textbf{0.143} & 4.45 & 0.34 & \textbf{0.287} \\
LLM (Opus 5, $n=159$)    & 3.39 & 0.19 & 0.140 & 5.74 & 0.30 & 0.255 \\
NLA ($n=500$)         & \textbf{1.99} & 0.17 & 0.055 & \textbf{3.00} & 0.37 & 0.105 \\
\midrule
rare-feature LoRA        & 4.17 & 0.05 & 0.075 & 7.26 & 0.08 & 0.145 \\
rare-weighted arm        & 4.24 & 0.04 & 0.070 & 7.37 & 0.07 & 0.142 \\
cluster arm ($n=1{,}805$) & 4.42 & 0.02 & 0.051 & 7.58 & 0.03 & 0.112 \\
\midrule
MAEM, texts cut to 32 tokens & 2.22 & 0.30 & --- & 3.16 & 0.57 & --- \\
\bottomrule
\end{tabular}
\caption{Diversity in meaning, both embedders; medians over features, $k=8$. Bold marks the extreme
of each column among the sources above the first rule: the MAEM is the most specific, NLA and
the MAEM the least spread, and the corpus top-8 covers the feature's other windows best. NLA's
coverage is not comparable, since it writes descriptions of text rather than text like the corpus
windows. Below the rule are the three rare-feature training runs, all read on features they were not
trained on: each is spread only because it is barely specific. On the LLM's $159$ features alone the MAEM's Vendi is $2.32$
(bge-small) and its raw coverage $0.591$ against the LLM's $0.664$.}
\label{tab:verb-diversity-embed}
\end{table}

\paragraph{Robustness.} The MAEM's median is flat across the set's four quartiles of pool peak
activation ($0.039$--$0.046$), so the score does not track how strongly a feature fires. The
quartiles in the right panel are formed by rank, because the rare-feature LoRA's scores tie heavily
near zero. Length checks for the Jaccard score cut every text to its first $32$ tokens, and to
tokens $8$--$40$ to remove the MAEM's shared opening words; for the embeddings, generated texts are
cut to their first $32$ tokens and corpus windows to $32$ tokens around their marked peak, since only
$56\%$ of peaks fall in a window's last $32$ tokens and a blind cut would drop them.

\end{document}